\documentclass[10pt]{article}

\usepackage[english]{babel}
\usepackage{amsmath}
\usepackage{hyperref}
\usepackage{amsthm}
\usepackage[shortlabels]{enumitem}
\usepackage[title]{appendix}
\usepackage{amsfonts}
\usepackage{mathtools}

\usepackage[ruled,vlined]{algorithm2e}
\usepackage{mathrsfs}
\usepackage{caption}
\usepackage[capbesideposition=outside,capbesidesep=quad]{floatrow}
\usepackage{tabu}
\usepackage{booktabs}
\usepackage{lipsum}
\usepackage{url}
\usepackage{subcaption}
\DeclareSymbolFont{bbold}{U}{bbold}{m}{n}
\DeclareSymbolFontAlphabet{\mathbbold}{bbold}
\usepackage{graphicx,xcolor} 
\usepackage[framemethod=tikz]{mdframed}
\usepackage{csquotes}
   \usepackage{tabu}
   \allowdisplaybreaks
\newtheorem{theorem}{Theorem}[section]
\newtheorem{lemma}{Lemma}[section]
\newtheorem{prop}{Proposition}[section]
\newtheorem{definition}{Definition}[section]
\newtheorem{corollary}{Corollary}[section]

\newcommand{\appropto}{\mathrel{\vcenter{
  \offinterlineskip\halign{\hfil$##$\cr
    \propto\cr\noalign{\kern2pt}\sim\cr\noalign{\kern-2pt}}}}}

\newcommand{\R}{\mathbb{R}}

\usepackage{authblk}

\DeclareMathOperator*{\argmin}{argmin} 

\newcommand{\ths}{\textsuperscript{th} }

\newcommand{\MELD}{\text{MELD}_\epsilon(X)}
\newcommand{\MELDemph}{\text{\emph{MELD}}_\epsilon(X)}

\usepackage{blindtext}

\title{A Multiscale Environment for Learning by Diffusion}

\author{James M. Murphy }
\author{Sam L. Polk\thanks{Corresponding Author. Present Address: 503 Boston Avenue, Medford, MA, USA. \newline \indent \; \textit{Email Addresses}: \url{JM.Murphy@Tufts.edu} (James M. Murphy), \url{Samuel.Polk@Tufts.edu} (Sam L. Polk)}}
\affil{Department of Mathematics, Tufts University}
\date{}

\begin{document}
\maketitle

\begin{abstract}
Clustering algorithms partition a dataset into groups of similar points. The clustering problem is very general, and different partitions of the same dataset could be considered correct and useful. To fully understand such data, it must be considered at a variety of scales, ranging from coarse to fine.  We introduce the Multiscale Environment for Learning by Diffusion (MELD) data model, which is a family of clusterings parameterized by nonlinear diffusion on the dataset. We show that the MELD data model precisely captures latent multiscale structure in data and facilitates its analysis.  To efficiently learn the multiscale structure observed in many real datasets, we introduce the Multiscale Learning by Unsupervised Nonlinear Diffusion (M-LUND) clustering algorithm, which is derived from a diffusion process at a range of temporal scales. We provide theoretical guarantees for the algorithm's performance and establish its computational efficiency. Finally, we show that the M-LUND clustering algorithm detects the latent structure in a range of synthetic  and real datasets. \end{abstract}


\section{Introduction} \label{sec: Intro 1}

Unsupervised machine learning algorithms detect structure in data given no known class labels~\cite{friedman2001elements}. Among the many branches of unsupervised learning, clustering is perhaps the most developed and widely used. A clustering algorithm partitions a dataset into groups. In a good partition, data points from the same group are ``similar'' to one another, while data points from distinct groups are ``dissimilar'' from one another. The specific notion of similarity used varies widely~\cite{vonluxburg2007spectralclustering, xu2005survey, ng2002spectralclustering, rodriguez2014clustering, lyzinski2016HSBM, shi2000ncut}. Often, cluster analysis is one of the first tasks performed by a user interested in learning more about an unexplored dataset. 

Given no further information, the clustering problem is quite general. One could easily imagine cases in which there are multiple correct separations in a single dataset, and the most useful clustering often depends on the specifics of how it will be applied in practice. A coarse separation of a dataset may be desired in one problem setting, while another problem setting may call for finer separation within the data. Thus, it may make sense to consider a dataset at various scales and analyze all of the many possible “correct” partitions. The property of data having multiple scales of relevant structure is readily observable in many empirical datasets; for example, in social networks (e.g., geographical community structures), protein-protein interaction networks (e.g., scales of chemical secondary structures), and gene interaction networks (e.g., co-expressed gene clusters) \cite{ahn2010link, delmotte2011protein, song2015multiscale}. Thus, to understand the structure of a dataset in its entirety, it is necessary to understand how it is structured at a multitude of scales.

Recent decades have brought significant advances in the development of clustering algorithms meant to detect multiple scales of separation~\cite{peixoto2014hierarchicalstructureSN, lyzinski2016HSBM,  liu2020MarkovStability, azran2006spectralclustering}. Typically, these approaches have relied upon a data model allowing for latent hierarchical structure in a dataset to provide performance guarantees on clustering algorithms. However, in many data models allowing for multiscale cluster structure, it is difficult to understand how separation at one scale relates to separation at another~\cite{lyzinski2016HSBM, peixoto2014hierarchicalstructureSN}. Diffusion geometry on graphs has been proposed to efficiently capture latent low-dimensional structure in high dimensional data, where the time scale of the diffusion process corresponds to a scale of separation---short time scales reflect fine, local structures in the data, while large time scales reflect coarse, global structures in the data~\cite{coifman2006diffusionmaps, coifman2005PNAS, nadler2006diffusion_maps_ACHA, murphy2019LUND}.  It is of interest to understand the precise nature of this time scaling and build clustering algorithms that allow for all time scales of interest to be considered simultaneously.

\subsection{Major Contributions}

This article makes two significant contributions. The first is the \emph{Multiscale Environment for Learning by Diffusion} (\emph{MELD}) data model. The MELD data model is a family of clusterings, parameterized by a diffusion time parameter.  For each of these clusterings, we show that diffusion distances (a data-dependent distance metric) between clusters are bounded away from the diffusion distances within clusters during an interval determined by the geometric properties of the underlying data and that clustering. We show that clusterings with coherent and well-separated clusters are more stable in the diffusion process and emphasized within the MELD data model. Finally, we show that when the clusterings in the MELD data model exhibit hierarchical structure, the number of latent clusters is monotonically non-increasing as a function of the diffusion time parameter.

The second major contribution is the \emph{Multiscale Learning by Unsupervised Nonlinear Diffusion} (\emph{M-LUND}) clustering algorithm. This algorithm is a multiscale generalization of the Learning by Unsupervised Nonlinear Diffusion (LUND) algorithm, which leverages diffusion distances' attractive theoretical properties to efficiently and accurately cluster high-dimensional data~\cite{murphy2019LUND}. The M-LUND algorithm extracts all clusterings in the MELD data model using the LUND algorithm. It then chooses the clustering that minimizes the variation of information (VI) between nontrivial extracted clusterings~\cite{meilua2007VI}. In this way, it is able to not only suggest a few salient clusterings but also output the one that best represents all the others from an information-theoretic perspective. In addition to theoretical guarantees, we show the strong empirical performance of M-LUND on synthetic datasets associated with poor performance of many popular clustering algorithms~\cite{murphy2019LUND, nadler2007failures}, as well as a range of real data~\cite{DuaUCI2019, gualtieri1999salinasA}.

\subsection{Notation and Outline}
\label{sec: paper structure 1}
Abbreviations are provided below\footnote[2]{
DGM: Diffusion Geometry Model. \hspace{1.25in} M-LUND: Multiscale Learning by Unsupervised Nonlinear Diffusion.

DPC: Density Peak Clustering. \hspace{1.46in} MELD: Multiscale Environment for Learning by Diffusion.  

GDM: Geometric Data Model. \hspace{1.5in} MMS: Multiscale Markov Stability.

HSBM: Hierarchical Stochastic Blockmodel. \hspace{0.81in} NMI: Normalized Mutual Information.

HSC: Hierarchical Spectral Clustering. \hspace{1.09in} SBM: Stochastic Blockmodel.

HSI: Hyperspectral Image \hspace{1.76in} SC: Spectral Clustering.

KDE: Kernel Density Estimate. \hspace{1.43 in} SLC: Single-Linkage Clustering.

LUND: Learning by Unsupervised Nonlinear Diffusion. \hspace{0.24in} VI: Variation of Information.}.  Notation used throughout this article appears in Table \ref{tab:notation}. In Section \ref{sec:Background}, we review preliminaries and introduce pertinent background on how graph diffusion is well-suited to the clustering problem.  In Section \ref{sec: MELD 3}, we introduce the MELD data model, which we will show efficiently captures multiscale cluster structure within a dataset.  In Section \ref{sec: MLUND 4}, we present and analyze the M-LUND algorithm, which leverages the theory of Section \ref{sec: MELD 3} to detect the most representative clustering among the many possible latent clusterings of a dataset.  In Section \ref{sec: Numerical 5}, we provide numerical corroboration of the theory developed in Sections \ref{sec: MELD 3} and \ref{sec: MLUND 4} on synthetic data and also present comparisons of the M-LUND clustering algorithm against related clustering schemes on eleven real-world benchmark datasets and one real-world HSI in Section \ref{sec: Numerical 5}. In Section \ref{sec: Conclusion}, we conclude and discuss future research.

\begin{table} 
    \centering
    \begin{tabular}{|p{0.94in}|p{4.83in}|}
         \hline
        \textbf{Notation} &  \textbf{Interpretation} \\
        \hline
        \hline
        $X=\{x_i\}_{i=1}^n$ & Data points to cluster \\ 
        \hline 
        $X_k$ / $X_k^{(\ell)}$ & The $k$\ths cluster of the clustering $\{X_k\}_{k=1}^{K}$ /
        $\{X_k^{(\ell)}\}_{k=1}^{K_\ell}$ \\
        \hline
        $\mathcal{C}$ & Estimated cluster assignments \\
        \hline 
        $\textbf{W}$ & Weight matrix  \\
        \hline 
        $\sigma$ & Diffusion scale parameter\\
        \hline 
        $\textbf{D}$ & Degree matrix \\
        \hline
        $\mathbf{P}$ & Markov transition matrix \\
        \hline
        $\pi$ & Stationary distribution of $\mathbf{P}$  \\
        \hline
        $\{(\psi_i,\lambda_i)\}_{i=1}^n$ & Right eigenvectors and eigenvalues of $\mathbf{P}$, sorted according to $|\lambda_i|$\\
        \hline
        $\Phi(x)$ & Laplacian eigenmap, evaluated at $x\in X$ \\
        \hline
        $D_t(x,y)$ & Diffusion distance between points $x$ and $y$ at diffusion time step $t$ \\
        \hline 
        $\Psi_t(x)$ & Diffusion map at time $t$, evaluated at $x\in X$ \\
        \hline
        $\mathbf{S}$ / $\mathbf{S}^{(\ell)}$ & Stochastic complement of $\mathbf{P}$ with respect to the clustering $\{X_k\}_{k=1}^K$  / $\{X_k^{(\ell)}\}_{k=1}^{K_\ell} $\\
        \hline
        $\mathbf{S}^\infty$ / $\mathbf{S}_\infty^{(\ell)}$ & $\lim_{t\rightarrow\infty}\mathbf{S}^t$ /   $\lim_{t\rightarrow\infty}[\mathbf{S}^{(\ell)}]^t$\\
        \hline
        $\textbf{Z}$ / $\textbf{Z}^{(\ell)}$& Invertible matrix diagonalizing $\mathbf{S}$ / $\mathbf{S}^{(\ell)}$ \\
        \hline 
        $\lambda_{K+1}$ / $\lambda_{K_\ell+1}^{(\ell)}$ & First non-unity eigenvalue of $\mathbf{S}$ / $\mathbf{S}^{(\ell)}$\\
        \hline
        $\delta$ / $\delta^{(\ell)}$ & $\|\mathbf{P}-\mathbf{S}\|_\infty$ / $\|\mathbf{P}-\mathbf{S}^{(\ell)}\|_\infty$\\
        \hline
        $\kappa$ / $\kappa^{(\ell)}$ & Infinity-norm condition number of diagonalizing $\mathbf{S}$ / $\mathbf{S}^{(\ell)}$ \\
        \hline 
        $\mathcal{I}_\epsilon$   /  $\mathcal{I}_\epsilon^{(\ell)}$  & Interval during which $\{X_k\}_{k=1}^K$ / $\{X_k^{(\ell)}\}_{k=1}^{K_\ell}$ is $\epsilon$-separable by diffusion distances\\
        \hline
        $D_t^{\text{in}}$ /  $D_t^{\text{in}}(\ell)$ & Maximum within-cluster diffusion distance for $\{X_k\}_{k=1}^{K}$ / $\{X_k^{(\ell)}\}_{k=1}^{K_\ell}$\\
        \hline
        $D_t^{\text{btw}}$ / $D_t^{\text{btw}}(\ell)$ & Minimum between-cluster diffusion distance for $\{X_k\}_{k=1}^{K}$ / $\{X_k^{(\ell)}\}_{k=1}^{K_\ell}$\\
        \hline
        $\gamma(t)$ / $\gamma^{(\ell)}(t)$  & Measure of how the $\ell^1$- and $\ell^2$-norm differ across rows of $\mathbf{P}^t-\mathbf{S}^\infty$ / $\mathbf{P}^t-\mathbf{S}_\infty^{(\ell)}$\\
        \hline
        $p(x)$ & Kernel density estimate, evaluated at $x\in X$  \\
        \hline
        $\sigma_0$ & KDE bandwidth  \\
        \hline 
        $NN(x,N)$ & Set of $N$ $\ell^2$-nearest neighbors of $x$ in the dataset $X$\\
        \hline 
        $\rho_t(x)$ & Diffusion distance between $x$ and the $D_t$-nearest neighbor of $x$ of higher density  \\
        \hline
        $\mathcal{D}_t(x)$ & $p(x)\rho_t(x)$ \\
        \hline
        $M$ & Number of distinct latent clusterings of $X$  \\
        \hline
        $\mathcal{C}_t$ & Latent clustering at time $t$\\
        \hline
        $\MELD$ & The MELD data model of $X$ for $\epsilon\in \Big(0,\frac{1}{\sqrt{n}}\Big)$\\
        \hline
        $A_\epsilon$ & $\bigcup_{\ell=1}^M \mathcal{I}_\epsilon^{(\ell)}$\\
        \hline
        $H(\mathcal{C})$ & Entropy of the clustering $\mathcal{C}$  \\
        \hline
        $I(\mathcal{C}, \mathcal{C}')$ & Mutual information between clusterings $\mathcal{C}$ and $\mathcal{C}'$\\
        \hline
        $VI(\mathcal{C}, \mathcal{C}')$ & Variation of information between clusterings $\mathcal{C}$ and $\mathcal{C}'$ \\
        \hline
        $\beta$ & Exponential sampling rate\\
        \hline 
        $T$ & $\Big\lceil\log_\beta\Big(\log_{|\lambda_2|}\Big(\frac{\tau\pi_{\min}}{2}\Big)\Big)\Big\rceil$ \\
        \hline
        $\tau$ & Stationarity threshold \\
        \hline
        $J$ & Times during which nontrivial clusterings are extracted by the LUND algorithm \\
        \hline
        $\text{VI}^{(\text{tot})}(\mathcal{C}_{t})$ & $\sum_{s\in J}\text{VI}(\mathcal{C}_{t}, \mathcal{C}_s)$\\
        \hline
        $\mathcal{M}_t$ & Cluster-wise empirical density maximizers for the latent clustering at time $t$\\
        \hline
        $\{x_{m_k}^{(t)}\}_{k=1}^n$ & The points in $X$, sorted according to $\mathcal{D}_t(x)$\\
        \hline
        $B_\epsilon$ & Transition regions between clusterings in $\MELD$\\
        \hline
        $\Delta(t)$ & Relative pointwise distance of $\mathbf{P}^t$ to its stationary distribution $\pi$\\
        \hline        
        $d$ & Doubling dimension of the dataset\\
        \hline
        $NMI(\mathcal{C}, \mathcal{C}')$ & Normalized mutual information between clusterings $\mathcal{C}$ and $\mathcal{C}'$\\\hline
        \end{tabular}
    \caption{Notation used in the article listed in order of its first appearance. We refer to the multiscale analogues of certain notations after a back-slash.}
    \label{tab:notation} 
\end{table}

\section{Background} \label{sec:Background}

\subsection{Background on Unsupervised Clustering }

In this section, background is provided on some of the many approaches to clustering data. Each of these algorithms has particular advantages and disadvantages, which are typically data- and task-dependent.  Clustering algorithms partition a dataset $X=\{x_i\}_{i=1}^n\subset\mathbb{R}^{D}$ into $K$ subsets $X_1,\dots,X_K$. The partition $\{X_k\}_{k=1}^K$ is called a \emph{clustering} of $X$, while each $X_k$ is called a \emph{cluster}. Clustering algorithms are typically \emph{unsupervised}, meaning that no expert annotations or labels are used in the partitioning of $X$. Thus, the number of clusters $K$ is often (though not always~\cite{murphy2019LUND, Fan2019_Unified, Little2020_Path}) a hyperparameter in clustering algorithms. Typically, we want a clustering to satisfy both a separation condition---that if $k\neq k'$, all points in $X_k$ are ``far'' from those in $X_{k'}$---and a coherence condition---that all points in each $X_k$ are ``close.''

\subsubsection{K-Means Clustering}\label{sec:kmeans 2}

The $K$-means algorithm is a classical clustering  algorithm that remains widely used. This algorithm learns clusters $\{X_k\}_{k=1}^K$ by optimizing the distance of points to cluster means:  $\mathcal{C} = \text{argmin}_{\{X_k\}_{k=1}^K} \sum_{k=1}^K \sum_{x\in X_k} \|x-~\bar{x}_k\|_2^{2}$, where $\bar{x}_k$ denotes the mean data point of a cluster $X_k$. Many variants of $K$-means exist~\cite{wagstaff2001constrainedkmeans, likas2003globalkmeans, arthur2006kmeans++, park2009kmediods}. However, it is easy to see that $K$-means is sensitive to outliers because of its use of Euclidean distances. One extension of $K$-means that is less sensitive to outliers is the $K$-medoids clustering algorithm, which replaces the cluster mean with a cluster medoid~\cite{park2009kmediods}. Nevertheless, $K$-means and its variants exhibit poor performance on data that do not resemble well-separated, near-spherical clusters of the same size~\cite{ng2002spectralclustering}.

\subsubsection{Dendrogram-Based Hierarchical Clustering} \label{sec:HC 2}

Dendogram-based clustering algorithms extract a family of partitions from a dataset $X$, varying from fine to coarse in scale, that can be expressed in a \emph{dendrogram}: a diagram representing a tree of clusterings~\cite{gower1969MST, friedman2001elements}. More formally, a dendrogram represents a family of $n$ clusterings $\big\{\{X_k^{(\ell)}\}_{k=1}^{\ell}\big\}_{\ell = 1}^n$ such that $\{X_k^{(n)}\}_{k=1}^{n}$ is the clustering consisting of $n$ singletons and $\{X_k^{(1)}\}_{k=1}^{1}$ is the clustering consisting of a single cluster.  Agglomerative hierarchical clustering algorithms initialize at $\ell=n$ and create intermediate clusterings $\{X_k^{(\ell)}\}_{k=1}^{\ell}$ by merging two clusters in $\{X_k^{(\ell+1)}\}_{k=1}^{\ell+1}$ found to minimize a linkage function~\cite{gower1969MST, friedman2001elements}. On the other hand, divisive hierarchical clustering algorithms initialize at $\ell=1$ and create intermediate clusterings $\{X_k^{(\ell+1)}\}_{k=1}^{\ell+1}$ by splitting a cluster at each scale.  One of the more popular hierarchical clustering algorithms is the single-linkage clustering (SLC) algorithm, which builds a hierarchy of partitions by iteratively merging clusters using $\mathcal{L}_{\text{SLC}}(X_1,X_2) =\min_{x_1\in X_1, x_2\in X_2} \|x_1~-~x_2\|_2$ as its linkage function~\cite{gower1969MST, friedman2001elements}. Despite its widespread use in practice, SLC has been shown to be statistically inconsistent if the dimension of the dataset is greater than 1~\cite{hartigan1981consistency}. Moreover, its linkage function's reliance on Euclidean distances makes it sensitive to small perturbations and outliers.

\subsubsection{Density Peak Clustering} 
\label{sec: density 2}

Density-based clustering algorithms learn regions of high and low empirical density to cluster a dataset~\cite{xu1998density, ester1996density, fukunaga1975meanshift, comaniciu2002meanshift, jisu2016statistical}. \emph{Density peak clustering} (DPC) is a widely-utilized example of a mode-based clustering algorithm~\cite{rodriguez2014clustering}. DPC labels high-density points that are far in Euclidean space from other high-density points as \emph{modes} of clusters in the dataset. Non-modal points are then paired with a labeled nearest neighbor iteratively.  Due to its use of Euclidean distances, DPC often fails on data with nonlinear structure~\cite{murphy2019LUND}.

\subsection{Background on Spectral Graph Theory and its Applications to Clustering} \label{sec: Spectral_Graph_Theory 2}

Spectral graph theory is widely used in clustering~\cite{shi2000ncut, ng2002spectralclustering, azran2006spectralclustering, vonluxburg2007spectralclustering, liu2020MarkovStability, Little2020_Path, murphy2019LUND}. Typically, spectral methods construct a local connectivity graph that stores information about the pairwise similarity between data points~\cite{ng2002spectralclustering, shi2000ncut}. The spectral decomposition of the adjacency matrix or graph Laplacian can then be used to locate highly connected regions within the graph~\cite{shi2000ncut}. Because spectral methods rely on nonlinear transformations derived from graph structure, they are highly effective at clustering datasets containing nonlinear or elongated structures~\cite{ng2002spectralclustering, vonluxburg2007spectralclustering}. This is in contrast to $K$-means and DPC, which may fail on datasets containing these structures~\cite{murphy2019LUND}. 

\subsubsection{Spectral Graph Theory}
In spectral graph theory, the points in $X$ may be represented as nodes in a graph. Let the edge weight between two nodes $x_i$ and $x_j$ be $\textbf{W}_{ij}$. Typically, $\textbf{W}_{ij}$ is computed using a symmetric, radial, and rapidly decaying similarity measure such as $\textbf{W}_{ij}= \exp\big(-\|x_i-x_j\|_2^2/\sigma^2\big)$ for some choice of scaling parameter $\sigma>0$ that reflects the interaction radius between points~\cite{shi2000ncut}. If $\sigma$ is large, then long-range interactions between points are considered, while if $\sigma$ is small, only short-range interactions are emphasized.

One can construct a Markov transition matrix $\textbf{P}\in\mathbb{R}^{n\times n}$ associated to $\textbf{W}\in\mathbb{R}^{n\times n}$ with an appropriate normalization~\cite{ng2002spectralclustering, vonluxburg2007spectralclustering, coifman2006diffusionmaps}. Let the degree matrix $\textbf{D}$ be the diagonal matrix with $\textbf{D}_{ii} = \sum_{j=1}^{n}\textbf{W}_{ij}$. We call $\textbf{D}_{ii}$ the \emph{degree} of the point $x_i\in X$.  Let $\textbf{P} = \textbf{D}^{-1}\textbf{W}.$  This matrix stores transition probabilities for a Markov diffusion process on the dataset, where $P_{ij}$ reflects the probability of transitioning from $x_i$ to $x_j$.\footnote[3]{We remark that, with an abuse of notation, $P_{ij}$ denotes the entries of $\textbf{P}$, while $\textbf{P}_{ij}$ shall denote block submatrices of $\textbf{P}$.} We assume that the Markov chain described by $\mathbf{P}$ is reversible, irreducible (i.e., the graph is connected), and aperiodic. Hence, $\mathbf{P}$ has a unique stationary distribution $\pi$ satisfying $\pi \textbf{P} = \pi$~\cite{levin2017markov_mixing}. The eigendecomposition of $\mathbf{P}$ is strongly associated with connectivity in $X$, making it useful for clustering. Let $\{\psi_i\}_{i=1}^n$ be the right eigenfunctions of $\mathbf{P}$ with corresponding eigenvalues $\{\lambda_i\}_{i=1}^n$. We will order eigenvalues according to $|\lambda_i|$ in non-increasing order; so when we say the ``first $k$ eigenfunctions,'' we refer to the $k$ eigenfunctions $\psi_i(x)$ corresponding to $|\lambda_i|$ closest to 1.   In general, the multiplicity of the unity eigenvalue is the number of connected components in the graph~\cite{vonluxburg2007spectralclustering}, which is 1 by our assumption that $\mathbf{P}$ is irreducible. 

Each eigenfunction $\psi_i(x)$ of $\mathbf{P}$ is also an eigenfunction of the random walk graph Laplacian $\textbf{L}_\text{rw}~=~\textbf{I}-\textbf{D}^{-1}\textbf{W}$ corresponding to the eigenvalue $1-\lambda_i$. The graph Laplacian is a discrete approximation of the Laplacian operator, so the eigenvectors of $\textbf{L}_{\text{rw}}$ (and therefore $\mathbf{P}$) are discrete approximations of the continuous eigenfunctions of the Laplace operator~\cite{mohar1991laplacian, szlam2005diffusion}. Each eigenfunction $\psi_i(x)$ has a frequency related to the corresponding eigenvalue $\lambda_i$. Hence, we will say that an eigenfunction $\psi_i(x)$ of $\mathbf{P}$ is \emph{low-frequency} if $\lambda_i$ is close to 1 and \emph{high-frequency} if $\lambda_i\ll 1$. In particular, the $K$ lowest-frequency eigenfunctions of the graph Laplacian of $X$ tend to concentrate on the $K$ components of the graph that are most highly connected. This property has been used to cluster data with nonlinear structure~\cite{lafon2006diffusion, ng2002spectralclustering, shi2000ncut, vonluxburg2007spectralclustering, Little2020_Path}. 

\subsubsection{Spectral Clustering} \label{sec: Spectral_Clustering}

Many classical clustering algorithms perform well when applied to certain classes of well-behaved data but fail on datasets with nonlinear structure~\cite{vonluxburg2007spectralclustering,lafon2006diffusion, ng2002spectralclustering, murphy2019LUND}. Applying the eigenmap $\Phi(x)=(\psi_{1}(x),\psi_{2}(x),\dots,\psi_{K}(x))$ for $K\le n$ as a preprocessing step before the application of $K$-means often produces better separation in a new data-dependent feature space independent of nonlinear structure in $X$~\cite{ng2002spectralclustering, shi2000ncut}. This is, in its essence, the spectral clustering (SC) algorithm. Typically (but not always~\cite{azran2006spectralclustering}), the number of clusters $K$ is assumed a priori and the first $K$ eigenvectors of $\mathbf{P}$ are extracted to compute $\Phi(X)$. A simple clustering algorithm like $K$-means is then applied to $\Phi(X)$ rather than $X$, usually after a normalization step~\cite{ng2002spectralclustering}. Since SC was first introduced~\cite{shi2000ncut, ng2002spectralclustering}, its theoretical properties have been investigated~\cite{rohe2011spectral, arias2011clustering, shi2000ncut, schiebinger2015geometry}. It was shown that, when $K=2$, SC produces an approximate solution to the normalized graph cut problem~\cite{shi2000ncut}. However, there are some classes of data for which SC has been observed to fail; for example, datasets with structure of varying in scale and/or density~\cite{nadler2007failures}. 

\subsection{Background on Diffusion Geometry}\label{sec: Diffusion Maps 2}

The matrix $\mathbf{P}$ is the transition matrix for a Markov diffusion process on a graph generated from the dataset $X$. Diffusion distances capture the structure encoded in $\mathbf{P}$ as a data-dependent distance metric between points~\cite{coifman2006diffusionmaps, coifman2005PNAS, nadler2006diffusion_maps_ACHA}.  

\begin{definition}
Let $\mathbf{P}$ be an irreducible, aperiodic Markov transition matrix on $X\subset \R^D$ with stationary distribution $\pi$. For points $x_i,x_j\in X$ and $t\geq 0$, let $p_t(x_i,x_j) = (P^t)_{ij}$. The \emph{diffusion distance} at time $t$ between $x_i$ and $x_j$ is defined to be  $D_t(x_i,x_j)  = \| p_t(x_i,:) - p_t(x_j,:)\|_{\ell^2(1/\pi)} = \sqrt{\sum_{u\in X} [p_t(x_i,u) - p_t(x_j,u)]^2\frac{1}{\pi(u)} }$.
\label{def: diffusiondistance}
\end{definition}

Importantly, diffusion distances are data-dependent, enabling the detection of nonlinear structure in data \cite{nadler2006diffusion_maps_ACHA, coifman2006diffusionmaps, coifman2005PNAS}. Moreover, diffusion distances have a natural connection with the clustering problem. The diffusion distance at time $t$ can be identified as the Euclidean distance between rows of $\mathbf{P}^t$, weighted according to $1/\pi$. If each cluster in a clustering of $X$ is highly-connected, irreducible, and well-separated from other clusters, then $p_t(x_i,:)$ will be nearly equal to $p_t(x_j,:)$ for any pair of points $x_i$ and $x_j$ in the same cluster $X_k$, implying a low diffusion distance between points within the same cluster. Conversely, if $x_i$ and $x_j$ are in distinct clusters, $p_t(x_i,:)$ is expected to be very different from $p_t(x_j,:)$.  This will be formalized in Section \ref{sec: recent_results}.

\begin{definition}\label{def: diffusionmap}
Let $\{(\psi_i, \lambda_i)\}_{i=1}^n$ be the right eigenvector-eigenvalue pairs of an irreducible, aperiodic  transition matrix $\mathbf{P}$, sorted according to $|\lambda_i|$ in non-increasing order. The \emph{diffusion map} at time $t\geq 0$ is defined to be $ \Psi_t(x) = (\psi_1(x), \lambda_2^t \psi_2(x), \dots, \lambda_n^t\psi_n(x))$. \end{definition}

\begin{figure}[t]
\centering
    \begin{subfigure}[t]{0.333\textwidth}
        \centering
        \includegraphics[height=1.9in]{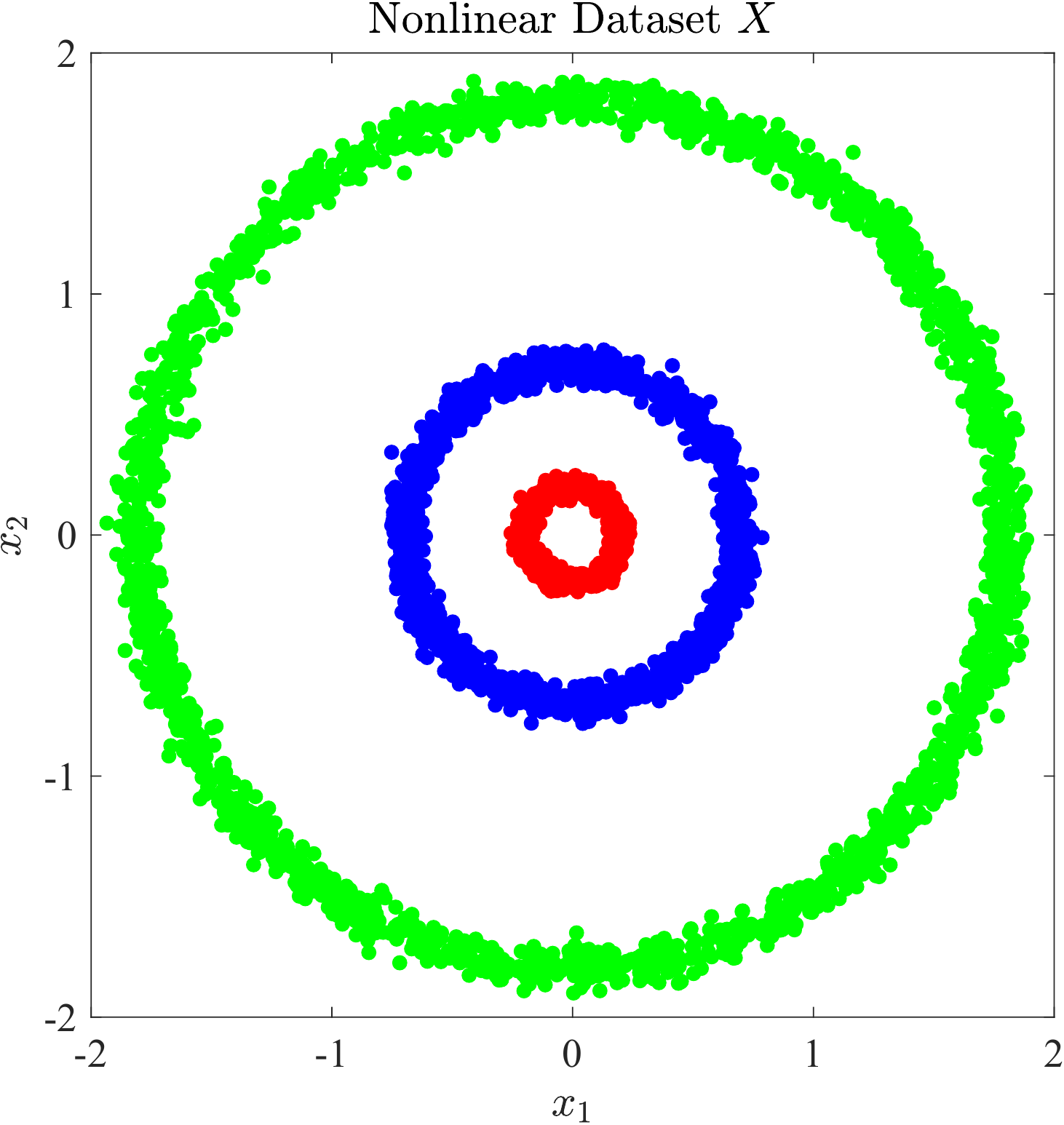}
    \end{subfigure}%
    \begin{subfigure}[t]{0.333\textwidth}
        \centering
        \includegraphics[height=1.9in]{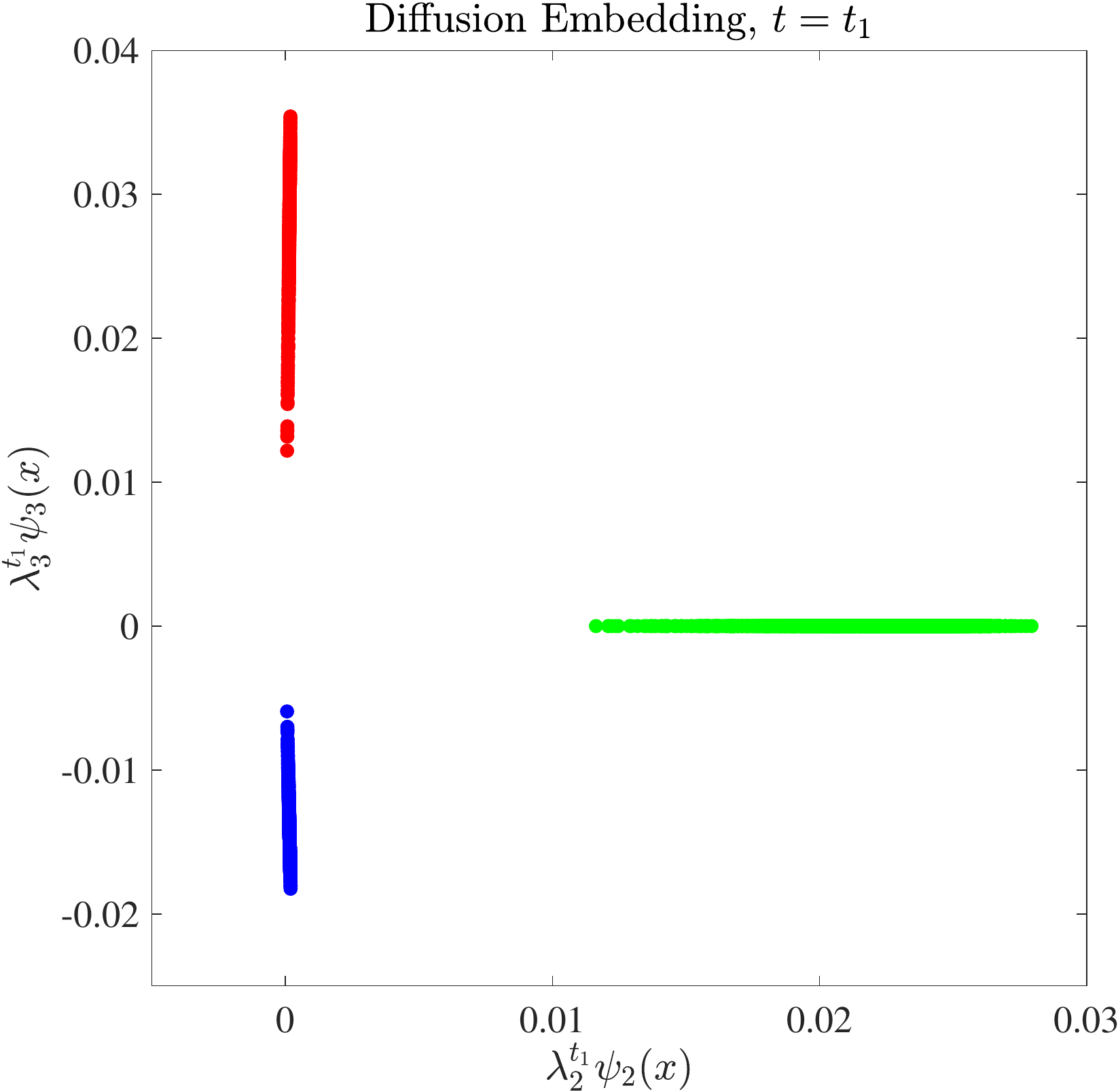}
    \end{subfigure}%
    \begin{subfigure}[t]{0.333\textwidth}
        \centering
        \includegraphics[height=1.9in]{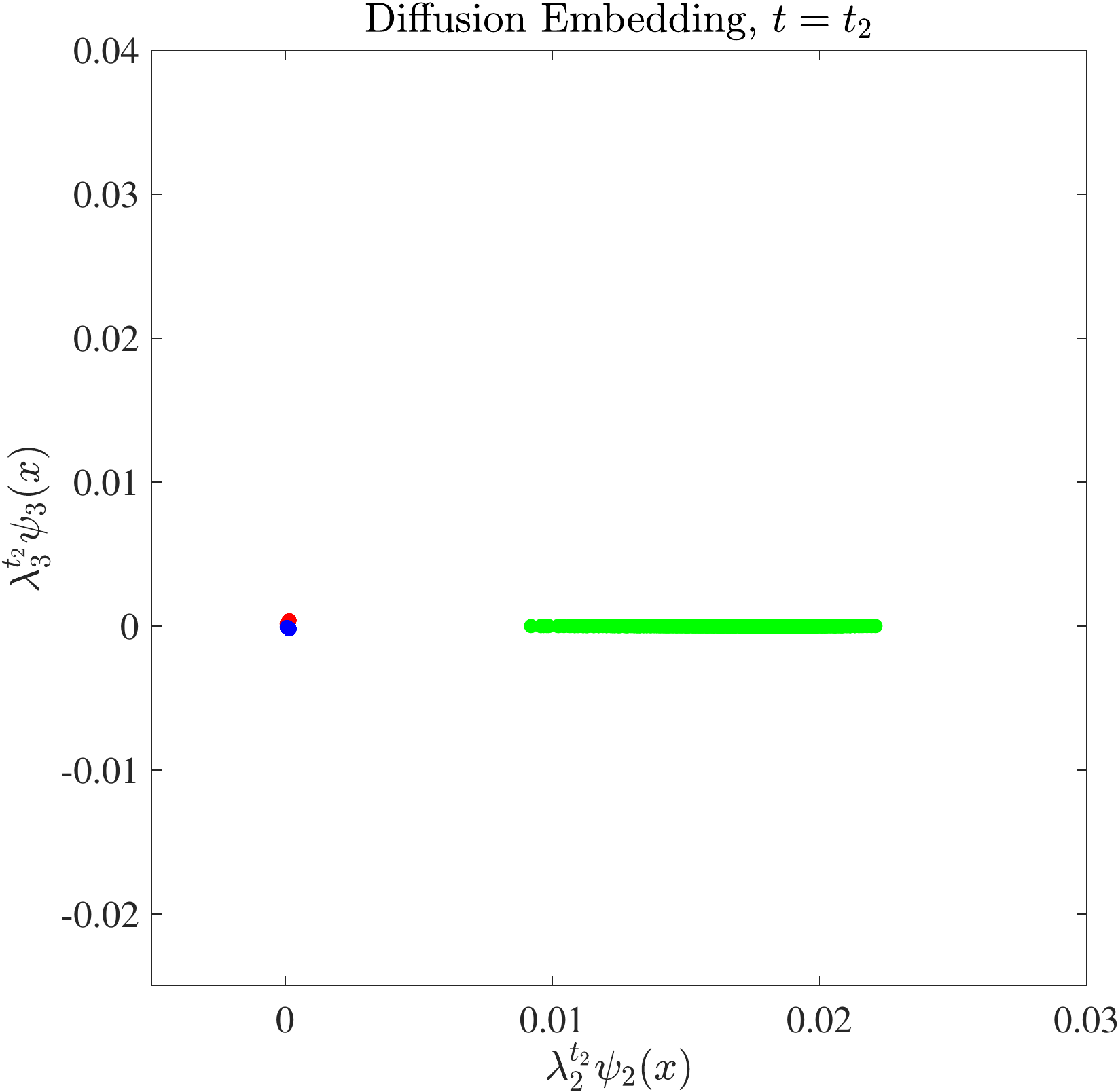}
    \end{subfigure}%
  \label{fig:gaussiandiff}
    \caption{Diffusion embedding of a nonlinear dataset. We plot the second and third diffusion map coordinates, which are the first coordinates of $\Psi_t(x)$ that depend on $t$ and the data; $\lambda_{1}=1$ and $\psi_{1}$ is constant by construction. When  $t$ is small ($t=t_1$), the diffusion map sends each ring in the dataset $X$ to a different cluster. Each ring is well-separated in the new data-dependent feature space. When $t$ becomes large ($t = t_2$), the diffusion map sends the inner two rings to a one-point mass. This corresponds to a different scale of separation by diffusion distances. Thus, the diffusion map exhibits multiscale structure as a function of $t$.} 
    \label{fig: multiscale Diffusion Map}
\end{figure}

Diffusion maps and diffusion distances are related as $D_t(x,y) = \|\Psi_t(x)-\Psi_t(y)\|_2$~\cite{coifman2006diffusionmaps}. In particular, diffusion distances can be identified as Euclidean distances in a new data-dependent feature space consisting of the coordinates of the diffusion map. The diffusion map can be identified as a natural extension of the eigenmap $\Phi(x)$ defined in Section \ref{sec: Spectral_Clustering}.  In $\Phi(x)$, each of the first $K$ eigenfunctions is weighted equally in the new feature space~\cite{nadler2006diffusion_maps, ng2002spectralclustering}. Conversely, in $\Psi_t(x)$, the $i$\ths eigenfunction is weighted according to $\lambda_i^t$. Thus, as $t$ increases, the coordinates of $\Psi_t(x_i)$ corresponding to higher-frequency eigenfunctions become vanishingly small. Because lower-frequency eigenfunctions of $\mathbf{P}$ tend to concentrate on highly-connected regions in the data, this fact may facilitate the detection of different scales of structure in the data for different values of the time parameter $t$~\cite{coifman2005PNAS}, as observed in Figure \ref{fig: multiscale Diffusion Map}.

\subsection{Background on Nearly Reducible Markov Chains}
\label{sec: SC}

Suppose that $X$ admits a latent clustering $\{X_k\}_{k=1}^K$.  Write $\mathbf{P}$, possibly after permuting the indices of data points, as
\begin{equation}
   \textbf{P} = \begin{bmatrix} \textbf{P}_{11} & \textbf{P}_{12}& \dots & \textbf{P}_{1K}\\ 
    \textbf{P}_{21} & \mathbf{P}_{22} & \dots & \mathbf{P}_{2K} \\
    \vdots & \vdots & \ddots & \vdots\\
    \mathbf{P}_{K 1}& \mathbf{P}_{K 2} & \dots & \mathbf{P}_{KK}
    \end{bmatrix}, \label{eq:Transition_Matrix_Basic}
\end{equation}
where the block $\mathbf{P}_{kk'}$ reflects the probability of transitioning from points $x\in X_k$ to points $y\in X_{k'}$. Thus, if the mass of $\mathbf{P}$ is centered on its block diagonal, diffusion is unlikely to exit any given cluster in the latent clustering of $X$. Moreover, if these blocks are in some sense irreducible, diffusion will explore a cluster quickly and diffuse within it for a long period of time. The stochastic complement, defined below, provides some formalism for this intuition. 

\begin{definition}
Let $\mathbf{P}$ be an irreducible, aperiodic Markov transition matrix on $X$, partitioned as in (\ref{eq:Transition_Matrix_Basic}). Let $\textbf{P}_k$ be the principal block submatrix generated by deleting the $k$\ths row and column of blocks from (\ref{eq:Transition_Matrix_Basic}). Similarly, define the matrices
$\mathbf{P}_{*k}=~[\mathbf{P}_{1,k}\; \mathbf{P}_{2,k}\dots \mathbf{P}_{k-1,k} \; \mathbf{P}_{k+1, k} \dots \mathbf{P}_{n,k}]^\top$ and $ \mathbf{P}_{k*} = [\mathbf{P}_{k,1} \; \mathbf{P}_{k, 2} \dots \mathbf{P}_{k,k-1} \; \mathbf{P}_{k,k+1}\dots \mathbf{P}_{k,n}]$. 
The \emph{stochastic complement} of the submatrix $\mathbf{P}_{kk}$ of $\mathbf{P}$ is defined to be $\mathbf{S}_{kk}=\mathbf{P}_{kk} + \mathbf{P}_{k*}(\textbf{I}-\textbf{P}_k)^{-1}\mathbf{P}_{*k}$. The \emph{stochastic complement} of $\mathbf{P}$ with respect to the clustering $\{X_k\}_{k=1}^{K}$ is defined to be the completely reducible, row-stochastic, block-diagonal matrix consisting of the stochastic complements of the diagonal blocks of $\mathbf{P}$:
\begin{equation*}
   \textbf{ S }= \begin{bmatrix} \mathbf{S}_{11} & 0 & \dots & 0\\ 
    0 & \mathbf{S}_{22} & \dots & 0 \\
    \vdots & \vdots & \ddots & \vdots\\
    0 & 0 & \dots & \mathbf{S}_{KK}
    \end{bmatrix}. 
\end{equation*}
\end{definition}

The stochastic complement $\mathbf{S}_{kk}$ consists of two terms: $\mathbf{P}_{kk}$ and $\mathbf{P}_{k*}(\mathbf{I}-\mathbf{P}_k)^{-1}\mathbf{P}_{*k}$. The term $\mathbf{P}_{kk}$ captures the probability of directly transitioning between points in $X_k$, while the term $\mathbf{P}_{k*}(\mathbf{I}-\mathbf{P}_k)^{-1}\mathbf{P}_{*k}$ captures the probability of transitioning into $X_k$ indirectly after first moving through the points in other clusters. Indeed,  $(\mathbf{I}-\mathbf{P}_k)^{-1}$ can be expanded as $\sum_{t=0}^\infty \textbf{P}_k^t$. Thus, the stochastic complement of $\mathbf{P}_{kk}$ encodes the probability of transitioning within $X_k$ after a path of arbitrary length from inside or outside of $X_k$. The stochastic complement can be viewed as an approximation of the transition matrix $\mathbf{P}$ that contains information about a latent clustering $\{X_k\}^{K}_{k=1}$ of $X$. The following theorem illustrates when this approximation is successful~\cite{meyer1989SC}. Recall $\|\textbf{A}\|_\infty = \max_{1\leq i \leq n}\sum_{j=1}^n |\textbf{A}_{ij}|$. 

\begin{theorem} \cite{meyer1989SC}
   Let $\mathbf{P}$ be an irreducible, aperiodic Markov transition matrix on $X$, partitioned as in (\ref{eq:Transition_Matrix_Basic}). Let $\mathbf{S}$ be the stochastic complement of $\mathbf{P}$ with respect to the clustering $\{X_k\}_{k=1}^{K}$. Suppose each stochastic complement $\mathbf{S}_{kk}$ is primitive (i.e., non-negative, irreducible, and aperiodic) so that the eigenvalues of $\mathbf{S}$ are  $1=\lambda_1 = \dots = \lambda_K>|\lambda_{K+1}|>\dots >|\lambda_n|\geq 0$.  Suppose that $\textbf{Z}$ diagonalizes $\mathbf{S}$, let $\delta = \|\mathbf{P}-\mathbf{S}\|_\infty$, and let $\kappa = \|\textbf{Z}\|_\infty \|\textbf{Z}^{-1}\|_\infty$. Finally, let $\mathbf{S}^\infty = \lim_{t\rightarrow\infty}\mathbf{S}^t$. Then for any $t\geq0$, $\|\mathbf{P}^t - \mathbf{S}^\infty\|_\infty \leq \delta t + \kappa |\lambda_{K+1}|^t$. Moreover, if for $\epsilon>0$, $t\in  \Big[\frac{\log(2\kappa/\epsilon)}{\log(1/|\lambda_{K+1}|)},\frac{\epsilon}{2\delta}\Big]$,
    then $\|\mathbf{P}^t-\mathbf{S}^\infty\|_\infty<\epsilon$.
    \label{thm:Meyer}
\end{theorem}

We will henceforth refer to the interval referenced in Theorem \ref{thm:Meyer} as $\mathcal{I}_\epsilon=\Big
[\frac{\log(2\kappa/\epsilon)}{\log(1/|\lambda_{K+1}|)},\frac{\epsilon}{2\delta}\Big]$. The interval $\mathcal{I}_\epsilon$ is dependent not only on $\epsilon$, but also on data-driven quantities derived from the transition matrix $\mathbf{P}$ and its stochastic complement $\mathbf{S}$: $\lambda_{K+1}$, $\delta$, and $\kappa$. These parameters will be of importance in Section \ref{sec: MELD 3}, where we develop a theory of multiscale clustering based on graph diffusion. If we assume---as in Theorem \ref{thm:Meyer}---that the stochastic complement $\mathbf{S}$ of $\mathbf{P}$ is block primitive and diagonalizable \cite{murphy2019LUND, meyer1989SC}, then these parameters may be interpreted as follows:

\begin{itemize}
 
     \item $\lambda_{K+1}$: Clearly, $\mathbf{S}$ is primitive if and only if each $\mathbf{S}_{kk}$ is primitive, so $|\lambda_{K+1}| = \max_{1\leq k \leq K}|\lambda_2(\mathbf{S}_{kk})|$. The second eigenvalue of an irreducible, row-stochastic matrix like $\mathbf{S}_{kk}$ is related to the conductance of the subgraph $X_k$ of $X$~\cite{sinclair1989Markov, levin2017markov_mixing}. Indeed, if all clusters are highly connected, $|\lambda_{K+1}|$ will be small \cite{meyer1989SC, sinclair1989Markov, levin2017markov_mixing}. Conversely, $|\lambda_{K+1}|$ will be near 1 if any cluster is only loosely connected.

    \item  $\delta$: Note that $\delta  = \|\mathbf{P}-\mathbf{S}\|_\infty= 2\max_{1\leq k \leq K}\|\mathbf{P}_{k*}\|_\infty$~\cite{meyer1989SC}.  Thus, the parameter $\delta$ can be interpreted as the maximum probability across all points in $X$ of transitioning from one cluster to another in a single time step. If transitions between any pair of clusters are likely, then $\delta$ will be large. Conversely, if transitions between all pairs of clusters are unlikely, then $\delta$ will be small. In this sense, $\delta$ measures the separation between clusters in $X$~\cite{murphy2019LUND}. Since $\delta$ is the maximal probability of transitioning between clusters, it is somewhat pessimistic in datasets in which outliers from one cluster overlap with outliers from another~\cite{murphy2019LUND}. In such datasets, $\delta$ will be large, but the probability of transitioning between points in cluster cores is still small. 
    
    \item $\kappa$:  By definition, $\kappa$ tells us how difficult it is to diagonalize the stochastic complement $\mathbf{S}$ of $\mathbf{P}$. Suppose that the latent clustering of $X$ is the one consisting of $n$ singletons. Clearly, the stochastic complement would be the identity matrix, so at this extreme $\kappa=1$. If $X$ is sampled from a common manifold, then $\kappa= O(1)$ with respect to $n$~\cite{murphy2019LUND}. If each cluster is sampled from a different common manifold, a similar result is expected to hold. However, the parameter $\kappa$ is admittedly not well-studied, and research on it is still ongoing. 
    
\end{itemize}

\subsection{Background on Diffusion Distances in Clustering}
\label{sec: recent_results}

Define the (worst-case) within-cluster and between-cluster diffusion distance at time $t$ with respect to the clustering $\{X_k\}_{k=1}^K$ by $D_t^{\text{in}} = \max_{1\leq k \leq K}\max_{x,y\in X_k}D_t(x,y)$ and  $
D_t^{\text{btw}} = \min_{1\leq k<k' \leq K}\min_{x\in X_k, y\in X_{k'}}D_t(x,y)$ respectively~\cite{murphy2019LUND}.  We desire a clustering of $X$ which will yield $ D_t^{\text{in}}$ small (a coherence condition) and $ D_t^{\text{btw}}$ large (a separation condition). In this section, we review a result bounding diffusion distances within and between clusters in terms of the underlying statistical and geometric properties of $\mathbf{P}$~\cite{murphy2019LUND}. The following piece of machinery will prove useful in this analysis:

\begin{definition}\label{def: gamma} Let $X$, $\mathbf{P}$, and $\mathbf{S}^\infty$ be as in Theorem \ref{thm:Meyer}, let $p_t(x_i,x_j) = (P^t)_{ij}$, and let $s^\infty(x_i,x_j) = (S^\infty)_{ij}$. Define 
$$\gamma(t) = \displaystyle\max_{x\in X} \left( 1- \frac{1}{2}\sum_{u\in X} \left| \frac{|p_t(x,u)  - s^\infty(x,u)|}{\|p_t(x,u)  - s^\infty(x,u)\|_2} - \frac{1}{\sqrt{n}}\right|^2 \right)^{-1}.$$
\end{definition}

For any vector $u\in\R^n$, we can write $\|u\|_2 = \frac{c_u}{\sqrt{n}}\|u\|_1$, where $c_u =\big( 1- \frac{1}{2}\sum_{i=1}^n \big| \frac{|u_i|}{\|u\|_2}- \frac{1}{\sqrt{n}}\big|^2 \big)^{-1}$~\cite{botelho2017exact}. Thus, $\gamma(t)$ can be identified as the maximum $c_u$, where the vectors $u$ are chosen from the rows of $\mathbf{P}^t-\mathbf{S}^\infty$. In this sense, $\gamma(t)$ indicates how much the $\ell^1$- and $\ell^{2}$ norms of the rows of $\mathbf{P}^t-\mathbf{S}^\infty$ differ. Diffusion distances are written using the $\ell^2$-norm, which gives the spectral decomposition. However, diffusion distances are arguably more natural in an $\ell^1$-norm framework: the setting of Theorem \ref{thm:Meyer}~\cite{meyer1989SC, murphy2019LUND, cowen2020diffusion}. The function $\gamma(t)$ bridges this disconnect and enables bounding diffusion distances using results that are written in the $\ell^1$-norm (e.g., Theorem \ref{thm:Meyer})~\cite{meyer1989SC, murphy2019LUND}. 

\begin{theorem}\label{thm:LUND}\cite{murphy2019LUND} Let  $\{X_k\}_{k=1}^K$ be a partition of $X = \{x_i\}_{i=1}^n$. Let  $\mathbf{P}$, $\delta$, $\kappa$, and $\lambda_{K+1}$ be as in Theorem \ref{thm:Meyer}, and define $s^\infty(x_i,x_j) = (S^\infty)_{ij}$. Then for any $t\geq 0$, 
\begin{equation*}
    D_t^{\text{\emph{in}}}\leq \frac{2\gamma(t)}{\sqrt{n}} \Big(\delta t + \kappa |\lambda_{K+1}|^t\Big), \; \;\;\;\;
    D_t^{\text{\emph{btw}}} \geq 2\min_{w\in X}\|s^\infty(w, . )\|_{\ell^2(1/\pi)} - \frac{2\gamma(t)}{\sqrt{n}}\Big(\delta t + \kappa |\lambda_{K+1}|^t\Big).
\end{equation*}
Moreover, if, for $\epsilon>0$, $t\in\mathcal{I}_\epsilon,$ then
\begin{equation*}
    D_t^{\text{\emph{in}}}\leq \frac{2\gamma(t)}{\sqrt{n}}\epsilon, \; \;\;\;
    D_t^{\text{\emph{btw}}} \geq 2\min_{w\in X}\|s^\infty(w, . )\|_{\ell^2(1/\pi)} - \frac{2\gamma(t)}{\sqrt{n}}\epsilon.
\end{equation*}
\end{theorem}

For $\epsilon>0$,  if $t\in  \mathcal{I}_\epsilon$, all clusters are of equal size $n/K$, and $s^\infty$ is uniform on each $X_k$, then Theorem \ref{thm:LUND} implies that $D_t^{\text{in}}/D_t^{\text{btw}} = O(\epsilon)$~\cite{murphy2019LUND}. For $\epsilon$ small, this indicates that the maximum within-cluster diffusion distance at time $t$ will be much less than the minimum between-cluster diffusion distance at time $t$. Notably, there is tension between the assumption that $t\in\mathcal{I}_\epsilon$ and the conclusion that diffusion distances at time $t$ induce a good separation among the clusters of the clustering $\{X_k\}_{k=1}^{K}$~\cite{murphy2019LUND}. If $\epsilon$ is large, the assumption that $t\in \mathcal{\mathcal{I}_\epsilon}$ may be lax, but the conclusion of Theorem \ref{thm:LUND} may be weak or even trivial. Conversely, if $\epsilon$ is small, Theorem \ref{thm:LUND} implies that $D_t^{\text{in}}/D_t^{\text{btw}}$ will also be small, but this strong result comes at the expense of narrowing the interval of time during which it can be attained. For fixed $\kappa$, $\delta$, and $\lambda_{K+1}$, $\mathcal{\mathcal{I}_\epsilon}$ shrinks to the empty set as $\epsilon\rightarrow 0^{+}$~\cite{murphy2019LUND}. In the idealized case in which there are no between-cluster transitions (so that $\delta = 0$) and each cluster is a point mass (so that $|\lambda_{K+1}|=0$), then $\mathcal{I}_\epsilon=[0,\infty)$~\cite{murphy2019LUND}.

\begin{algorithm}[t]
\SetAlgoLined
\caption{Learning by Unsupervised Nonlinear Diffusion (LUND)}
\KwIn{ $X$ (dataset),  $\sigma$~(diffusion scale parameter), $\sigma_0$ (KDE bandwidth), $t$ (diffusion time parameter)}
\KwOut{$\mathcal{C}$ (clustering), $K$ (no. clusters)}
Construct transition matrix $\mathbf{P}$ with a Gaussian kernel and diffusion scale parameter $\sigma$\;
Compute the KDE $p(x)$ with KDE bandwidth $\sigma_0$ for each $x\in X$\;
Compute $\rho_t(x)$ according to Definition \ref{def:rho}  for each $x\in X$\;
Store $\mathcal{D}_t(x) = p(x) \rho_t(x)$ for each $x\in X$\;
Sort $X$ in non-increasing order according to $\mathcal{D}_t(x)$. Denote this sorting $\{x_{m_k}\}_{k=1}^n$\;
Solve $K = \text{argmin}\Big\{\frac{\mathcal{D}_t(x_{m_k})}{\mathcal{D}_t(x_{m_{k+1}})}\Big\}_{k=1}^{n-1}$ and label  each cluster mode, $x_{m_k}$ ($k=1,\dots, K)$ by $\mathcal{C}(x_{m_k})=k$\;
Sort $X$ in non-increasing order according to $p(x)$. Denote this sorting $\{x_{\ell_k}\}_{k=1}^n$\;
\For{$k=1:n$}{
    \If{$\mathcal{C}(x_{\ell_k})=0$}{
        $x^* = \text{argmin}_{y\in X}\{D_t(x_{\ell_k}, y)\mid p(y)\geq p(x_{\ell_k}), \text{ $y$ is labeled}\}$\;
        $\mathcal{C}(x_{\ell_k}) = \mathcal{C}(x^*)$\;
        }
    }
\label{alg: LUND}
\end{algorithm}

\subsection{The LUND Clustering Algorithm} \label{sec: LUND 2}

The LUND algorithm (Algorithm \ref{alg: LUND}) was introduced to leverage diffusion distances to cluster data~\cite{murphy2019LUND}.  This clustering algorithm locates high-density points that are far in diffusion distance from other high-density points and labels them as cluster modes. Non-modal points are then paired with a labeled point iteratively.  More precisely, the LUND algorithm captures density using a kernel density estimate (KDE) $p(x) = \frac{1}{Z}\sum_{y\in NN(x,N)}\exp\big(-\|x-y\|_2^2/\sigma_0^2\big)$, where $\sigma_0$ is a KDE bandwidth, $NN(x,N)$ is the set of $N$ $\ell^2$-nearest neighbors of $x$, and $Z$ is a normalization constant such that $p(x)$ sums to one~\cite{murphy2019LUND}. To capture diffusion geometry, we introduce a different function: 

\begin{definition}
Let $X$ and $\mathbf{P}$ be as in Theorem \ref{thm:Meyer}, and let $p(x)$ be a KDE of $X$. Define 
\begin{equation*}
    \rho_t(x) = \begin{cases} \min_{y\in X} \{ D_t(x,y) \mid p(y)\geq p(x)\} & x \neq \text{\emph{argmax}}_{y\in X} p(y),\\
    \max_{y\in X} D_t(x,y) & x = \text{\emph{argmax}}_{y\in X} p(y)
    \end{cases}.
\end{equation*}
\label{def:rho}
\end{definition}

Thus, $\rho_t(x)$ assigns $x$ the diffusion distance between $x$ and its $D_t$-nearest neighbor of higher density. The LUND algorithm then analyzes $\mathcal{D}_t(x) = p(x) \rho_t(x)$. The maximizers of $\mathcal{D}_t(x)$ tend to be high in empirical density and far in diffusion distance from other high-density points, making them suitable choices as cluster modes. The function $\mathcal{D}_t(x)$ can also be used to estimate the number of latent clusters in $X$~\cite{murphy2019LUND}.  While $K$-means and SC algorithms can estimate the number of latent clusters $K$ via the scree plot~\cite{cattell1966scree} and eigengap \cite{little2015multiscale}, respectively, these estimates of $K$ have been shown to fail on data classes in which the LUND estimate succeeds (e.g., datasets with nonlinear structure for the scree plot and datasets with multimodal bottleneck structure for the eigengap~\cite{murphy2019LUND, Little2020_Path}). It has been shown that, under plausible assumptions on cluster structure and density, the estimate provided by the LUND algorithm on the number of clusters in the dataset is accurate, even for datasets with these problematic structures~\cite{murphy2019LUND}. We review theoretical guarantees on the performance of the LUND algorithm in Section \ref{sec: LUND guarantees}.

The LUND algorithm relies on the diffusion time parameter $t$ when calculating the diffusion distance between points. As discussed in Section \ref{sec: Diffusion Maps 2}, this parameter tends to affect the scale of a clustering separable by diffusion distances~\cite{coifman2006diffusionmaps, nadler2006diffusion_maps_ACHA}.  Thus, as $t$ varies, the clustering that the LUND algorithm estimates will change.  To improve the LUND algorithm, it is necessary to understand how its cluster assignments change as a function of $t$.  A better theoretical understanding of how the diffusion process changes may enable the elimination of the dependence on $t$ as well as a deeper understanding of which time scale yields the most representative clustering of $X$. 

\section{A Multiscale Environment for Learning by Diffusion} \label{sec: MELD 3}

Theorem \ref{thm:LUND} states that diffusion distances induce strong separation on a latent clustering during an interval in the diffusion process. However, it is limited in that it only considers a fixed scale and a single latent clustering. Many datasets exhibit multiscale structure with many partitions that could be considered ``correct'' and useful~\cite{coifman2005PNAS, ahn2010link, delmotte2011protein, song2015multiscale}. In this section, we will generalize Theorem \ref{thm:LUND} by allowing multiple latent clusterings, varying in scale, to exist within the same dataset~\cite{murphy2019LUND}. We will then introduce the MELD data model, which parameterizes the clusterings of $X$ by $t$.

Suppose there are $M$ latent clusterings of $X$, denoted $\{X_k^{(\ell)}\}_{k=1}^{K_\ell}$ for $\ell\in\{1,\dots,M\}$. We will not require that these clusterings are hierarchical but consider that special case in Section \ref{sec: Hierarchy 3}. For $1\leq \ell \leq M$, define the submatrices $\mathbf{P}_{kk'}^{(\ell)}$ of the transition matrix $\mathbf{P}$, possibly after permuting the indices of data points, implicitly by
\begin{equation}
   \mathbf{P}^{(\ell)} = \begin{bmatrix} \mathbf{P}_{11}^{(\ell)} & \mathbf{P}_{12}^{(\ell)} & \dots & \mathbf{P}_{1K_\ell}^{(\ell)}\\ 
    \mathbf{P}_{21}^{(\ell)} & \mathbf{P}_{22}^{(\ell)} & \dots & \mathbf{P}_{2K_\ell}^{(\ell)} \\
    \vdots & \vdots & \ddots & \vdots\\
    \mathbf{P}_{K_\ell 1}^{(\ell)} & \mathbf{P}_{K_\ell 2}^{(\ell)} & \dots & \mathbf{P}_{K_\ell K_\ell}^{(\ell)}
    \end{bmatrix}, \label{eq:TransitionMatrix_Multiscale}
\end{equation}
where the block $\mathbf{P}_{kk'}^{(\ell)}$ reflects the probability of transitioning from points $x\in X_k^{(\ell)}$ to points $y\in X_{k'}^{(\ell)}$. In particular, the block matrix $\mathbf{P}^{(\ell)}_{kk}$ reflects the probability of remaining in the cluster $X_k^{(\ell)}$ in the $\ell$\ths latent clustering of $X$, while $\mathbf{P}_{kk'}^{(\ell)}$ reflects the probability of transitioning from the cluster $X_k^{(\ell)}$ to $X_{k'}^{(\ell)}$ in the $\ell$\ths latent clustering of $X$. 

The stochastic complement of $\mathbf{P}$ depends on the clustering assumed a priori. Thus, for each of the latent clusterings of $X$, a different stochastic complement can be extracted. We will refer to the stochastic complement of the submatrix $\mathbf{P}_{kk}^{(\ell)}$ as $\mathbf{S}_{kk}^{(\ell)}$ and the stochastic complement of $\mathbf{P}$ with respect to the $\ell$\ths clustering of $X$ as $\mathbf{S}^{(\ell)}$.  Let $\mathbf{S}_\infty^{(\ell)} =~\lim_{t\rightarrow\infty}[\mathbf{S}^{(\ell)}]^t$.  Similar to the case in which there was only one latent clustering, the stochastic complement $\mathbf{S}^{(\ell)}_{kk}$ may be interpreted as capturing the probability of transitioning into the cluster $X_k^{(\ell)}$, either directly from inside of $X_k^{(\ell)}$ or indirectly after a path of arbitrary length starting outside of $X_k^{(\ell)}$~\cite{meyer1989SC,murphy2019LUND}.  As before, we require $\mathbf{S}^{(\ell)}_{kk}$ to be primitive and diagonalizable  for each $k\in \{1,\dots, K_\ell\}$ and $\ell\in \{1,\dots,M\}$. Denote the invertible $n\times n$ matrix that diagonalizes $\mathbf{S}^{(\ell)}$ as $\mathbf{Z}^{(\ell)}$~\cite{murphy2019LUND, meyer1989SC}. 

The interval $\mathcal{I}_\epsilon$ is regulated by three constants---$\lambda_{K+1}$, $\delta$, and $\kappa$---each derived from the stochastic complement of $\mathbf{P}$ corresponding to a clustering. Therefore, the interval $\mathcal{I}_\epsilon$ will change as a function of the scale of clustering.

\begin{definition}
\label{def: interval}
 Let $\mathbf{P}$ be an aperiodic, irreducible Markov transition matrix on $X$, partitioned as in (\ref{eq:TransitionMatrix_Multiscale}). Let $\mathbf{S}^{(\ell)}$ be the stochastic complement of $\mathbf{P}$ with respect to the $\ell$\ths clustering of $X$. Define $\lambda_{K_\ell + 1}^{(\ell)} = \lambda_{K_\ell+1}[\mathbf{S}^{(\ell)}]$, $\delta^{(\ell)} = \|\mathbf{P}-\mathbf{S}^{(\ell)}\|_\infty$ and $\kappa^{(\ell)}~=~\|\mathbf{Z}^{(\ell)}\|_\infty \|[\mathbf{Z}^{(\ell)}]^{-1}\|_\infty$. For $\epsilon>0$, define the interval $\mathcal{I}^{(\ell)}_\epsilon = \bigg[\frac{\log(2\kappa^{(\ell)}/\epsilon)}{\log(1/|\lambda_{K_\ell+1}^{(\ell)}|)},\frac{\epsilon}{2\delta^{(\ell)}}\bigg]$. 
\end{definition}

We will refer to the maximum within-cluster and minimum between-cluster diffusion distance at time $t$ for the clustering  $\{X_k^{(\ell)}\}_{k=1}^{K_\ell}$ as  $D_t^{\text{{in}}}(\ell)$ and $D_t^{\text{{btw}}}(\ell)$ respectively. We argued in Section \ref{sec: Diffusion Maps 2} that the dependence of diffusion distances on the diffusion time parameter affects what scales of structure can be uncovered by diffusion distances~\cite{coifman2006diffusionmaps, nadler2006diffusion_maps_ACHA}.  Thus, for any fixed $t$, the ratio $D_t^{\text{in}}(\ell)/D_t^{\text{btw}}(\ell)$ may be large for some $\ell$ and small for others. In the following definition, the notion of strong separation by diffusion distances at a given time scale is formalized.

\begin{definition}\label{def:epsilon-separation}
Let $\epsilon>0$. We will say that $\{X_k^{(\ell)}\}_{k=1}^{K_\ell}$ is \emph{$\epsilon$-separable by diffusion distances at time $t$} if $\frac{D_t^{\text{in}}(\ell)}{D_t^{\text{btw}}(\ell)}\leq\frac{\epsilon}{1/\sqrt{n}-\epsilon}$.
\end{definition}

The criterion for $\epsilon$-separation by diffusion distances is related to the notion of a perfect clustering. A clustering $\{X_k\}_{k=1}^K$ is said to be \emph{perfect} under the metric $m(\cdot,\cdot)$ if there is an $r>0$ for which the maximum within-cluster distance 
is at most $r$ and the minimum between-cluster distance is at least $4r$, where distance is measured using $m$~\cite{mcsherry2001spectral,vu2018simple}. If $\epsilon \ll 1/\sqrt{n}$, the clustering $\{X_k^{(\ell)}\}_{k=1}^{K_\ell}$ is $\epsilon$-separable by diffusion distances at time $t$ if and only if $D_t^{\text{{in}}}(\ell)/D_t^{\text{{btw}}}(\ell)= O(\epsilon)$.  More precisely, if $0<\epsilon\leq \frac{1}{5\sqrt{n}}$, a clustering that is $\epsilon$-separable by diffusion distances at time $t$ is also perfect under the metric $D_t$. In datasets with a perfect partition, cluster structure may be detected with $K$-means using the metric $D_{t}$ (if $r$ is unknown) or by thresholding a minimum spanning tree (if $r$ is known)~\cite{mcsherry2001spectral,  vu2018simple}.  

Let $\gamma^{(\ell)}(t)$ be the multiscale extension of $\gamma(t)$, where $s^\infty(x_i,x_j)$ is replaced by $s_\infty^{(\ell)}(x_i,x_j) = (S^{(\ell)}_\infty)_{ij}$ in Definition \ref{def: gamma}.  This function measures how much the $\ell^1$-norm of rows in $\mathbf{P}^t-\mathbf{S}^{(\ell)}_\infty$ differs from the $\ell^2$-norm of rows in $\mathbf{P}^t-\mathbf{S}^{(\ell)}_\infty$. As was the case for $\gamma(t)$, $1\leq \gamma^{(\ell)}(t)\leq \sqrt{n}$ for any $t$ and $\ell\in\{1,\dots,M\}$. Using the established notation, we are able to provide the following corollary, which serves as a multiscale extension of Theorem \ref{thm:LUND}.

\begin{corollary}
 Let $\mathbf{P}$ be an aperiodic, irreducible Markov transition matrix on a dataset $X$, partitioned as in (\ref{eq:TransitionMatrix_Multiscale}). Let $\ell\in~\{1,\dots, M\}$ be a fixed clustering scale, and let $\mathbf{S}^{(\ell)}$ be the stochastic complement of $\mathbf{P}$ with respect to the clustering $\{X_k^{(\ell)}\}_{k=1}^{K_\ell}$. Let $\delta^{(\ell)}$, $\kappa^{(\ell)}$, and $\lambda_{K_\ell+1}^{(\ell)}$ be the geometric constants introduced in Definition \ref{def: interval}, and let  $s_\infty^{(\ell)}(x_i,x_j) = (S_\infty^{(\ell)})_{ij}$. 
 \begin{enumerate}[(a)]
    \item For any $t\geq 0$, 
\begin{align*}
  D_t^{\text{\emph{in}}}(\ell) \leq \frac{2\gamma^{(\ell)}(t)}{\sqrt{n}}\big(\delta^{(\ell)}t + \kappa^{(\ell)}|\lambda_{K_\ell+1}^{(\ell)}|^t\big);\;\;\;  D_t^{\text{\emph{btw}}}(\ell)  \geq  \min_{w\in X}\|s_\infty^{(\ell)}(w,:)\|_{\ell^2(1/\pi)} - \frac{2\gamma^{(\ell)}(t)}{\sqrt{n}}\big(\delta^{(\ell)}t + \kappa^{(\ell)}|\lambda_{K_\ell+1}^{(\ell)}|^t\big).
\end{align*}
Moreover, if, for $\epsilon>0$, $t\in \mathcal{I}_\epsilon^{(\ell)}$, then
\begin{equation*}
    D_t^{\text{\emph{in}}}(\ell) \leq \frac{2\gamma^{(\ell)}(t)}{\sqrt{n}}\epsilon; \; \;\;
     D_t^{\text{\emph{btw}}}(\ell)  \geq  2\min_{w\in X}\|s_\infty^{(\ell)}(w,:)\|_{\ell^2(1/\pi)}- \frac{2\gamma^{(\ell)}(t)}{\sqrt{n}}\epsilon.
\end{equation*}
\item If $\epsilon<1/\sqrt{n}$, then the clustering $\{X_k^{(\ell)}\}_{k=1}^{K_\ell}$ is $\epsilon$-separable by diffusion distances at times $t\in\mathcal{I}_\epsilon^{(\ell)}$.  
\end{enumerate}
\label{cor:diff_dist_bound}
\end{corollary}
\begin{proof}

Theorem \ref{thm:LUND} gives (a) immediately~\cite{murphy2019LUND}. To obtain (b), note first that $\min_{w\in X}\|s_\infty^{(\ell)}(w,:)\|_{\ell^2(1/\pi)}$ can be bounded from below:
$\frac{1}{\sqrt{n}}= \frac{1}{\sqrt{n}}\min_{w\in X}\|s^{(\ell)}_\infty(w,:)\|_1 \leq \min_{w\in X}\|s^{(\ell)}_\infty(w,:)\|_2\leq \min_{w\in X}\|s_\infty^{(\ell)}(w,:)\|_{\ell^2(1/\pi)}$. The assumption $\epsilon<\frac{1}{\sqrt{n}}$ implies $\epsilon<~\frac{1}{\sqrt{n}}\leq~\frac{1}{\gamma^{(\ell)}(t)}\leq \frac{\sqrt{n}}{\gamma^{(\ell)}(t)}\min_{w\in X}\|s_\infty^{(\ell)}(w,:)\|_{\ell^2(1/\pi)}$. Rearranging yields $ 2\min_{w\in X}\|s_\infty^{(\ell)}(w,:)\|_{\ell^2(1/\pi)}> \frac{2\gamma^{(\ell)}(t)}{\sqrt{n}}\epsilon$, so the lower bound on $D_t^{\text{btw}}(\ell)$ given in (a) is positive for $t\in \mathcal{I}_\epsilon^{(\ell)}$. Thus,
\begin{align*}
    \frac{D_t^{\text{in}}(\ell)}{D_t^{\text{btw}}(\ell)}
    &\leq \frac{2\gamma^{(\ell)}(t)\epsilon/\sqrt{n}}{2\min_{w\in X}\|s_\infty^{(\ell)}(w,:)\|_{\ell^2(1/\pi)}-2\gamma^{(\ell)}(t)\epsilon/\sqrt{n}}\\
    &=\frac{\gamma^{(\ell)}(t)\epsilon}{\sqrt{n}\min_{w\in X}\|s_\infty^{(\ell)}(w,:)\|_{\ell^2(1/\pi)}-\gamma^{(\ell)}(t)\epsilon}\\
    &\leq \frac{\epsilon}{\min_{w\in X}\|s_\infty^{(\ell)}(w,:)\|_{\ell^2(1/\pi)}-\epsilon}\\
    &\leq \frac{\epsilon}{1/\sqrt{n}-\epsilon},
\end{align*}
where $\gamma^{(\ell)}(t) \leq \sqrt{n}$ was used to obtain the second to last inequality and $\min_{w\in X}\|s_\infty^{(\ell)}(w,:~)\|_{\ell^2(1/\pi)}\geq 1/\sqrt{n}$ was used to obtain the last inequality. 
\end{proof}

The proof of Corollary \ref{cor:diff_dist_bound} suggests that the notion of $\epsilon$-separation by diffusion distances is somewhat pessimistic, as it relies on worst-case assumptions on the behavior of the parameters $\gamma^{(\ell)}(t)$ and $\min_{w\in X}\|s_\infty^{(\ell)}(w,:)\|_{\ell^2(1/\pi)}$. In practice, if $t\in \mathcal{I}_\epsilon^{(\ell)}$, the rows of $\mathbf{P}^t-\textbf{S}_\infty^{(\ell)}$ tend to be nearly uniform, so $\gamma^{(\ell)}(t) = O(1)$ with respect to $n$~\cite{murphy2019LUND}. Similarly, $\min_{w\in X}\|s_\infty^{(\ell)}(w,:)\|_{\ell^2(1/\pi)} \geq 1/\sqrt{n}$ is a worst-case lower bound. If the rows of $\mathbf{P}^t-\textbf{S}_\infty^{(\ell)}$ are completely uniform, then $\gamma^{(\ell)}(t)=1$~\cite{murphy2019LUND}. Assuming this is the case for each $t\in\mathcal{I}_\epsilon^{(\ell)}$, where $\epsilon\in (0,1)$, the lower bound of $D_t^{\text{btw}}(\ell)$ in (a) of Corollary \ref{cor:diff_dist_bound} is positive. Hence,
$$\frac{D_t^{\text{in}}(\ell)}{D_t^{\text{btw}}(\ell)}\leq \frac{\gamma^{(\ell)}(t)\epsilon}{\sqrt{n}\min_{w\in X}\|s_\infty^{(\ell)}(w,:)\|_{\ell^2(1/\pi)}-\gamma^{(\ell)}(t)\epsilon}= \frac{\epsilon}{\sqrt{n}\min_{w\in X}\|s_\infty^{(\ell)}(w,:)\|_{\ell^2(1/\pi)}-\epsilon}\leq \frac{\epsilon}{1-\epsilon},$$
where we have used $\min_{w\in X}\|s_\infty^{(\ell)}(w,:)\|_{\ell^2(1/\pi)}\geq 1/\sqrt{n}$ to obtain the last inequality. This is clearly a tighter bound than the one required for $\epsilon$-separation and is notably independent of $n$. For $\epsilon\ll 1$, this new inequality implies that diffusion distances at times $t\in\mathcal{I}_\epsilon^{(\ell)}$ will induce excellent separation on the clusters in the $\ell$\ths clustering.

By Corollary \ref{cor:diff_dist_bound}, there are $M$ intervals in the diffusion process, during each of which a different clustering is $\epsilon$-separable by diffusion distances. If two distinct intervals $\mathcal{I}_\epsilon^{(\ell)}$ and $\mathcal{I}_\epsilon^{(\ell')}$ ever overlapped for some fixed $\epsilon\in\Big(0,\frac{1}{\sqrt{n}}\Big)$, time steps $t$ would exist during which multiple clusterings are $\epsilon$-separable by diffusion distances at the same time. We therefore make the simplifying assumption that, if $\ell\neq \ell'$ and $\epsilon$ are fixed, $\mathcal{I}^{(\ell)}_\epsilon \bigcap \mathcal{I}^{(\ell')}_\epsilon =\emptyset$ so that the intervals $\mathcal{I}_\epsilon^{(\ell)}$ do not intersect. Thus, at times $t\in \mathcal{I}_\epsilon^{(\ell)}$, $\{X_k^{(\ell)}\}_{k=1}^{K_\ell}$ is the unique clustering that is $\epsilon$-separable by diffusion distances as a result of Corollary \ref{cor:diff_dist_bound}. This is the basis of the MELD data model, wherein the unique latent partitions of $X$ are parameterized by the diffusion time parameter. 

\begin{definition}
\label{def: MELD}
Let $X$ be a dataset with $M$ distinct latent clusterings $\{X_k^{(\ell)}\}_{k=1}^{K_\ell}$ for $1\leq \ell\leq M$. Fix $\epsilon\in\Big(0,\frac{1}{\sqrt{n}}\Big)$, and assume that the intervals $\mathcal{I}^{(\ell)}_\epsilon$ are nonintersecting. For each $t\geq 0$, if $t\in \mathcal{I}_\epsilon^{(\ell)}$ for some $\ell\in\{1,\dots,M\}$, then we define $\mathcal{C}_t = \{X_k^{(\ell)}\}_{k=1}^{K_\ell}$ to be the clustering that is $\epsilon$-separable by diffusion distances at time $t$ as a result of Corollary \ref{cor:diff_dist_bound}. Define the \emph{Multiscale Environment for Learning by Diffusion (MELD)} data model for this choice of $\epsilon$ to be $\MELDemph =~\big\{ \mathcal{C}_t\;\big| \;t\in \mathcal{I}_{\epsilon}^{(\ell)} \text{ \emph{for some} $\ell\in\{1,\dots,M\}$}\big\}.$ 
\end{definition}

The MELD data model is similar in spirit to the \emph{diffusion geometry model} (\emph{DGM}), which also assumes the existence of $M$ latent clusterings of $X$: $\{\mathcal{C}^{(\ell)}\}_{\ell=1}^M$~\cite{cowen2020diffusion}. In the DGM, the stochastic complement of $\mathbf{P}$ with respect to each of the $M$ clusterings is extracted, along with the corresponding geometric constants: $\{(\lambda_{K_\ell+1}^{(\ell)},\delta^{(\ell)},\kappa^{(\ell)})\}_{\ell=1}^M$. The DGM was used to aggregate information about the many latent scales of structure in the dataset into a single distance metric~\cite{cowen2020diffusion}. On the other hand, the MELD data model provides a fine-scale view of how latent cluster structure changes as a function of the diffusion process. Corollary \ref{cor:diff_dist_bound} guarantees that a clustering $\mathcal{C}_t\in\MELD$ is $\epsilon$-separable by diffusion distances at time $t$. In this sense, a clustering $\mathcal{C}_t$ in the MELD data model can be interpreted as the latent clustering of $X$ at time $t$ in the diffusion process. 

Notably, the union of the intervals $\mathcal{I}_\epsilon^{(\ell)}$ may not be $[0,\infty)$. If there are time steps $t$ not contained in any interval $\mathcal{I}_\epsilon^{(\ell)}$, then diffusion distances at those time steps may not induce an $\epsilon$-separation on any clustering whatsoever. The time steps between any two intervals $\mathcal{I}_\epsilon^{(\ell)}$  and $\mathcal{I}_\epsilon^{(\ell')}$ can be thought of as transition regions between two latent clusterings. Transition regions are, in datasets with very well-defined multiscale cluster structure, short intervals during which $\mathbf{P}$ is rapidly mixing and transitioning between states. During these transition regions, there is not necessarily a ``true'' latent clustering of $X$. Therefore, the MELD data model naturally only captures the time steps at which diffusion distances yield strong separation of clusters in a clustering of $X$. We will refer to the time steps during which a latent clustering of $X$ is $\epsilon$-separable by diffusion distances as $A_\epsilon=\bigcup_{\ell=1}^M\mathcal{I}_\epsilon^{(\ell)}$.

\subsection{Stability in the MELD Data Model}
\label{sec: stability}

A cluster can be viewed as a region of the graph on which diffusion is unlikely to exit~\cite{meila2001neurIPS}. The duration the random walk is ``trapped'' on the cluster, aggregated across clusters, can be interpreted as a clustering's stability.
\begin{definition}\label{def:stability}
 Fix $\epsilon\in\Big(0,\frac{1}{\sqrt{n}}\Big)$ and let $\MELDemph$ be as in Definition \ref{def: MELD}. Let $\mathcal{C}_t$ and $\mathcal{C}_s$ be clusterings in $\MELDemph$ with $t\in \mathcal{I}_\epsilon^{(\ell)}$ and $s\in \mathcal{I}_{\epsilon}^{(\ell')}$. We say that $\mathcal{C}_t$ is more \emph{$\epsilon$-stable} than $\mathcal{C}_s$ if $$\log\left[ \frac{\epsilon}{2\delta^{(\ell)}}- \frac{\log(2\kappa^{(\ell)}/\epsilon)}{\log(1/|\lambda_{K_\ell+1}^{(\ell)}|)}\right] \geq \log\left[ \frac{\epsilon}{2\delta^{(\ell')}}- \frac{\log(2\kappa^{(\ell')}/\epsilon)}{\log(1/|\lambda_{K_{\ell'}+1}^{(\ell')}|)}\right].$$
\end{definition}

Thus, a clustering $\mathcal{C}_t$ is considered more $\epsilon$-stable than $\mathcal{C}_s$ if the interval of time during which $\mathcal{C}_t$ is $\epsilon$-separable by diffusion distances is longer on a logarithmic scale than the interval of time during which $\mathcal{C}_s$ is $\epsilon$-separable by diffusion distances. We examine stability on a logarithmic scale because of the exponential dependence of diffusion distances on the spectrum of $\mathbf{P}$. Transitions between clusterings of $X$ typically occur after a component of $\Psi_t(x)$ is sent to zero. Because each component of $\Psi_t(x)$ converges exponentially to zero, the length of intervals $\mathcal{I}_\epsilon^{(\ell)}$ later in the diffusion process tends to be exponentially longer than that of intervals early in the diffusion process. Thus, if we choose to examine stability on a linear scale (rather than logarithmic), a clustering that is $\epsilon$-separable by diffusion distances later in the diffusion process will tend to be more $\epsilon$-stable than a clustering that is $\epsilon$-separable by diffusion distances early in the diffusion process. Taking logarithms allows for a more fair comparison between clusterings that are $\epsilon$-separable at different stages in the diffusion process. The connection between the stability and geometry of a clustering is explored in Proposition \ref{prop:stability}.  

\begin{prop}
Fix $\epsilon\in\Big(0,\frac{1}{\sqrt{n}}\Big)$, and
let $\mathcal{C}_t,\mathcal{C}_s\in \MELDemph$ for $t\in \mathcal{I}_\epsilon^{(\ell)}$ and $s\in \mathcal{I}_\epsilon^{(\ell')}$. If $|\lambda_{K_\ell+1}^{(\ell)}|\leq |\lambda_{K_{\ell'}+1}^{(\ell')}|$, $\delta^{(\ell)}\leq \delta^{(\ell')}$, and
    $\kappa^{(\ell)} = \kappa^{(\ell')}$, then  $\mathcal{C}_t$ is more $\epsilon$-stable than $\mathcal{C}_s$. 
    \label{prop:stability}
\end{prop}
\begin{proof}
If $\delta^{(\ell)}\leq\delta^{(\ell')}$, then $\frac{\epsilon}{2\delta^{(\ell)}}\geq \frac{\epsilon}{2\delta^{(\ell')}}$. Similarly, if $|\lambda^{(\ell')}_{K_{\ell'}+1}|\geq |\lambda^{(\ell)}_{K_{\ell}+1}|$, then $\frac{1}{\log(1/|\lambda_{K_{\ell'}+1}^{(\ell')}|)}\geq \frac{1}{\log(1/|\lambda_{K_\ell+1}^{(\ell)}|)}$. By our assumption that $\kappa^{(\ell)}=\kappa^{(\ell')}$, this clearly implies that $\frac{\log(2\kappa^{(\ell')}/\epsilon)}{\log(1/|\lambda_{K_{\ell'}+1}^{(\ell')}|)}\geq \frac{\log(2\kappa^{(\ell)}/\epsilon)}{\log(1/|\lambda_{K_\ell+1}^{(\ell)}|)}$. In particular, $\frac{\epsilon}{2\delta^{(\ell)}}-\frac{\log(2\kappa^{(\ell)}/\epsilon)}{\log(1/|\lambda_{K_\ell+1}^{(\ell)}|)}\geq~\frac{\epsilon}{2\delta^{(\ell')}}-~\frac{\log(2\kappa^{(\ell')}/\epsilon)}{\log(1/|\lambda_{K_{\ell'}+1}^{(\ell')}|)}$. Taking logarithms on both sides yields the result.  
\end{proof}

Proposition \ref{prop:stability} implies that if the clusters in a clustering $\mathcal{C}_t\in\MELD$ with $t\in \mathcal{I}_\epsilon^{(\ell)}$ are better separated (so that $\delta^{(\ell)}$ is small) and more coherent (so that $|\lambda_{K_\ell+1}^{(\ell)}|$ is small) than the clusters in a different clustering $\mathcal{C}_s\in \MELD$, then $\mathcal{C}_t$ will be more $\epsilon$-stable and appear more frequently than $\mathcal{C}_s$ in the MELD data model. This implies that clusterings with well-separated, coherent clusters are emphasized within the MELD data model. 

\subsection{Applications of the MELD Data Model to Hierarchical Clustering} \label{sec: Hierarchy 3}

In this section, we investigate the relationship between the clusterings in the MELD data model and the diffusion time parameter in the case that the MELD data model exhibits hierarchical structure.

\begin{definition}
Let $\mathcal{C} = \{X_k\}_{k=1}^K$ and $\mathcal{C}' = \{X_k'\}_{k=1}^{K'}$ be clusterings of $X$.  The clustering $\mathcal{C}$ is a \emph{refinement} of the clustering $\mathcal{C}'$ if $K\geq K'$ and if, for every cluster $X_k'\in \mathcal{C}'$, $X_k' = \bigcup_{j=1}^m X_{k_j}$ for some subsequence $\{k_j\}_{j=1}^m$ of $ \{1,2\dots, K\}$.  The family of clusterings $\mathscr{C} = \{\mathcal{C}_\alpha\}_{\alpha\in A}$ \emph{exhibits hierarchical structure} if, for each pair $\mathcal{C}_\alpha$ and $\mathcal{C}_\beta$ in $\mathscr{C}$, either  $\mathcal{C}_\alpha$  is a refinement of  $\mathcal{C}_\beta$  or $\mathcal{C}_\beta$  is a refinement of  $\mathcal{C}_\alpha$. 
\end{definition}

Thus, the MELD data model exhibits hierarchical structure if, for each pair of coarse and fine clusterings in the model, any cluster in the coarse clustering can be expressed as the union of clusters from the fine clustering. In general, the MELD data model does not assume hierarchical structure because not all multiscale cluster structure is hierarchical. Indeed, one of the advantages of the MELD data model is its ability to capture non-hierarchical multiscale structure in data. Nevertheless, the assumption that the MELD data model exhibits hierarchical structure does provide us with the ability to provide concrete analysis about the structure of the MELD data model. 

\begin{lemma}\label{lemma:delta bound}
Fix $\epsilon\in\Big(0,\frac{1}{\sqrt{n}}\Big)$, and
let $\mathcal{C}_t,\mathcal{C}_s\in \MELDemph$, where $t\in \mathcal{I}_\epsilon^{(\ell)}$ and $s\in \mathcal{I}_\epsilon^{(\ell')}$. If $\mathcal{C}_t$ is a refinement of $\mathcal{C}_s$,  then $\delta^{(\ell)}\leq \delta^{(\ell')}$. 
\end{lemma}
\begin{proof} 
Let $ \mathbf{P}_{k*}^{(\ell)} = [\mathbf{P}_{k,1}^{(\ell)} \; \mathbf{P}_{k, 2}^{(\ell)} \dots \mathbf{P}_{k,k-1}^{(\ell)} \; \mathbf{P}_{k,k+1}^{(\ell)}\dots \mathbf{P}_{k,n}^{(\ell)}]$ for each $k\in\{1,\dots, K_{\ell}\}$ and $\ell\in\{1,\dots, M\}$. By assumption, the clusters in $\{X_k^{(\ell)}\}_{k=1}^{K_\ell}$ can be merged to form the any of the clusters in $\{X_k^{(\ell')}\}_{k=1}^{K_{\ell'}}$. So, for all block rows $k$, any $\mathbf{P}_{k*}^{(\ell)}$ is a submatrix of some $\mathbf{P}_{j*}^{(\ell')}$. Therefore, for each $k\in \{1,\dots,K_\ell\}$, $\|\mathbf{P}_{k*}^{(\ell)}\|_\infty\leq \|\mathbf{P}_{j*}^{(\ell')}\|_\infty$ for some $j\in \{1\dots, K_{\ell'}\}$. Thus, $\delta^{(\ell')} = 2\max_{1\leq k\leq K_{\ell'}} \|\mathbf{P}_{k*}^{(\ell')}\|_\infty \leq 2\max_{1\leq k\leq K_{\ell}} \|\mathbf{P}_{k*}^{(\ell)}\|_\infty = \delta^{(\ell)}$.
\end{proof}

As diffusion progresses, diffusion distances separate ever coarser structure within the dataset. Because the $i$\ths component of the diffusion map $\Psi_t(x)$ is weighted by $\lambda_i^t$, each eigenfunction's contribution to diffusion distances will decay exponentially with $t$. As low-frequency eigenfunctions are annihilated, different mesoscopic equilibria will arise, during which diffusion distances induce different clusterings on $X$. Moreover, because fewer low-frequency eigenfunctions contribute to diffusion distances as $t$ increases, it is reasonable to expect the latent structure separated by diffusion distances to go from fine to coarse in scale.

\begin{prop} Fix $\epsilon\in\Big(0,\frac{1}{\sqrt{n}}\Big)$ and let $K_t$ denote the number of clusters in the clustering $\mathcal{C}_t\in \MELDemph$. If $\MELDemph$ exhibits hierarchical structure, then $K_t$ is monotonically non-increasing during $A_\epsilon$.\label{prop:MELD_K}
\end{prop}
\begin{proof} 
Let $\mathcal{C}_t,\mathcal{C}_s\in \MELD$ be any two distinct clusterings of $X$. We will show that if $K_t>K_s$, then $t<s$. Because $\mathcal{C}_t,\mathcal{C}_s\in \MELD$, there are intervals $\mathcal{I}_\epsilon^{(\ell)}$ and $\mathcal{I}_\epsilon^{(\ell')}$ such that $t\in\mathcal{I}_\epsilon^{(\ell)}$ and $s\in\mathcal{I}_\epsilon^{(\ell')}$. Since $\MELD$ exhibits hierarchical structure and $K_t> K_s$, $\mathcal{C}_t$ is a refinement of $\mathcal{C}_s$. By Lemma \ref{lemma:delta bound}, $\delta^{(\ell')}\geq  \delta^{(\ell)}$, which reduces to $\frac{\epsilon}{2\delta^{(\ell)}} \leq  \frac{\epsilon}{2\delta^{(\ell')}}$. By the assumption that $\mathcal{I}_\epsilon^{(\ell)}\cap\mathcal{I}_\epsilon^{(\ell')}=\emptyset$, this implies that  $\mathcal{I}_\epsilon^{(\ell)}$ is earlier in the diffusion process than $\mathcal{I}_\epsilon^{(\ell)}$. In particular, $t < s$, as desired. Thus, for any pair of clusterings $\mathcal{C}_t,C_s\in\MELD$, if $K_s<K_t$, then $\mathcal{C}_t$ is $\epsilon$-separable by diffusion distances earlier in the diffusion process than $\mathcal{C}_s$. We conclude that $K_t$ is monotonically non-increasing as a function of the diffusion time parameter $t$ during $A_\epsilon$, as desired.
\end{proof}

\subsection{Comparison to Related Models}\label{sec: data model comparison 3}

In order to understand MELD in greater detail, we compare to two related data models.

\subsubsection{Geometric Data Model} \label{sec: geometric data model 3}

The geometric data model (GDM) models $X$ by assuming points are sampled from a probability measure $\mu = \sum_{k=1}^K w_{k}\mu_k,$ where each $\mu_k$ is itself a probability measure on $X$ and $\sum_{k=1}^K w_k=1$. Each $\mu_k$ is assumed to be supported on some subset of $\R^D$, which is allowed to be nonlinear, nonconvex, and multimodal.  Typically, separation and coherence conditions are imposed on $\{\mu_k\}_{k=1}^K$; e.g., if $k\neq k'$, the support of $\mu_k$ does not overlap too much with that of $\mu_{k'}$, but connections are strong between data points sampled from each $\mu_k$. 

The GDM is non-parametric and assumes very little about the distributions $\{\mu_k\}_{k=1}^K$. Moreover, the assumption that each cluster is sampled from a different distribution leads to a simple interpretation of the clustering problem: to recover the correct distribution $\mu_k$ from which each $x_i\in X$ was sampled given solely the information provided by the dataset. However, the GDM requires there to be but one latent clustering to be learned, even though many datasets exhibit multiscale structure in practice. In contrast, the MELD data model allows for many different scales of cluster analysis, in some sense generalizing the GDM. Moreover, the GDM assumes a latent distribution on the data itself. While this generality offers significant theoretical advantages, it is also not always clear which clustering algorithm can best recover the correct distribution $\mu_k$ from which each data point was sampled~\cite{schiebinger2015geometry, arias2017spectralPCA, trillos2019geometric}. 

We note that the assumptions of the GDM can be modified to allow for hierarchical structure. More precisely, if each $\mu_k$ is itself of the form $\mu_k = \sum_{k'=1}^{K_k} w_{k,k'}\mu_{k,k'}$, where each $\mu_{k,k'}$ is a probability measure on $X$ and $\sum_{k=1}^{K'} w_{k,k'}=1$, data generated by the GDM have multiple scales of latent structure. One can interpret the goal of a coarse clustering of $X$ as recovering the correct distribution $\mu_k$ from which each $x_i\in X$ was sampled, given only the information in the dataset. On the other hand, one can interpret the goal of a finer clustering of $X$ as recovering the correct distribution $\mu_{k,k'}$ from which each $x_i\in X$ was sampled, given only the information in the dataset.  If assumptions on the separation and coherence of the supports of the measures $\mu_k$ and $\mu_{k,k'}$ are made, we expect the MELD data model to be able to be fit into the modified GDM framework described above. However, this is still a topic of ongoing research.

\subsubsection{The Stochastic Blockmodel } \label{sec: HSBM 3}

Another class of data models that remains widely used in clustering models the points in $X$ as nodes in a random network. The edges between nodes typically are sampled independently according to some probability distribution.  The stochastic blockmodel (SBM)~\cite{holland1983stochastic} is a random network model that assumes $K$ latent clusters $X_1,\dots, X_K$ exist in the graph and that there is a $K\times K$ matrix $\textbf{Q}$ storing between-cluster edge probabilities. More precisely, an edge will exist between $x\in X_i$ and $y\in X_j$ with probability $\textbf{Q}_{ij}$, independently of other edges. The SBM is a useful tool for proving performance guarantees on clustering algorithms because of its statistical construction \cite{rohe2011spectral}. 

By its definition, however, the SBM assumes a single scale of latent structure. The hierarchical stochastic blockmodel (HSBM) is a multiscale extension of the SBM~\cite{lyzinski2016HSBM}. The HSBM is similar to the SBM in that points are modeled as the nodes of a graph, the edges between which are sampled according to a probability distribution. However, unlike the SBM, the HSBM allows for multiple scales of separation to exist within the same graph. The benefits of the HSBM result from its statistical framework, which facilitates the analysis of hierarchical clustering algorithms. However, it is difficult to gain a geometric interpretation of the communities in HSBMs, because edges are generated independently. 

\section{Multiscale Learning by Unsupervised Nonlinear Diffusion} \label{sec: MLUND 4}

An advantage of the MELD data model is that it encapsulates a range of scales of separation within a single dataset. The possible separations range from coarse to fine, and we have shown that there is a natural relationship between the scale of latent cluster structure and the time parameter in a diffusion process. An important implication is that there are many ``correct'' clusterings of the same dataset.  It is then natural to ask: which among the many latent clusterings of a dataset contains the most information about its underlying structure? In this section, we introduce the M-LUND algorithm: a multiscale extension of the LUND algorithm. The M-LUND algorithm chooses the partition of $X$ that best represents all latent multiscale structure. In particular, it finds the barycenter among all nontrivial clusterings of $X$ learned by the LUND algorithm, where distance is measured using \emph{variation of information} (\emph{VI})~\cite{meilua2007VI}.

\subsection{Background on the Variation of Information Between Clusterings} \label{sec: VI 4}

Let $\mathcal{C}=\{X_k\}_{k=1}^K$ and  $\mathcal{C}' = \{X_k'\}_{k=1}^{K'}$ be two clusterings of $X$ with cluster sizes $|X_k| = n_k$ for $1\leq k\leq K$ and $|X_k'| = m_k$ for $1\leq k \leq K'$ respectively. A data point sampled from a uniform distribution over $X$ has probability $n_k/n$ of being a point from $X_k$. Hence, the clustering $\mathcal{C}$ can be associated with a discrete random variable taking $K$ values. A similar discrete random variable taking $K'$ values can be constructed for the clustering $\mathcal{C}'$~\cite{meilua2007VI}. 

The uncertainty associated with a random variable can be quantified by its \emph{entropy}.  The entropy of the clustering $\mathcal{C}$ is identified as the entropy of the random variable associated to $\mathcal{C}$: $H(\mathcal{C}) = -\sum_{i=1}^K \frac{n_i}{n}\log\big(\frac{n_i}{n}\big)$~\cite{meilua2007VI}. The entropy of a clustering will be zero whenever there is no uncertainty whatsoever about which cluster each point belongs to (i.e., the single-cluster clustering). Conversely, the entropy of $\mathcal{C}$ is maximal when it consists of $n$ singleton clusters. 

The random variables associated with $\mathcal{C}$ and $\mathcal{C}'$ also have a joint distribution: $\mathbb{P}(x\in X_i\bigcap X_j') = n_{ij}/n$, where $n_{ij}=|X_i\bigcap X_j'|$. Define the \emph{mutual information} between the clusterings $\mathcal{C}$ and $\mathcal{C}'$ by the mutual information between the random variables associated with them: $I(\mathcal{C}, \mathcal{C}') = -\sum_{i=1}^K\sum_{j=1}^{K'} \frac{n_{ij}}{n}\log\Big(\frac{n_{ij}/n}{(n_i/n)(m_j/n)}\Big)$~\cite{meilua2007VI}.
Mutual information quantifies the information gained about one random variable by observing another. In the context of clustering, $I(\mathcal{C}, \mathcal{C}')$ quantifies the information gained about the clustering $\mathcal{C}$  of $X$ from the observation of a different clustering $\mathcal{C}'$ of $X$~\cite{meilua2007VI}.

The \emph{VI} between $\mathcal{C}$ and $\mathcal{C}'$ can be defined in terms of the entropy of and mutual information between the clusterings $\mathcal{C}$ and $\mathcal{C}'$. The VI comparison scheme has the advantageous property of being a distance metric measuring how much information is maintained across two clusterings of the same dataset~\cite{meilua2007VI}.  

\begin{definition}\label{def: VI}  The \emph{VI} between two clusterings $\mathcal{C}$ and $\mathcal{C}'$ of X is defined to be $VI(\mathcal{C}, \mathcal{C}') = H(\mathcal{C}) + H(\mathcal{C}') - 2 I(\mathcal{C}, \mathcal{C}').$
\end{definition}

\subsection{The M-LUND Clustering Algorithm}
\label{sec: Algorithm 4}

In Section \ref{sec: MELD 3}, we noted that a clustering $\mathcal{C}_t\in \MELD$ can be interpreted as the latent clustering of $X$ at time $t$. Under assumptions on cluster density and diffusion at time $t$, the LUND algorithm with input $t$ is guaranteed to recover the latent clustering $\mathcal{C}_t$~\cite{murphy2019LUND}. Thus, the MELD data model and the LUND algorithm are closely linked, and the LUND algorithm can be interpreted as an algorithm to find the MELD clustering at a fixed time step. In this section, we leverage this relationship for a multiscale extension of the LUND algorithm based on the MELD data model. 

We begin by considering how the LUND algorithm's cluster assignments behave at very large time. We note that, when diffusion is close to stationarity, these clusterings become independent of $t$.  Indeed, for $x,y\in X$,
\begin{align*}
    D_t(x,y)^2 &= \lambda_2^{2t}(\psi_2(x)-\psi_2(y))^2+ \sum_{k=3}^n \lambda_k^{2t}(\psi_k(x)-\psi_k(y))^2\\ &= \lambda_2^{2t}\bigg[(\psi_2(x)-\psi_2(y))^2 +\sum_{k=3}^n \bigg(\frac{\lambda_k}{\lambda_2}\bigg)^{2t}(\psi_k(x)-\psi_k(y))^2\bigg].
\end{align*}
If there is a gap between $|\lambda_2|$ and $|\lambda_3|$, then $\big|\frac{\lambda_k}{\lambda_2}\big|<1$ for each $k\geq 3$. Therefore, while all eigenfunctions' contributions to diffusion distances converge to zero as $t\rightarrow\infty$, they do not do so at the same rate. When diffusion is near stationarity, higher-frequency eigenfunctions' contributions to diffusion distances are nearly zero relative to the contribution of the second eigenfunction. This implies that the clustering generated by the second eigenfunction of $\mathbf{P}$ will persist until diffusion distances are numerically zero. The persistence of that clustering does not reflect its stability, as the diffusion process will have effectively arrived at stationarity.

To avoid artificially increasing the stability of the clustering generated by the second eigenfunction of $\mathbf{P}$, we choose to terminate cluster analysis at a maximum time step. We will cluster $X$ using the LUND algorithm for all $t$ in the set $\{0, 1, \beta,  \dots, \beta^T\}$, where $\beta>1$ is an exponential sampling rate and $T= \big\lceil\log_\beta\big[\log_{|\lambda_2|}\big(\frac{\tau\pi_{\min}}{2}\big)\big]\big\rceil$ is a maximum time index depending on the quantity $\pi_{\min} = \min_{u\in X} \pi(u)$ and a stationarity threshold $\tau\in(0,1)$. Intuitively, smaller $\beta$ corresponds to a finer sampling of this interval and more precision when recovering latent multiscale structure in $X$. However, each additional time sample would correspond to another run of the LUND algorithm, so $\beta$ should be tuned according to the size of the dataset and available computational resources. The quantity $\tau$ is a threshold for how close to stationarity diffusion should be to end cluster analysis and is typically small ($\tau\ll 10^{-2}$) in practice. The quantity $T$ will be justified further in our theoretical guarantees in Section \ref{sec:tau}. 

To find the optimal clustering of $X$, we solve $\mathcal{C}_{t^*} = \text{argmin} \big\{\text{VI}^{(\text{tot})}(\mathcal{C}_{t}) \big| t\in J\big\}$, where the \emph{total VI} of the clustering $\mathcal{C}_t$ is defined to be  $\text{VI}^{(\text{tot})}(\mathcal{C}_{t})= \sum_{s\in J}\text{VI}(\mathcal{C}_{t}, \mathcal{C}_s)$ and $J = \big\{t=\beta^j\;|j\in \{-\infty,0,1,\dots, T\}, K_t\in~[2,\frac{n}{2})\big\}$. We restrict our analysis to clusterings sampled during $J$ because it is possible that some clusterings extracted by the LUND algorithm are not meaningful; for example if the LUND algorithm is evaluated during a transition region. We will refer to a clustering $\mathcal{C}_t$ as \emph{nontrivial} if $K_t\in [2,\frac{n}{2})$ and \emph{trivial} otherwise. Thus, $J$ corresponds to the time steps during which the LUND algorithm extracts a nontrivial clustering.  We choose a lower bound of $K_t=2$ because the single-cluster clustering yields no meaningful information about the dataset. We choose an upper bound of $K_t = \frac{n}{2}$ to avoid singleton clusters. Thus, the clustering $\mathcal{C}_{t^*}$ is the partition of $X$ that best represents the nontrivial multiscale structure detected by the LUND algorithm across the diffusion process. The M-LUND algorithm is provided in Algorithm \ref{alg: MLUND}. 

\begin{figure}[t] 
\begin{algorithm}[H]\label{alg: MLUND}
\SetAlgoLined
\KwIn{$X$ (dataset),  $\sigma$~(diffusion scale), $\sigma_0$ (KDE bandwidth), $\beta$ (sampling rate), \newline $\tau$ (diffusion stationarity threshold)}
\KwOut{$\{\mathcal{C}_{t_i} |t_i= 0,1,\beta, \dots, \beta^T \}$ (multiscale clusterings), $\mathcal{C}_{t^*}$ (optimal clustering), \newline $K_{t^*}$ (optimal no. clusters)}
Construct the transition matrix $\mathbf{P}$ and its stationary distribution $\pi$ with a Gaussian kernel and diffusion scale $\sigma$\;
Calculate $T =\big\lceil\log_\beta\big[\log_{|\lambda_2(\mathbf{P})|}\big(\frac{\tau\min(\pi)}{2}\big)\big]\big\rceil$\;
\For{$t_i\in \{0, 1,\beta, \beta^2, \dots, \beta^T\}$}{
    $[\mathcal{C}_{t_i}, K_{t_i}] = \text{LUND}(X, \sigma_0, \sigma, t_i)$\;
}
$J = \{t_i\;|1<K_{t_i} <\frac{n}{2} \}$ \;
\For{$t_i\in J$}{
    $\text{VI}^{(\text{tot})}(\mathcal{C}_{t_i})= \sum_{s\in J}\text{VI}(\mathcal{C}_{t_i}, \mathcal{C}_s)$\;
}

$\mathcal{C}_{t^*} = \text{argmin} \big\{\text{VI}^{(\text{tot})}(\mathcal{C}_{t_i}) \big| t_i\in J\big\}$\;
$K_{t^*}$ = number of unique clusters in $\mathcal{C}_{t^*}$\;
 \caption{Multiscale Learning by Unsupervised Nonlinear Diffusion (M-LUND)}
\end{algorithm}
\end{figure}

Two important factors contribute to the minimizer $\mathcal{C}_{t^*}$. The first is $VI(\mathcal{C}_{t^*}, \mathcal{C}_t)$ for $t\neq t^*$. The clustering $\mathcal{C}_{t^*}$ minimizes total VI across all nontrivial clusterings of $X$ learned by the LUND algorithm across the diffusion process. Thus, $\mathcal{C}_{t^*}$ is the clustering that contains the most information about the latent multiscale structure in $X$.  The second important factor in the M-LUND algorithm's decision-making is the stability of latent clusterings.  Assume that the LUND algorithm extracts $L$ unique nontrivial clusterings $\mathcal{C}_\ell$ and define $J_\ell = \{s\in J \ | \ \mathcal{C}_s = \mathcal{C}_\ell\}$ to be the time samples during which the LUND algorithm extracted the clustering $\mathcal{C}_\ell$.  We can write the total VI of a nontrivial clustering $\mathcal{C}_t$ as $VI^{(\text{tot})}(\mathcal{C}_t) = \sum_{\ell=1}^L |J_\ell| VI(\mathcal{C}_t, \mathcal{C}_\ell)$. The value $|J_\ell|$ can be interpreted as a proxy for the log-length of the interval $\mathcal{I}_\epsilon^{(\ell)}$. Thus, stable clusterings are emphasized by the M-LUND minimization scheme. Stability in the diffusion process is highly related to attractive properties in a clustering. For example, Proposition \ref{prop:stability} implies that clusterings that consist of well-separated and coherent clusters are, all else equal, $\epsilon$-separable by diffusion distances for a longer interval of time on a logarithmic scale. Therefore, by emphasizing clusterings' stability in its optimization, the M-LUND algorithm weights representative clusterings with coherent and well-separated clusters higher. 

 In our performance guarantees (Section \ref{sec: guarantees}), we show that, under assumptions on entropy and mutual information of the clusterings in the MELD data model, the $VI^{(\text{tot})}$-minimizer will be in $\MELD$. In this sense, the M-LUND algorithm chooses the stability-weighted VI barycenter of the MELD data model.

\subsubsection{The Role of Diffusion Stability in the Output of the M-LUND Algorithm}\label{sec: Toy Example 4}

\begin{figure}[t] 
    \centering 
    \begin{subfigure}[t]{0.25\textwidth}
        \centering
         \includegraphics[height = 1.65in]{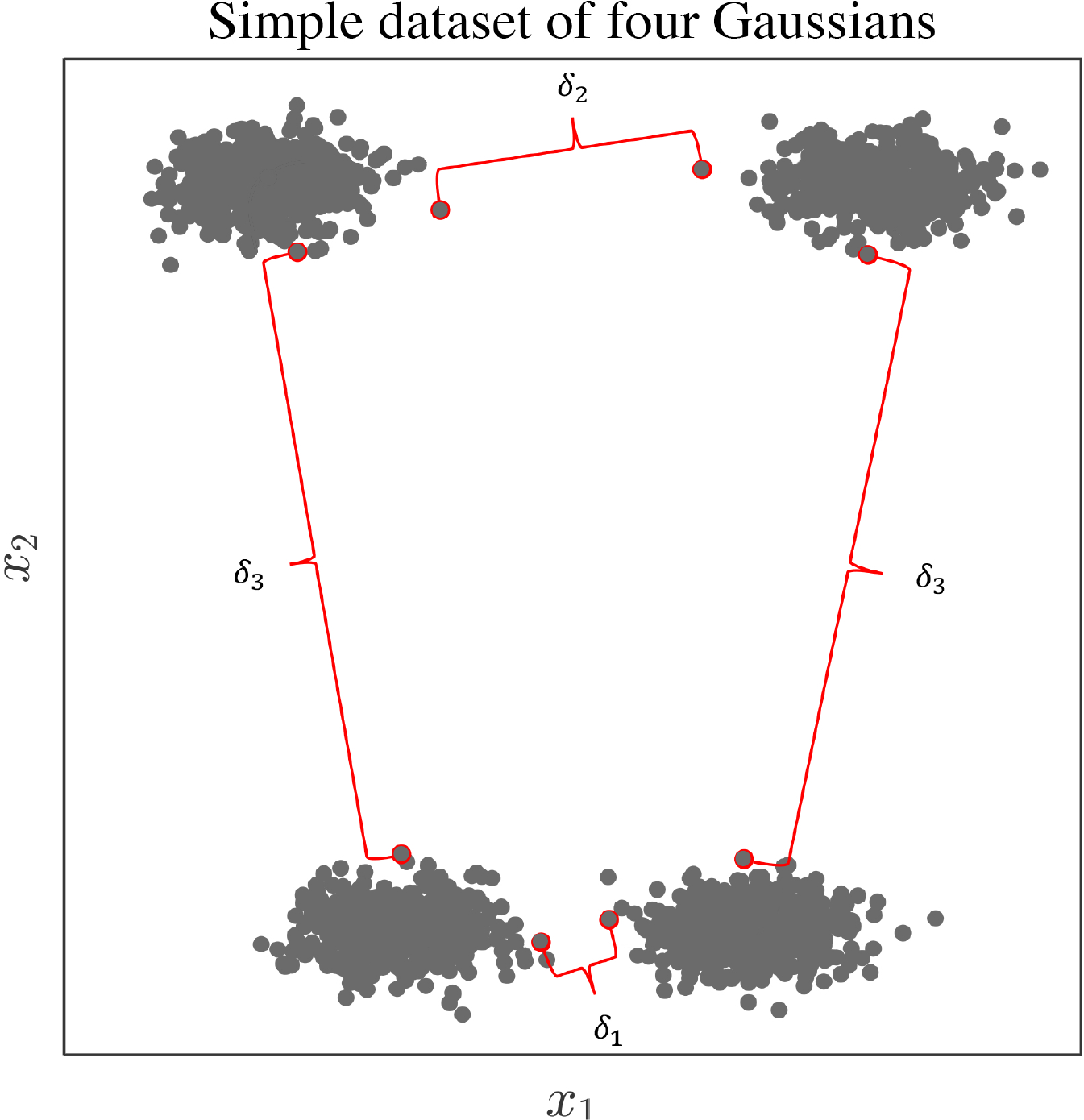} 
     \end{subfigure}%
    \begin{subfigure}[t]{0.25\textwidth}
        \centering
        \includegraphics[height = 1.65in]{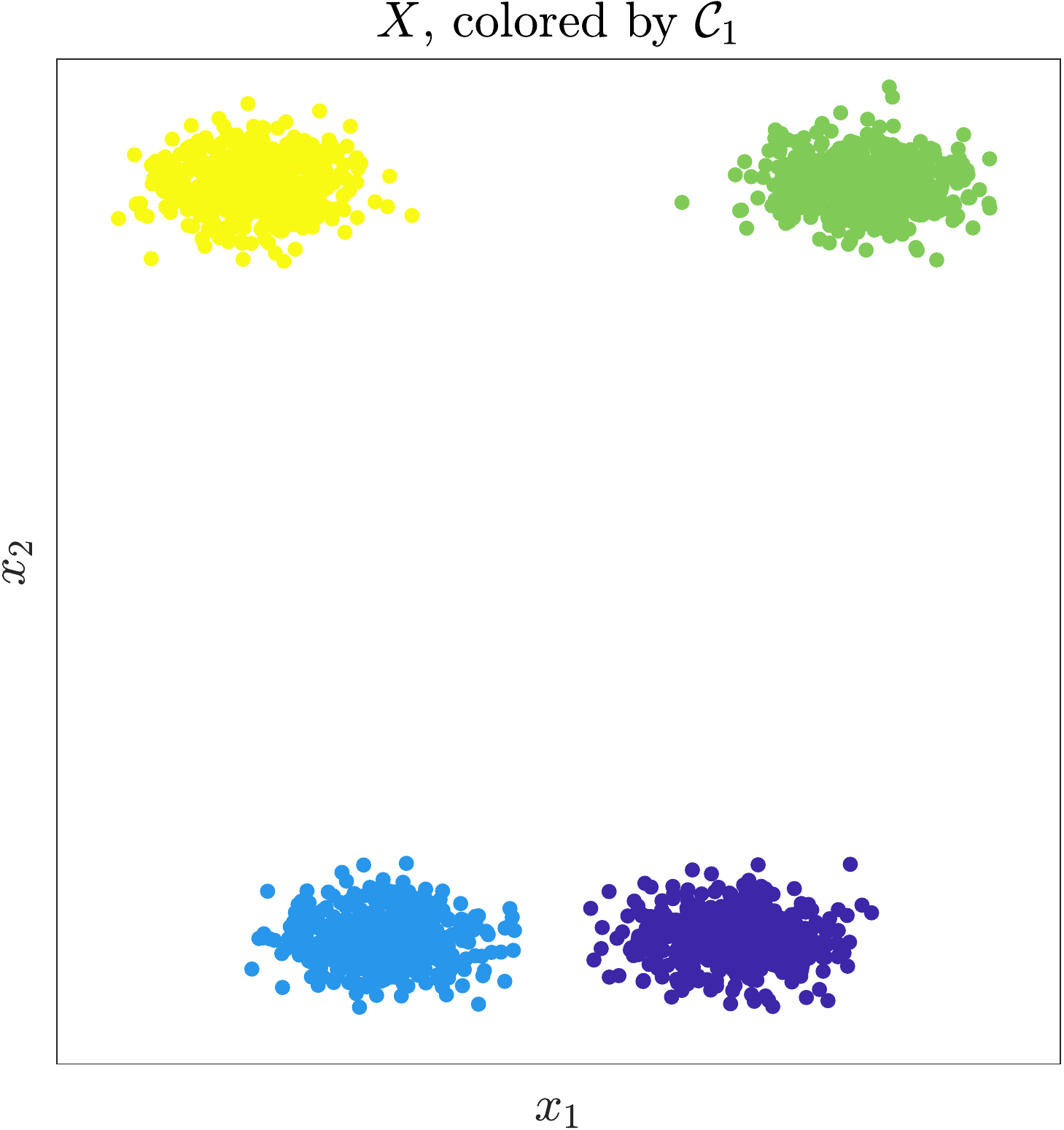}
    \end{subfigure}%
    \begin{subfigure}[t]{0.25\textwidth}
        \centering
        \includegraphics[height = 1.65in]{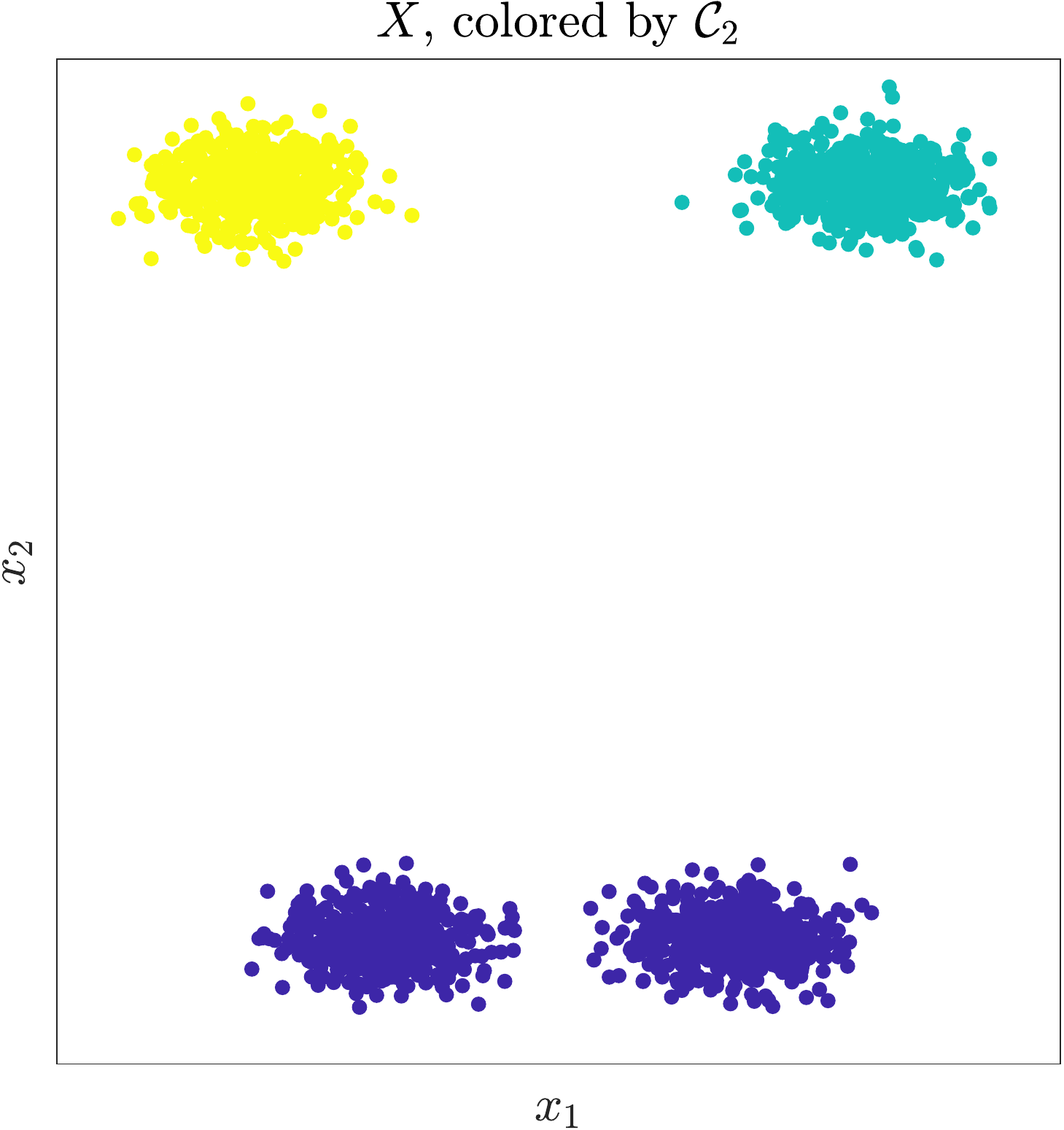}
    \end{subfigure}%
    \begin{subfigure}[t]{0.25\textwidth}
        \centering
        \includegraphics[height = 1.65in]{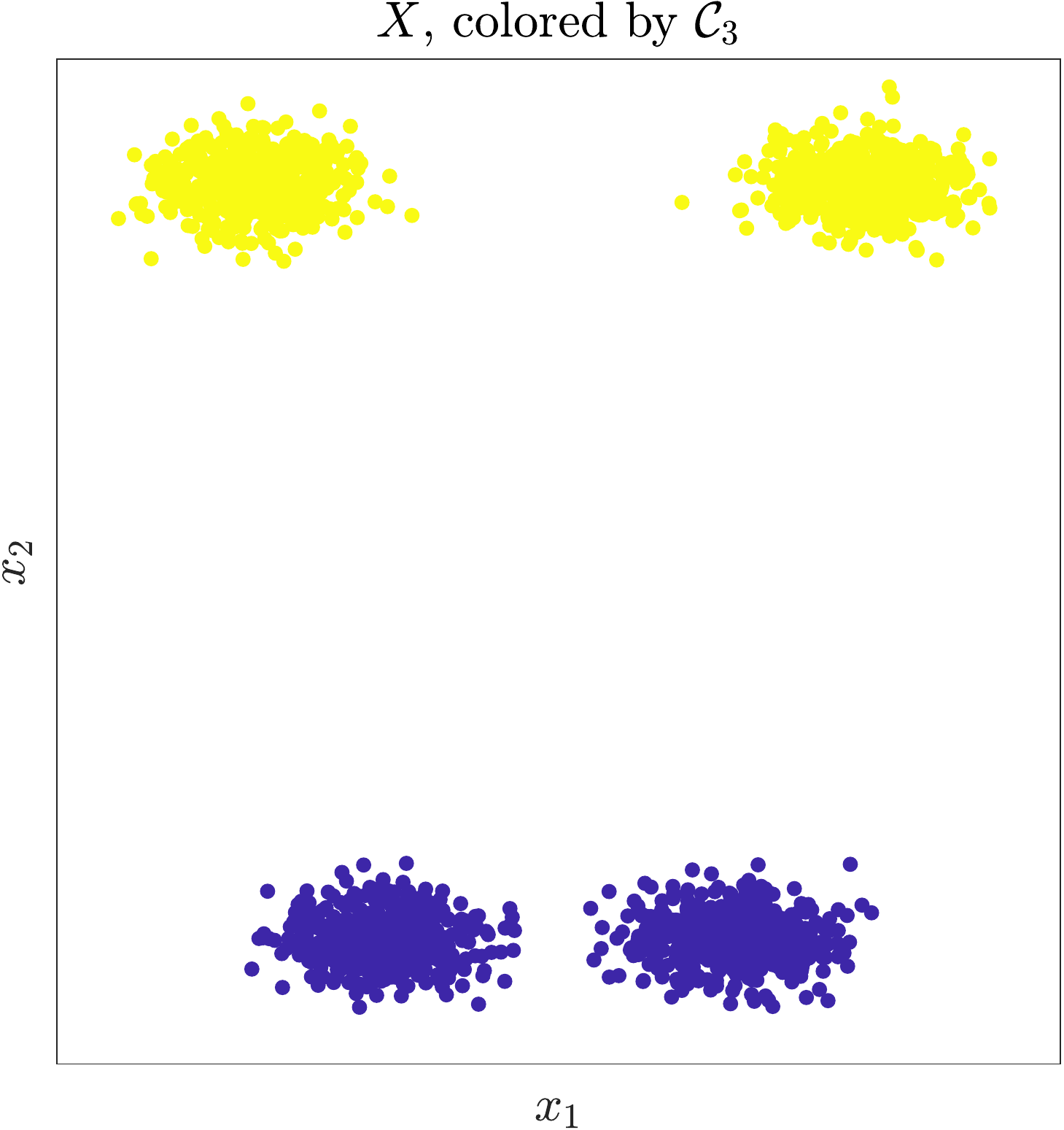}
    \end{subfigure}%
     \caption{A mixture of Gaussians example considered in this section. In the leftmost panel, the distances $\delta_1$, $\delta_2$, and $\delta_3$ are indicated on the dataset, with between-cluster Euclidean distance minimizer indicated in red. The clusterings $\mathcal{C}_1$, $\mathcal{C}_2$, and $\mathcal{C}_3$ are visualized in the right three panels.}     \label{fig:toy}
\end{figure}

Consider four well-separated clusters in $\R^D$ of equal size $n/4$, arranged on the vertices of a trapezoid (see Figure \ref{fig:toy}). Mathematically, we let $X = \bigcup_{k=1}^4 X_k$, where $X_k$ consists of points from the $k$\ths cluster. We define $\delta_1 = \min_{x\in X_1, y\in X_2}\|x-y\|_2$, $\delta_2 = \min_{x\in X_3, y\in X_4} \|x-y\|_2$, and assume that $\delta_3= \min_{x\in X_1, y\in X_3} \|x-y\|_2= \min_{x\in X_2, y\in X_4} \|x-y\|_2$ so that the clusters $X_1$ and $X_3$ are as well-separated as the clusters $X_2$ and $X_4$.  Supposing $0\ll \delta_1<\delta_2< \delta_3$, consider three distinct nontrivial clusterings: $\mathcal{C}_1 = \{X_1, X_2, X_3, X_4\}$, $\mathcal{C}_2 = \{X_1\bigcup X_2, X_3, X_4$\}, and $\mathcal{C}_3 = \{X_1\bigcup X_2, X_3\bigcup X_4\}$. 

The separation parameters $\delta_k$ are related to the stability of the clusterings $\mathcal{C}_\ell$.  If the $\delta_k$ are large and nearly equal, then transitions between any pair of clusters will tend to be unlikely. In this case, $\mathcal{C}_1$ will be more stable in the diffusion process than $\mathcal{C}_2$ and $\mathcal{C}_3$. Conversely, if $\delta_1$ and $\delta_2$ are very small compared to $\delta_3$, then $\mathcal{C}_3$ will be more stable in the diffusion process than $\mathcal{C}_1$ and $\mathcal{C}_2$. The M-LUND algorithm learns multiscale cluster structure by evaluating the LUND algorithm at an exponential sampling of the diffusion process. Thus, more stable clusterings will be extracted more frequently and weighted higher in the M-LUND minimization problem. We can explicitly derive when the M-LUND algorithm will choose one clustering over another as a function of the stability of those clusterings to graph diffusion.

\begin{prop}\label{prop:toy}
For $\ell=1,2,3$, assume $\mathcal{C}_\ell$ is extracted by the M-LUND algorithm $m_\ell\in\mathbb{N}$ times. If $\mathcal{C}_1$, $\mathcal{C}_2$, and $\mathcal{C}_3$ are the only nontrivial clusterings extracted by the M-LUND algorithm, then
\begin{enumerate}
    \item $\mathcal{C}_1$ is chosen by the M-LUND algorithm if and only if $m_1\geq m_2 + m_3$. 
    \item $\mathcal{C}_2$ is chosen by the M-LUND algorithm  if and only if $m_2\geq |m_1-m_3|$.
    \item $\mathcal{C}_3$ is chosen by the M-LUND algorithm if and only if $m_3\geq m_1+m_2$. 
\end{enumerate}
\end{prop}
\begin{proof} The stated assumptions imply $VI^{(tot)}(\mathcal{C}_\ell) = \sum_{k=1}^3 m_\ell VI(\mathcal{C}_\ell, \mathcal{C}_k)$. By definition, $VI(\mathcal{C}_1,\mathcal{C}_2) =~0.5 \log(2)$, $VI(\mathcal{C}_1,\mathcal{C}_3) =~\log(2)$, and $ VI(\mathcal{C}_2,\mathcal{C}_3) = 0.5 \log(2)$. Since $VI(\mathcal{C}_\ell, \mathcal{C}_\ell) = 0 $ for all $\ell$, total VI is calculated to be
\begin{align*}
    VI^{(\text{tot})}(\mathcal{C}_1) &= m_2VI(\mathcal{C}_1, \mathcal{C}_2) + m_3VI(\mathcal{C}_1, \mathcal{C}_3)
    = 0.5\log (2)[m_2+ 2m_3]\\
    VI^{(\text{tot})}(\mathcal{C}_2) &= m_1VI(\mathcal{C}_1, \mathcal{C}_2) + m_3VI(\mathcal{C}_1, \mathcal{C}_3)
    = 0.5\log (2)[m_1 + m_3]\\
    VI^{(\text{tot})}(\mathcal{C}_3) &= m_1VI(\mathcal{C}_1, \mathcal{C}_3) + m_2VI(\mathcal{C}_2, \mathcal{C}_3)
    = 0.5\log (2)[2m_1+ m_2 ].
\end{align*}
Algebra comparing $VI^{(\text{tot})}(\mathcal{C}_\ell)$ across $\ell\in\{1,2,3\}$ yields the result.
\end{proof}

Proposition \ref{prop:toy} suggests that stability in the diffusion process is critical in M-LUND's optimization scheme. If $m_\ell = m_{\ell'}$ for all $1\leq \ell,\ell' \leq 3$, Proposition \ref{prop:toy} implies that the M-LUND algorithm will output the intermediate clustering $\mathcal{C}_2$.  Conversely, if $\mathcal{C}_1$ is extracted $k>2$ times more frequently than $\mathcal{C}_2$ and $\mathcal{C}_3$ so that $m_1> km_\ell$ for $\ell=2,3$, then $\mathcal{C}_1$ will be the minimizer of total VI. Thus, even though $\mathcal{C}_2$ is an intermediate clustering closest in VI to both $\mathcal{C}_1$ and $\mathcal{C}_3$, the M-LUND algorithm chooses $\mathcal{C}_1$ because of its relatively higher stability in the diffusion process. 
 
\subsection{Performance Guarantees for Unsupervised Clustering}\label{sec: guarantees}

In this section, we provide performance guarantees on the M-LUND algorithm. We begin by reviewing guarantees on the performance of the LUND algorithm in Section \ref{sec: LUND guarantees} and extend these to performance guarantees of the M-~LUND algorithm in Section \ref{sec:MLUND guarantees}. In Section \ref{sec:tau}, we provide theoretical justification for the termination of the first for-loop in the M-LUND algorithm at time $\beta^T$ by showing that, for any $t>\beta^T$ and pair of points $x,y\in X$, $D_t(x,y)<\tau$. 

\subsubsection{Performance Guarantees on the LUND Clustering Algorithm} \label{sec: LUND guarantees}

In this section, we review previously-introduced guarantees on the performance of the LUND algorithm at recovering latent cluster structure at a fixed time step~\cite{murphy2019LUND}. We will assume that there is a latent clustering  $\mathcal{C}_t = \{X_k^{(t)}\}_{k=1}^{K_t}$ of $X$ at time $t\geq 0$ and refer to the maximum within-cluster and minimum between-cluster diffusion distance at time $t$ for the clustering $\mathcal{C}_t$ as $D_t^{\text{{in}}}(\mathcal{C}_t)$ and $D_t^{\text{{btw}}}(\mathcal{C}_t)$ respectively. We aim to show that, under plausible assumptions on density and diffusion at time $t$, the LUND algorithm with input $t$ recovers the latent clustering $\mathcal{C}_t$~\cite{murphy2019LUND}.
 
\begin{definition}\label{def: density maximizers 3}For a latent clustering  $\mathcal{C}_t = \{X_k^{(t)}\}_{k=1}^{K_t}$ of $X$ at time $t\geq 0$, define the set of \emph{cluster density maxima} at time $t$ by $\mathcal{M}_t = \Big\{ p(x) \Big| \exists k\in \{1,\dots, K_t\} : x = \text{\emph{argmax}}_{x\in X_k^{(t)}} p(x)\Big\}$. 
 \end{definition}

The LUND algorithm estimates the modes of clusters in the latent clustering at time  $t\geq 0$ as the maximizers of $\mathcal{D}_t(x)$. The following theorem guarantees that the cluster modes learned by the LUND algorithm are the highest-density points within clusters in the latent clustering at time $t$~\cite{murphy2019LUND}. 

\begin{theorem} \label{thm: density maximizers} \cite{murphy2019LUND}
    For a latent clustering $\mathcal{C}_t = \{X_k^{(t)}\}_{k=1}^{K_t}$ of $X$ at time $t\geq 0$, denote the $K_t$ maximizers of $\mathcal{D}_t(x)$ as $\big\{x_i^{(t)*}\big\}_{k=1}^{K_t}$. If $\frac{D_t^{\text{\emph{in}}}(\mathcal{C}_t)}{D_t^{\text{\emph{btw}}}(\mathcal{C}_t)}<\frac{\min(\mathcal{M}_t)}{\max(\mathcal{M}_t)}$, there is a permutation $(k_1,\dots, k_{K_t})$ of $(1,\dots, K_t)$ such that $x_i^{(t)*}$ maximizes empirical density among points in the cluster $X_{k_i}^{(t)}$.
    \end{theorem}

Thus, the cluster modes estimated by the LUND algorithm are cluster-wise empirical density maximizers. The LUND algorithm estimates the number of clusters at time $t$ using the ratio of the sorted values taken by $\mathcal{D}_t(x)$.

\begin{corollary} \label{cor: modes} \cite{murphy2019LUND}
For a latent clustering  $\mathcal{C}_t = \{X_k^{(t)}\}_{k=1}^{K_t}$ of $X$ at time $t\geq 0$, let $\{x_{m_i}^{(t)}\}_{i=1}^{n}$  be the points in $X$ sorted in non-increasing order by $\mathcal{D}_t(x)$. Then, for $j< K_t$, $\frac{\mathcal{D}_t(x_{m_j}^{(t)})}{\mathcal{D}_t(x_{m_{j+1}}^{(t)})}\leq \frac{\max(\mathcal{M}_t)}{\min(\mathcal{M}_t)}\frac{\max_{1\leq k \leq K}\rho_t(x_{m_k}^{(t)})}{\min_{1\leq k \leq K}\rho_t(x_{m_k}^{(t)})}$. Conversely, $\frac{\mathcal{D}_t(x_{m_{K_t}}^{(t)})}{\mathcal{D}_t(x_{m_{K_t+1}}^{(t)})}\geq \frac{\min(\mathcal{M}_t)}{\max(\mathcal{M}_t)}\frac{D_t^{\text{\emph{btw}}}(\mathcal{C}_t)}{D_t^{\text{\emph{in}}}(\mathcal{C}_t)}$.
\end{corollary}

By Corollary \ref{cor: modes}, under reasonable assumptions on the data and the latent clustering, $\mathcal{D}_t(x_{m_k}^{(t)})/\mathcal{D}_t(x_{m_{k+1}}^{(t)})$ will be large for the first $K_t$ values and small thereafter, yielding an accurate estimation of the number of clusters at time $t$.  In Corollary \ref{cor: LUND labeling}, these assumptions imply that the LUND algorithm perfectly recovers the latent clustering at time $t$~\cite{murphy2019LUND}. 

\begin{corollary}\label{cor: LUND labeling} 
For a latent clustering  $\mathcal{C}_t = \{X_k^{(t)}\}_{k=1}^{K_t}$ of $X$ at time $t\geq 0$, let $\{x_{m_i}^{(t)}\}_{i=1}^{n}$  be the points in $X$ sorted in non-increasing order by $\mathcal{D}_t(x)$. The LUND algorithm with input $t$ will recover the latent clustering $\mathcal{C}_t$ if 
\[    \frac{D_t^{\emph{\text{in}}}}{D_t^{\emph{btw}}} <\min\left\{\frac{\min_{1\leq i \leq K_t}\rho_t(x_{m_i}^{(t)})}{\max_{1\leq i \leq K_t}\rho_t(x_{m_i}^{(t)})} \bigg(\frac{\min(\mathcal{M}_t)}{\max(\mathcal{M}_t)}\bigg)^2,  \frac{\min_{y\in X} p(y)}{\max(\mathcal{M}_t)} \frac{\min_{1\leq k\leq K_t}\min_{x\neq y\in X_k^{(t)}}D_t(x,y)}{D_t^{\text{\emph{in}}}(\mathcal{C}_t)}\right\}.
\]
\end{corollary}
\begin{proof} First, we prove that the LUND algorithm correctly recovers the number of clusters $\mathcal{C}_t$, denoted $K_t$. For $j<K_t$,
\[\frac{\mathcal{D}_t(x_{m_j}^{(t)})}{\mathcal{D}_t(x_{m_{j+1}}^{(t)})}\leq~\frac{\max(\mathcal{M}_t)}{\min(\mathcal{M}_t)}\frac{\max_{1\leq k\leq K_t}\rho_t(x_{m_k}^{(t)})}{\min_{1\leq k\leq K_t}\rho_t(x_{m_k}^{(t)})} <~ \frac{\min(\mathcal{M}_t)}{\max(\mathcal{M}_t)}\frac{D_t^{\text{{btw}}}(\mathcal{C}_t)}{D_t^{\text{{in}}}(\mathcal{C}_t)} \leq \frac{\mathcal{D}_t(x_{m_{K_t}}^{(t)})}{\mathcal{D}_t(x_{m_{K_t+1}}^{(t)})},\]
where Corollary \ref{cor: modes} was used to gain the first and last inequalities and  $\frac{D_t^{\text{{in}}}(\mathcal{C}_t)}{D_t^{\text{{btw}}}(\mathcal{C}_t)}\frac{\max_{1\leq i \leq K_t}\rho_t(x_{m_i}^{(t)})}{\min_{1\leq i \leq K_t}\rho_t(x_{m_i}^{(t)})}<~\Big(\frac{\min(\mathcal{M}_t)}{\max(\mathcal{M}_t)}\Big)^2$ was used to gain the second. Next, let $j>K_t$. 
Because $D_t^{\text{in}}\leq D_t^{\text{btw}}$ and  $p(x_{m_j}^{(t)})\leq \min(\mathcal{M}_t)$ by Theorem \ref{thm: density maximizers}, we clearly have that $\frac{\rho_t(x_{m_j}^{(t)})}{\rho_t(x_{m_{j+1}}^{(t)})} \leq \frac{D_t^{\text{in}}(\mathcal{C}_t)}{\min_{1\leq k \leq K_t} \min_{x,y\in X_k^{(t)}}D_t(x,y)}$. Thus,  \[\frac{\mathcal{D}_t(x_{m_j}^{(t)})}{\mathcal{D}_t(x_{m_{j+1}}^{(t)})} \leq \frac{\min(\mathcal{M}_t)}{\min_{y\in X}p(y)} \frac{D_t^{\text{in}}(\mathcal{C}_t)}{\min_{1\leq k \leq K_t} \min_{x,y\in X_k^{(t)}}D_t(x,y)} < \frac{\min(\mathcal{M}_t)}{\max(\mathcal{M}_t)}\frac{D_t^{\text{{btw}}}(\mathcal{C}_t)}{D_t^{\text{{in}}}(\mathcal{C}_t)} \leq \frac{\mathcal{D}_t(x_{m_{K_t}}^{(t)})}{\mathcal{D}_t(x_{m_{K_t+1}}^{(t)})},\]  
where  $\frac{D_t^{{\text{in}}}(\mathcal{C}_t)}{\min_{1\leq k\leq K_t}\min_{x\neq y\in X_k^{(t)}}D_t(x,y)} < \frac{\min_{y\in X} p(y)}{\max(\mathcal{M}_t)}  \frac{D_t^{\text{{btw}}}(\mathcal{C}_t)}{D_t^{\text{{in}}}(\mathcal{C}_t)}$ was used to gain the second inequality, and Corollary \ref{cor: modes} was used to gain the last. So, the LUND algorithm correctly estimates $K_t = \text{argmax}_{1\leq k \leq n-1}  \mathcal{D}_t(x_{m_k}^{(t)})/\mathcal{D}_t((x_{m_{k+1}}^{(t)})$ and labels cluster modes $\mathcal{C}(x_{m_k}^{(t)}) = k$ ($k=1,\dots, K_t$). Lastly, we show that non-modal labels are assigned correctly. Let $x\in X_k^{(t)}$ be any unlabeled, non-modal point. Because $D_t^{\text{in}}(\mathcal{C}_t)\leq D_t^{\text{btw}}(\mathcal{C}_t)$, the data point $x^* = \text{argmin}_{y\in X}\{D_t(x,y) | p(y)\geq p(x), \text{$y$ is labeled}\}$ must be a point in $X_k^{(t)}$. So,  $\mathcal{C}(x) = \mathcal{C}(x^*)= k$. By induction, all non-modal points are labeled correctly.
\end{proof}

Corollary \ref{cor: LUND labeling} relies on a technical assumption that is sufficient (though not necessary) for successful recovery of $\mathcal{C}_t$ by LUND. The first assumption, that  $\frac{D_t^{\text{{in}}}(\mathcal{C}_t)}{D_t^{\text{{btw}}}(\mathcal{C}_t)}<\frac{\min_{1\leq i \leq K_t}\rho_t(x_{m_i}^{(t)})}{\max_{1\leq i \leq K_t}\rho_t(x_{m_i}^{(t)})} \Big(\frac{\min(\mathcal{M}_t)}{\max(\mathcal{M}_t)}\Big)^2$, holds if $p(x)$ yields comparable values for cluster modes and between-cluster mode diffusion distances are roughly constant. For such datasets,  $\min_{1\leq i \leq K_t}\rho_t(x_{m_i}^{(t)})/\max_{1\leq i \leq K_t}\rho_t(x_{m_i}^{(t)})$ will be insignificant, and $\min(\mathcal{M}_t)/\max(\mathcal{M}_t)$ will be close to 1. The second assumption, that $\frac{D_t^{\text{{in}}}(\mathcal{C}_t)}{D_t^{\text{{btw}}}(\mathcal{C}_t)}<\frac{\min_{y\in X} p(y)}{\max(\mathcal{M}_t)} \frac{\min_{1\leq k\leq K_t}\min_{x\neq y\in X_k^{(t)}}D_t(x,y)}{D_t^{\text{\emph{in}}}(\mathcal{C}_t)}$, holds for datasets in which $p(x)$ has low variance and for which $\Psi_t(x)$ sends each cluster approximately to a point mass (e.g., in Figure \ref{fig: multiscale Diffusion Map}). For such datasets, within-cluster diffusion distances are nearly constant, so $D_t^{\text{in}}(\mathcal{C}_t)/D_t^{\text{btw}}(\mathcal{C}_t)$ is small compared to $\min_{1\leq k\leq K_t}\min_{x\neq y\in X_k^{(t)}}D_t(x,y)/D_t^{\text{in}}(\mathcal{C}_t)$. 

\subsubsection{Performance Guarantees on the M-LUND Clustering Algorithm}\label{sec:MLUND guarantees}

 We now provide performance guarantees for the M-LUND algorithm, all of which rely on the following setup:
  
\begin{definition}\label{def: setup}
We refer to the following as \emph{the usual setup}: let $\epsilon\in\Big(0,\frac{1}{\sqrt{n}}\Big)$, $\beta>1$, and $\tau\in(0,1)$. For each $t\in \big\{0,1,\beta, \dots, \beta^T\}\bigcap A_\epsilon$, let $\mathcal{C}_t~\in~\MELD$ be the latent clustering of $X$ at time $t$, and let 
$\{x_{m_i}^{(t)}\}_{i=1}^{n}$  be the points in $X$ sorted in non-increasing order by $\mathcal{D}_t(x)$. Assume $\displaystyle\min_{1\leq \ell \leq M}\delta^{(\ell)}>\frac{\epsilon}{2}\log_{\frac{\tau \pi_{\min}}{2}}(|\lambda_2|)$ and that, for each  $t\in~\big\{0,1,\beta, \dots, \beta^T\}\bigcap A_\epsilon$,  
\begin{align}
    \frac{\epsilon}{1/\sqrt{n}-\epsilon} <\min\left\{\frac{\min_{1\leq i \leq K_t}\rho_t(x_{m_i}^{(t)})}{\max_{1\leq i \leq K_t}\rho_t(x_{m_i}^{(t)})} \bigg(\frac{\min(\mathcal{M}_t)}{\max(\mathcal{M}_t)}\bigg)^2,  \frac{\min_{y\in X} p(y)}{\max(\mathcal{M}_t)} \frac{\min_{1\leq k\leq K_t}\min_{x\neq y\in X_k^{(t)}}D_t(x,y)}{D_t^{\text{\emph{in}}}(\mathcal{C}_t)}\right\}.\label{eq: assumption}
\end{align}
\end{definition}

There are two main assumptions in the usual setup. The first, that $\displaystyle\min_{1\leq \ell \leq M}\delta^{(\ell)}>\frac{\epsilon}{2}\log_{\frac{\tau \pi_{\min}}{2}}(|\lambda_2|)$, requires that the separation between clusters not be too strong. To gain some intuition for this condition, consider the idealized case in which, for some clustering scale $\ell$, $\delta^{(\ell)}=0$ so that the clusters $X_k^{(\ell)}$ are perfectly separated.  Then, the upper limit of $\mathcal{I}_\epsilon^{(\ell)}$ is infinite; to accurately estimate $\mathcal{I}_\epsilon^{(\ell)}$, the M-LUND algorithm would need to sample an infinite number of time steps.  Thus, separation must not be so strong that diffusion spreads within clusters of a single clustering ad infinitum.

\begin{lemma} \label{lemma: endpoint}
    Let $\epsilon>0$, $\beta>1$, and $\tau\in(0,1)$. If $\displaystyle\min_{1\leq \ell \leq M}\delta^{(\ell)}>\frac{\epsilon}{2}\log_{\frac{\tau \pi_{\min}}{2}}(|\lambda_2|)$, then $A_\epsilon \subset [0,\beta^T]$.
\end{lemma}
\begin{proof}
If $\displaystyle\min_{1\leq \ell \leq M}\delta^{(\ell)}>\frac{\epsilon}{2}\log_{\frac{\tau \pi_{\min}}{2}}(|\lambda_2|)$, then $\displaystyle\max_{1\leq \ell\leq M}\frac{\epsilon}{2\delta^{(\ell)}}< \log_{|\lambda_2|}\left(\frac{\tau \pi_{\min}}{2}\right)\leq\beta^T$. Since $A_\epsilon\subset \left[0,\displaystyle\max_{1\leq \ell \leq M}\frac{\epsilon}{2\delta^{(\ell)}}\right]$, it follows that $A_\epsilon\subset [0,\beta^T]$.
\end{proof}

The M-LUND algorithm extracts the latent clusterings of $X$ by implementing the LUND algorithm at different choices of $t$.  However, because cluster analysis is terminated at time $t=\beta^T$, the cluster extraction stage of the M-LUND algorithm may end before the end of the last interval $\mathcal{I}_\epsilon^{(\ell)}$. In this case, important information about the latent structure of $X$ will be lost, and the performance of the M-LUND algorithm will correspondingly worsen.  Lemma \ref{lemma: endpoint} guarantees that M-LUND samples all relevant time scales in the diffusion process when extracting cluster structure. We remark that, if there exists a $\delta^{(\ell)}$ near zero, the $\ell$\ths MELD clustering would be easy to find by conventional means; e.g., running $K$-means on the rows of $\mathbf{P}^t$ for $t$ very large. Moreover, the technical assumption of Lemma \ref{lemma: endpoint} is lax (e.g.. if $\lambda_2 = 1-10^{-5}$, $\tau = 10^{-5}$, $\pi_{\min} = 10^{-2}$, and $\epsilon = 10^{-2}$, it holds if $\min_{1\leq\ell\leq M}\delta^{(\ell)}>10^{-8}$).   

The second major assumption, that (\ref{eq: assumption}) holds for each $t$ sampled from $ A_\epsilon$, links the MELD data model and the M-LUND clustering algorithm. Indeed, when this condition holds, it implies that diffusion distances at any sampled $t\in A_\epsilon$ will induce sufficiently strong separation on the clusterings $\mathcal{C}_t~\in \MELD$ that these clusterings can be learned by the M-LUND algorithm. In this sense, the M-LUND clustering algorithm is guaranteed to recover the MELD data model. The condition (\ref{eq: assumption}) is easier to satisfy when the variance of $p$ is low and diffusion maps send clusters of MELD clusterings to coherent, well-separated clusters. For example, if density is uniform and $\Psi_t(x)$ maps each cluster in $\mathcal{C}_t$ to a point mass for $t\in \{0,1,\beta, \dots, \beta^T\}\bigcap A_\epsilon$, the right hand side of (\ref{eq: assumption}) will be 1. In this idealized case, (\ref{eq: assumption}) will be satisfied by any $\epsilon\in\Big(0,\frac{1}{2\sqrt{n}}\Big)$. Conversely, (\ref{eq: assumption}) is more difficult to satisfy when the variance of $p$ is high, or if diffusion distances do not separate cluster structure well. Proposition \ref{prop: M-LUND Labeling} summarizes the recovery of the MELD data model under the usual setup.

\begin{prop} \label{prop: M-LUND Labeling}
Under the usual setup, the M-LUND algorithm extracts a superset of an exponential sampling of $\MELDemph$. 
\end{prop}
\begin{proof}
By Lemma \ref{lemma: endpoint}, $A_\epsilon\subset [0,\beta^T]$.  For each $t\in \big\{0,1,\beta, \dots, \beta^T\}\bigcap A_\epsilon$, $\frac{D_t^{\text{in}}(\mathcal{C}_t)}{D_t^{\text{btw}}(\mathcal{C}_t)}\leq \frac{\epsilon}{1/\sqrt{n}-\epsilon}$ by Corollary \ref{cor:diff_dist_bound}, so the assumptions of Corollary \ref{cor: LUND labeling} are satisfied. Hence, the LUND algorithm perfectly recovers  $\mathcal{C}_t\in\MELD$. This yields a superset of an exponential sampling of $\MELD$.
\end{proof}

In the M-LUND algorithm, the $\epsilon$-stability of a clustering is approximated by exponentially sampling the interval $[0,\beta^T]$. In particular, if a clustering $\{X_k^{(\ell)}\}_{k=1}^{K_\ell}$ is more $\epsilon$-stable,  the interval $\mathcal{I}_\epsilon^{(\ell)}$ will be sampled more frequently. The M-LUND algorithm may obtain a fine-scale perspective of the $\epsilon$-stability of the clusterings in $\MELD$ by decreasing the exponential sampling rate $\beta$. However, this requires implementations of the LUND algorithm at more time steps, increasing computational complexity.  On the other hand, if $\beta$ is too large, the M-LUND algorithm may not sample a MELD clustering of $X$. It is important to understand what choices of $\beta$ are suitable for M-LUND. 

\begin{prop}\label{prop: beta} If $\beta\in \Big(1,\frac{\epsilon}{2\delta^{(\ell^*)}}\Big/\frac{\log(2\kappa^{(\ell^*)}/\epsilon)}{\log(1/|\lambda_{K_{\ell^*}+1}^{(\ell^*)}|)}\Big]$, where $\ell^*= \argmin\limits_{ 1\leq \ell \leq M}\Big[\log_\beta\Big(\frac{\epsilon}{2\delta^{(\ell)}}\Big)-\log_\beta\Big(\frac{\log(2\kappa^{(\ell)}/\epsilon)}{\log(1/|\lambda_{K_\ell+1}^{(\ell)}|)}\Big)\Big]$ and $\epsilon\in\Big(0,\frac{1}{\sqrt{n}}\Big)$, then under the usual setup, the M-LUND algorithm extracts each of the clusterings in $\MELD$ at least once.
\end{prop}
\begin{proof}
By Lemma \ref{lemma: endpoint}, $\mathcal{I}_\epsilon^{(\ell)}\subset [0,\beta^T]$ for each $\ell\in\{1,\dots, M\}$. It therefore suffices to show that there exists a sample $\beta^{k_\ell}\in \mathcal{I}_\epsilon^{(\ell)}$ ($k_\ell\in\{0,\dots,T\}$) for every scale $\ell\in\{1,\dots, M\}$.  
If $\beta\in \Big(1,\frac{\epsilon}{2\delta^{(\ell^*)}}\Big/\frac{\log(2\kappa^{(\ell^*)}/\epsilon)}{\log(1/|\lambda_{K_{\ell^*}+1}^{(\ell^*)}|)}\Big]$, then for each $\ell\in\{1,\dots,M\}$, 
\[1\leq \log_\beta\Big(\frac{\epsilon}{2\delta^{(\ell^*)}}\Big)-\log_\beta\left(\frac{\log(2\kappa^{(\ell^*)}/\epsilon)}{\log(1/|\lambda_{K_{\ell^*}+1}^{(\ell^*)}|)}\right)\leq \log_\beta\Big(\frac{\epsilon}{2\delta^{(\ell)}}\Big)-\log_\beta\left(\frac{\log(2\kappa^{(\ell)}/\epsilon)}{\log(1/|\lambda_{K_\ell+1}^{(\ell)}|)}\right).\] 
Thus, for each $\ell\in \{1,\dots, M\}$, there is a $k_\ell\in\{0,\dots, T\}$ such that $\log_\beta\Big(\frac{\log(2\kappa^{(\ell)}/\epsilon)}{\log(1/|\lambda_{K_\ell+1}^{(\ell)}|)}\Big)~\leq~k_\ell \leq~\log_\beta\Big(\frac{\epsilon}{2\delta^{(\ell)}}\Big)$, implying $\beta^{k_\ell}\in\mathcal{I}_\epsilon^{(\ell)}$.  
\end{proof}

Proposition \ref{prop: beta} illustrates that there is a tension between finding a $\beta$ that will sample all intervals $\mathcal{I}_\epsilon^{(\ell)}$ and satisfying (\ref{eq: assumption}) for all time steps $t\in A_\epsilon\bigcap \{0,1,\beta, \dots, T\}$. If $\epsilon$ is large, then there will be a wide range of exponential sampling rates $\beta$ that can be used to sample all intervals $\mathcal{I}_\epsilon^{(\ell)}$. However, if $\epsilon$ is too large, $\epsilon$-separation by diffusion distances might not guarantee strong enough separation of clusters to satisfy (\ref{eq: assumption}) at all sampled time steps. On the other hand, if $\epsilon$ is small, then (\ref{eq: assumption}) is easier to satisfy because of strong $\epsilon$-separation by diffusion distances. However, because the intervals $\mathcal{I}_\epsilon^{(\ell)}$ shrink as $\epsilon$ becomes smaller, $\beta$ must be decreased to guarantee that the M-LUND algorithm samples each interval $\mathcal{I}_\epsilon^{(\ell)}$.  Proposition \ref{prop: beta} also illustrates how $\epsilon$-stability affects the range of suitable choices of $\beta$. For fixed $\epsilon\in \Big(0,\frac{1}{\sqrt{n}}\Big)$, if each interval $\mathcal{I}_\epsilon^{(\ell)}$ is large on a logarithmic scale, $\beta$ can be chosen to be large and the M-LUND algorithm will still recover all clusterings in $\MELD$. On the other hand, if one of the clusterings $\{X_k^{(\ell)}\}_{k=1}^{K_\ell}$ in $\MELD$ is unstable so that $\mathcal{I}_\epsilon^{(\ell)}$ is small on a logarithmic scale, $\beta$ must be decreased to guarantee that M-LUND samples $\mathcal{I}_\epsilon^{(\ell)}$.

Because the intervals $\mathcal{I}_\epsilon^{(\ell)}$ are not known a priori, the entire time domain $[0,\beta^T]$ must be sampled to learn the clusterings in the MELD data model. Thus, it is possible that the minimizer of total VI is not within $\MELD$ and is instead a clustering obtained during a transition region, i.e. intervals of time during which the transition matrix is rapidly mixing, and there is no ``true'' latent clustering. Because no latent clustering exists during transition regions, the VI between a clustering sampled during a transition region and a MELD clustering is expected to be high.  Proposition \ref{prop: transitionregion} provides a lax technical assumption that guarantees a MELD clustering is the minimizer of total VI. 

\begin{prop} \label{prop: transitionregion}
Assume the usual setup, and let  $B_\epsilon = [0,\beta^T] \setminus A_\epsilon$ be the transition regions between clusterings in $\MELD$. If there is a $t~\in~J\bigcap A_\epsilon$ such that for any $r~\in~J\bigcap B_\epsilon$, $$\frac{1}{|J|}\sum_{s\in J} \big[I(\mathcal{C}_r, \mathcal{C}_s) - I(\mathcal{C}_t, \mathcal{C}_s)\big]~<~\frac{1}{2}\big[ H(\mathcal{C}_r)-~H(\mathcal{C}_t)\big],$$ then the M-LUND algorithm outputs a clustering from $\MELDemph$ as the minimizer of total VI. 
\end{prop}
\begin{proof}
 By Lemma \ref{lemma: endpoint}, $B_\epsilon$ is well-defined. Moreover, by Corollary \ref{cor: LUND labeling}, at each time step $t\in\{0,1,\beta, \dots, \beta^T\}\bigcap A_\epsilon$, the LUND algorithm extracts the latent clustering at time $t$: $\mathcal{C}_t\in\MELD$. It suffices to show that the total VI will be lower for a clustering sampled during $A_\epsilon$ than for any sampled during $B_\epsilon$. By the stated assumption, there is a $t\in J\bigcap A_\epsilon$ such that for any $r\in J \bigcap B_\epsilon$,
\begin{alignat*}{2}
 & & \quad \frac{1}{|J|}\sum_{s\in J} \big[I(\mathcal{C}_r, \mathcal{C}_s) - I(\mathcal{C}_t, \mathcal{C}_s)\big] &< \frac{1}{2}\big[ H(\mathcal{C}_r)-H(\mathcal{C}_t)\big]\\ 
&\iff & \quad |J|H(\mathcal{C}_t) - 2\sum_{s\in J}I(\mathcal{C}_t, \mathcal{C}_s) &< |J|H(\mathcal{C}_r) - 2\sum_{s\in J}I(\mathcal{C}_r, \mathcal{C}_s)\\
&\iff &\quad    \sum_{s\in J} \Big[H(C_t) + H(C_s) - 2 I(C_t,C_s)\Big] &<  \sum_{s\in J} \Big[ H(\mathcal{C}_r) + H(\mathcal{C}_s) - 2 I(\mathcal{C}_r, \mathcal{C}_s)\Big]\\
&\iff &\quad \sum_{s\in J} VI(\mathcal{C}_t, \mathcal{C}_s) &<  \sum_{s\in J} VI(\mathcal{C}_r, \mathcal{C}_s),
\end{alignat*}
Since the total VI of $\mathcal{C}_t$ is less than that of $\mathcal{C}_r$ where $r\in B_\epsilon$, the minimizer of total VI must be sampled during $A_\epsilon$.
\end{proof}

Proposition \ref{prop: transitionregion} relies on a technical assumption on the entropy of and mutual information between nontrivial clusterings extracted by the LUND algorithm. The quantity
 $\frac{1}{|J|}\sum_{s\in J} \big[I(\mathcal{C}_r, \mathcal{C}_s) - I(\mathcal{C}_t, \mathcal{C}_s)\big]$ is the average difference in mutual information encoded in $\mathcal{C}_r$ and $\mathcal{C}_t$, where the average is across all nontrivial extracted clusterings $\mathcal{C}_s$.  The assumption of Proposition \ref{prop: transitionregion} is easier to satisfy if this quantity is small; i.e., if a clustering in $\MELD$ stores more information about the latent structure in $X$ than the clusterings sampled during transition regions. On the other hand, the quantity $\frac{1}{2}\big[ H(\mathcal{C}_r)-H(\mathcal{C}_t)\big]$ is half the difference in entropy between $\mathcal{C}_r$ and $\mathcal{C}_t$.  The assumption of Proposition \ref{prop: transitionregion} is easier to satisfy if this quantity is large. The entropy of a clustering is maximal if it consists of $n$ singleton clusters, so the constraint on the entropy of $\mathcal{C}_t$ can be viewed as regularization: downweighting complicated clusterings that may not actually correspond to meaningful structure. Thus, a simple partition that shares high levels of mutual information with the other nontrivial extracted clusterings of $X$ tends to satisfy the assumption of Proposition \ref{prop: transitionregion}.

\subsubsection{Diffusion Near Equilibrium}\label{sec:tau}

In this section, we will justify the termination of the first for-loop of the M-LUND algorithm at time $t=\beta^T$. If $|\lambda_2|>|\lambda_3|$, then for any $\eta>0$, there exists $t$ such that 
$\max_{x,y\in X}\left|D_t(x,y) - |\lambda_2|^{t}|\psi_2(x)-\psi_2(y)| \right|\leq \eta .$
This leads the LUND algorithm to continue to label the clustering generated by the second eigenfunction of $\mathbf{P}$ until diffusion distances are numerically zero. However, the persistence of that clustering may not reflect its stability, as the diffusion process will have effectively arrived at its stationary distribution. For this reason, cluster analysis is terminated once diffusion is sufficiently close to stationarity in the M-LUND scheme. The following quantity will prove useful in measuring how close the diffusion process is to its stationary distribution:

\begin{definition} \label{def:pointwise distance} 
Let $\mathbf{P}$ be a reversible, irreducible, and aperiodic transition matrix of a Markov chain on state space $X$ with stationary distribution $\pi$. The \emph{relative pointwise distance} of $\mathbf{P}^t$ to $\pi$ at time $t$ is $\Delta(t) = \max_{1\leq i,j\leq n} |(P^t)_{ij}-\pi_j|/\pi_j$.
\end{definition}

It is known that $\Delta(t) \leq |\lambda_2(\mathbf{P})|^t/\pi_{\min}$ ~\cite{jerrum1989approximating, sinclair1989Markov}.  This yields a uniform bound on $D_{t}$.
\begin{prop}
    For any $\tau\in(0,1)$ and $x,y\in X$, if $t> \log_{|\lambda_2|}\big(\frac{\tau\pi_{\min}}{2}\big)$, then $D_t(x,y) < \tau$. \label{prop:Tau} 
\end{prop}
\begin{proof}
Let $\epsilon>0$ and $x,y\in X$ be given. By the definition of diffusion distances,
    \begin{align*}
        D_t(x,y) &= \|p_t(x,:) - p_t(y,:)\|_{\ell^2(1/\pi)}\\
        &\leq \|p_t(x,:) - \pi\|_{\ell^2(1/\pi)}+\| p_t(y,:) - \pi\|_{\ell^2(1/\pi)}\\
        &\leq 2\sqrt{\sum_{u\in X} \max_{z\in X} \frac{|p_t(z,u)-\pi(u)|^2}{\pi(u)^2}\pi(u) }\\
        &\leq 2\Delta(t) \sqrt{\sum_{u\in X}\pi(u)}\\
        &= 2\Delta(t)\\
        &\leq 2|\lambda_2|^t/\pi_{\min} ,
    \end{align*}
   Thus, if $t> \log_{|\lambda_2|}\big(\frac{\tau\pi_{\min}}{2}\big)$, then  $D_t(x,y)<\tau$.  
\end{proof}

The value $\log_{|\lambda_2|}\big(\frac{\tau\pi_{\min}}{2}\big)$ is determined by quantities pertaining to the original graph and how diffusion spreads on it. In a graph with coherent components that have few edges between each other, $|\lambda_2|$ is close to 0. Hence, coherent cluster structure in the dataset indicates that a longer time horizon is needed. Similarly, a smaller $\tau$ indicates that more time is needed before the threshold for stationarity is met. One can interpret the dependence of $T$ on $\pi_{\min}$ as capturing the fact that more time is needed for diffusion to reach points of lower degree. 

\subsection{Computational Complexity}
\label{sec: complexity}

We will now analyze the computational complexity of the M-LUND algorithm, which is essentially linear when nearest neighbor searches are performed using the \emph{cover tree} indexing structure~\cite{beygelzimer2006covertrees, murphy2019LUND}. Often, high-dimensional datasets $X\subset \R^D$ lie on or near intrinsically low-dimensional sets (e.g. subspaces or manifolds).  The \emph{doubling dimension} of $X$ quantifies this notion of latent low-dimensionality~\cite{beygelzimer2006covertrees}. Let $c>0$ be the minimum value such that any ball $B(p,r) = \{q\in X \ |\; \|p-q\|_2\leq r\}$ can be covered by $c$ balls of half the radius. The doubling dimension of $X$ is defined to be $d = \log_{2} c$.  Note that a uniform sample on a $d$-dimensional manifold has covering dimension $d$.  If $X\subset\mathbb{R}^{D}$ has covering dimension $d$, the calculation of all $N$ nearest neighbors for $X$ using cover trees has a computational complexity of $O(N D C^d n \log(n) )$\cite{beygelzimer2006covertrees} with $C$ a constant independent of $n,N,d,D$.

\begin{theorem}\label{thm: complexity}
    Let $d$ be the doubling dimension of $X$. Suppose $\mathbf{P}$ is built using a KNN graph with $O(\log(n))$ nearest neighbors, cover trees are used for nearest neighbor searches, and $O(1)$ eigenfunctions are used to compute diffusion distances. If  $T = \big\lceil\log_\beta\big[\log_{|\lambda_2|}\big(\frac{\tau \pi_{\min}}{2}\big)\big]\big\rceil$, the complexity of the M-LUND algorithm is  $O(TDC^d n\log^2(n)  + T^2 n\log(n))$ with $C$ a constant independent of $n,N,d,D$.\end{theorem}
\begin{proof}
Under the stated assumptions, a single run of the LUND algorithm has complexity $O(DC^d \log(n)^2n)$ \cite{murphy2019LUND, beygelzimer2006covertrees}.  Thus, the first for-loop in the M-LUND algorithm has complexity $O(T DC^dn\log^2(n))$. Computing $VI(\mathcal{C}_t,\mathcal{C}_s)$ costs $O(n\log(n))$ operations~\cite{meilua2007VI}. Since $|J|\leq T+2$, the complexity of the second for-loop in the M-LUND algorithm is $O(T^2 n\log(n) )$. Combining these two results, the overall complexity is 
$O(TDC^dn\log(n)^2~+~T^2n\log(n))$.  
\end{proof}

 Note that if $\beta$ is replaced with $\beta^{1/m}$, the complexity of the first for-loop increases by a factor of $m$, and the complexity of the second increases by a factor of $m^2$. This is because a finer sampling frequency is used; i.e., the LUND algorithm must be evaluated more frequently. Similarly, $\tau$ indicates how close to stationarity $\mathbf{P}^t$ is required to be before terminating cluster analysis. So, if $\tau$ is decreased, the M-LUND algorithm's complexity will increase. More precisely, if $\tau$ is replaced by $\tau^{m}$, the value of $T$ will increase slightly to $\Big\lceil\log_\beta\big[\log_{|\lambda_2|}\big(\frac{ \pi_{\min}}{2}\big) + m\log_{|\lambda_2|}\tau\big]\Big\rceil$.  

 We expect that $T=O(1)$ with respect to $n$ because  $T$ reflects the length of the interval for which diffusion distances remain bounded away from $\tau$.  If $T = O(1)$ with respect to $n$, the following simplification holds:

\begin{corollary}\label{cor: complexity}
Under the assumptions of Theorem \ref{thm: complexity}, if $T =O(1)$ with respect to $n$, M-LUND has complexity $O(DC^d \log(n)^2n)$.
\end{corollary}

Thus, no further complexity (with respect to $n$) is added to a single implementation of the LUND algorithm in its multiscale extension. Importantly, because the $T+2$ implementations of the LUND algorithm are independent of each other, the dominant for-loop in which the LUND clusterings are computed is embarrassingly parallelizable, as is computing total VI in the second dominant for-loop.

\subsection{Comparisons With Other Multiscale Clustering Algorithms}

In this section, we compare the M-LUND algorithm to related hierarchical and multiscale clustering schemes. 

 \subsubsection{Comparison with Dendrogram-Based Hierarchical Clustering Algorithms} 
 
In Section \ref{sec:HC 2}, we described classical hierarchical clustering algorithms, which extract a family of partitions of the dataset using a linkage function and express them as a dendrogram. Most linkage functions use Euclidean distances to compare clusters, causing poor performance on outliers. Conversely, because LUND relies on the function $\mathcal{D}_t(x) = p(x)\rho_t(x)$ to compute modes, it downweights low-density points that are high in diffusion distance from their $D_t$-nearest neighbor. Thus, the M-LUND algorithm is able to capture latent multiscale structure while remaining robust to outliers. Moreover, linkage-based clustering algorithms are typically greedy, optimizing for the best split at each iteration. This makes the output of these algorithms prone to small perturbations in the data. Conversely, all scales of clusterings extracted by the M-LUND algorithm arise from the same graph, making it more robust to minor variations in the data.

 \subsubsection{Comparison with Hierarchical Spectral Clustering} 

 SC has the disadvantage of requiring a priori knowledge of the number of clusters $K$.  However, $K$ may be estimated from $\mathbf{P}$ using the number of eigenvalues close to 1~\cite{vonluxburg2007spectralclustering}. A similar fact is true of $\mathbf{P}^t$: the number of eigenvalues with $|\lambda_i|^t$ near 1 may be descriptive of the number of latent clusters at time $t$~\cite{azran2006spectralclustering}. Hierarchical Spectral Clustering (HSC) leverages this property of $\mathbf{P}$ in a multiscale adaptation of classical SC~\cite{azran2006spectralclustering}. Define $\mathcal{S}_t=\{\lambda_{k}^t-\lambda_{k+1}^t\}_{k=1}^{n-1}$. If each of the $K$ clusters in a latent clustering of $X$ is a complete graph of equal size and if the effective rank of $\mathbf{P}^t$ is $K$, then the first $K-1$ entries of $\mathcal{S}_t$ will be small because $\lambda_1^t\approx \lambda_2^t \approx \dots\approx  \lambda_K^t\approx 1$. Similarly, the last $n-K-1$ entries of $\mathcal{S}_t$ will be small because $\lambda_{K+1}^t\approx \lambda_{K+2}^t \approx \dots,\lambda_n^t\approx 0$. Thus, $\mathcal{S}_t$ is expected to be a sequence of nearly-zero numbers in all but the $K$\ths entry. The quantity $ \lambda_K^t-\lambda_{K+1}^t$ is called the \emph{eigengap at time $t$}~\cite{vonluxburg2007spectralclustering, azran2006spectralclustering}.  As $t$ varies, $K_t = \text{argmin}_{k}(\lambda_{k}^t-\lambda_{k+1}^t)$ varies as well, so $t$ can be interpreted as a scaling parameter~\cite{azran2006spectralclustering}. HSC finds the $K_t$ corresponding to local maxima of $\Delta_t = \displaystyle\max_{1\leq k\leq n-1}\{\lambda_{k}^t-\lambda_{k+1}^t\}$ and uses these as inputs for the SC algorithm (Section \ref{sec: Spectral_Clustering}). HSC is provided in Algorithm \ref{alg: Multiscale_Spectral_Clustering}. 
 
 \begin{figure}[t]
\begin{algorithm}[H]
\SetAlgoLined
\caption{Hierarchical Spectral Clustering (HSC)~\cite{azran2006spectralclustering}}
\label{alg: Multiscale_Spectral_Clustering}
\KwIn{ $X$ (dataset), $\sigma$ (diffusion scale), $T_{\max}$ (maximum time step)}
\KwOut{$M$ partitions of $X$ with stability measure and eigengap $\{(\mathcal{C}^{(\ell)}, \alpha_\ell, \beta_\ell)\}_{\ell=1}^M$}
Construct transition matrix $\mathbf{P}$ with a Gaussian kernel and diffusion scale $\sigma$\;
Calculate the eigenvalues of $\mathbf{P}$: $\{\lambda_i\}_{k=1}^n$\;
For $t\in \{1,2, \dots, T_{\max}\}$, compute $\Delta_t = \max_{1\leq k \leq n-1} |\lambda_k^t - \lambda_{k+1}^t|$ and $K_t = \text{argmax}_{1\leq k \leq n-1} |\lambda_k^t - \lambda_{k+1}^t|$\;
Find the $M$ local maxima of $\Delta_t$ and denote them $\Delta_{t_1}, \Delta_{t_2} \dots, \Delta_{t_M}$. Set $t_0=0$.\;
\For{$\ell = 1:M$}{
$\mathcal{C}^{(\ell)} = \text{SC}[X,\sigma, K_{t_\ell}]$\;
Store $(\mathcal{C}^{(\ell)}, \alpha_\ell, \beta_\ell )$, where $\alpha_\ell = (t_\ell - t_{\ell-1})/T_{\max}$ and $\beta_\ell = \Delta_{t_\ell}$\;
}
\end{algorithm}
\end{figure}

There are similarities between M-LUND and HSC. For example, both algorithms rely on a Markov diffusion process to extract multiscale cluster structure. However, there are some key differences between them, the most important being that HSC does not directly incorporate density into predictions.  SC exhibits fundamental limitations on datasets with clusters that are not of uniform density and scale~\cite{nadler2007failures}; these limitations persist in the multiscale implementation of SC. In contrast, the M-LUND algorithm has performance guarantees for recovering the correct clusterings on datasets of varying scale and density.

Another difference between the M-LUND algorithm and HSC is the latter algorithm's reliance on the eigengap to estimate the number of clusters at time $t$. While the eigengap is effective at uncovering the effective rank of $\mathbf{P}^t$ in some idealized cases, it may fail when Euclidean distances are used to extract $\mathbf{P}$ from a dataset that does not consist of well-separated spherical clusters~\cite{Little2020_Path, arias2011clustering}. Conversely, there is strong empirical and theoretical evidence to support the use of the function $\mathcal{D}_t(x)$ to measure the number of latent clusters at time $t$~\cite{murphy2019LUND}.

\subsubsection{Comparison with Multiscale Markov Stability Clustering} \label{sec:mms}

Another diffusion-based approach to multiscale community detection uses the \emph{Markov stability} of a clustering at time $t$ as a quality measure for multiscale clustering~\cite{lambiotte2008laplacian, lambiotte2014random, liu2020MarkovStability}. Markov stability is derived from the autocovariance matrix $\textbf{B}(t)=\Pi \mathbf{P}^t-\pi^\top\pi$,  where $\Pi$ is the diagonal matrix with $\Pi_{ii} = \pi_i$~\cite{lambiotte2008laplacian, lambiotte2014random}. The quantity $\textbf{B}(t)$ reflects the probability of a random walk beginning in a cluster $X_k$ and being in that cluster after $t$ steps, minus the probability that two independent random walks end in $X_k$, evaluated at stationarity~\cite{liu2020MarkovStability}. Thus, $\textbf{B}(t)$ is expected to be large for stable clusterings and nearly zero for intermediate, less stable clusterings. Nevertheless, to our knowledge, this relationship is more an intuition than a theoretical result. The Markov stability of a clustering $\mathcal{C}=\{X_k\}_{k=1}^K$ of $X$ is defined to be $r(t,\mathcal{C}) = \sum_{k=1}^K\; \sum_{x_i,x_j\in X_k} \textbf{B}(t)_{ij}$~\cite{lambiotte2008laplacian, lambiotte2014random, liu2020MarkovStability}. The Markov stability $r(t,\mathcal{C})$ will be large if a $t$-step random walk is likely to terminate in the cluster in which it began, and it is likely to be small if a between-cluster transition is likely~\cite{liu2020MarkovStability}. 

The Multiscale Markov Stability (MMS) clustering algorithm optimizes $r(t,\mathcal{C})$ across partitions $\mathcal{C}$ using a modified Louvain algorithm: $\mathcal{C}_t = \text{argmax}\big\{r(t,Z) \ \big| \ \text{$Z$ is a partition of $X$}\big\}$~\cite{liu2020MarkovStability, blondel2008optimization}.  This optimization is performed across across an exponential sampling of the diffusion process to learn multiscale structure. For each pair of times $s$ and $t$, $VI(\mathcal{C}_s, \mathcal{C}_{t})$ is calculated~\cite{meilua2007VI, liu2020MarkovStability}. If $\mathcal{C}_t$ is stable in the diffusion process, it is likely to be close in VI to other extracted clusterings~\cite{liu2020MarkovStability}. The authors therefore look for large diagonal blocks with small values in the $VI(\mathcal{C}_s, \mathcal{C}_{t})$ matrix. 

The M-LUND clustering algorithm is similar in some crucial ways to the MMS algorithm~\cite{liu2020MarkovStability}. Both algorithms rely on a Markov diffusion process to indicate scale~\cite{murphy2019LUND, liu2020MarkovStability}. Similarly, both algorithms use the VI between clusterings to determine which partition is most representative of the dataset on a whole~\cite{meilua2007VI, liu2020MarkovStability}. The main difference between the M-LUND and MMS algorithms is how clusterings are derived and the theoretical guarantees both clustering algorithms provide. The M-LUND algorithm uses the LUND algorithm to extract $K_t$ and $\mathcal{C}_t$ at times $t\geq0$. In Corollary \ref{cor: LUND labeling}, we showed that under reasonable assumptions on cluster density and diffusion at time $t$, the LUND algorithm will perfectly recover $K_t$ and $\mathcal{C}_t$. While the Markov stability-based clustering has been shown perform well on many benchmark datasets, it does not enjoy similar theoretical backing.

\section{Numerical Experiments} \label{sec: Numerical 5}

 In this section, we illustrate the performance of the M-LUND algorithm.  We compute a number of statistics on its performance on three synthetic datasets (overlapping Gaussians, concentric rings, and data with bottlenecks), eleven real, benchmark datasets, and the Salinas A HSI. For synthetic datasets, we implemented the M-LUND algorithm on 100 samples of the latent distribution and provide detailed analysis of its performance on a representative sample. Weight matrices were calculated using a Gaussian kernel with diffusion scale $\sigma>0$. KDEs were computed (as described as in Section \ref{sec: LUND 2}) with KDE bandwidth $\sigma_0$ and $N$ $\ell^2$-nearest neighbors. Diffusion maps were truncated to only include the first 10 eigenfunctions. Because the transition matrices analyzed in this section were approximately low rank, this resulted in a much-reduced computational complexity for M-LUND while retaining high levels of accuracy in diffusion distance computations. For all numerical experiments, we used a stationarity threshold of $\tau=~10^{-5}$ and sampling rate $\beta=2$, but we used different choices of $N$, $\sigma$, and $\sigma_0$ for different datasets.

We computed the stochastic complements of $\mathbf{P}$ with respect to the clusterings extracted using the LUND algorithm in order to measure the geometric constants $\lambda_{K_\ell+1}^{(\ell)}$, $\delta^{(\ell)}$, and $\kappa^{(\ell)}$ and the intervals $\mathcal{I}_\epsilon^{(\ell)}$. In the fifth column of the figures in this section, the lower and upper limits of the interval $\mathcal{I}_\epsilon^{(\ell)}$ are plotted as a function of $\epsilon$. The blue curve corresponds to the lower limit of $\mathcal{I}_\epsilon^{(\ell)}$, while the red curve corresponds to its upper limit. If, for a fixed $\epsilon>0$, the value taken by the blue curve is greater than the value taken by the red, then $\mathcal{I}_\epsilon^{(\ell)}=\emptyset$ and the clustering is not guaranteed to be $\epsilon$-separable by diffusion distances at any time in the diffusion process. On the other hand, if the red curve takes a greater value than the blue for some $\epsilon\in\Big(0,\frac{1}{\sqrt{n}}\Big)$, a range of time exists during which the clustering can be $\epsilon$-separable by diffusion distances. Hence, from the clusterings of $X$ extracted by the M-LUND algorithm, we recovered the MELD data model of $X$. We remark that the condition $t\in\mathcal{I}_\epsilon^{(\ell)}$ for $\epsilon\in\Big(0,\frac{1}{\sqrt{n}}\Big)$ is sufficient for the $\ell$\ths MELD clustering to be $\epsilon$-separable by diffusion distances at time $t$ but not necessary. Thus, even if $\mathcal{I}_\epsilon^{(\ell)}$ is empty, it is possible for the clustering $\{X_k^{(\ell)}\}_{k=1}^{K_\ell}$ to be $\epsilon$-separable by diffusion distances. 

\begin{figure}[t] 
\begin{minipage}{\textwidth}
    \centering
    \begin{subfigure}[t]{0.2\textwidth}
        \centering
        \includegraphics[height = 1.1in]{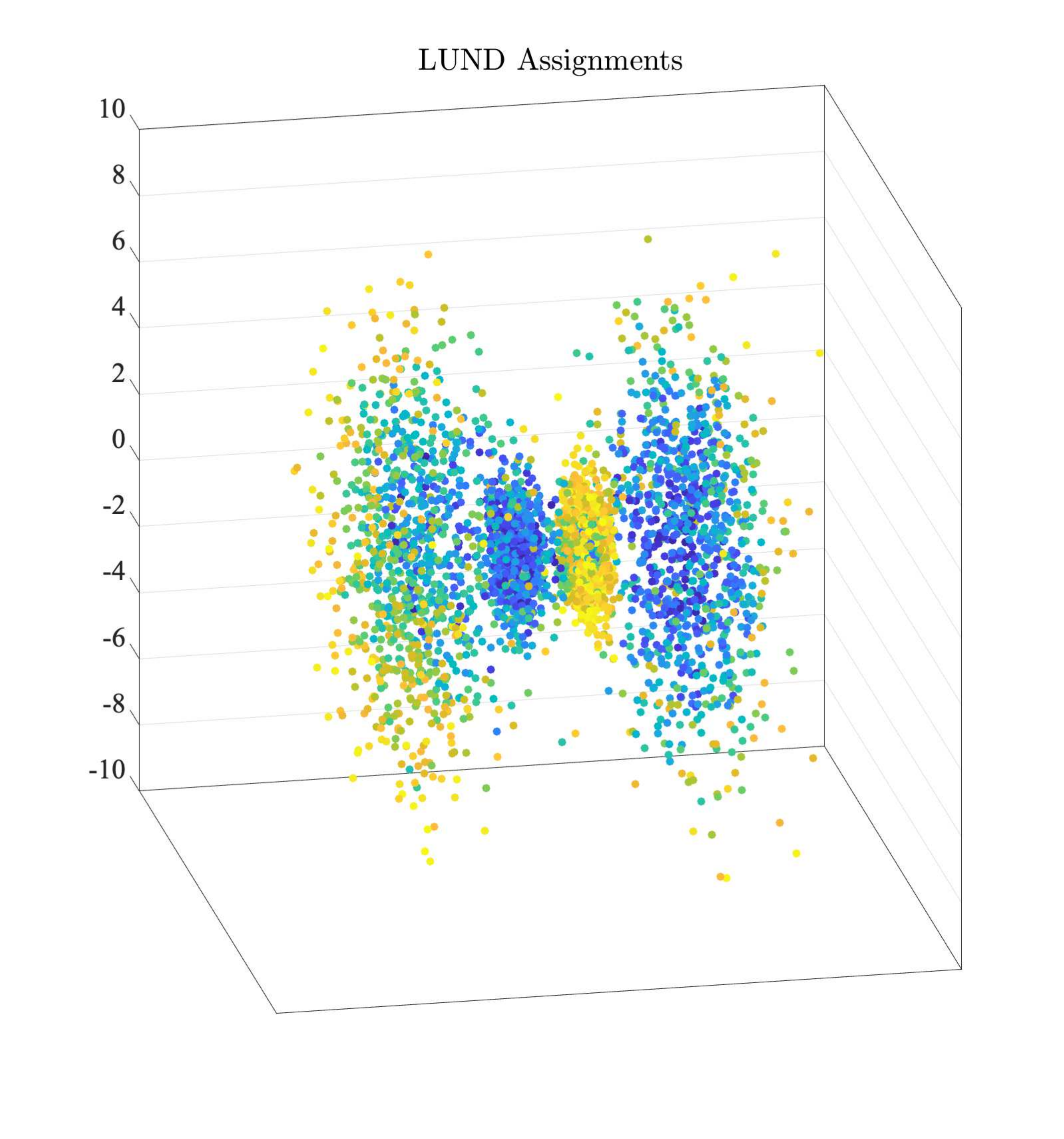}
    \end{subfigure}%
    \begin{subfigure}[t]{0.2\textwidth}
        \centering
        \includegraphics[height = 1.1in]{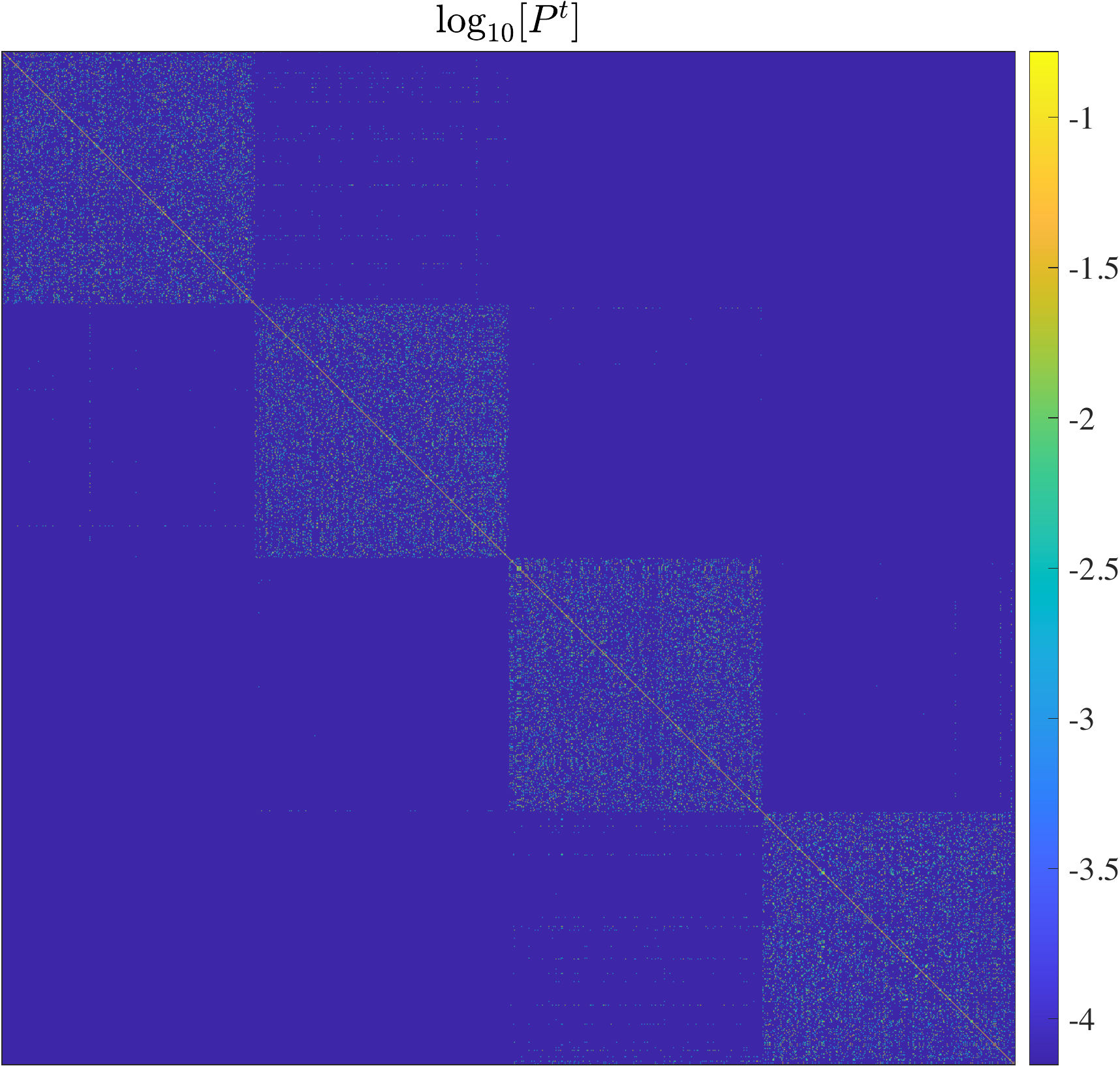}
    \end{subfigure}%
    \begin{subfigure}[t]{0.2\textwidth}
        \centering
        \includegraphics[height = 1.1in]{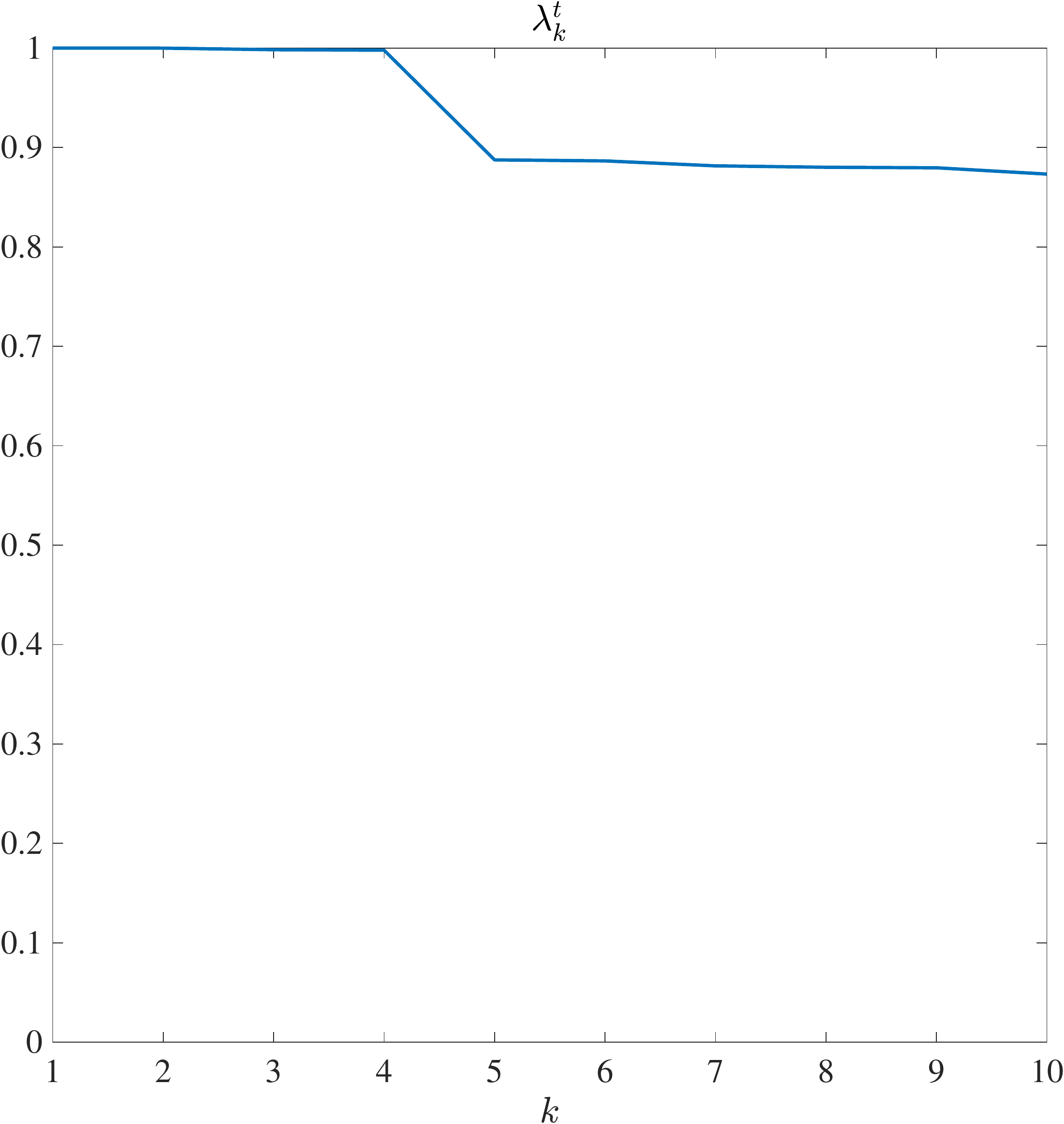}
    \end{subfigure}%
    \begin{subfigure}[t]{0.2\textwidth}
        \centering
        \includegraphics[height = 1.1in]{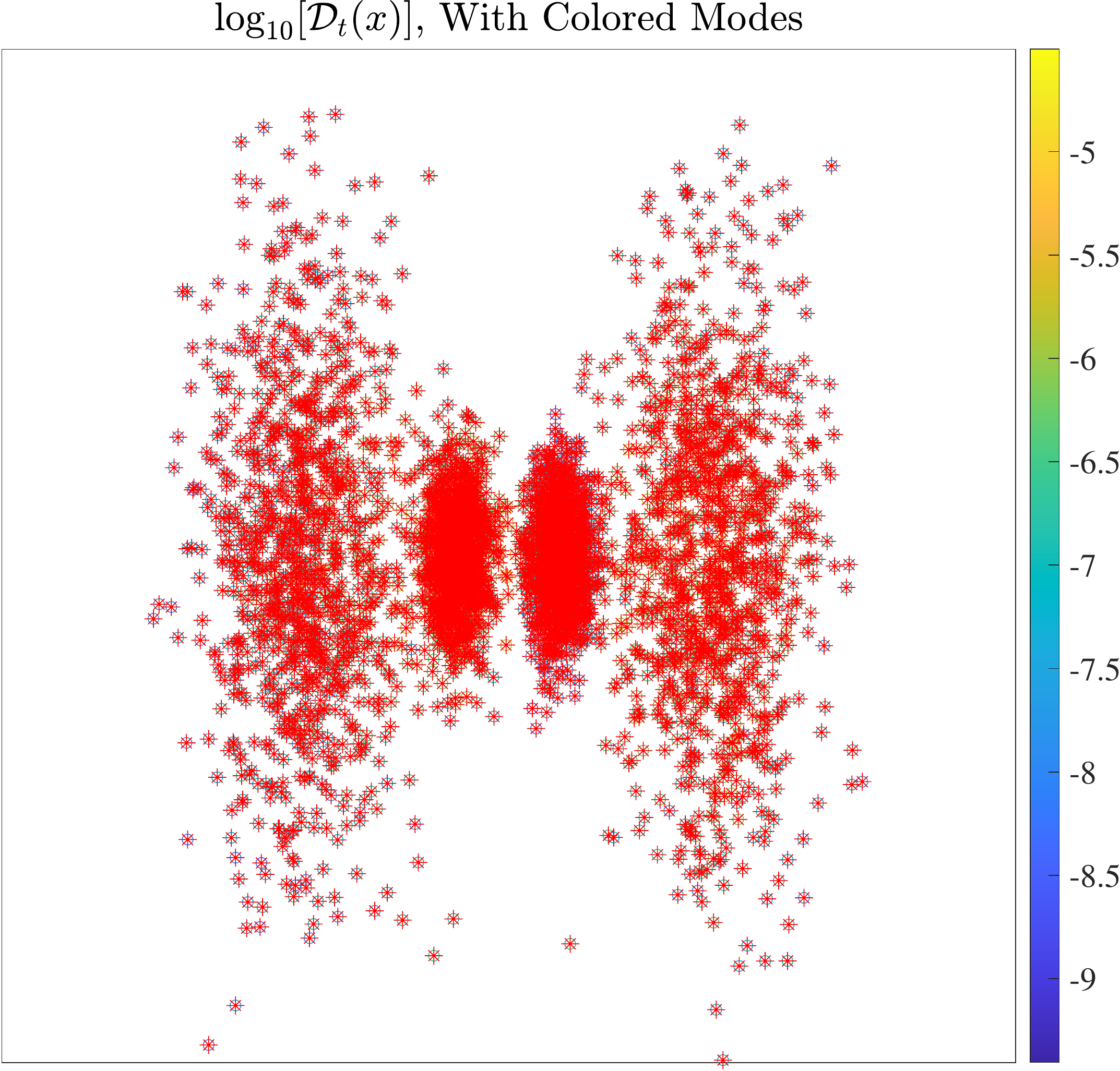}
    \end{subfigure}%
    \begin{subfigure}[t]{0.2\textwidth}
        \centering
        \includegraphics[height = 1.1in]{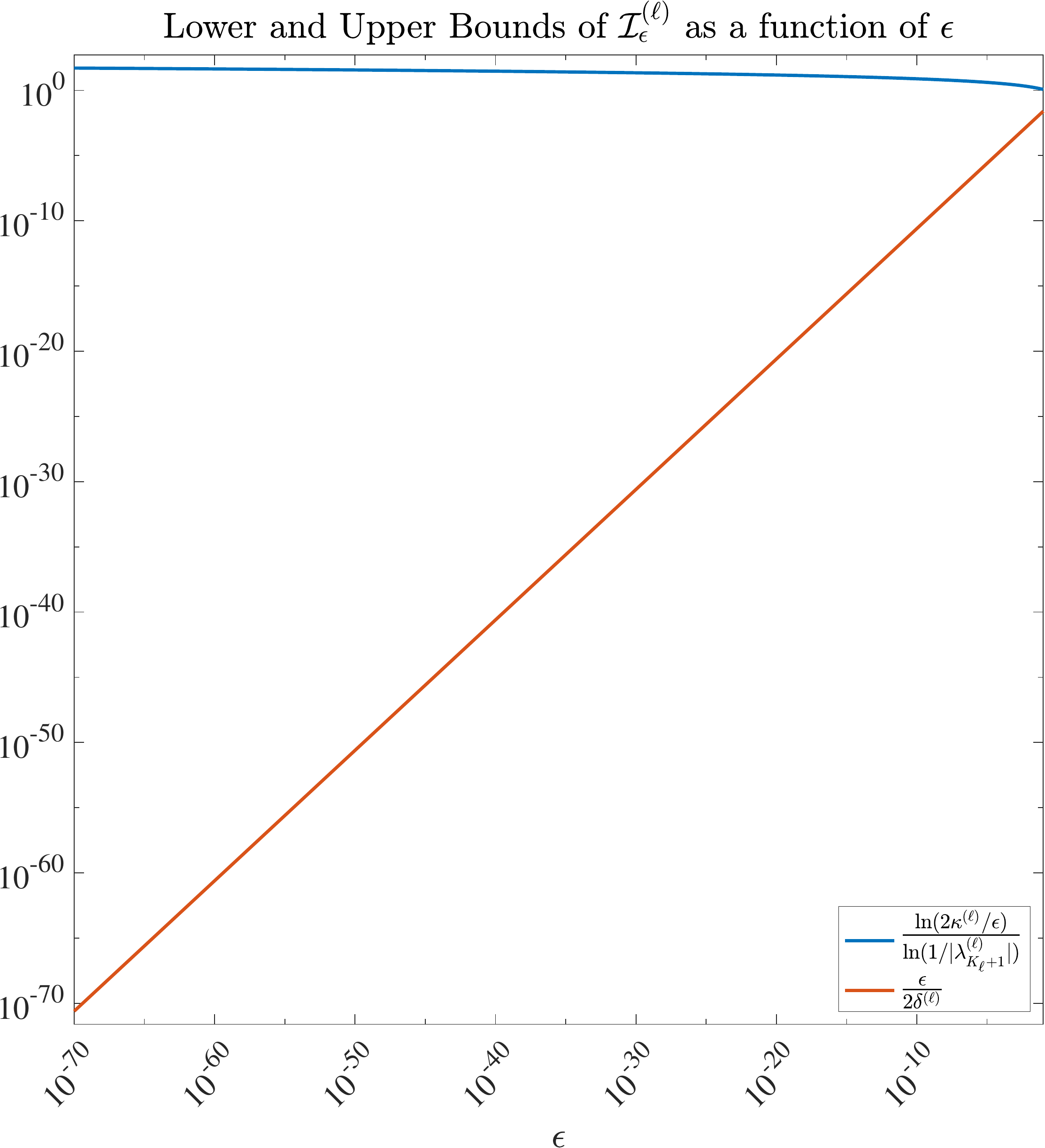}
    \end{subfigure}%
    \subcaption{LUND assignments, transition matrix, spectrum, $\mathcal{D}_t(x)$, and interval bounds for extracted clustering at time $t=2^1$.  3998 clusters.}
  \label{fig:gaussian1}\par\vspace{0.25in}
    \begin{subfigure}[t]{0.2\textwidth}
        \centering
        \includegraphics[height = 1.1in]{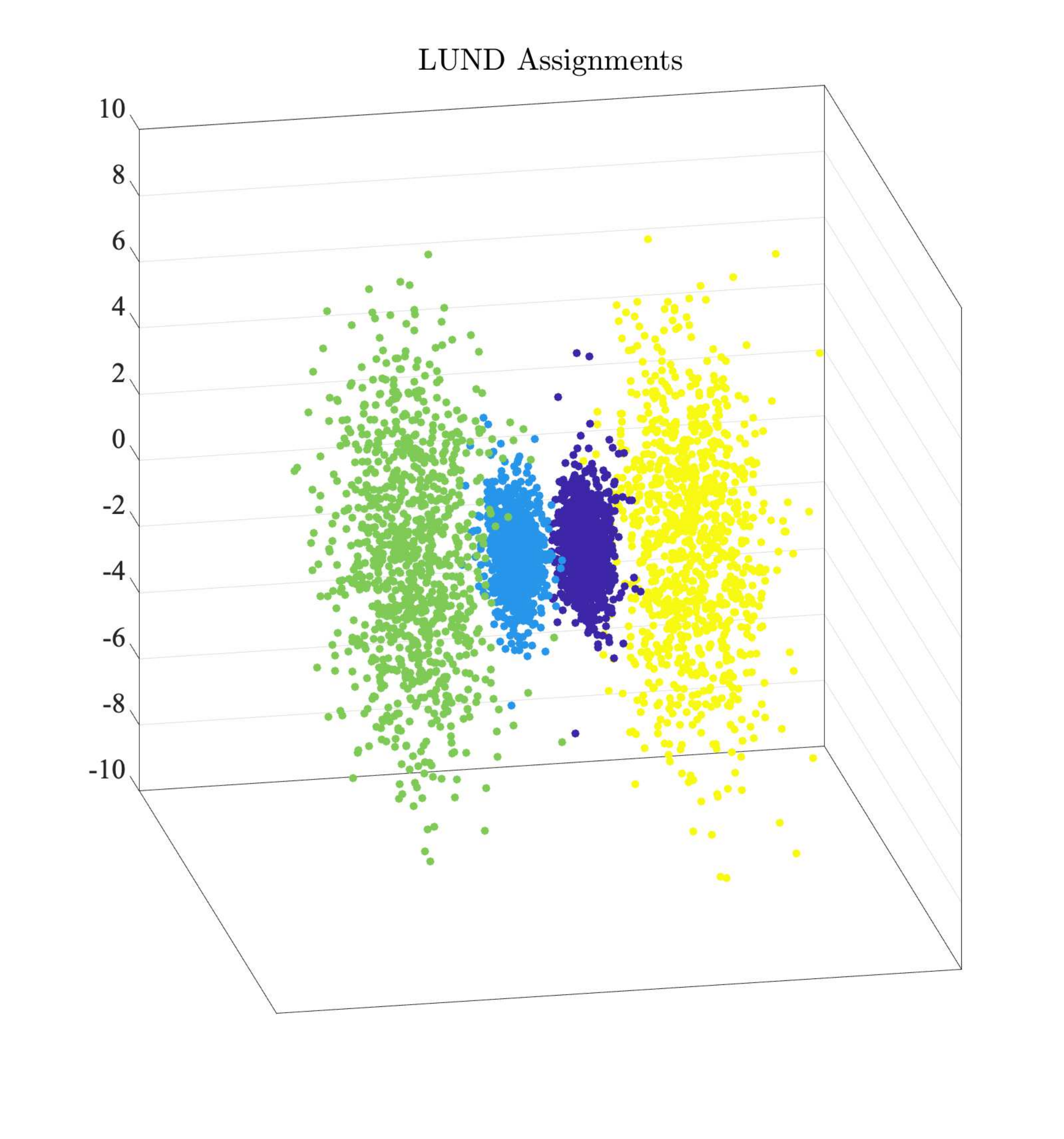}
    \end{subfigure}%
    \begin{subfigure}[t]{0.2\textwidth}
        \centering
        \includegraphics[height = 1.1in]{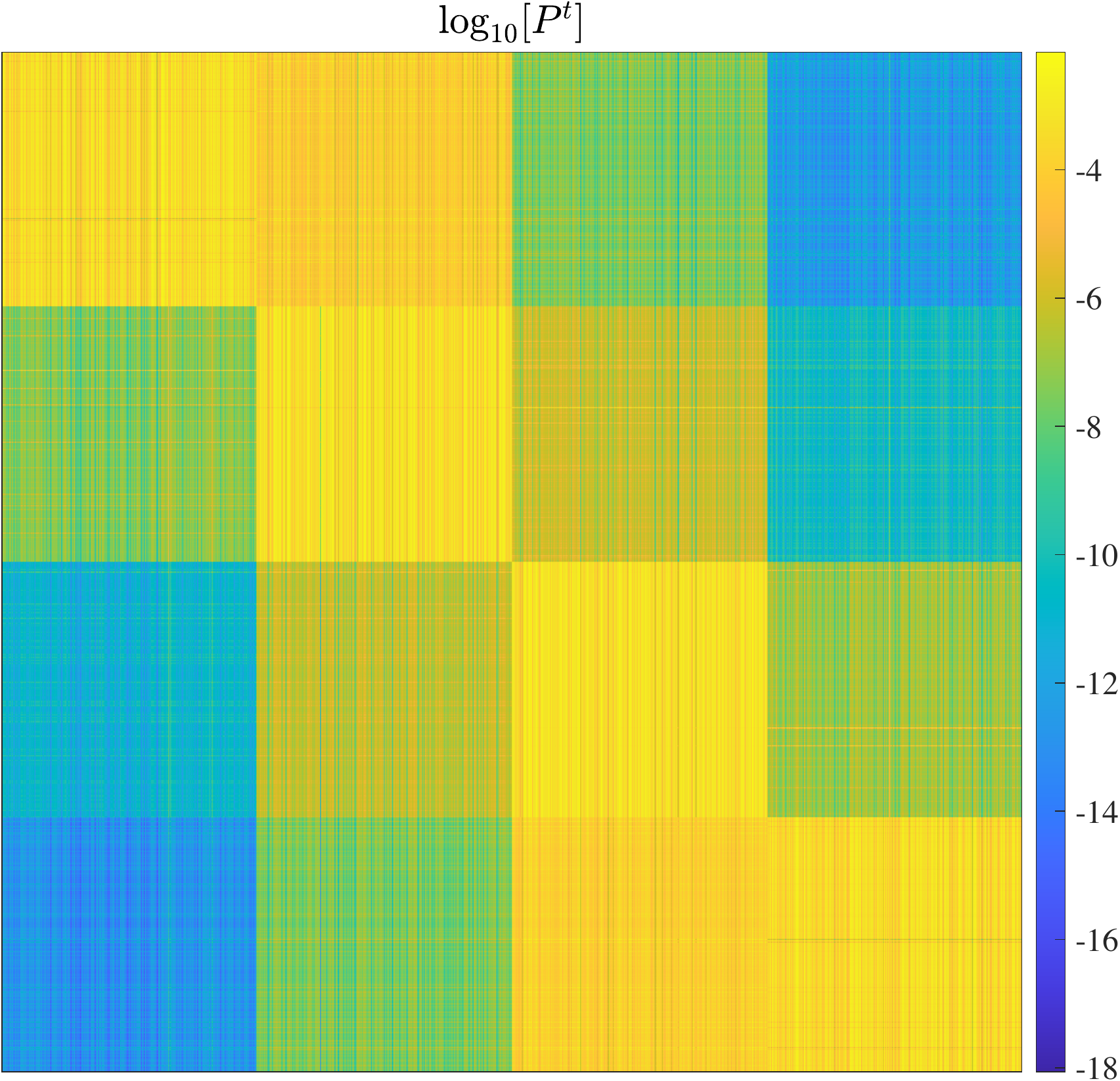}
    \end{subfigure}%
    \begin{subfigure}[t]{0.2\textwidth}
        \centering
        \includegraphics[height = 1.1in]{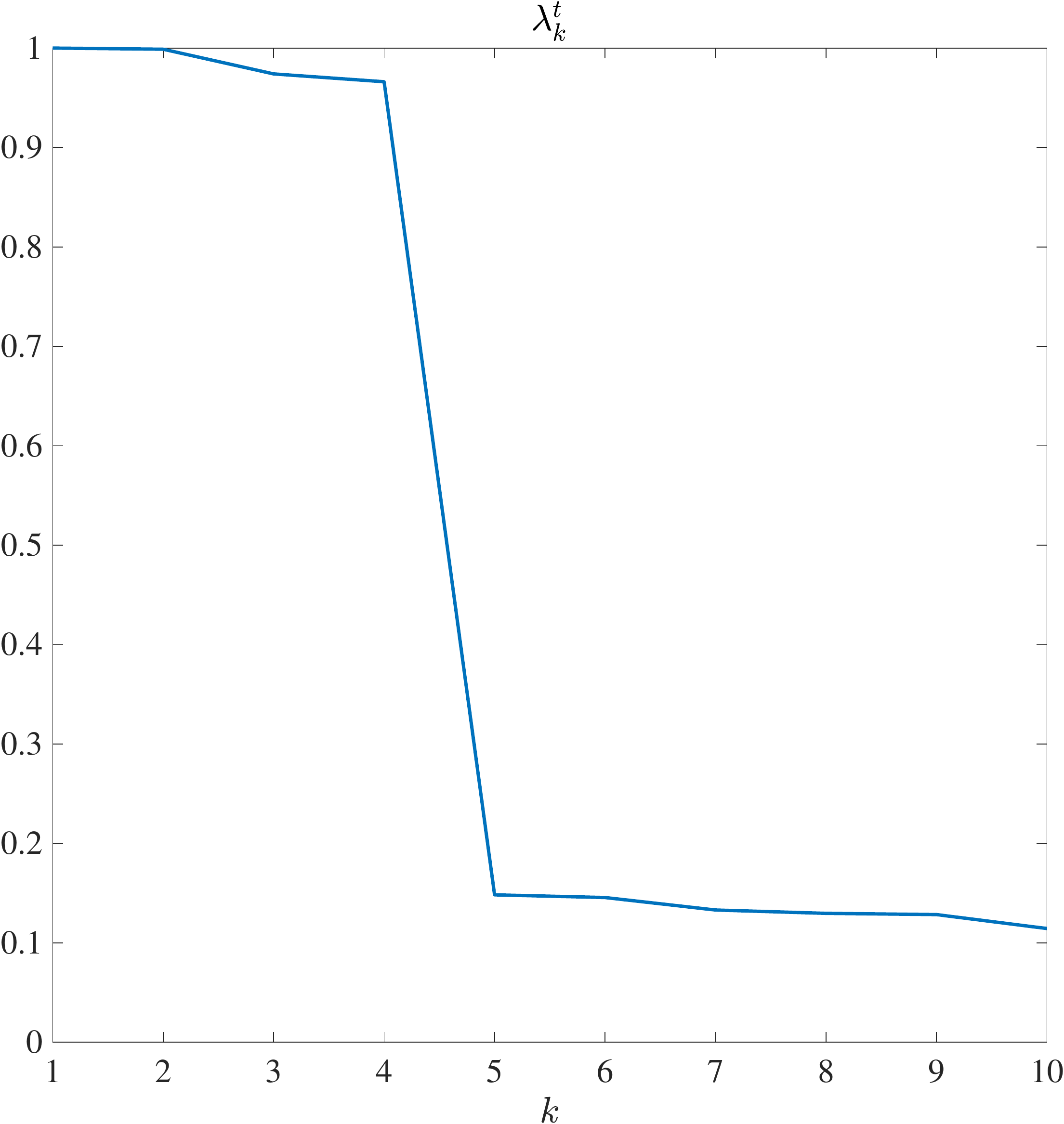}
    \end{subfigure}%
    \begin{subfigure}[t]{0.2\textwidth}
        \centering
        \includegraphics[height = 1.1in]{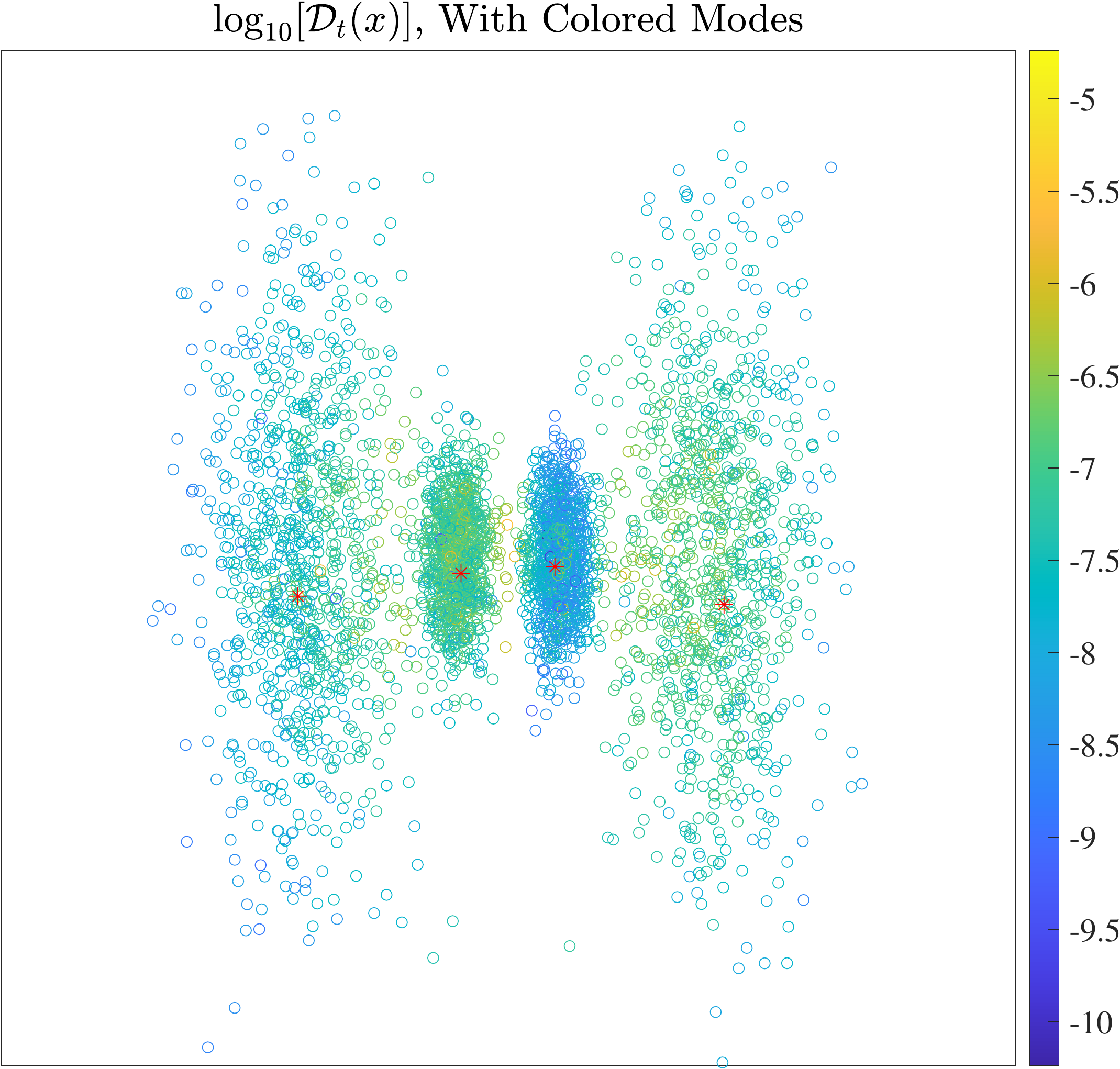}
    \end{subfigure}%
    \begin{subfigure}[t]{0.2\textwidth}
        \centering
        \includegraphics[height = 1.1in]{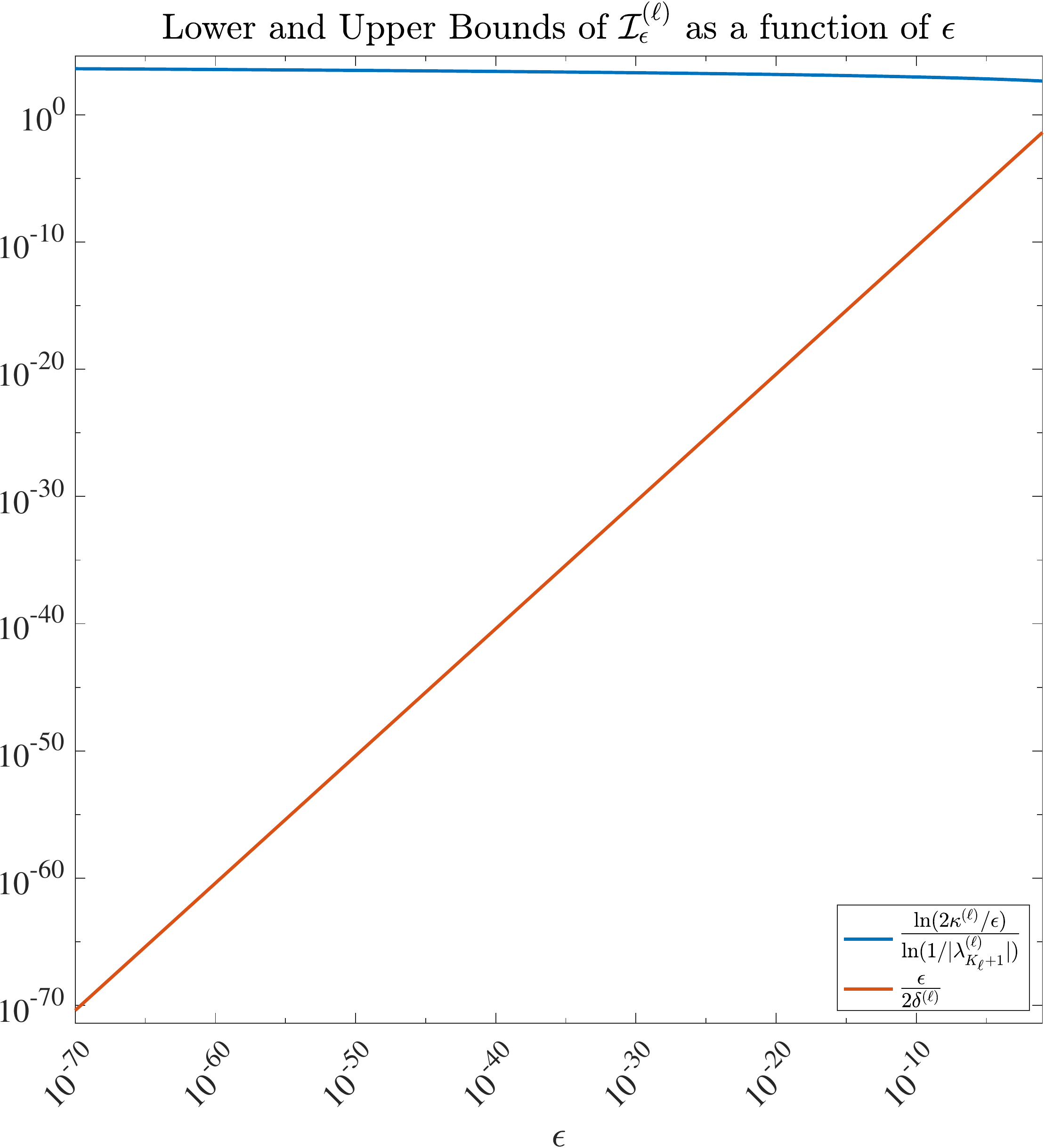}
    \end{subfigure}%
    \subcaption{LUND assignments, transition matrix, spectrum, $\mathcal{D}_t(x)$, and interval bounds for extracted clustering at time $t=2^6$.  4 clusters, total VI = 6.00.}
  \label{fig:gaussian2}\par\vspace{0.25in}    \centering
    \begin{subfigure}[t]{0.2\textwidth}
        \centering
        \includegraphics[height = 1.1in]{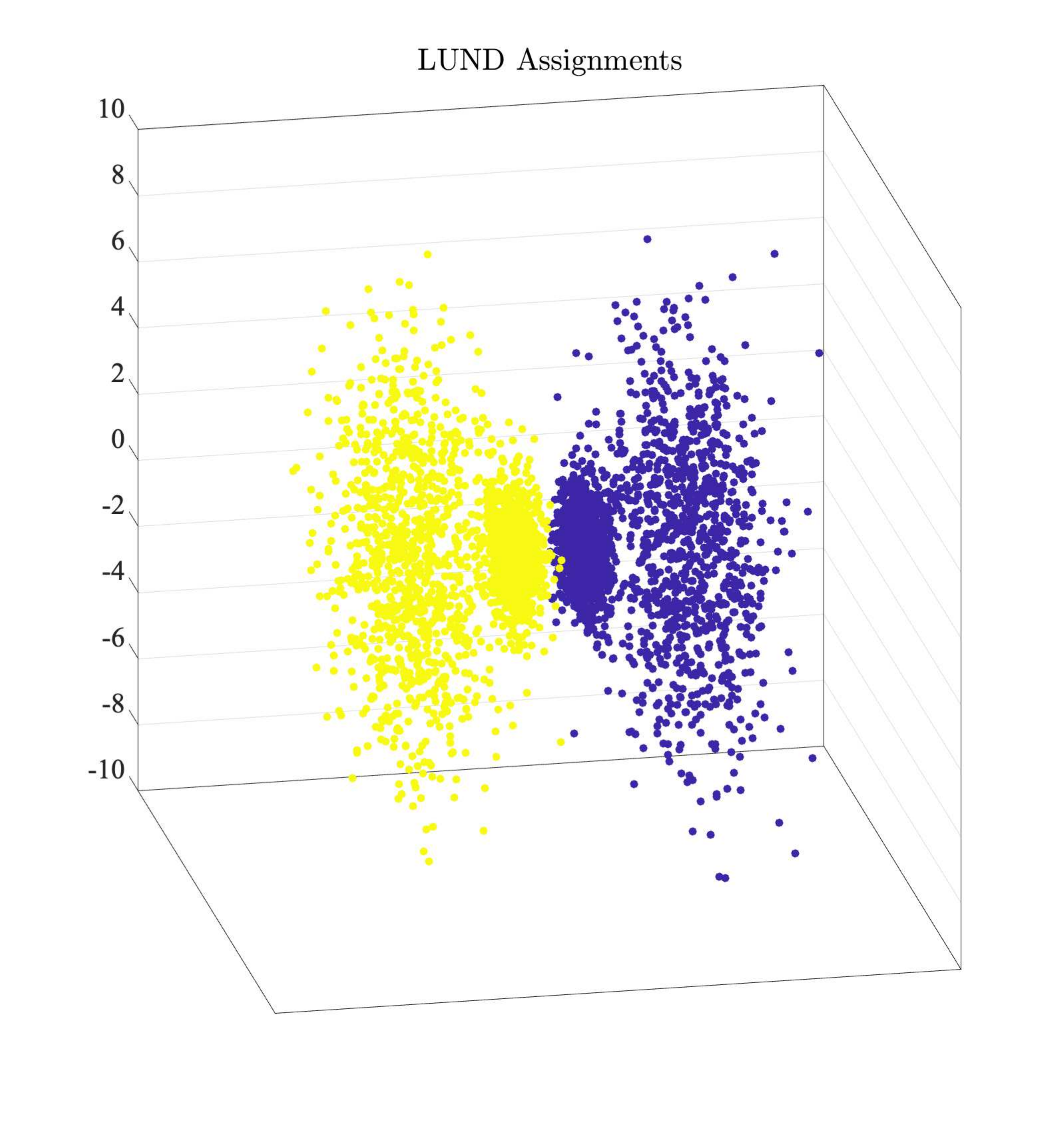}
    \end{subfigure}%
    \begin{subfigure}[t]{0.2\textwidth}
        \centering
        \includegraphics[height = 1.1in]{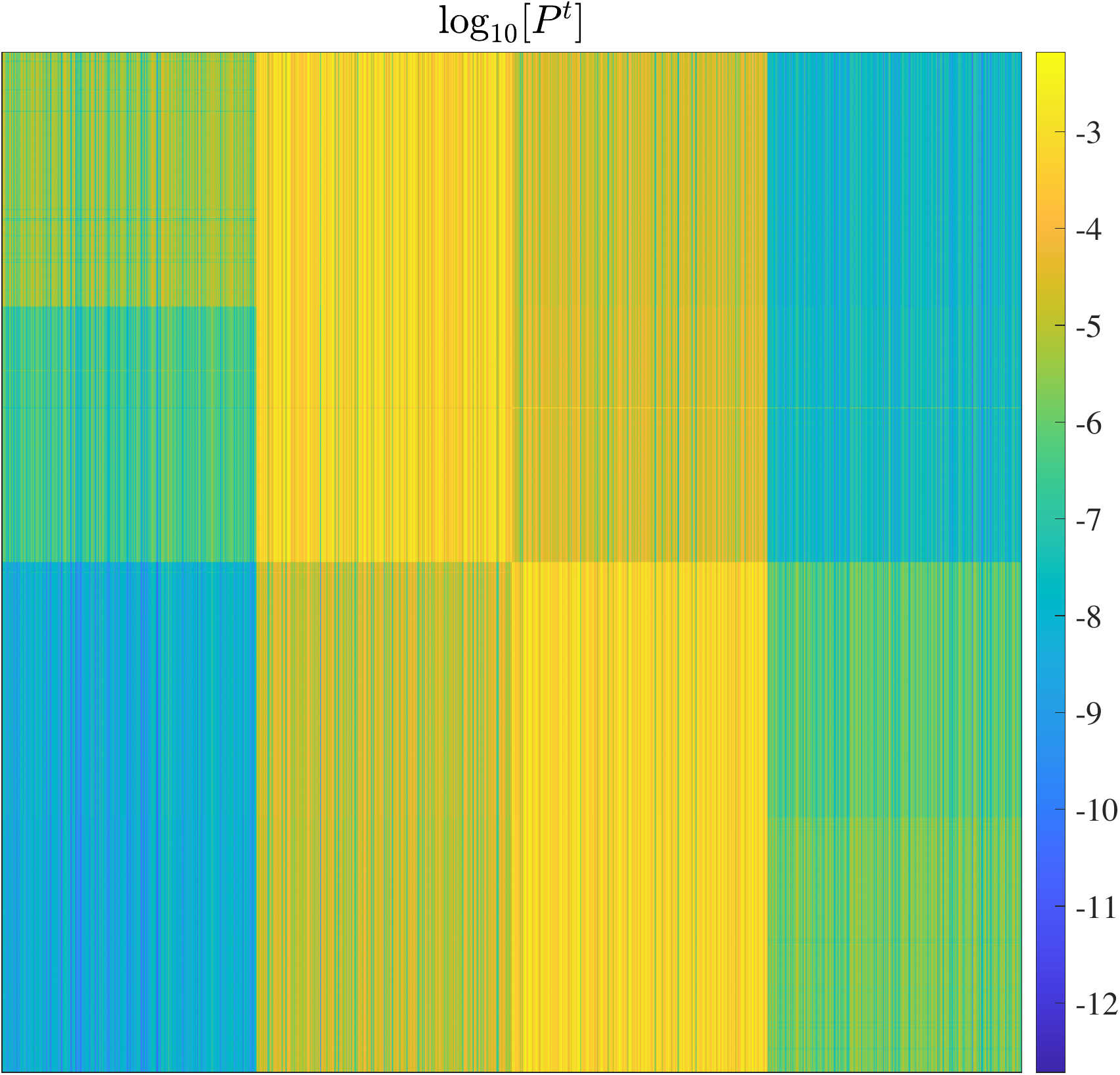}
    \end{subfigure}%
    \begin{subfigure}[t]{0.2\textwidth}
        \centering
        \includegraphics[height = 1.1in]{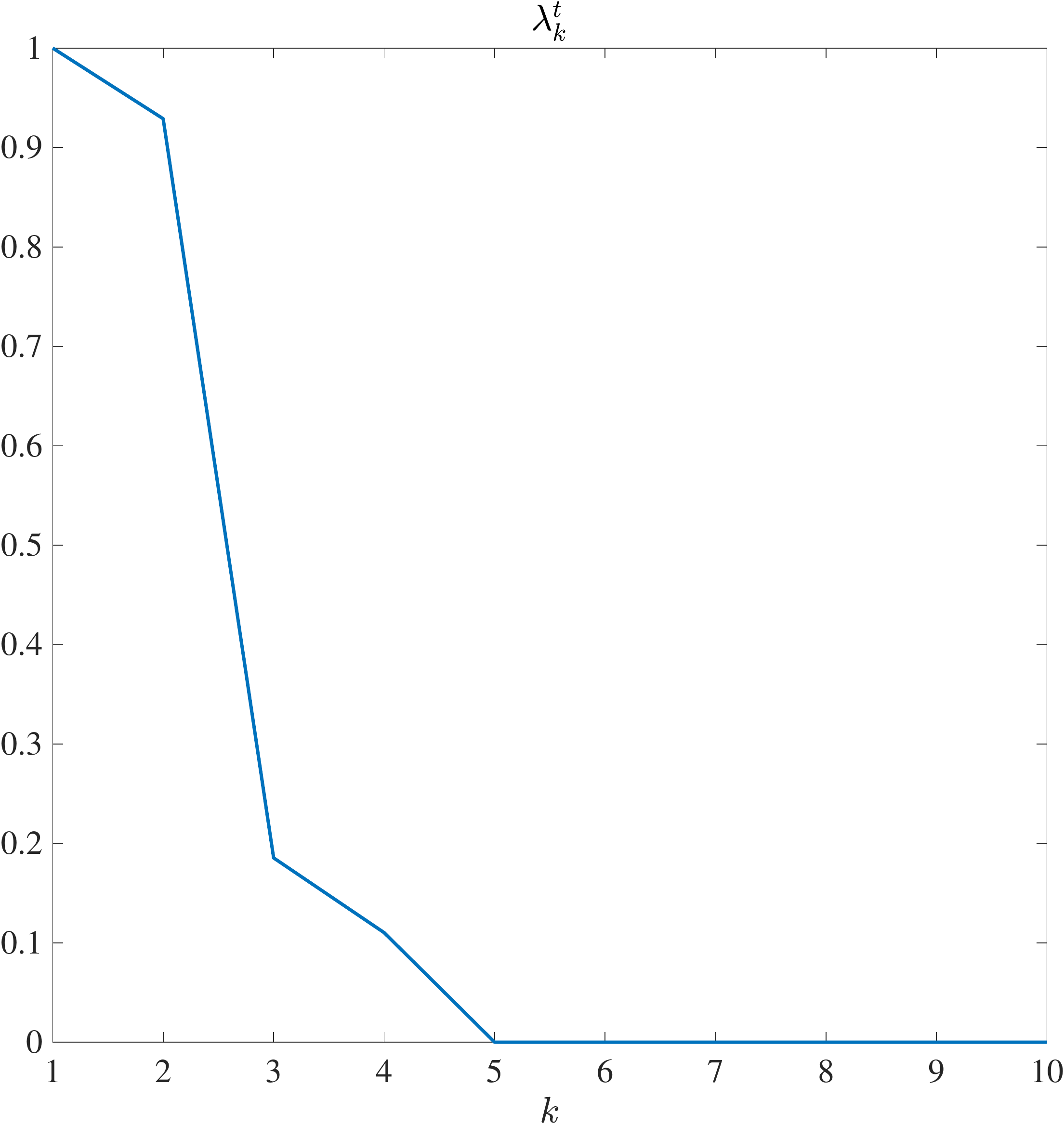}
    \end{subfigure}%
    \begin{subfigure}[t]{0.2\textwidth}
        \centering
        \includegraphics[height = 1.1in]{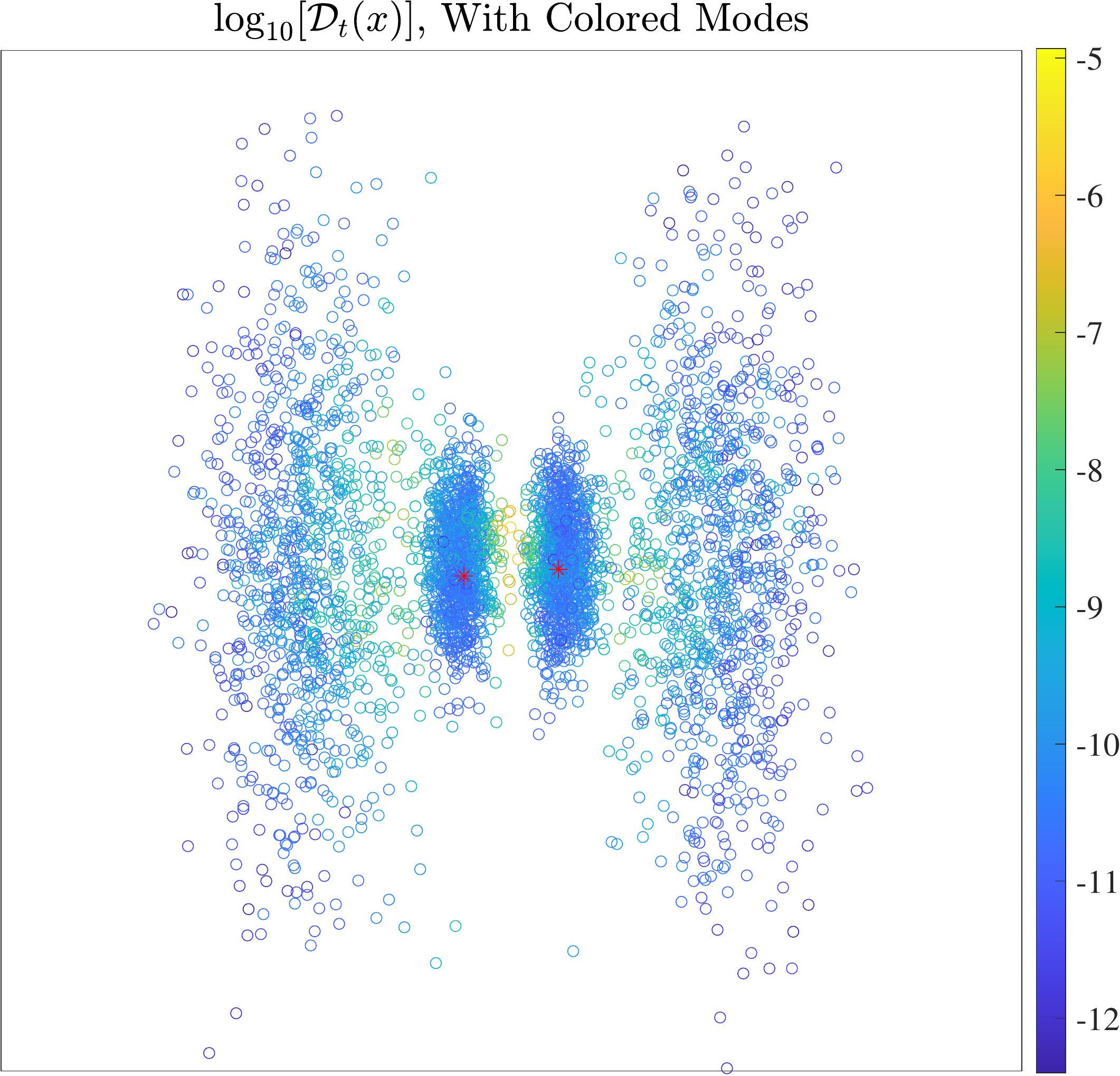}
    \end{subfigure}%
    \begin{subfigure}[t]{0.2\textwidth}
        \centering
        \includegraphics[height = 1.1in]{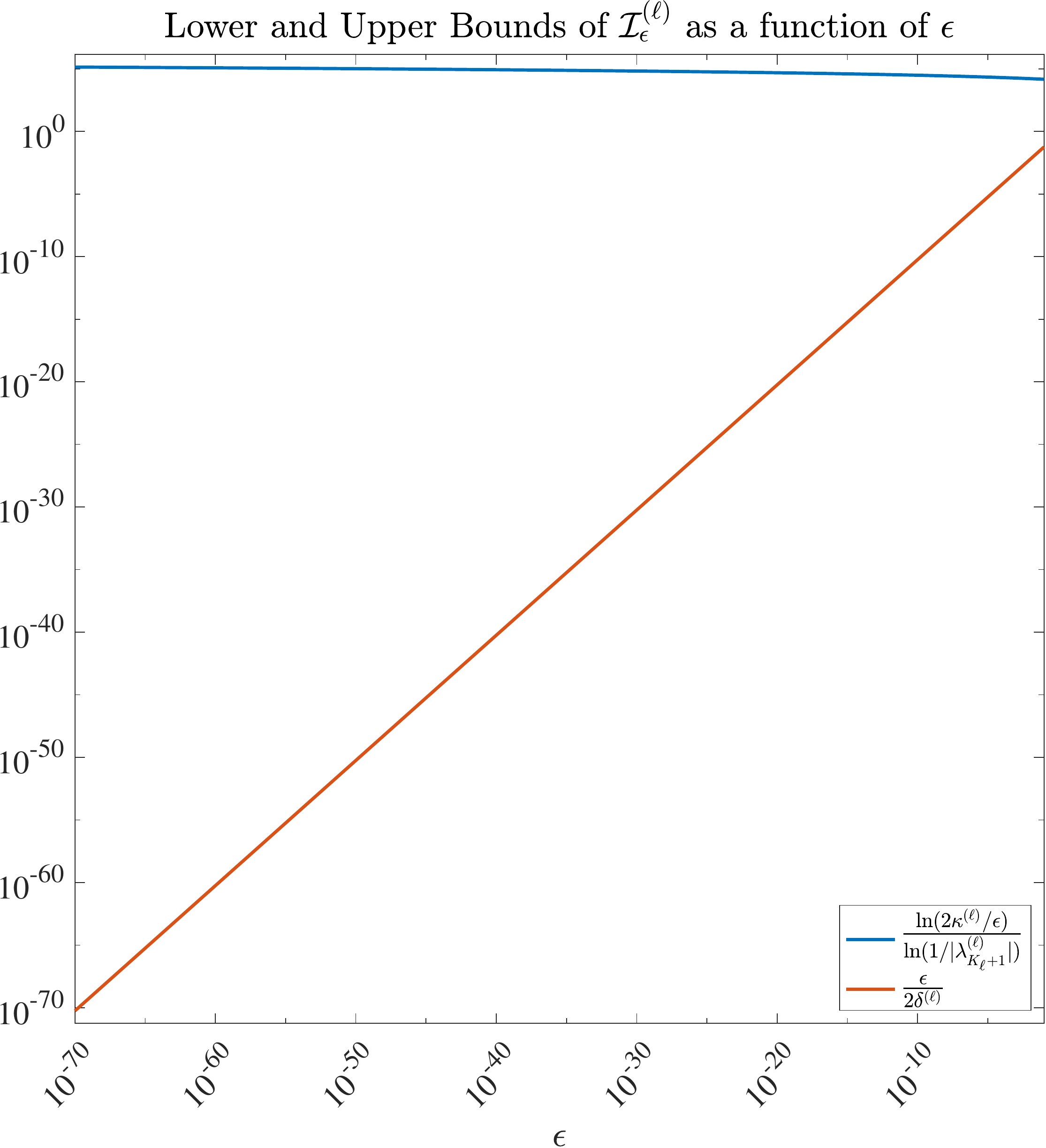}
    \end{subfigure}%
    \subcaption{LUND assignments, transition matrix, spectrum, $\mathcal{D}_t(x)$, and interval bounds for extracted clustering at time $t=2^{10}$.  2 clusters, total VI = 4.00. Optimal clustering. }
  \label{fig:gaussian3}
\end{minipage}
\caption{Diffusion on four three-dimensional Gaussians of variable density ($n=4000$). Red points indicate cluster modes. The number of estimated clusters monotonically decreases with $t$. Because of poor separation between Gaussians, $\MELD$ is empty for all choices of $\epsilon$.} \label{fig: Gaussian Diffusion}
\end{figure}

 \subsection{Synthetic Gaussians} \label{sec: 3D Gaussians 5}

In this section, we analyze a dataset sampled from four overlapping Gaussians in $\mathbb{R}^{3}$. The outer Gaussians have larger radii than the two closer to the origin, which are higher density. We implemented the M-LUND algorithm using a KNN graph with edges weighted with a Gaussian kernel. The parameters we used were $N=25$ nearest neighbors, diffusion scale $\sigma =3.10$ and KDE bandwidth $\sigma_0=1.45$. In Figure \ref{fig: Gaussian Diffusion}, we show how the labels assigned by the LUND algorithm change as a function of the diffusion time  parameter. Early in the diffusion process, higher-frequency eigenfunctions still contribute to diffusion distance computations and the LUND algorithm assigns a trivial singleton clustering (Figure \ref{fig:gaussian1}, $t\in [0,2^5]$). Later in the diffusion process, only the first four eigenfunctions contribute significantly to diffusion distances, and each of the four Gaussians is assigned to its own cluster (Figure \ref{fig:gaussian2}, $t\in [2^6,2^9]$). Finally, each large-radius Gaussian is merged with the nearest small-radius Gaussian (Figure \ref{fig:gaussian3}, $t\in [2^{10}, 2^{16}]$). The M-LUND algorithm assigns a total VI of 6.00 to the 4-cluster clustering and a total VI of 4.00 to the 2-cluster clustering, which is more stable and thus the total VI minimizer.

Figure \ref{fig: Gaussian Diffusion} makes clear that the MELD data model is highly sensitive to cluster overlap. If outliers of one cluster are close to outliers of another, then $\delta$ will be large~\cite{murphy2019LUND}. Because of the significant overlap between the four Gaussians in this dataset, the interval $\mathcal{I}_\epsilon^{(\ell)}$ is empty for both clusterings of $X$ across all choices of $\epsilon>0$. The numerical experiments given in this section therefore imply that the reliance of $\MELD$ on the separation parameter $\delta$ is somewhat pessimistic, and the M-LUND algorithm is able to detect latent structure at a variety of scales even when $\epsilon$-separation is not necessarily achieved.

\begin{figure}[t] 
\begin{minipage}{\textwidth}
  \vspace*{\fill}
  \centering
    \centering
    \begin{subfigure}[t]{0.2\textwidth}
        \centering
        \includegraphics[height = 1.1in]{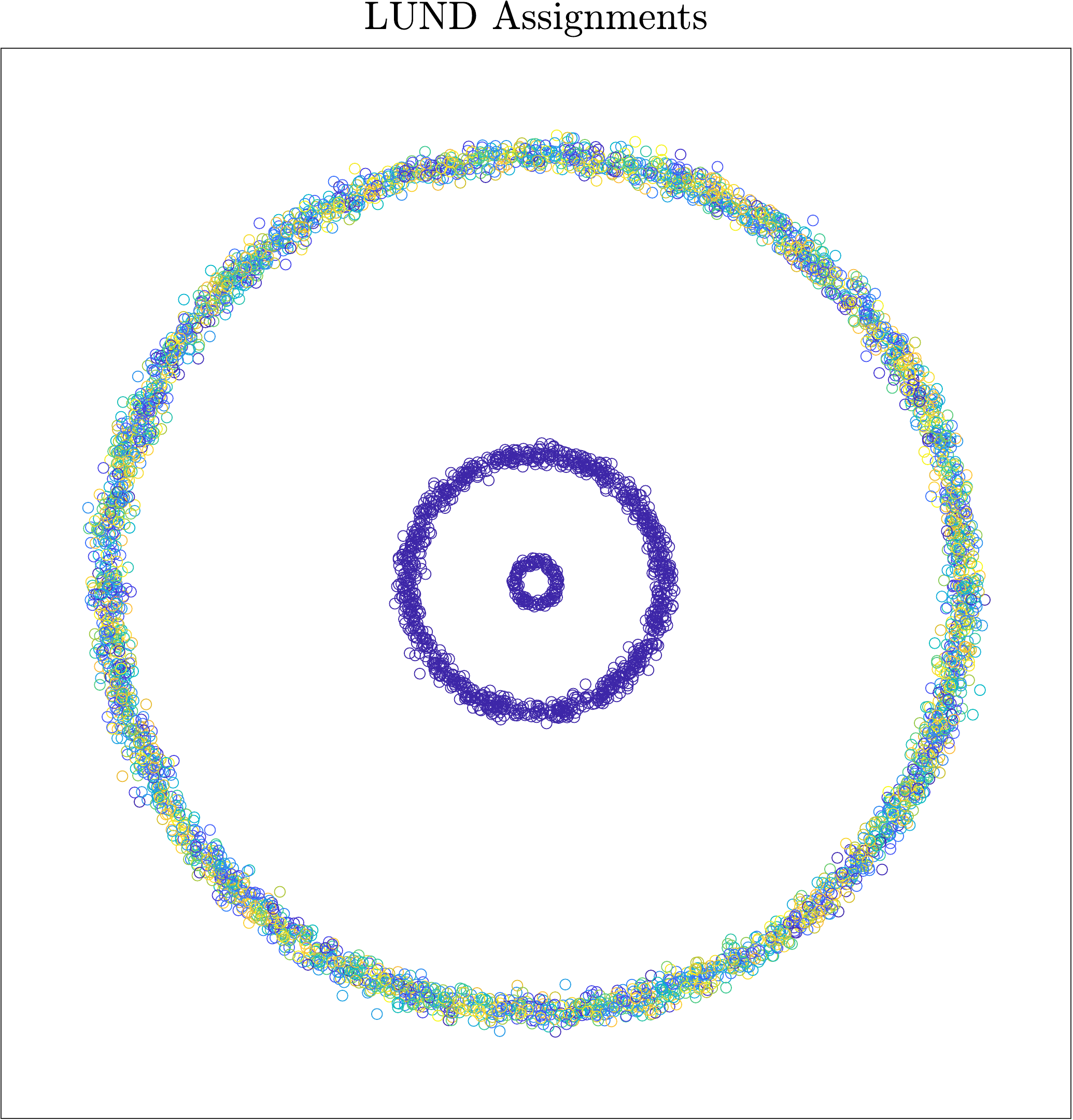}
    \end{subfigure}%
    \begin{subfigure}[t]{0.2\textwidth}
        \centering
        \includegraphics[height = 1.1in]{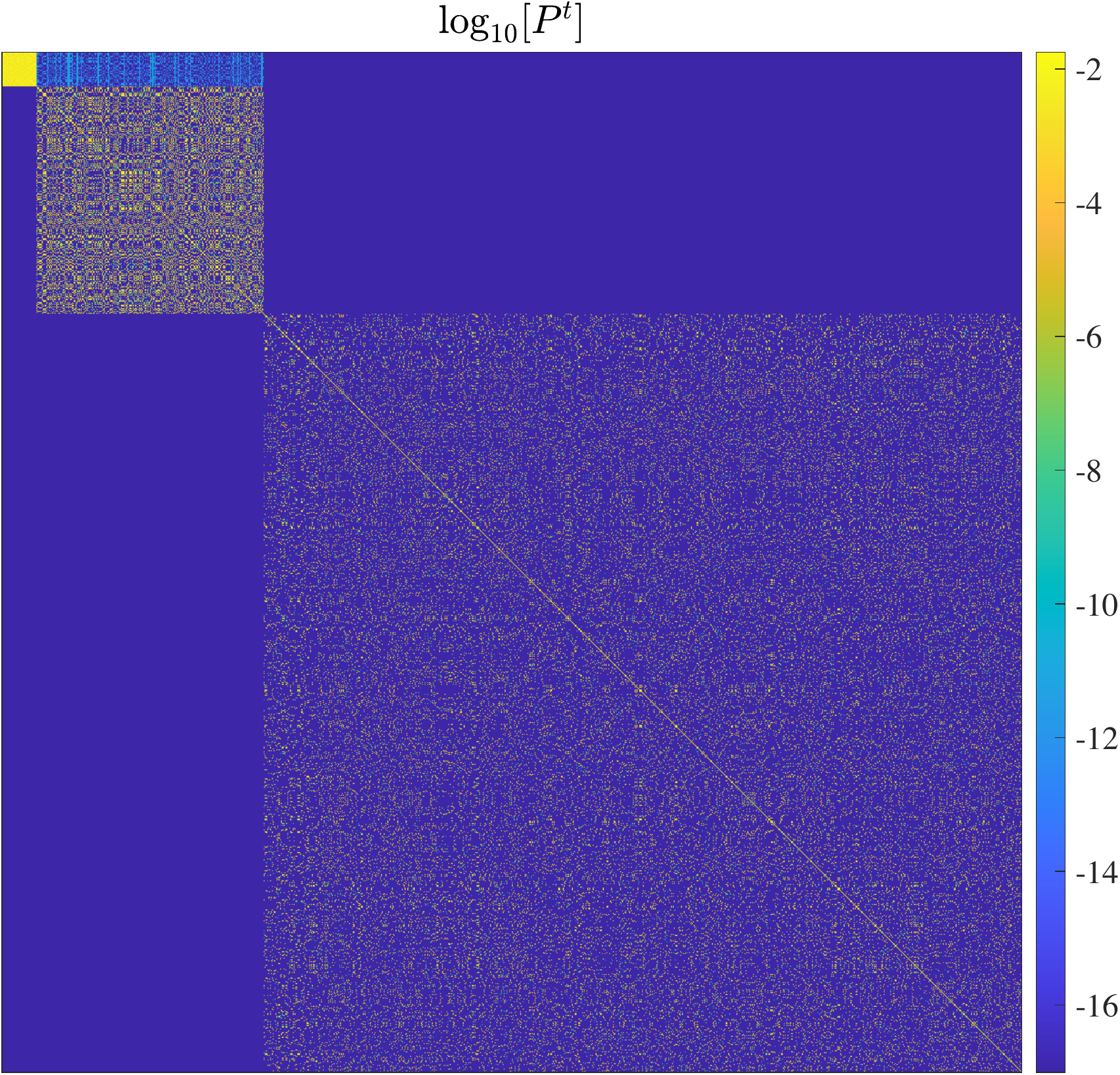}
    \end{subfigure}%
    \begin{subfigure}[t]{0.2\textwidth}
        \centering
        \includegraphics[height = 1.1in]{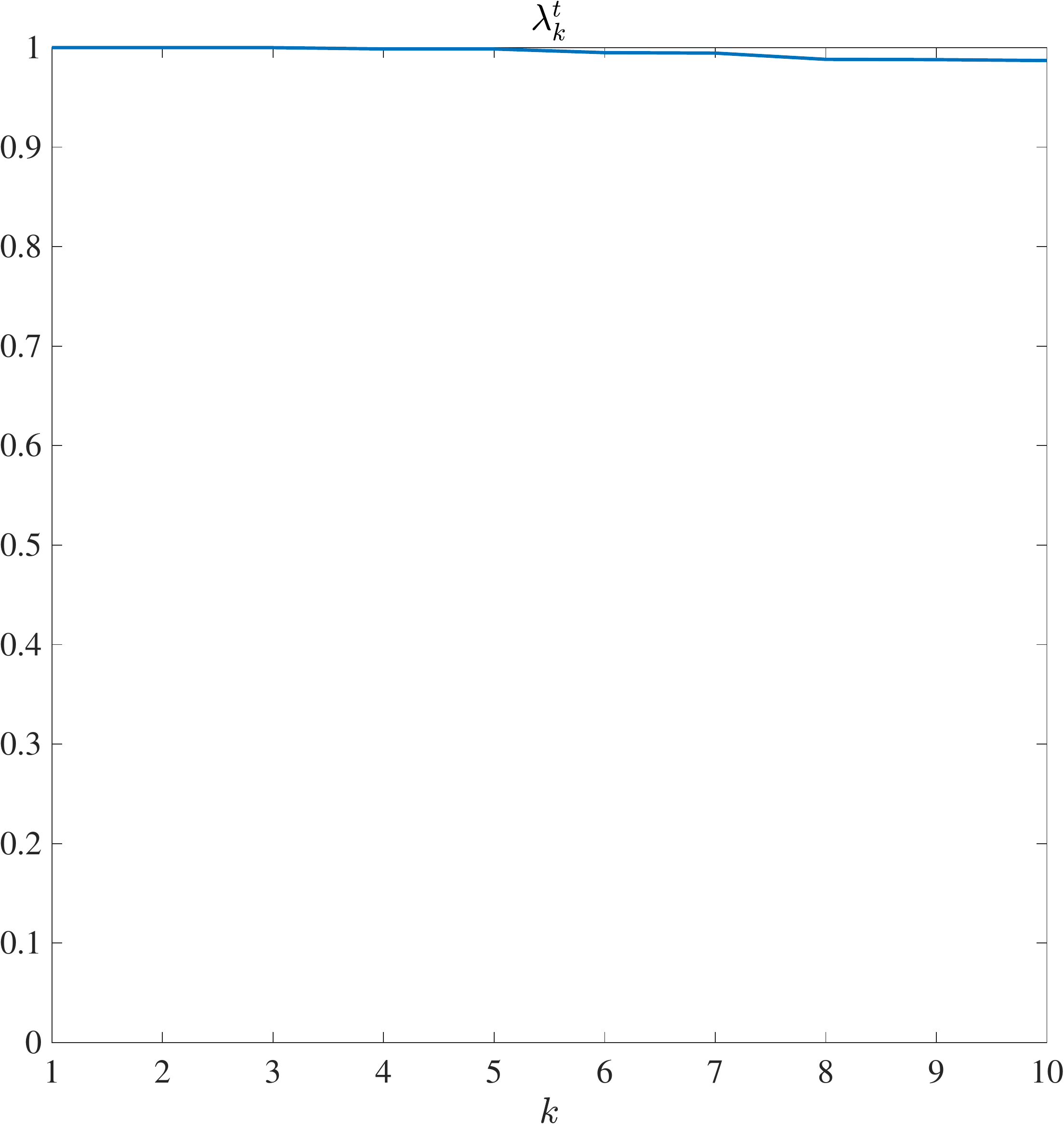}
    \end{subfigure}%
    \begin{subfigure}[t]{0.2\textwidth}
        \centering
        \includegraphics[height = 1.1in]{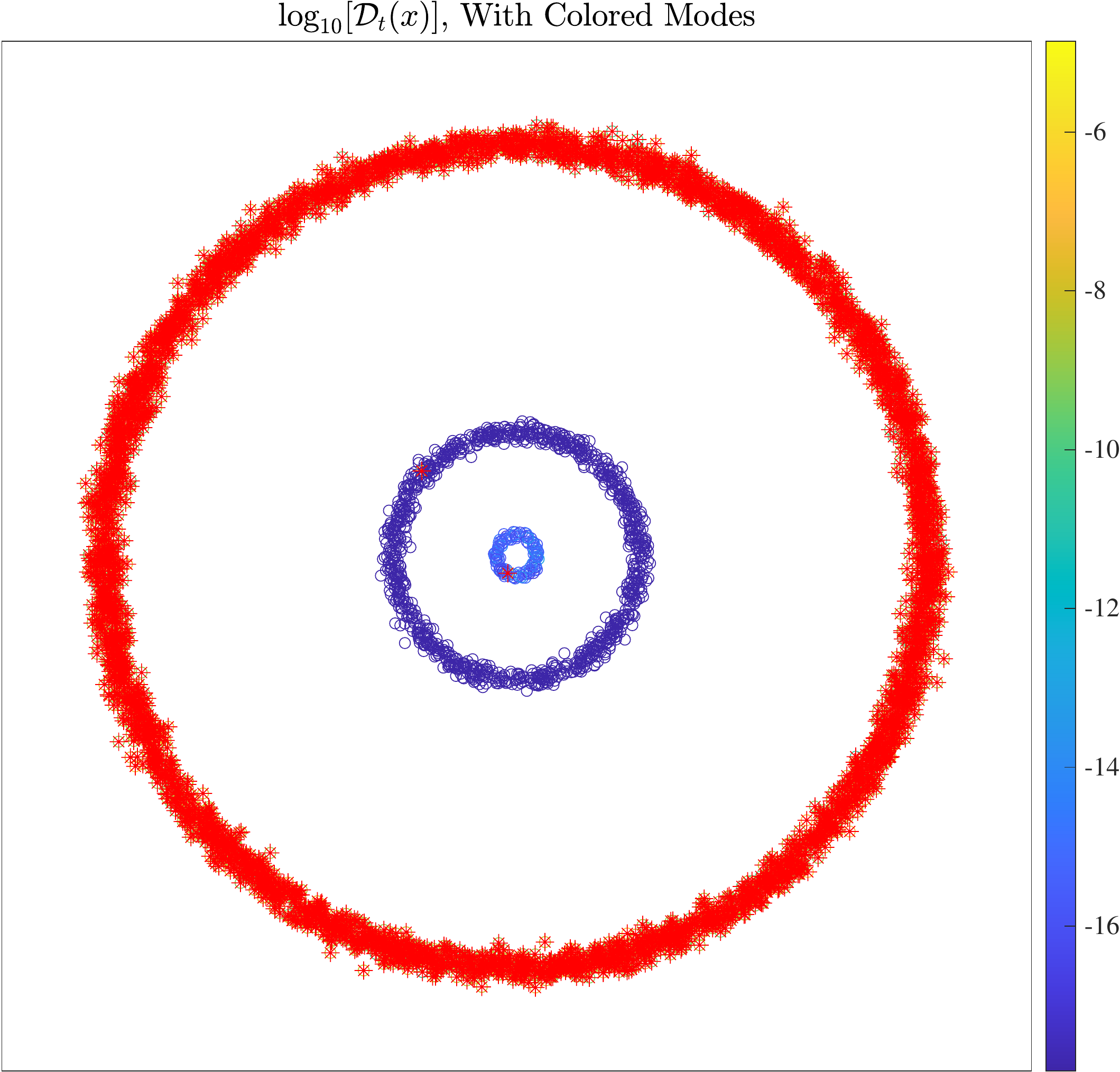}
    \end{subfigure}%
    \begin{subfigure}[t]{0.2\textwidth}
        \centering
        \includegraphics[height = 1.1in]{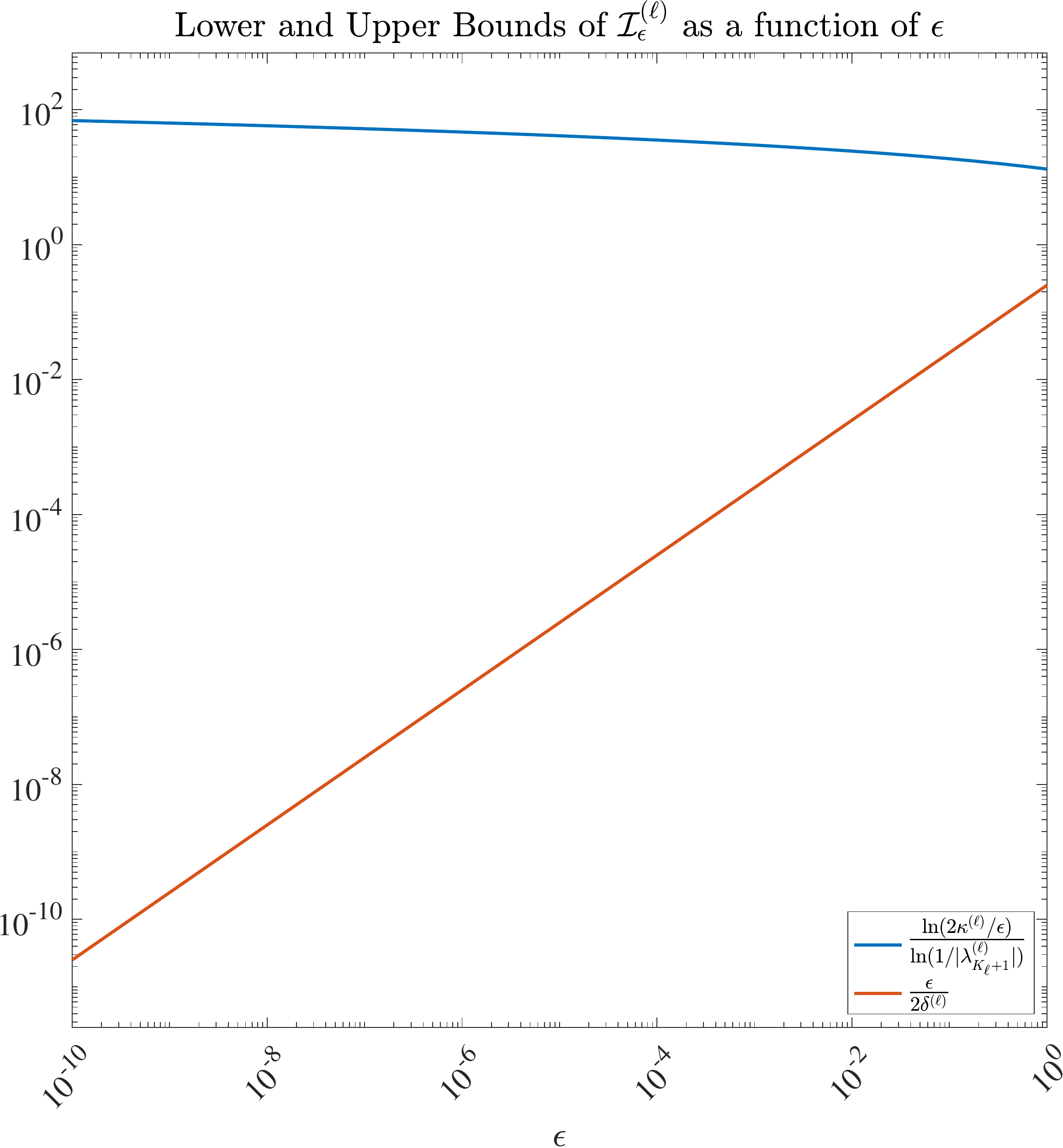}
    \end{subfigure}%
    \subcaption{LUND assignments, transition matrix, spectrum, $\mathcal{D}_t(x)$, and interval bounds for extracted clustering at time $t=2^1$. 5198 clusters.}
  \label{fig:nonlinear1}\par\vspace{0.0625in}
    \centering
    \begin{subfigure}[t]{0.2\textwidth}
        \centering
        \includegraphics[height = 1.1in]{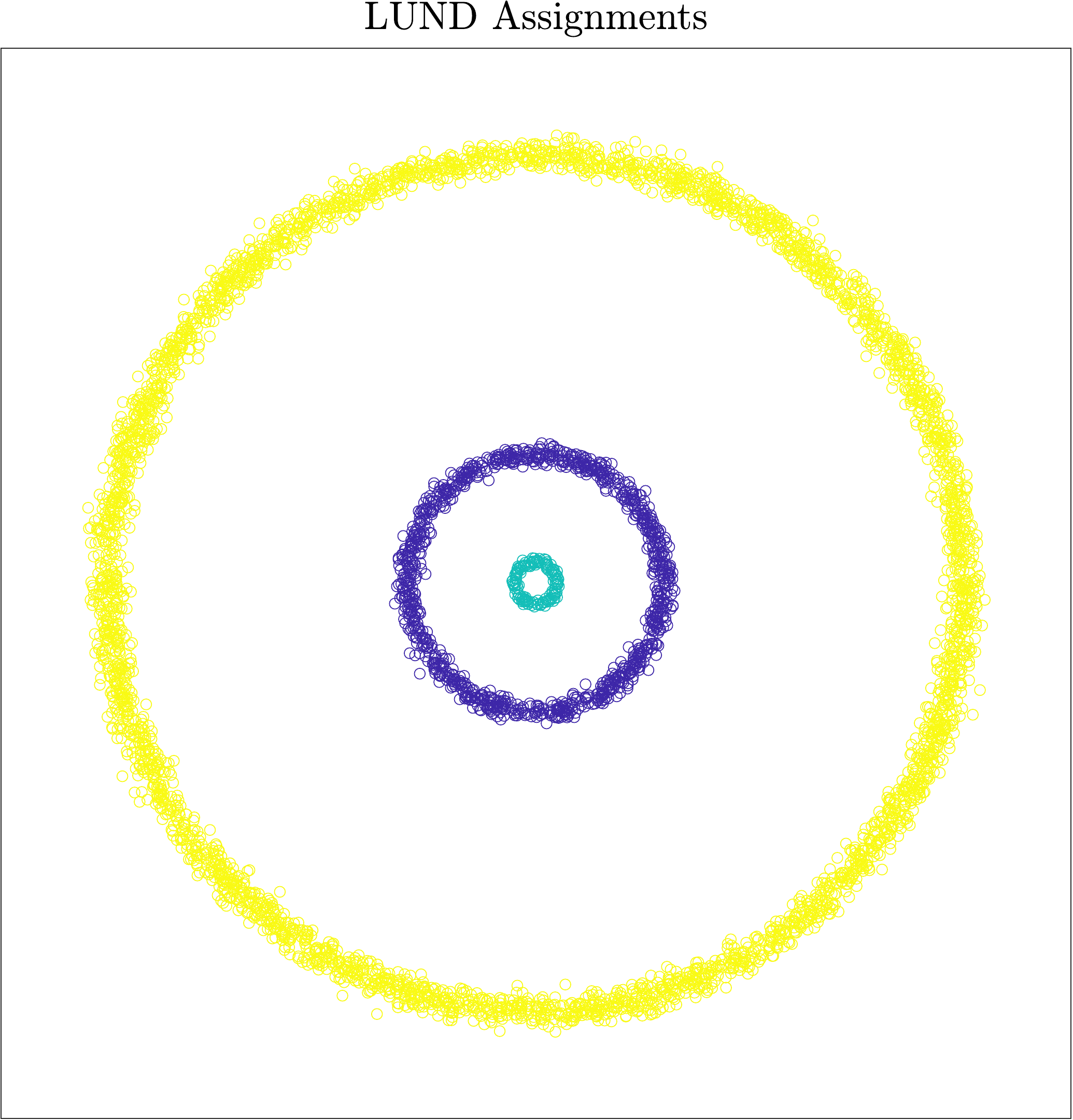}
    \end{subfigure}%
    \begin{subfigure}[t]{0.2\textwidth}
        \centering
        \includegraphics[height = 1.1in]{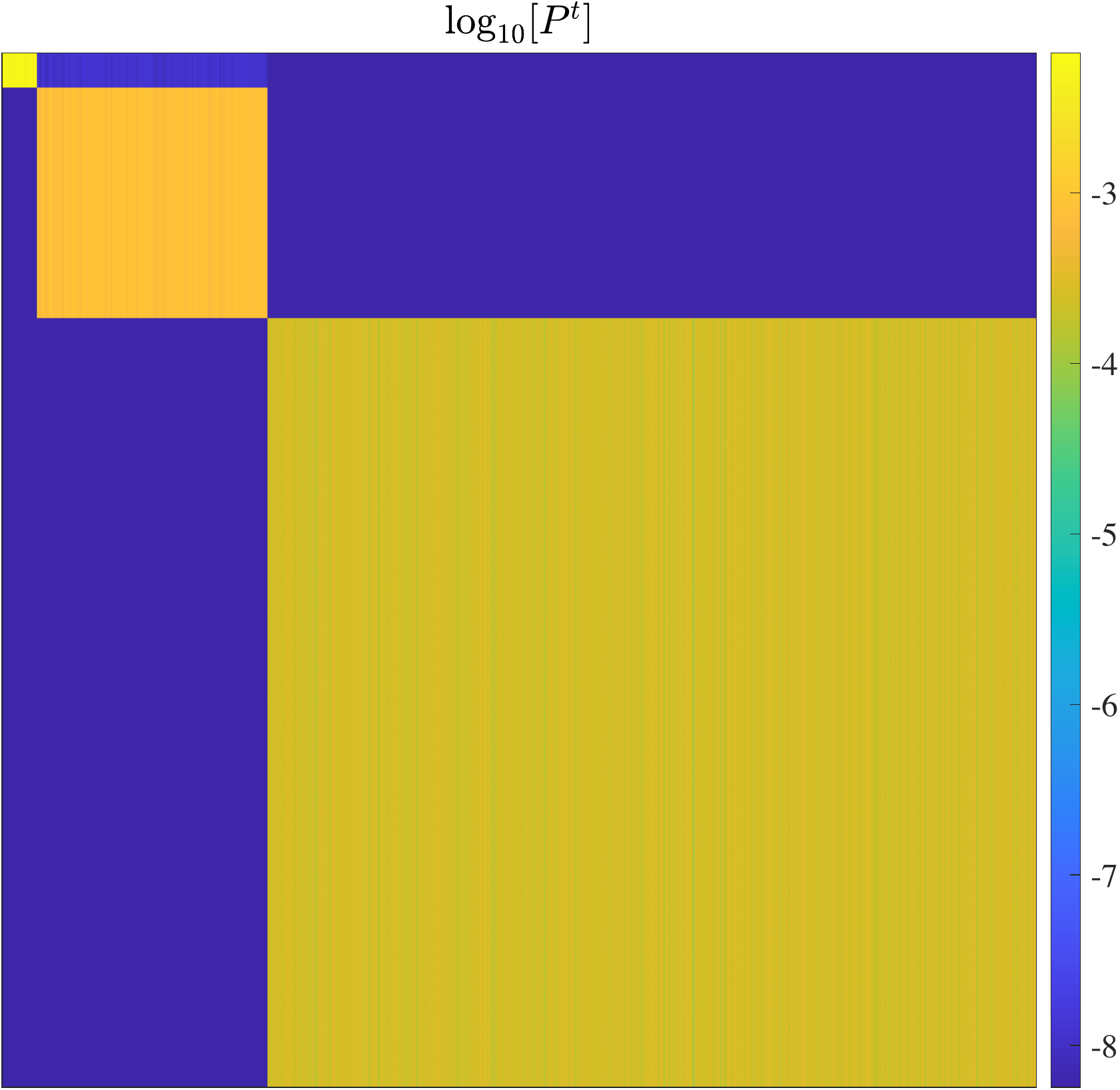}
    \end{subfigure}%
    \begin{subfigure}[t]{0.2\textwidth}
        \centering
        \includegraphics[height = 1.1in]{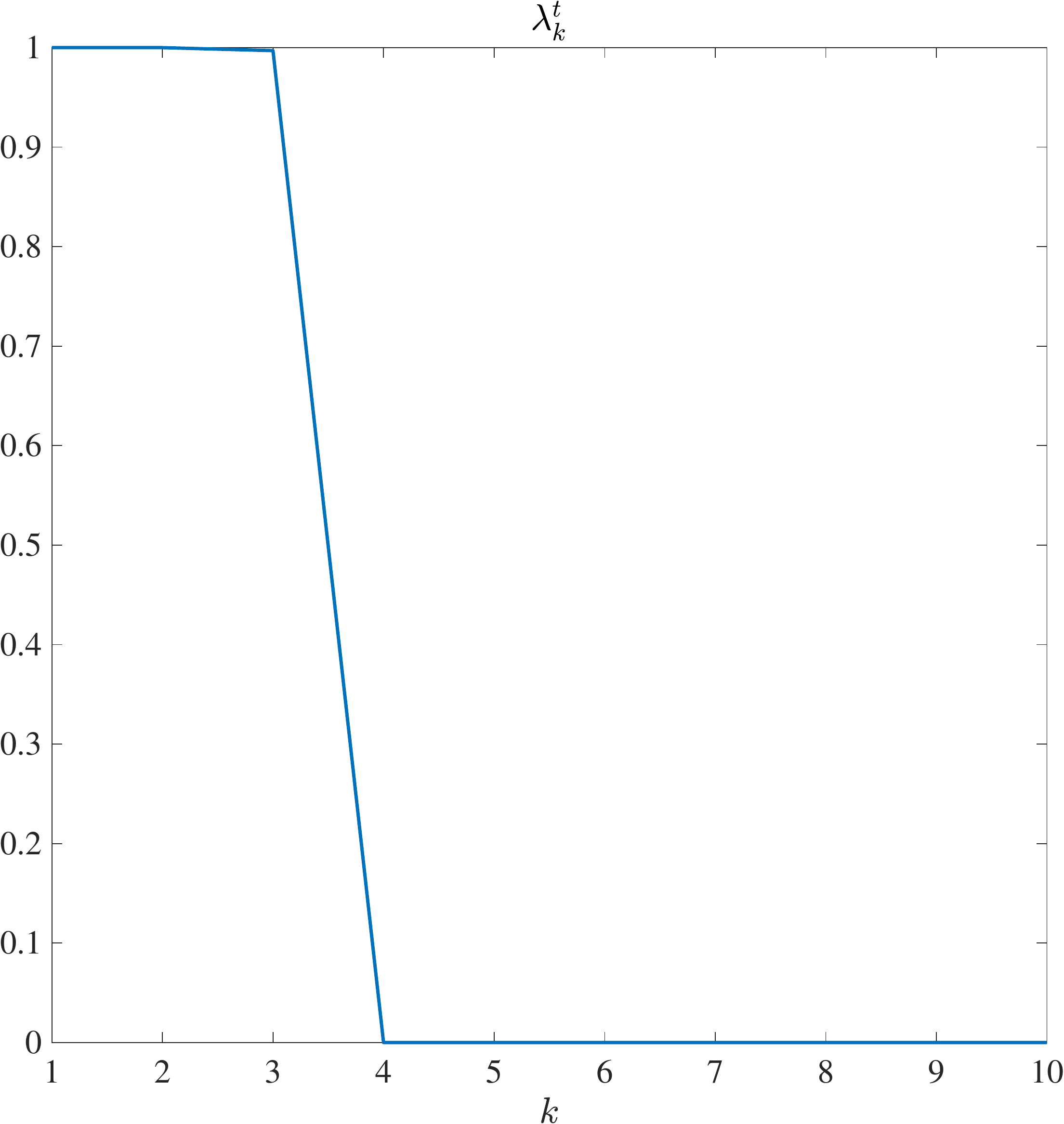}
    \end{subfigure}%
    \begin{subfigure}[t]{0.2\textwidth}
        \centering
        \includegraphics[height = 1.1in]{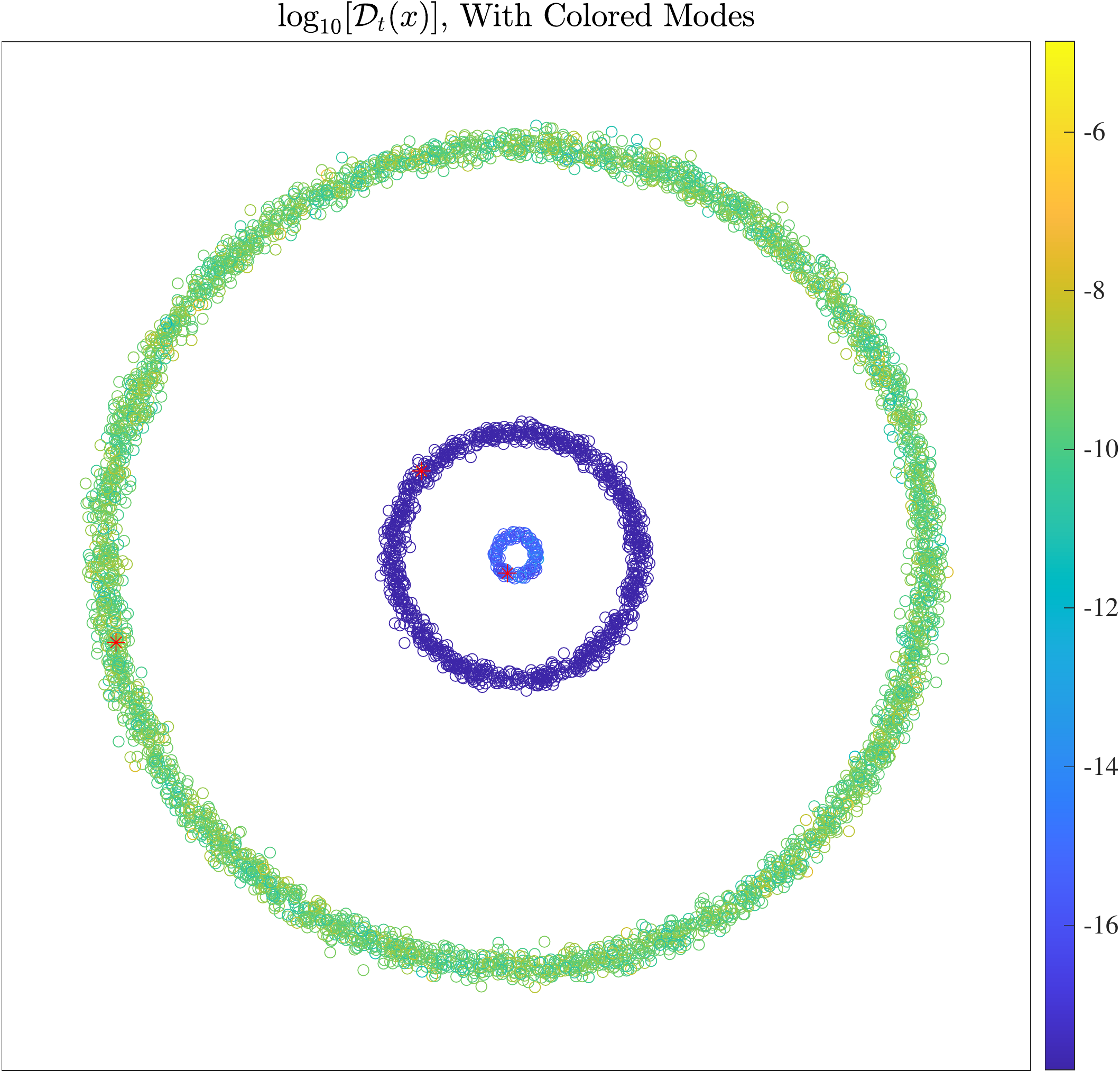}
    \end{subfigure}%
    \begin{subfigure}[t]{0.2\textwidth}
        \centering
        \includegraphics[height = 1.1in]{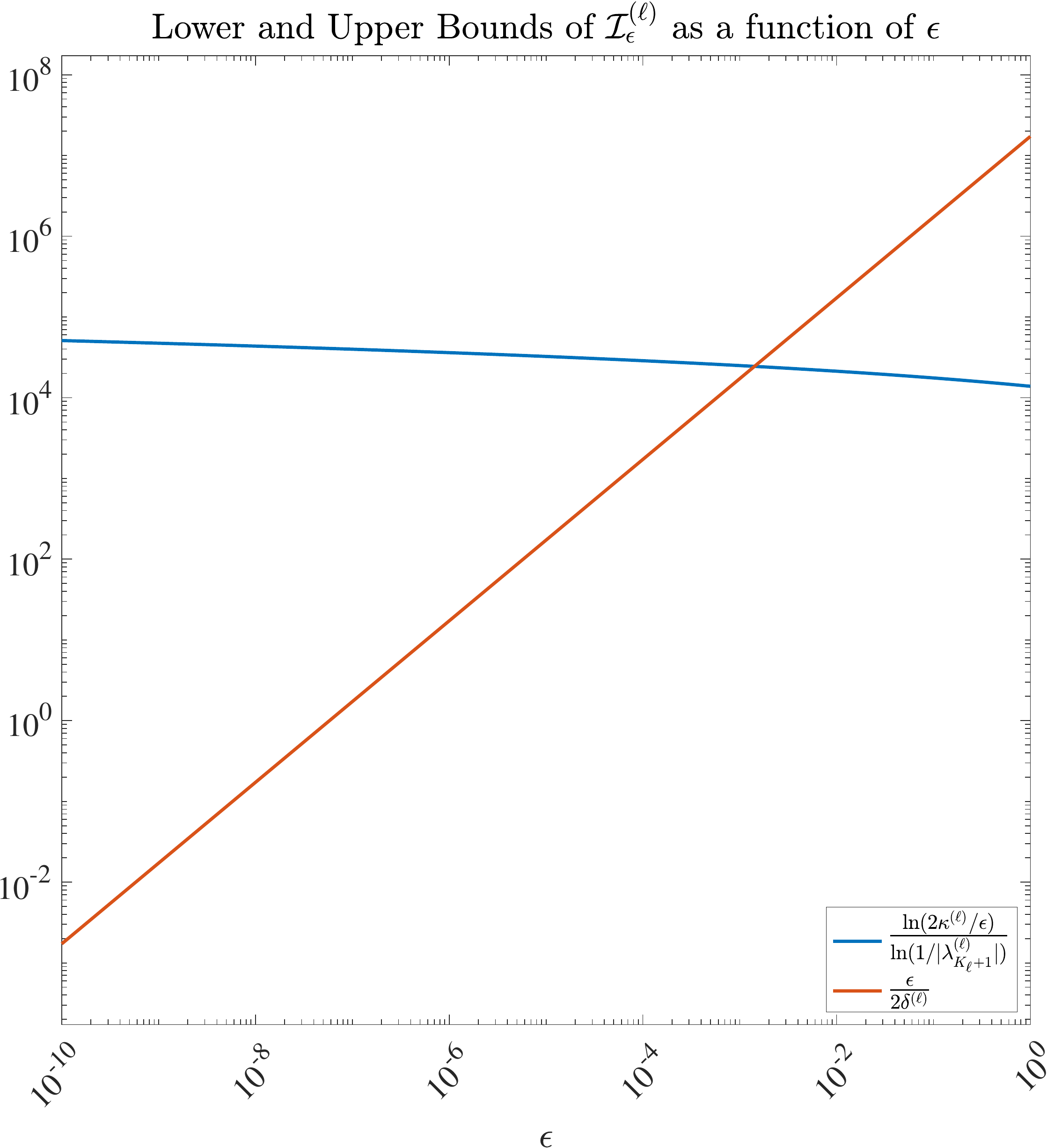}
    \end{subfigure}%
      \subcaption{LUND assignments, transition matrix, spectrum, $\mathcal{D}_t(x)$, and interval bounds for extracted clustering at time $t=2^{16}$. 3 clusters, total VI = 1.79. Optimal clustering.}
  \label{fig:nonlinear2}\par\vspace{0.0625in}
    \centering
    \begin{subfigure}[t]{0.2\textwidth}
        \centering
        \includegraphics[height = 1.1in]{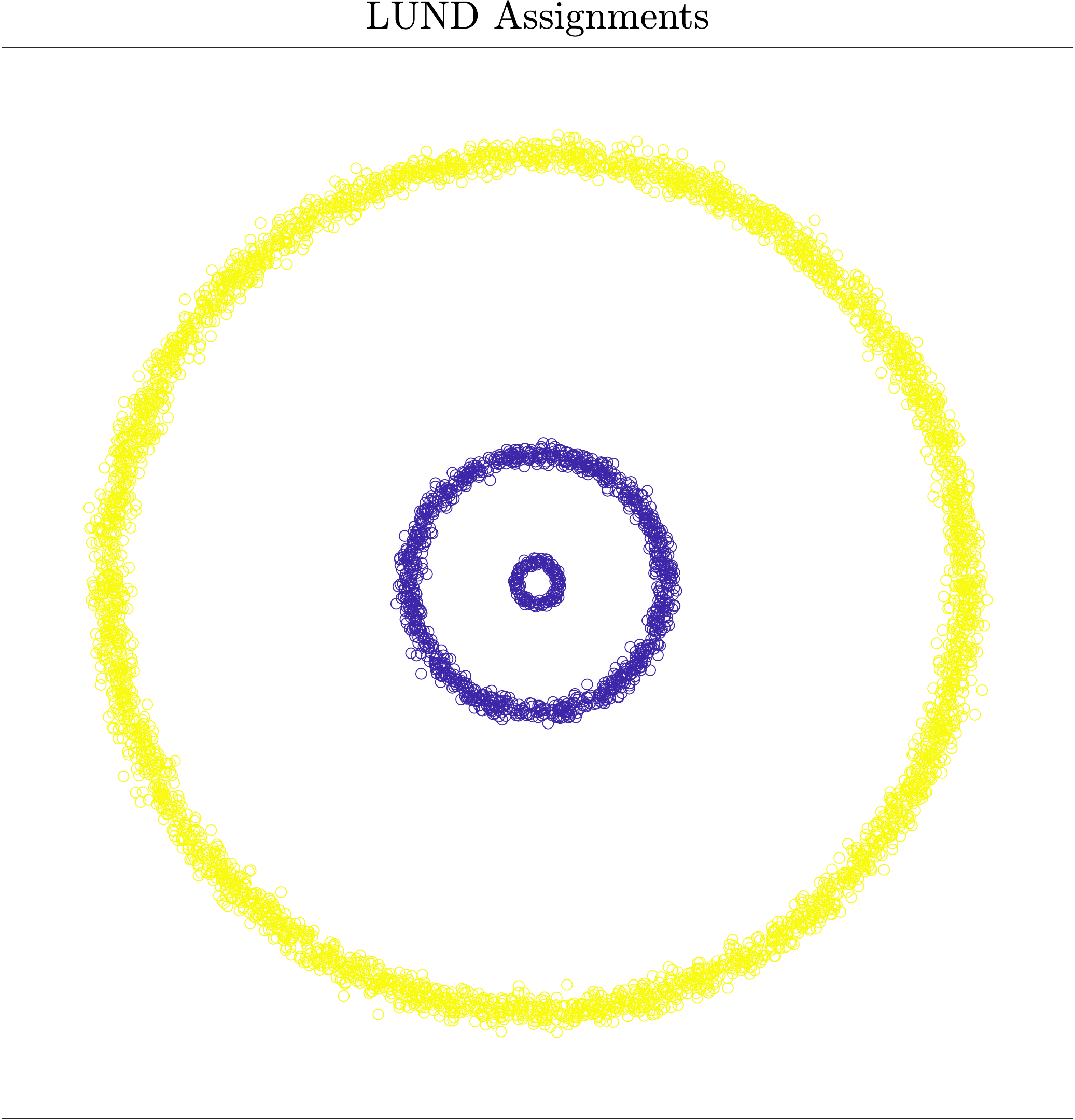}
    \end{subfigure}%
    \begin{subfigure}[t]{0.2\textwidth}
        \centering
        \includegraphics[height = 1.1in]{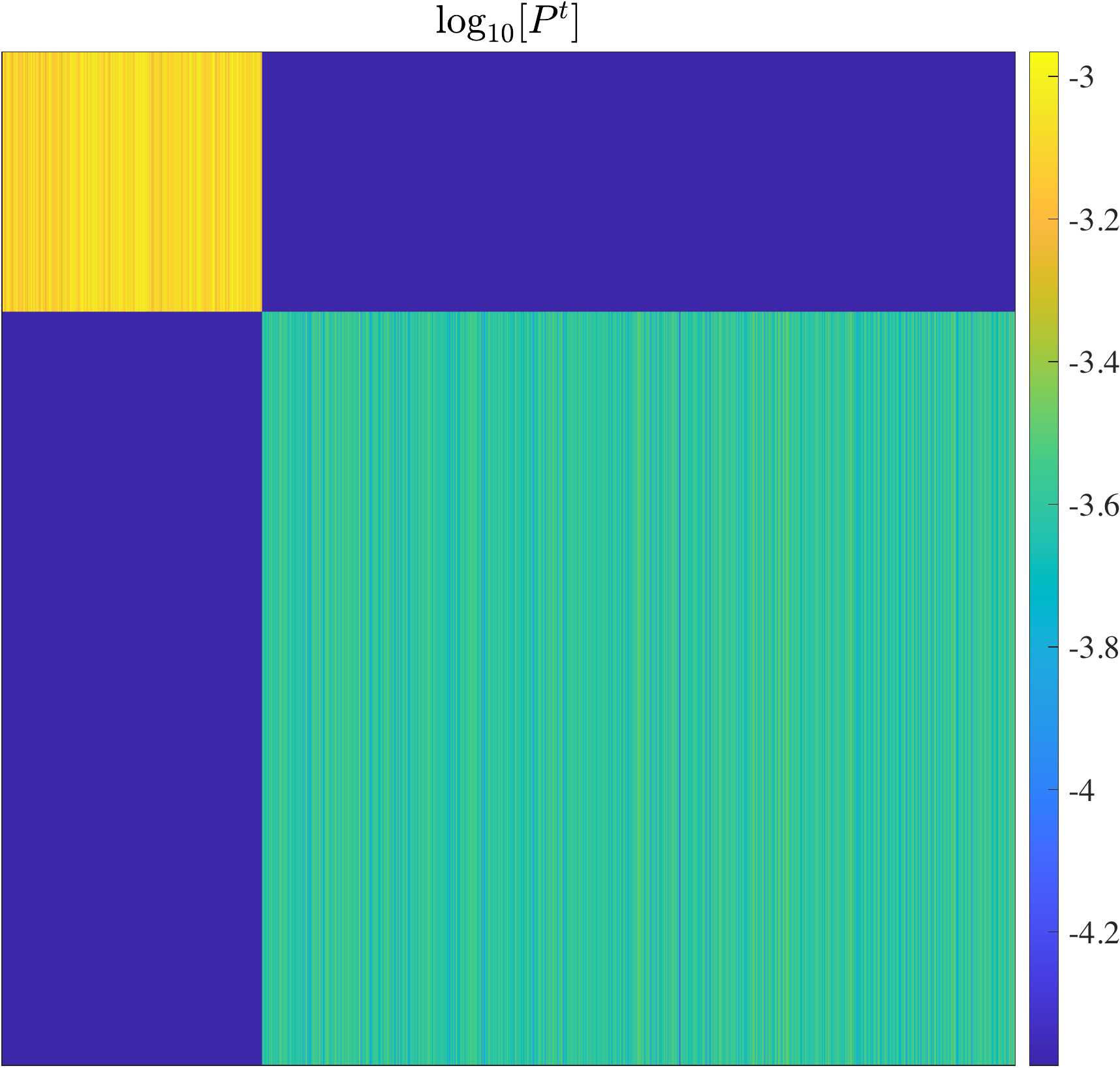}
    \end{subfigure}%
    \begin{subfigure}[t]{0.2\textwidth}
        \centering
        \includegraphics[height = 1.1in]{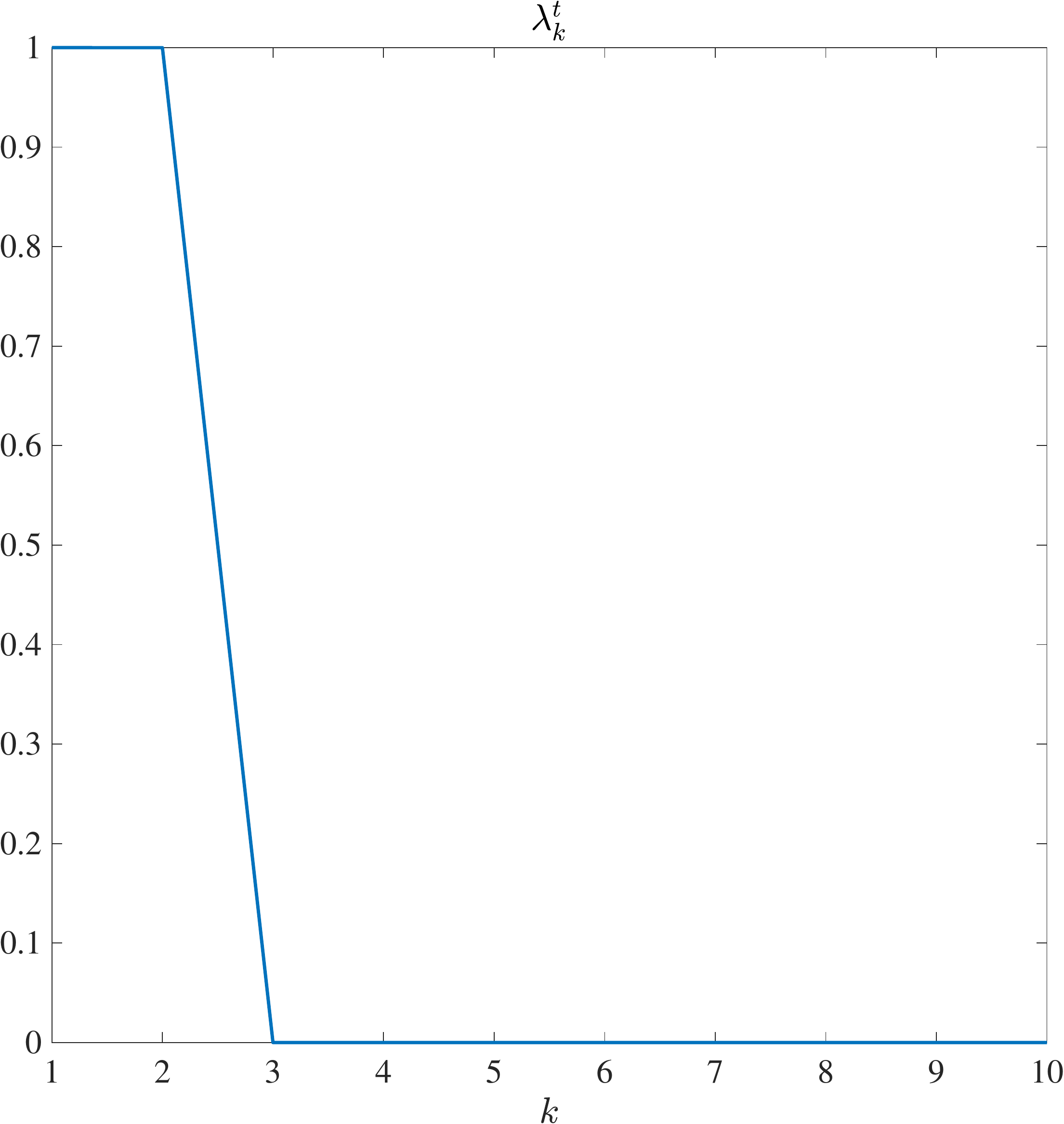}
    \end{subfigure}%
    \begin{subfigure}[t]{0.2\textwidth}
        \centering
        \includegraphics[height = 1.1in]{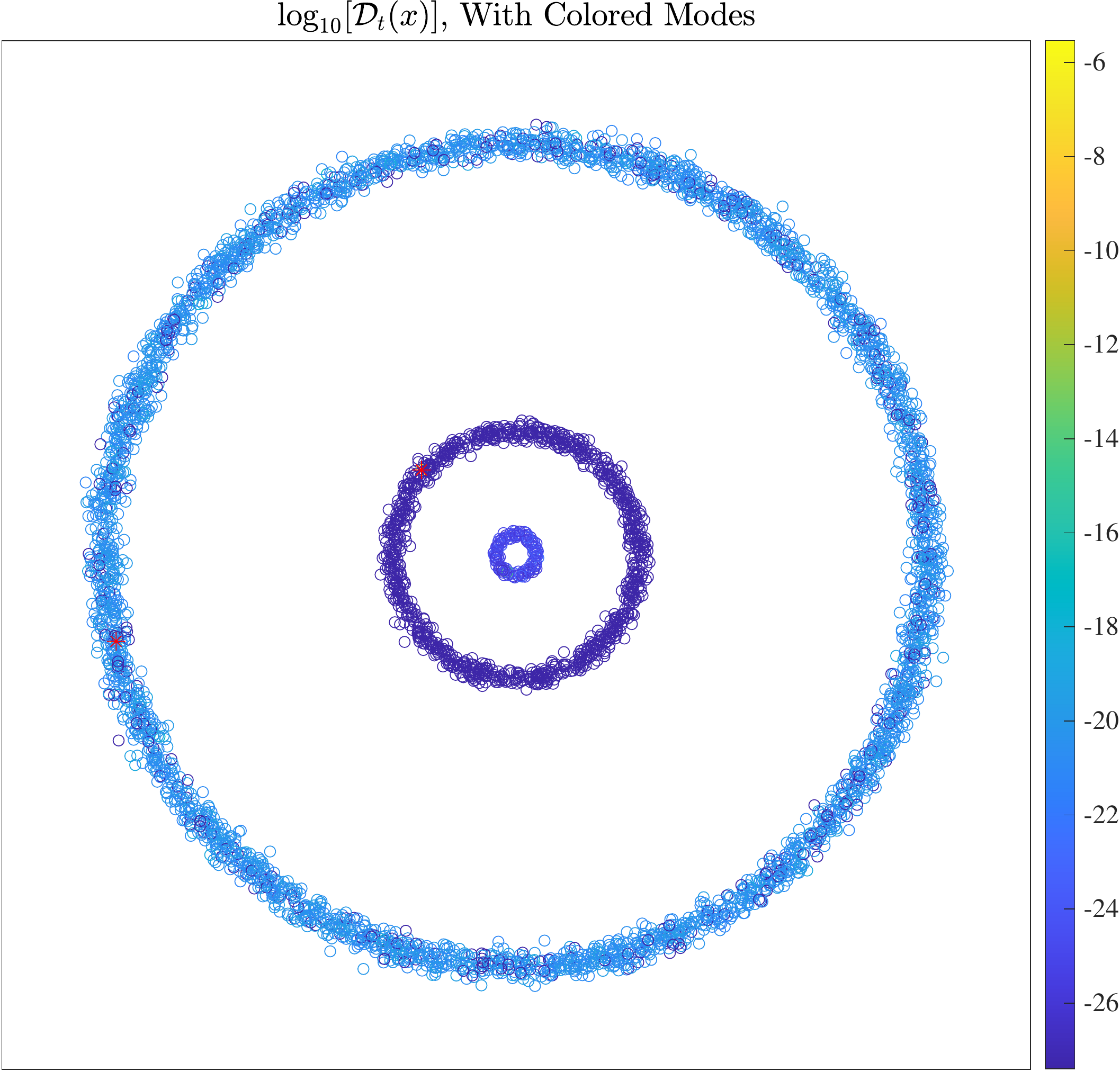}
    \end{subfigure}%
          \begin{subfigure}[t]{0.2\textwidth}
        \centering
        \includegraphics[height = 1.1in]{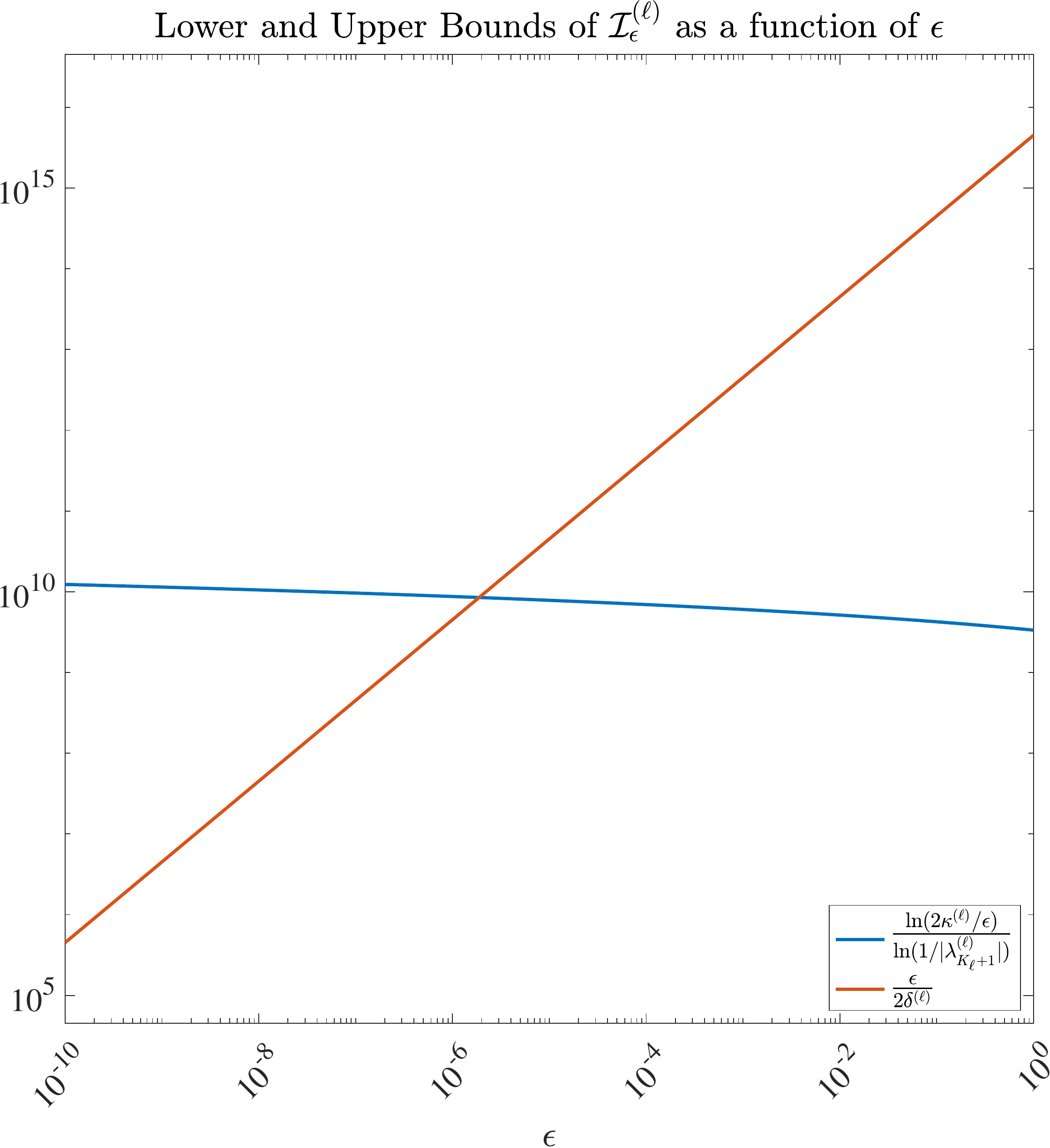}
    \end{subfigure}%
      \subcaption{LUND assignments, transition matrix, spectrum, $\mathcal{D}_t(x)$, and interval bounds for extracted clustering at time $t=2^{33}$. 2 clusters, total VI = 3.80.  }  \label{fig:nonlinear3}
\end{minipage}
\caption{Diffusion on three nested circles in $\R^2$ ($n=5380$).  Red points indicate cluster modes. Nonlinear structure is detected. The number of estimated clusters monotonically decrease with $t$. Clusterings are observed to transition as a function of when a given component of the diffusion map is annihilated. Notably, the intervals $\mathcal{I}_\epsilon^{(\ell)}$ do not intersect for any choice of $\epsilon>0$.}\label{fig: Nonlinear Diffusion}
\end{figure}

\subsection{Synthetic Nonlinear Data}
\label{sec: Nonlinear 5}

In this section, we analyze a dataset sampled from three nested rings of uniform density. We implemented the M-LUND algorithm using a complete graph. Edges were weighted using a Gaussian kernel with diffusion scale $\sigma =0.21$. The parameters we used for the KDE were $N=200$ nearest neighbors and a KDE bandwidth of $\sigma_0=3.00$.  The distance between the middle and inner rings is smaller than the distance between the outer and middle rings. In Figure \ref{fig: Nonlinear Diffusion}, we show how the labels assigned by the LUND algorithm change as a function of time. Many classical clustering algorithms (e.g., $K$-means, $K$-medoids, density peaks clustering~\cite{friedman2001elements, rodriguez2014clustering}) may not perform well on data with nonlinear structure, but the LUND algorithm returns reasonable clusterings for much of the diffusion process. For $t$ small (Figure \ref{fig:nonlinear1}, $t\in [0,2^{15}]$), diffusion has not passed a critical point at which enough higher-frequency eigenfunctions have been annihilated that diffusion distances can accurately separate cluster structure in the outer ring. Notably, because of poor separation between clusters, the interval $\mathcal{I}_\epsilon^{(\ell)}=\emptyset$ for any choice of $\epsilon>0$. In particular, this clustering is not included in $\MELD$ for any $\epsilon>0$.

When $t$ is sufficiently large, only the first four eigenfunctions contribute significantly to diffusion distances, and the LUND algorithm assigns each ring to its own cluster (Figure \ref{fig:nonlinear2}, $t\in [2^{16},2^{32}]$). Later in the diffusion process, the third and fourth coordinates of the diffusion map decay to zero as well, and the middle and inner ring clusters merge (Figure \ref{fig:nonlinear3}, $t\in [2^{33}, 2^{35}]$). Notice that, in Figures \ref{fig:nonlinear2}-\ref{fig:nonlinear3}, $\mathcal{D}_t(x)$ returns relatively small values on all $x\in X$ except cluster modes. On modal points, the value taken by $\mathcal{D}_t(x)$ is several orders of magnitude larger than that which is taken on the surrounding dataset, implying that the LUND estimate for $K_t$ is highly robust for these clusterings. Total VI was minimized for the 3-cluster clustering, which was assigned a total VI of 1.79. Conversely, the 2-cluster clustering was assigned a total VI of 3.80. For $\epsilon$ sufficiently large, there are non-intersecting intervals of time $\mathcal{I}_\epsilon^{(\ell)}$ during which each of these clusterings is $\epsilon$-separable by diffusion distances. In this sense, the MELD data model recovered from this nonlinear dataset consists of the 3-cluster and 2-cluster clusterings (Figures \ref{fig:nonlinear2}-\ref{fig:nonlinear3}).

\begin{figure}[!t] 
\begin{minipage}{\textwidth}
  \centering
    \centering
    \begin{subfigure}[t]{0.2\textwidth}
        \centering
        \includegraphics[height = 1.1in]{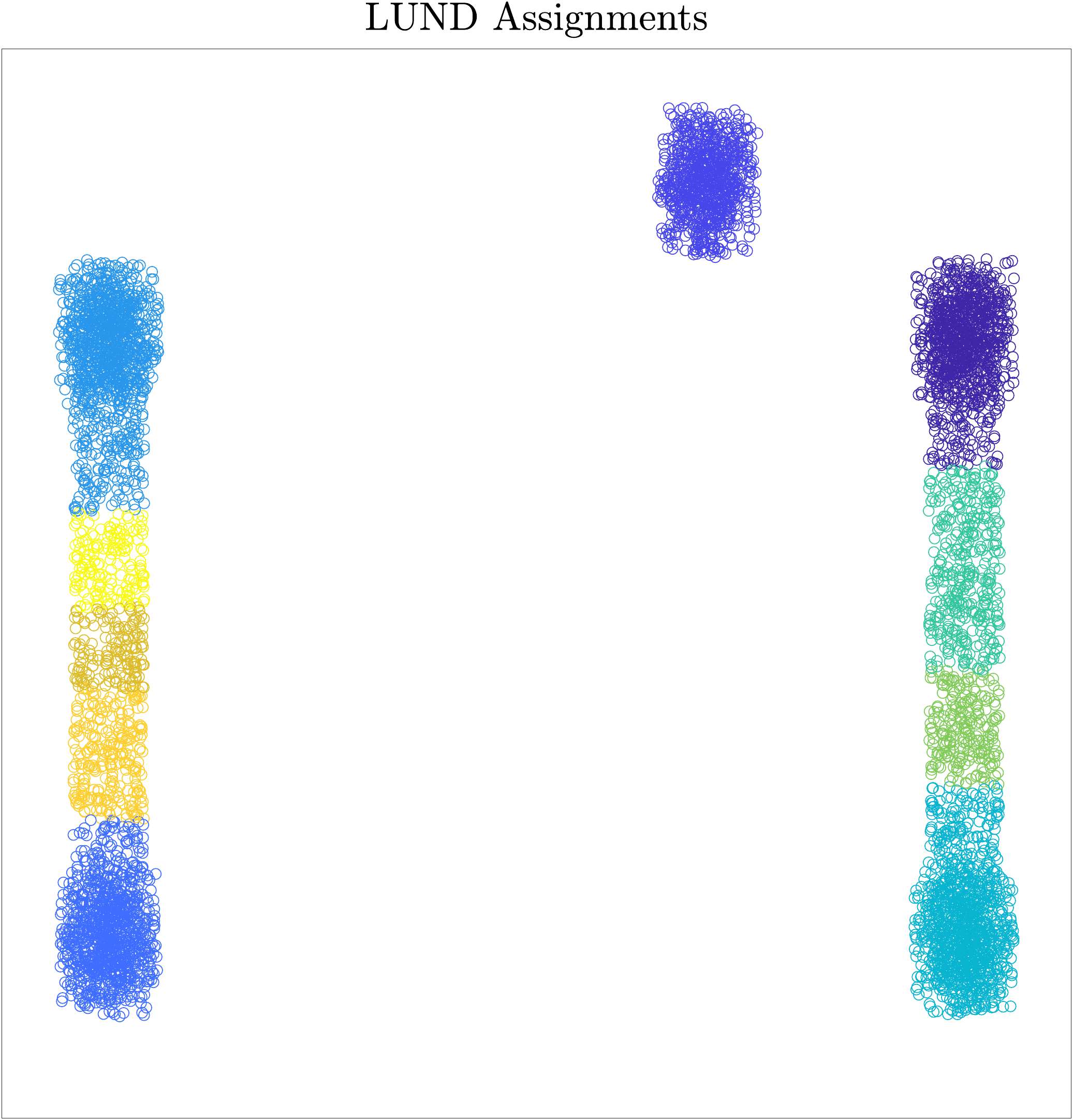}
    \end{subfigure}%
    \begin{subfigure}[t]{0.2\textwidth}
        \centering
        \includegraphics[height = 1.1in]{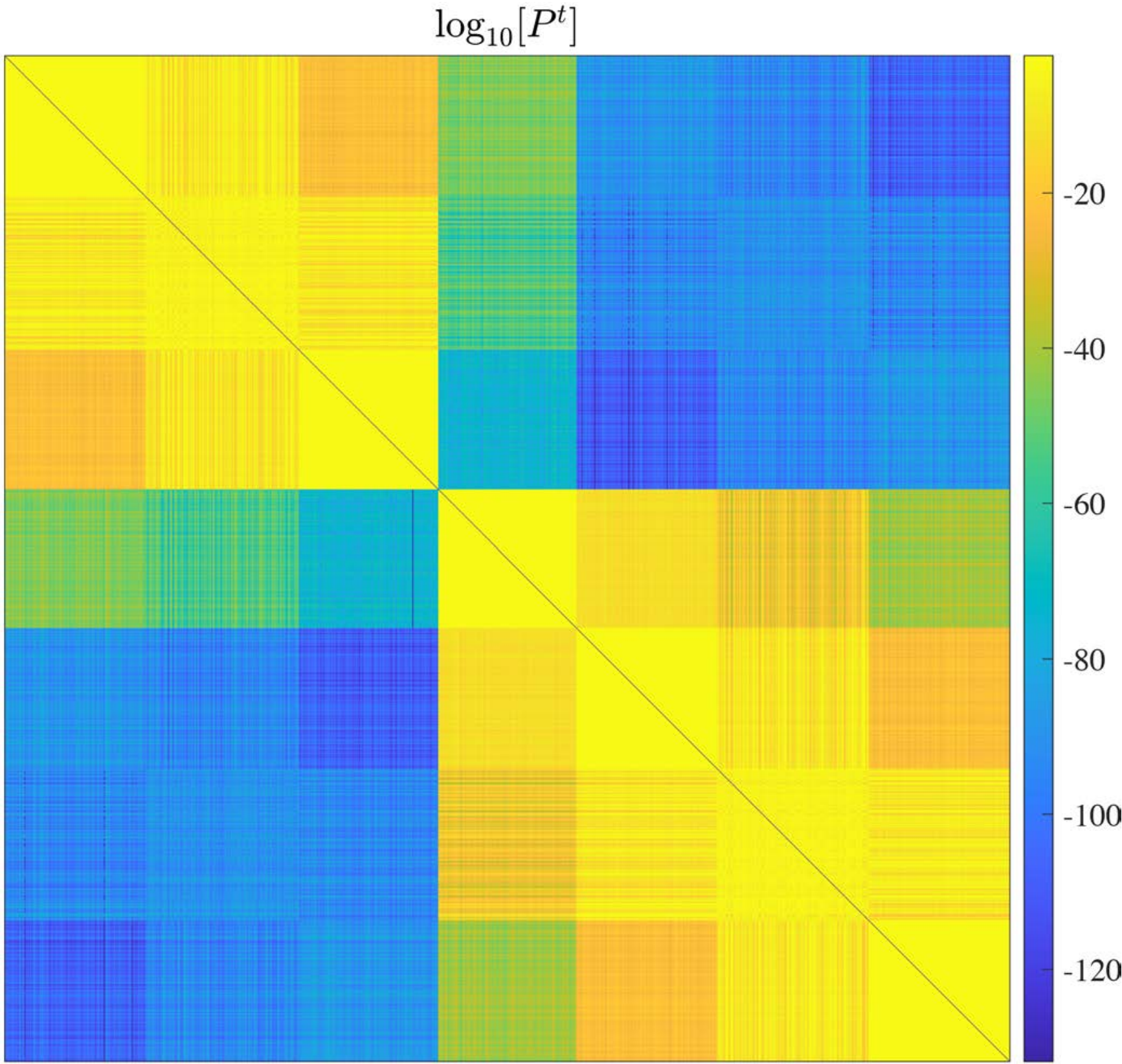}
    \end{subfigure}%
    \begin{subfigure}[t]{0.2\textwidth}
        \centering
        \includegraphics[height = 1.1in]{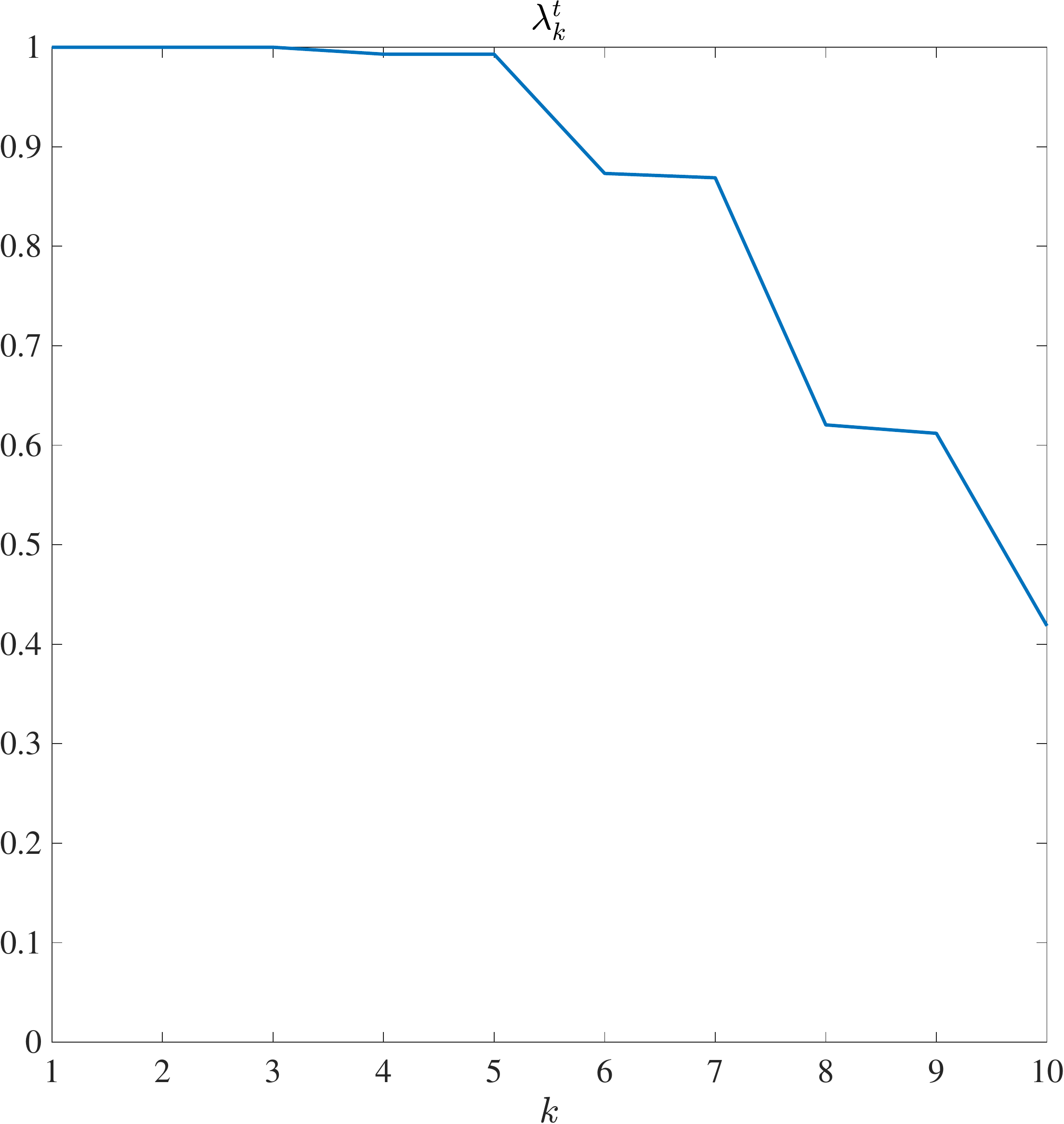}
    \end{subfigure}%
    \begin{subfigure}[t]{0.2\textwidth}
        \centering
        \includegraphics[height = 1.1in]{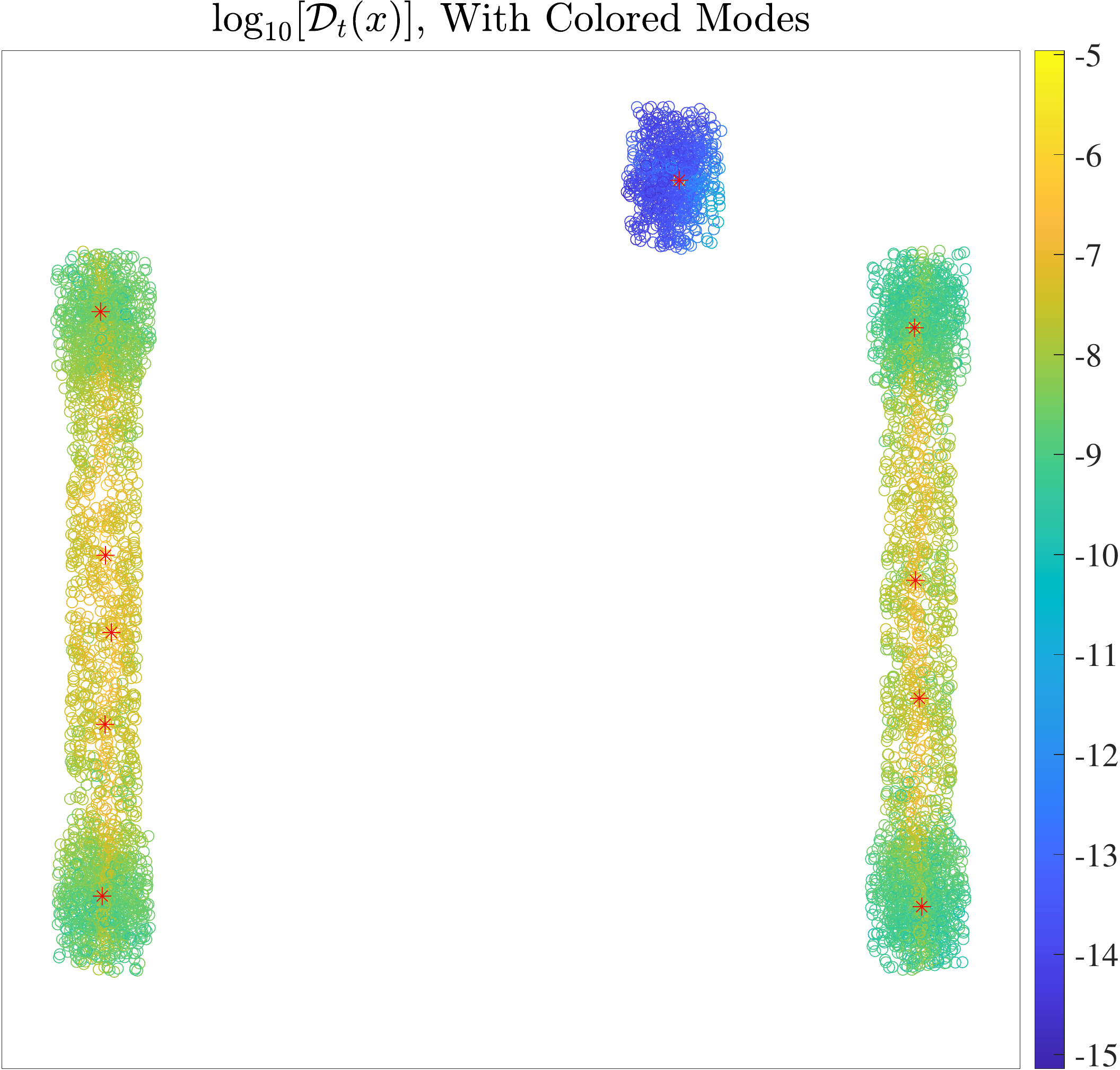}
    \end{subfigure}%
    \begin{subfigure}[t]{0.2\textwidth}
        \centering
        \includegraphics[height = 1.1in]{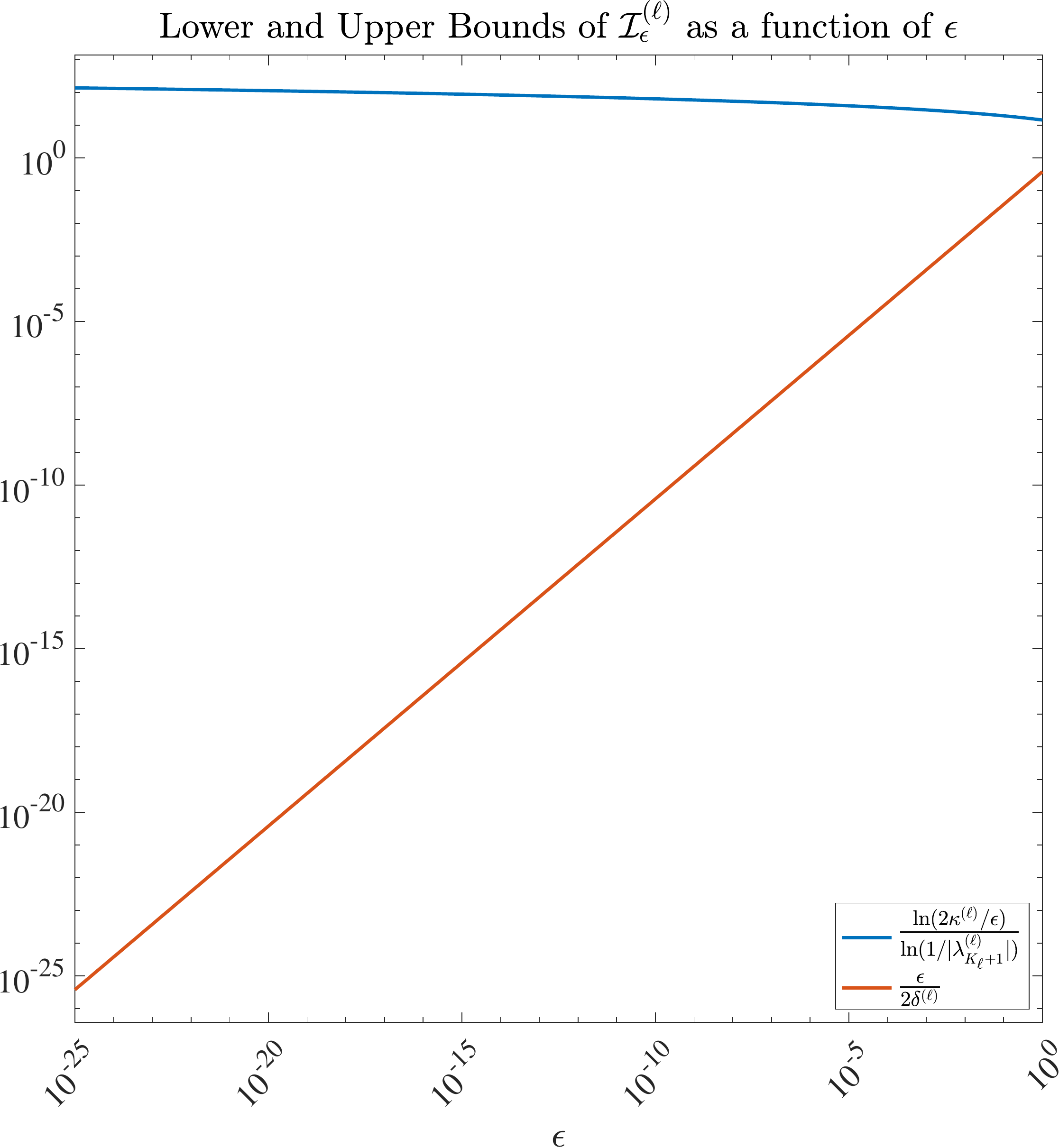}
    \end{subfigure}%
  \subcaption{LUND assignments, transition matrix, spectrum, $\mathcal{D}_t(x)$, and interval bounds for extracted clustering at time $t=2^1$. 10 clusters, total VI = 31.26.}
  \label{fig:bottleneck1}\par\vspace{0.0625in}
    \centering
       \begin{subfigure}[t]{0.2\textwidth}
        \centering
        \includegraphics[height = 1.1in]{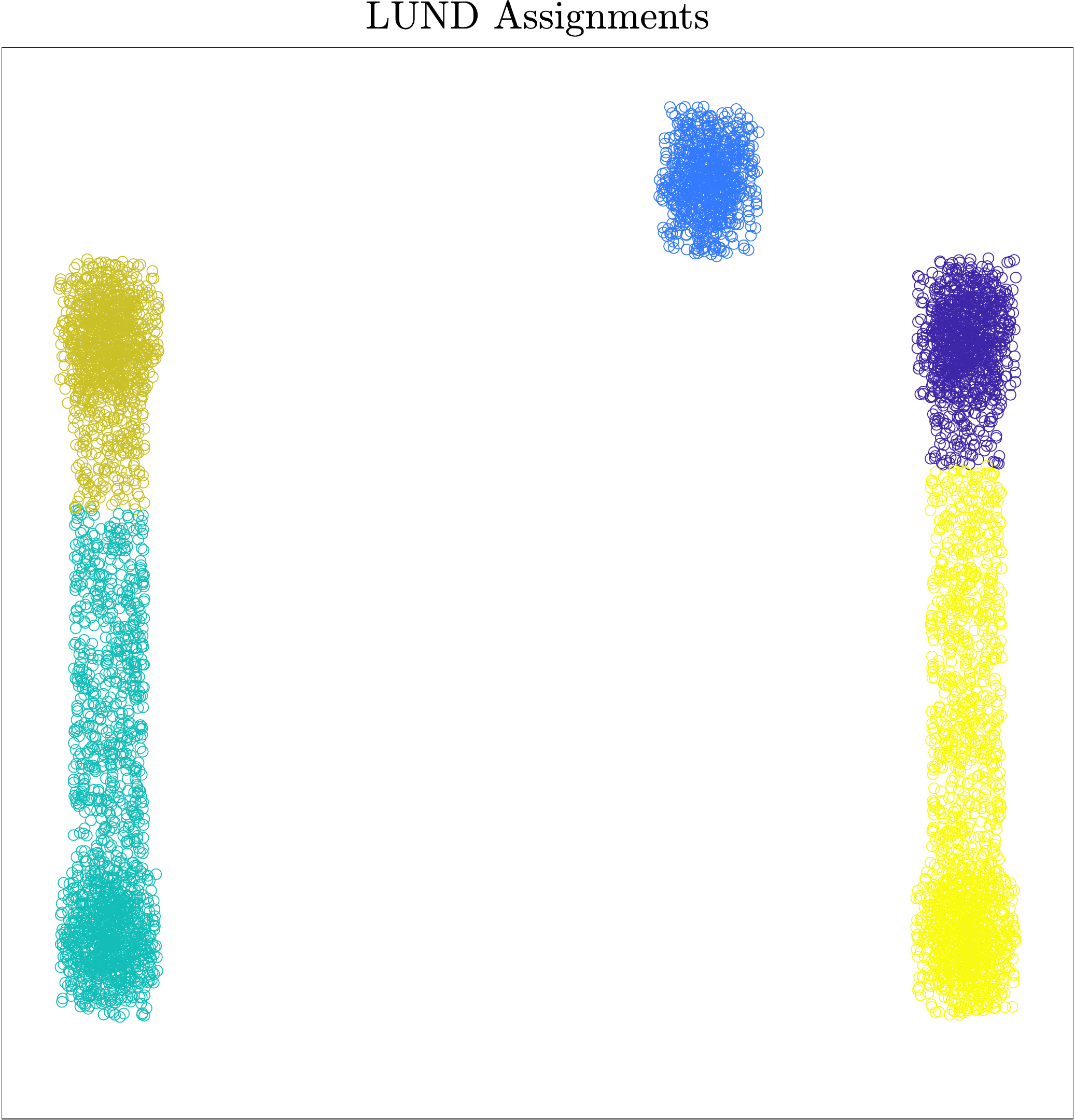}
    \end{subfigure}%
    \begin{subfigure}[t]{0.2\textwidth}
        \centering
        \includegraphics[height = 1.1in]{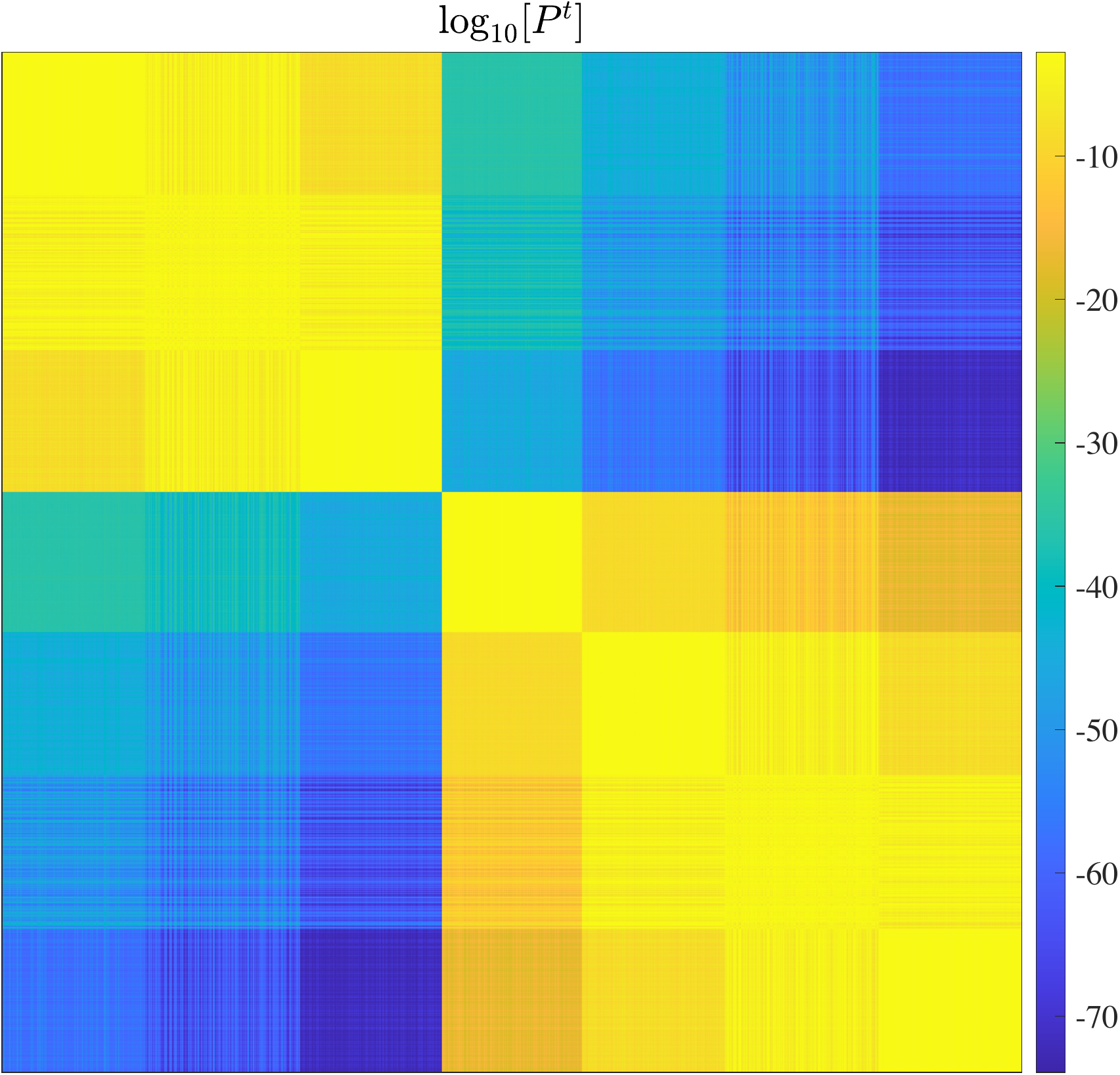}
    \end{subfigure}%
    \begin{subfigure}[t]{0.2\textwidth}
        \centering
        \includegraphics[height = 1.1in]{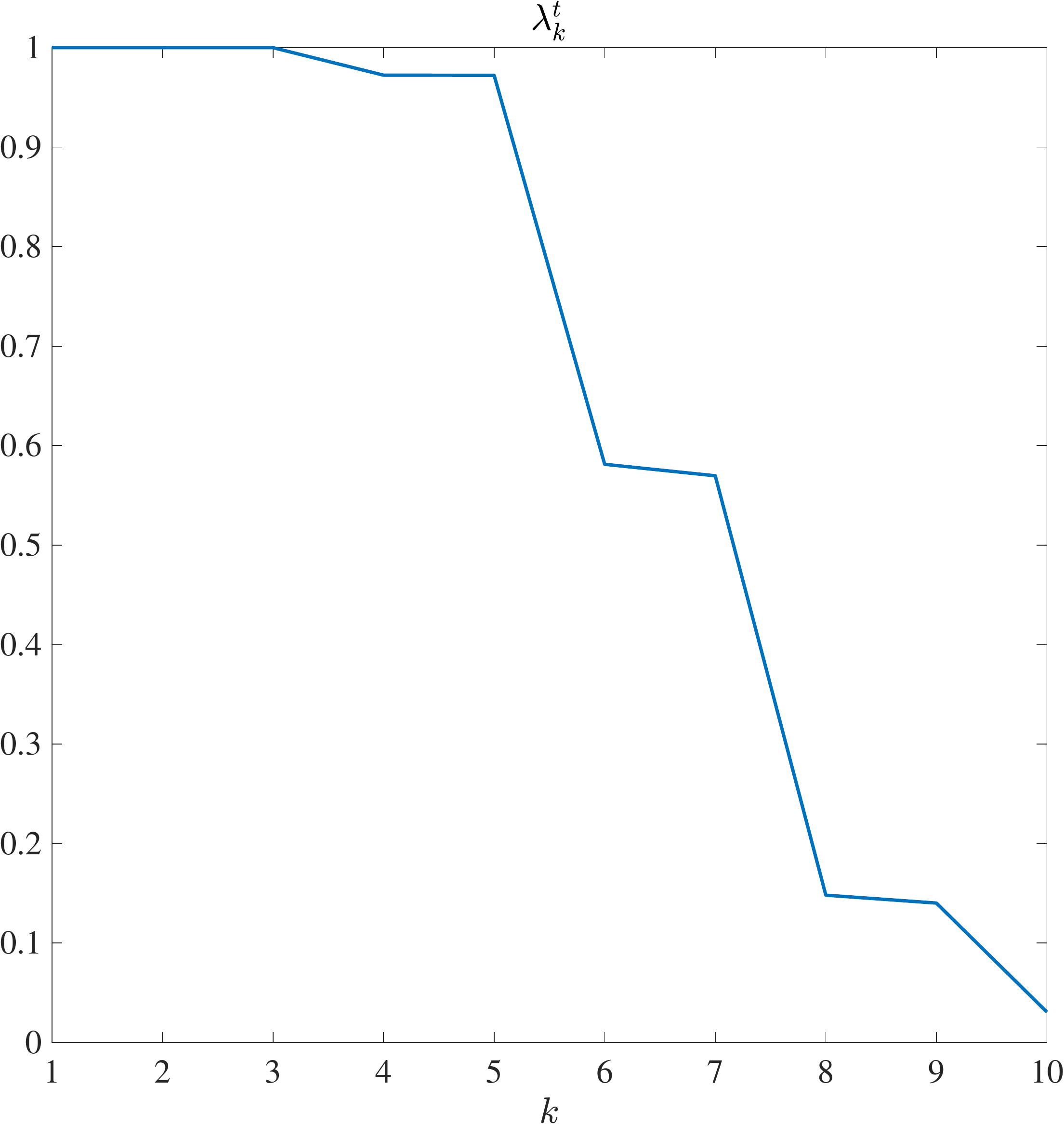}
    \end{subfigure}%
    \begin{subfigure}[t]{0.2\textwidth}
        \centering
        \includegraphics[height = 1.1in]{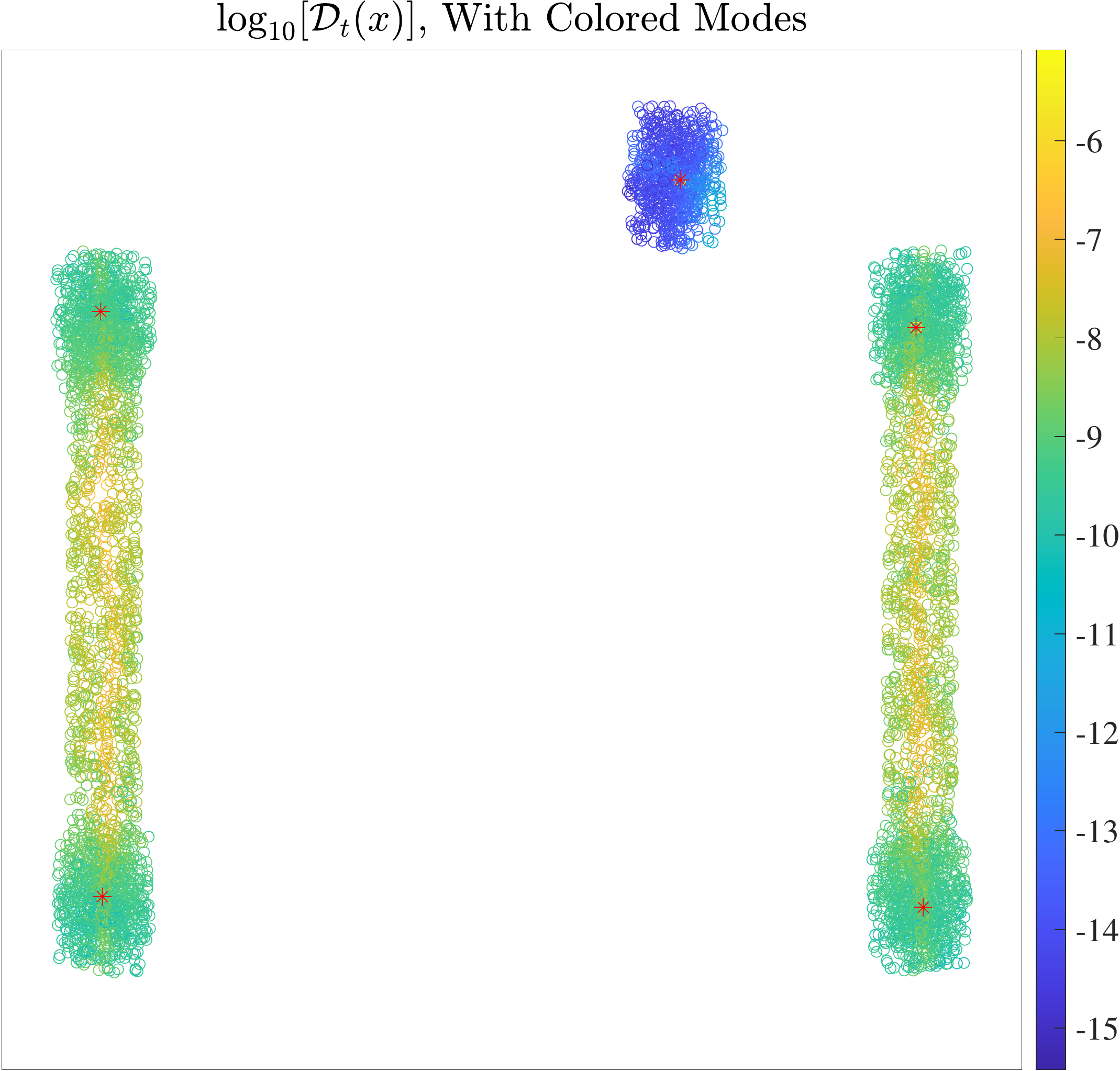}
    \end{subfigure}%
    \begin{subfigure}[t]{0.2\textwidth}
        \centering
        \includegraphics[height = 1.1in]{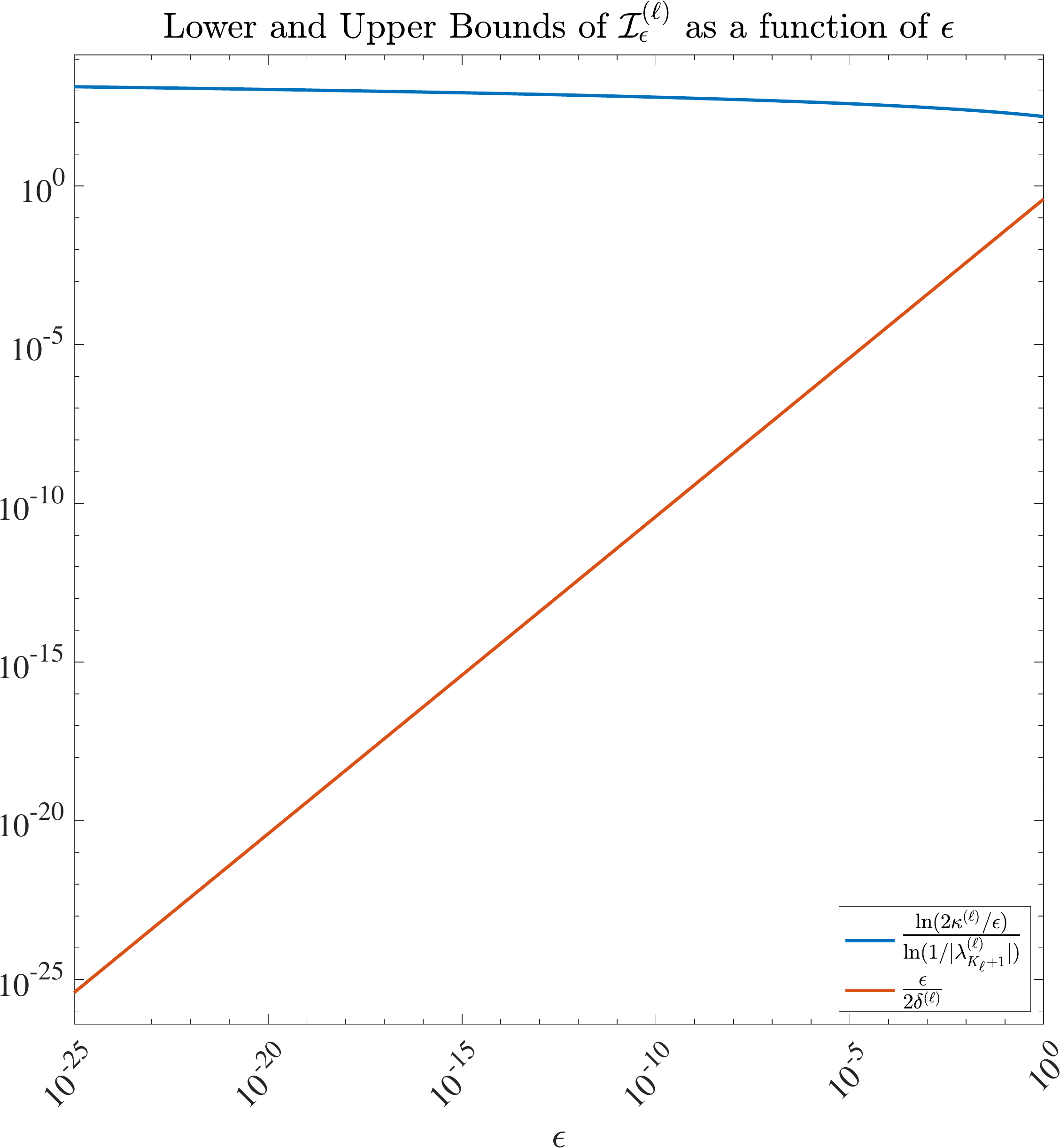}
    \end{subfigure}%
  \subcaption{LUND assignments, transition matrix, spectrum, $\mathcal{D}_t(x)$, and interval bounds for extracted clustering at time $t=2^2$. 5 clusters, total VI = 20.60.}
  \label{fig:bottleneck2}\par\vspace{0.0625in}
    \centering
    \begin{subfigure}[t]{0.2\textwidth}
        \centering
        \includegraphics[height = 1.1in]{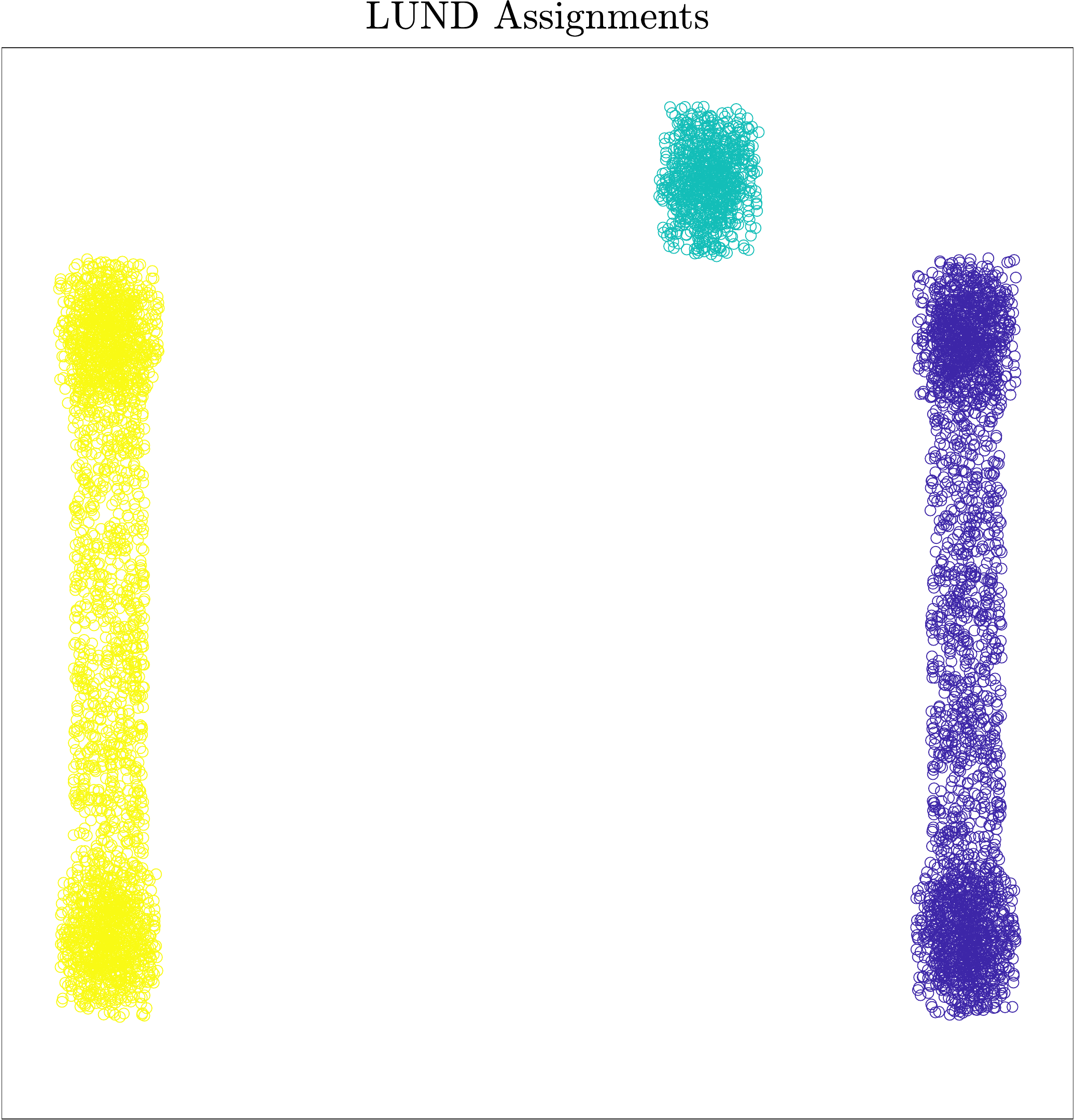}
    \end{subfigure}%
    \begin{subfigure}[t]{0.2\textwidth}
        \centering
        \includegraphics[height = 1.1in]{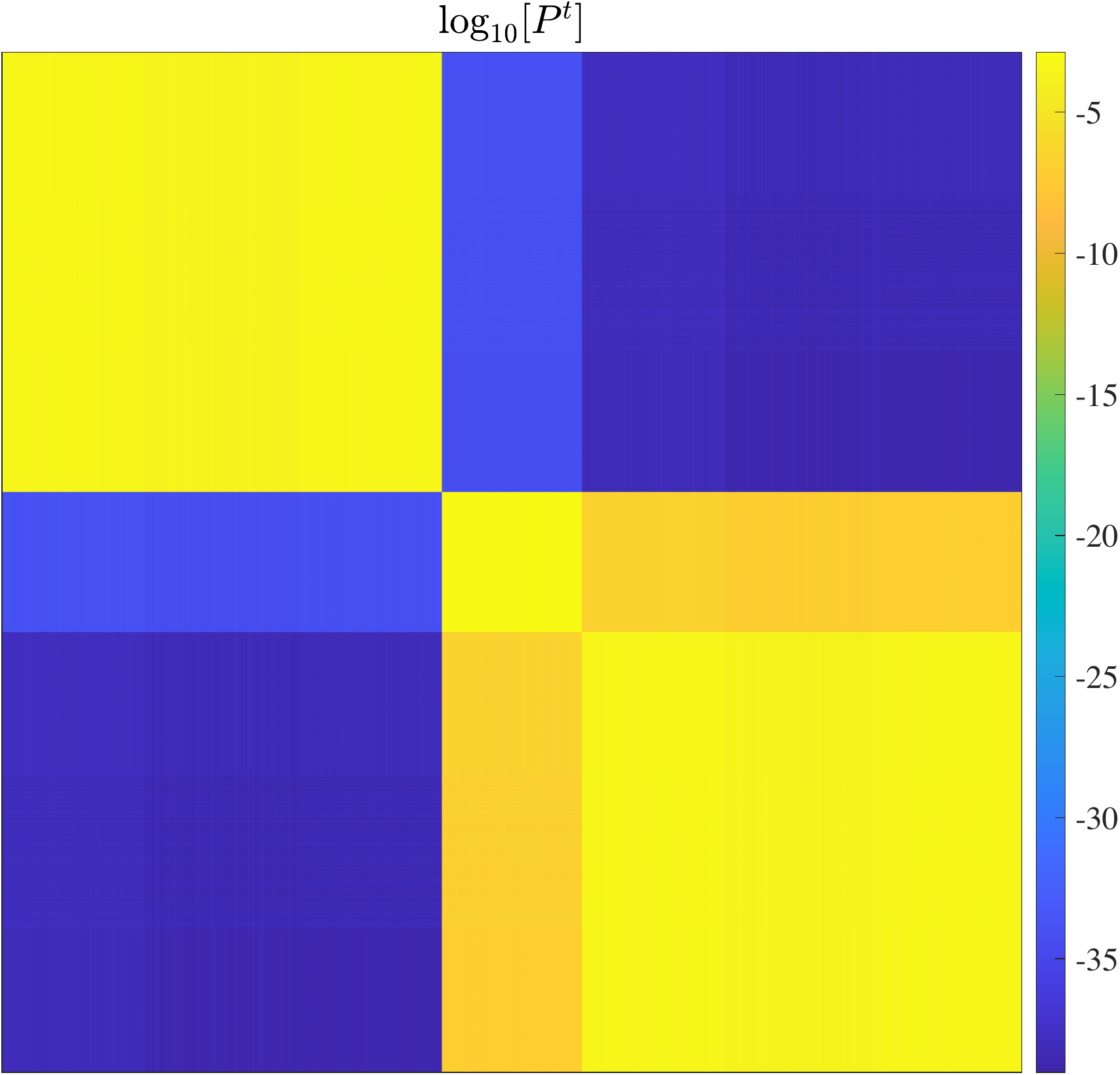}
    \end{subfigure}%
    \begin{subfigure}[t]{0.2\textwidth}
        \centering
        \includegraphics[height = 1.1in]{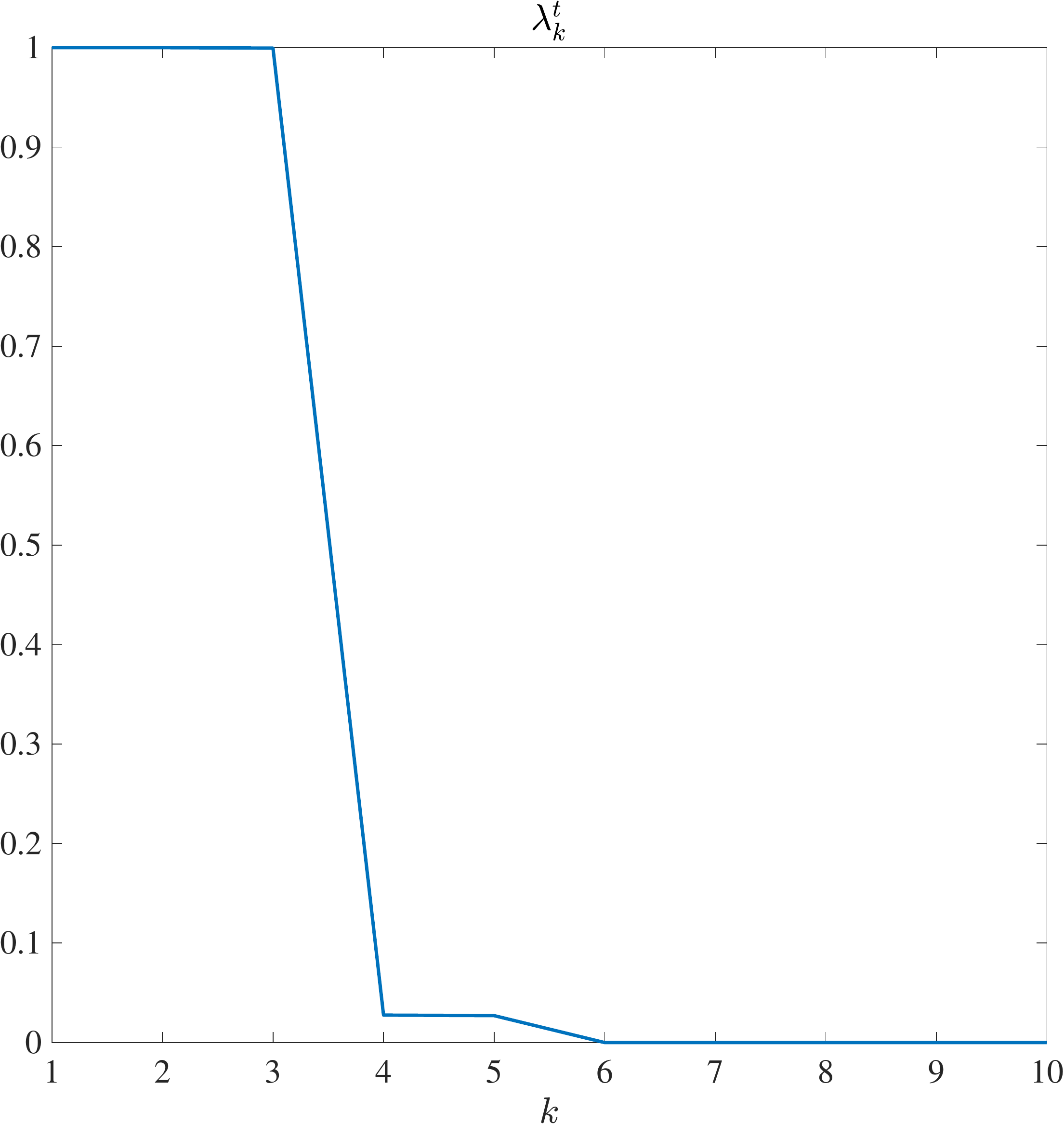}
    \end{subfigure}%
    \begin{subfigure}[t]{0.2\textwidth}
        \centering
        \includegraphics[height = 1.1in]{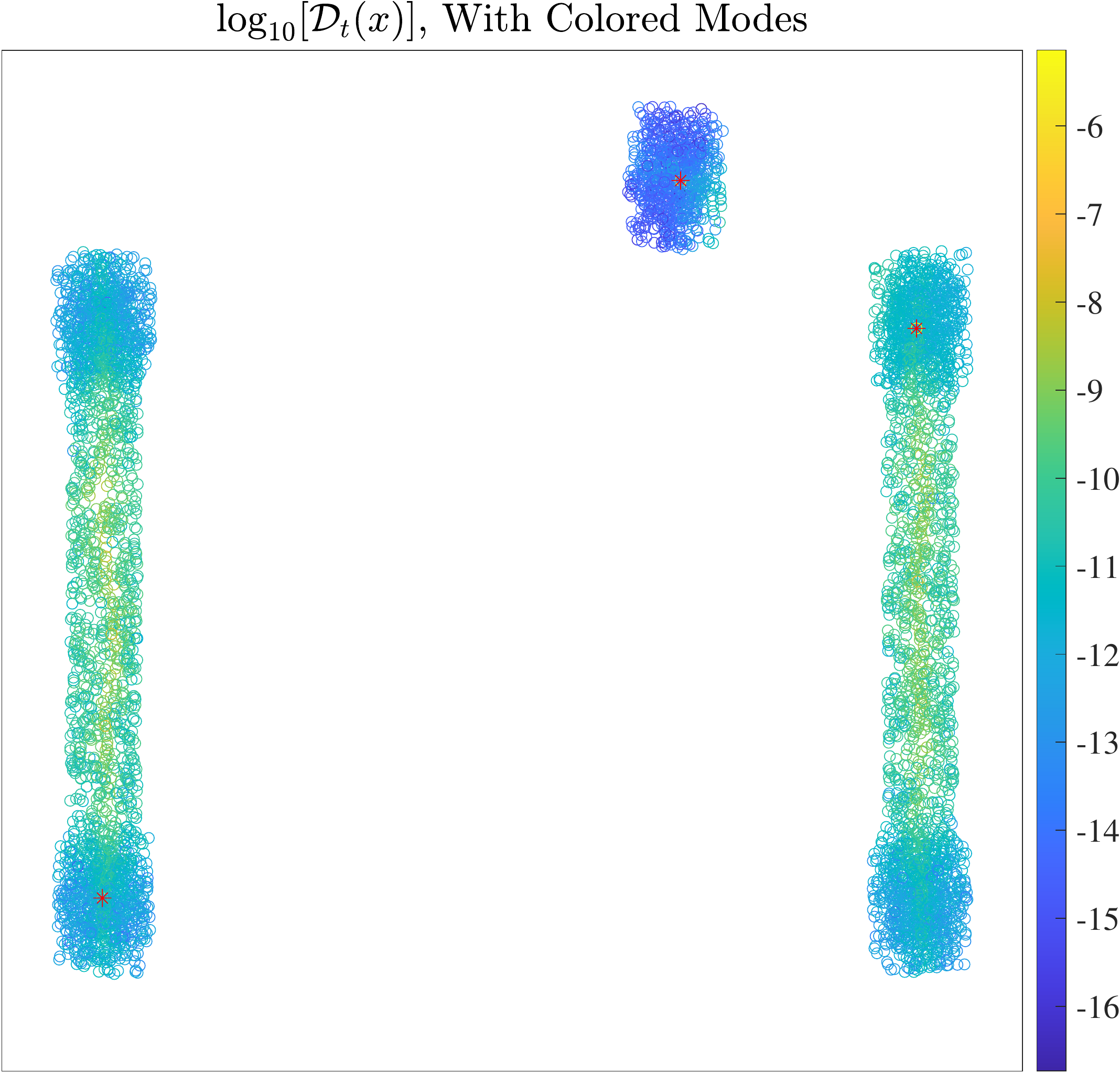}
    \end{subfigure}%
    \begin{subfigure}[t]{0.2\textwidth}
        \centering
        \includegraphics[height = 1.1in]{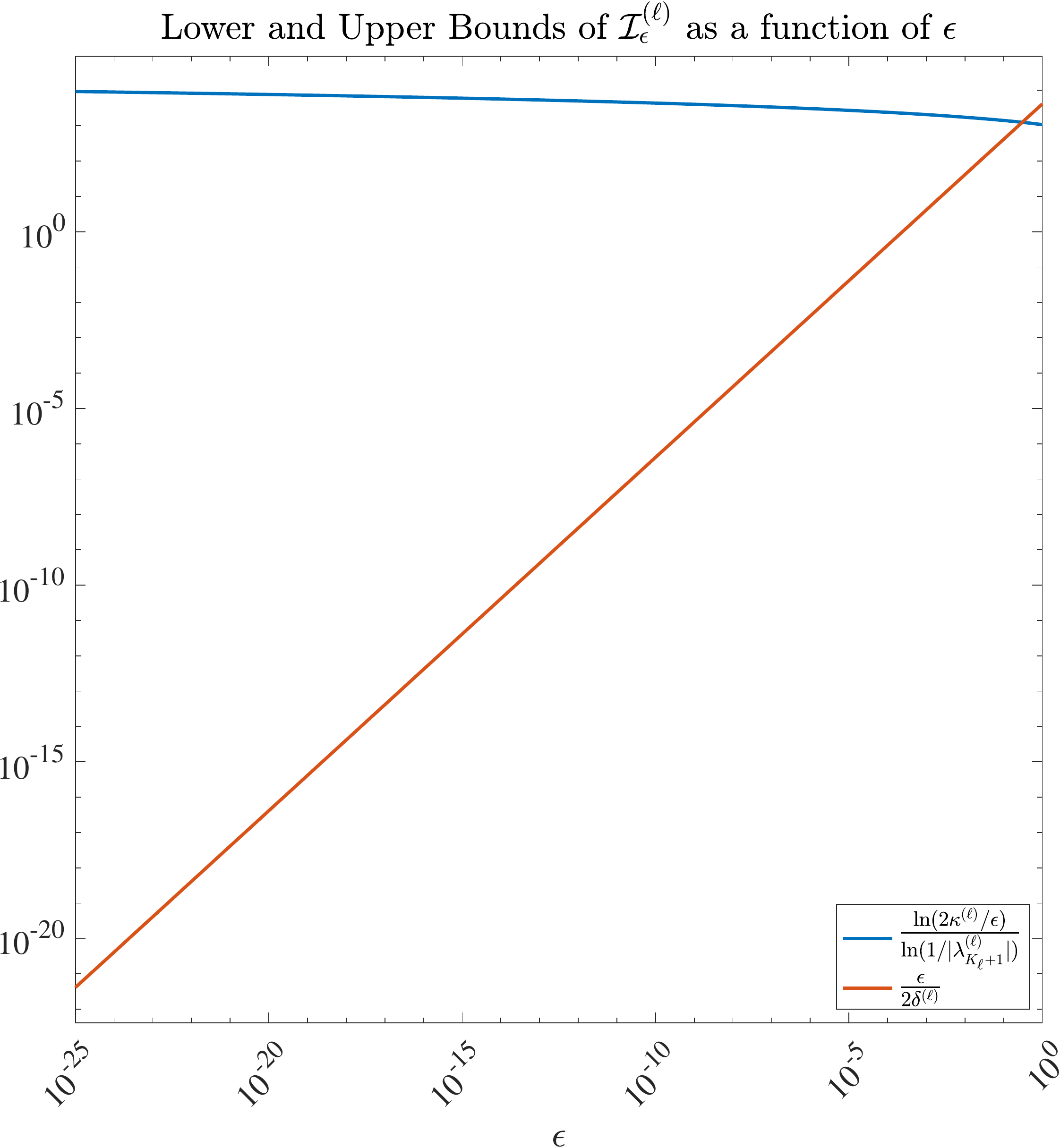}
    \end{subfigure}%
  \subcaption{LUND assignments, transition matrix, spectrum, $\mathcal{D}_t(x)$, and interval bounds for extracted clustering at time $t=2^9$. 3 clusters, total VI = 14.63. Optimal clustering.}
  \label{fig:bottleneck3}\par\vspace{0.0625in}
    \centering
    \begin{subfigure}[t]{0.2\textwidth}
        \centering
        \includegraphics[height = 1.1in]{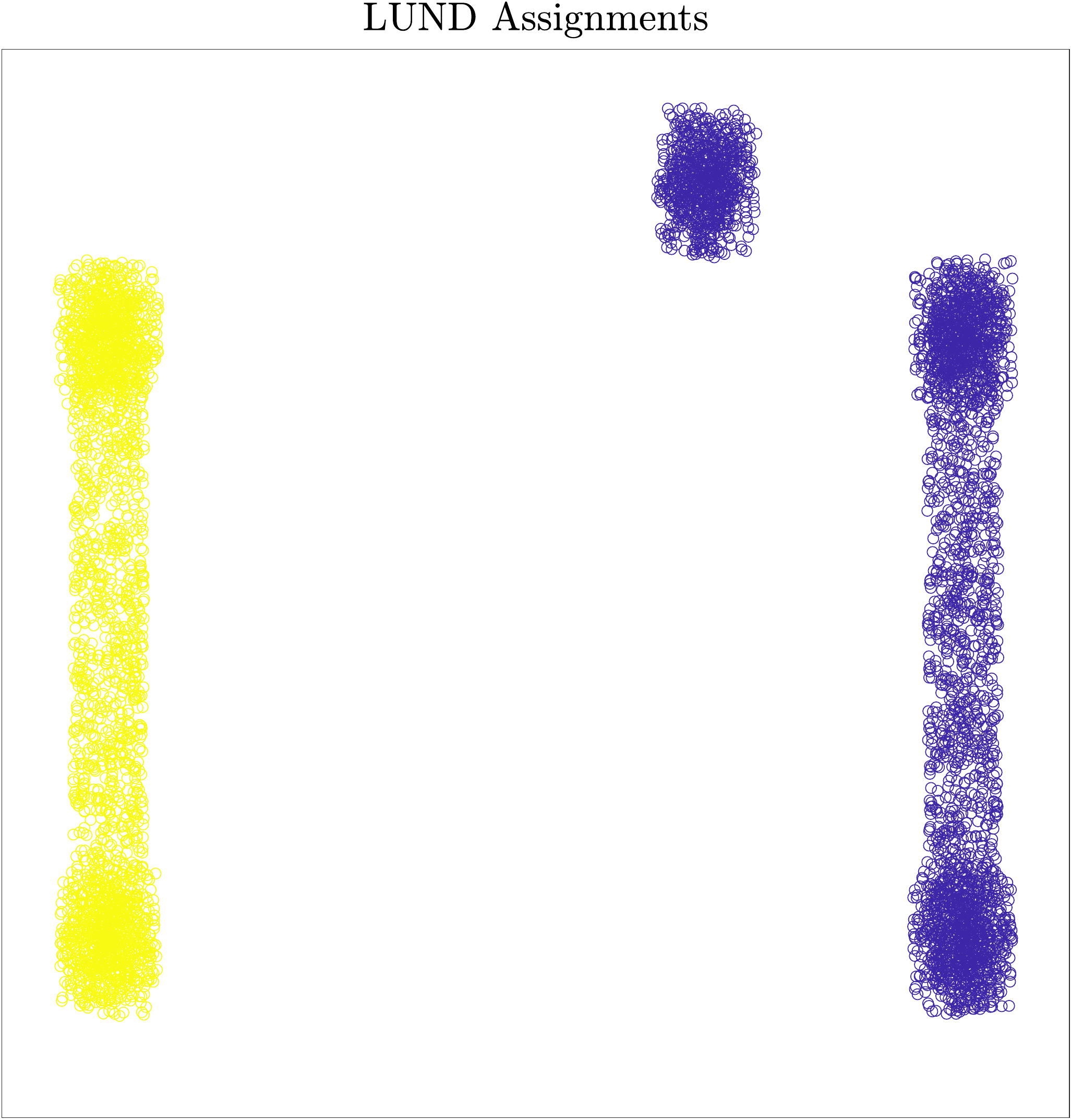}
    \end{subfigure}%
    \begin{subfigure}[t]{0.2\textwidth}
        \centering
        \includegraphics[height = 1.1in]{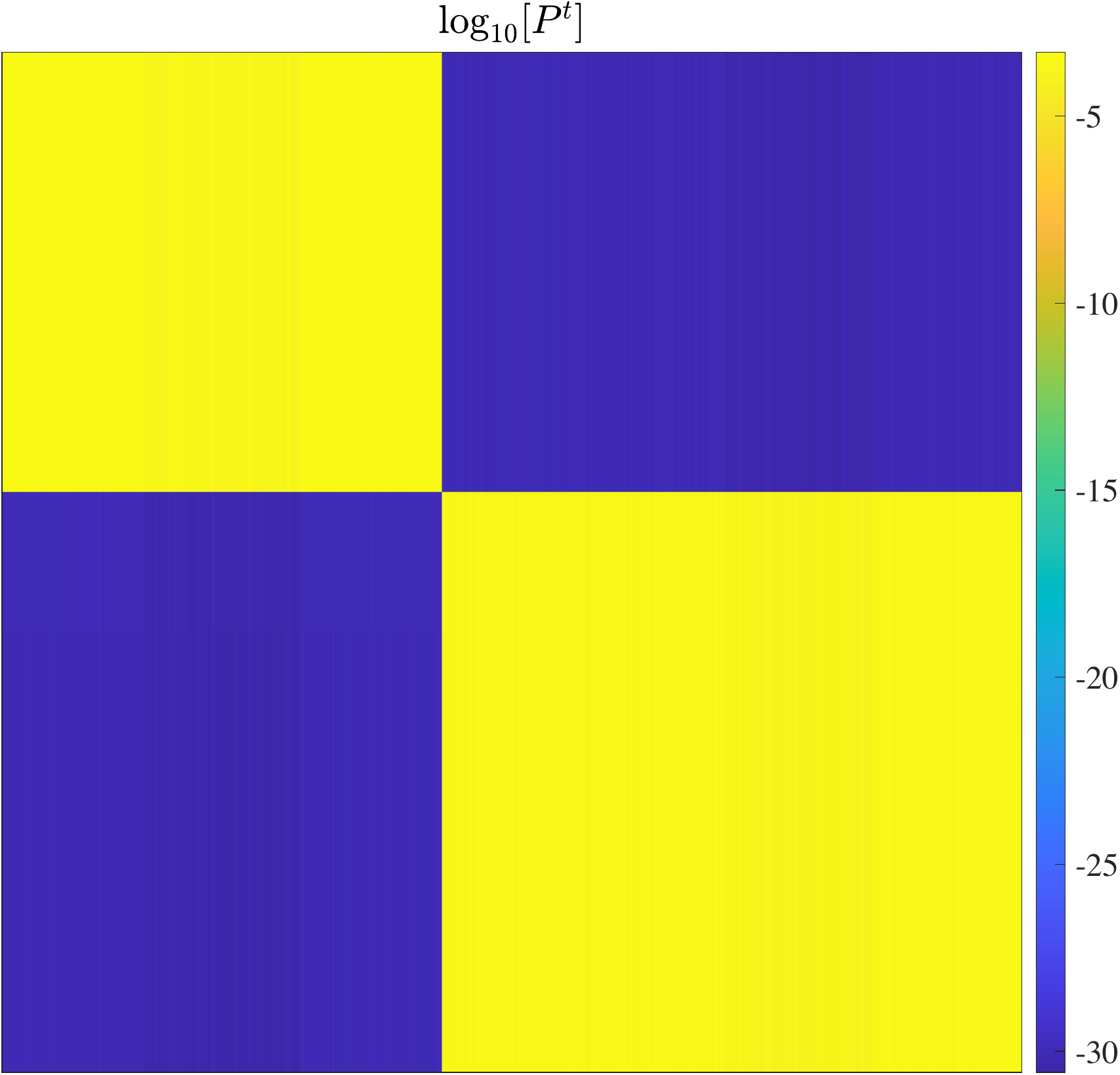}
    \end{subfigure}%
    \begin{subfigure}[t]{0.2\textwidth}
        \centering
        \includegraphics[height = 1.1in]{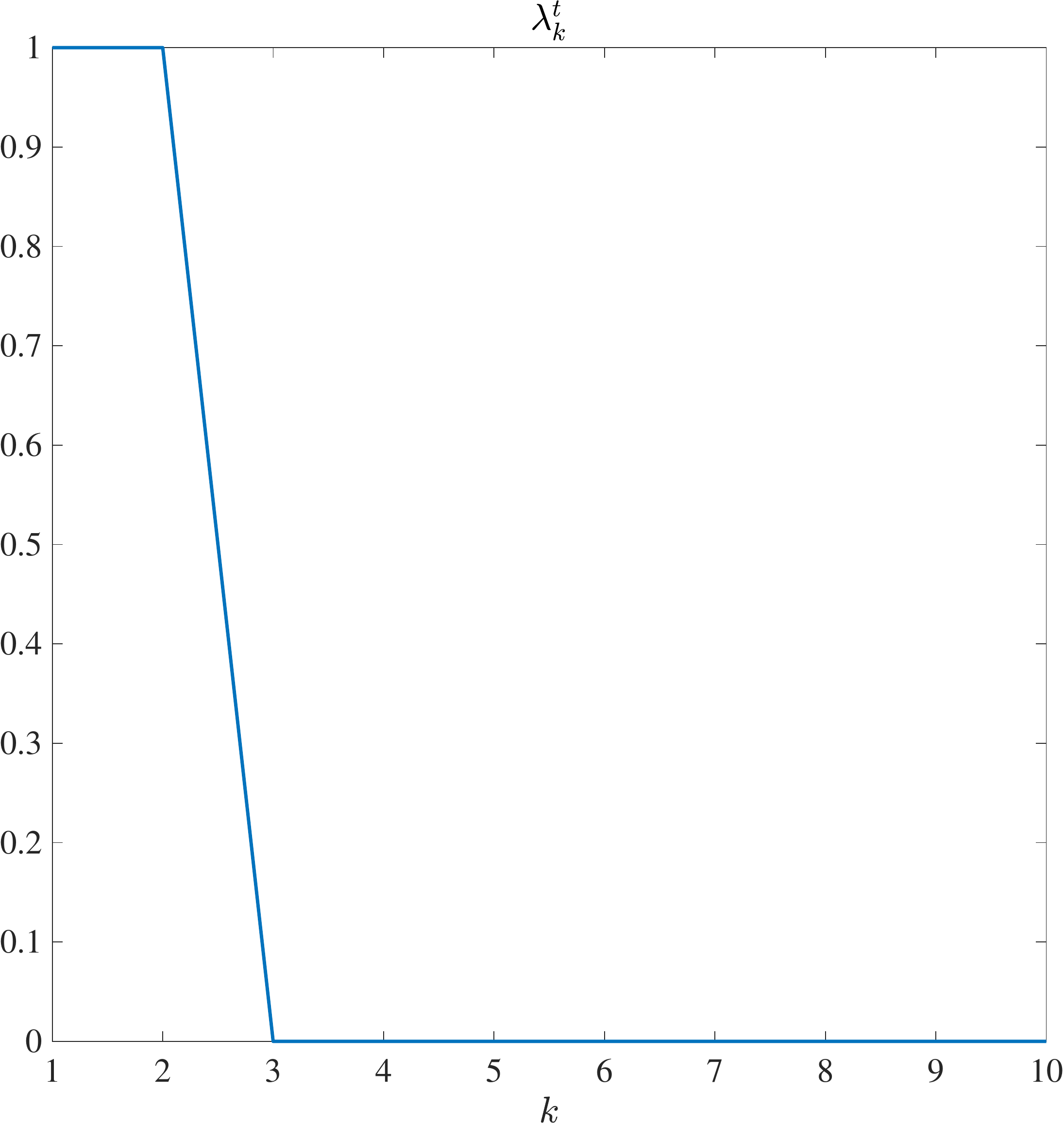}
    \end{subfigure}%
    \begin{subfigure}[t]{0.2\textwidth}
        \centering
        \includegraphics[height = 1.1in]{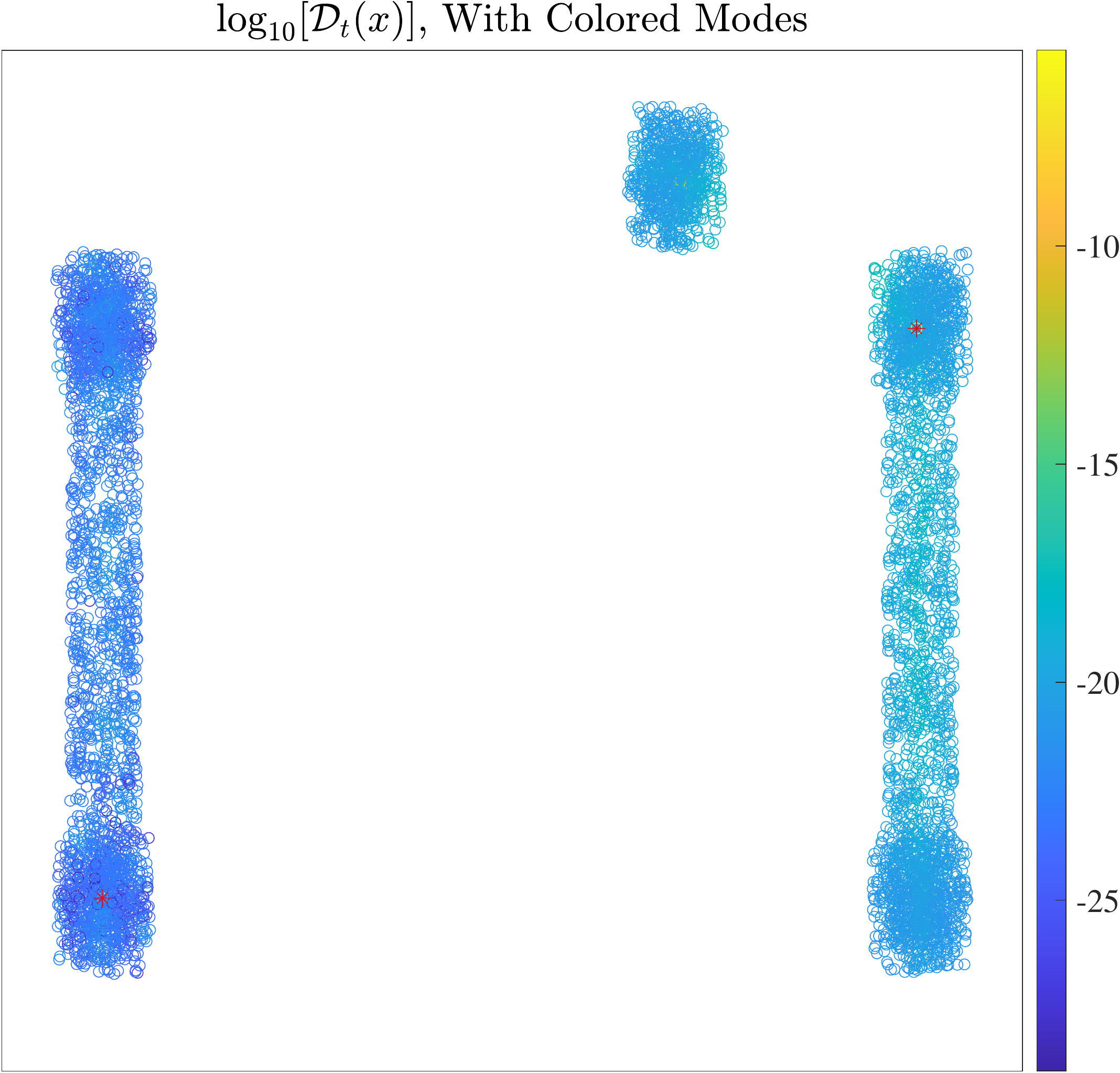}
    \end{subfigure}%
    \begin{subfigure}[t]{0.2\textwidth}
        \centering
        \includegraphics[height = 1.1in]{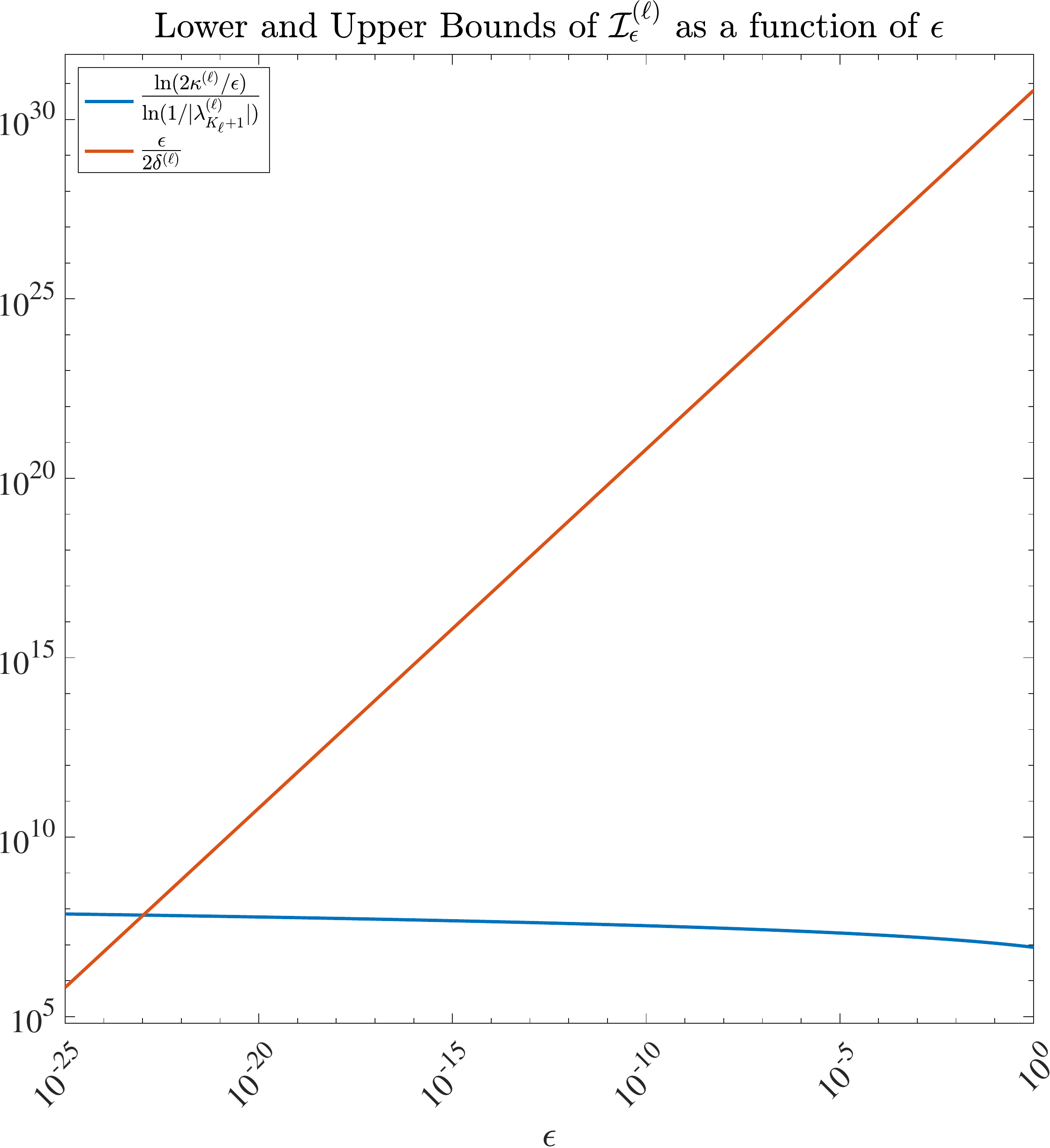}
    \end{subfigure}%
  \subcaption{LUND assignments, transition matrix, spectrum, $\mathcal{D}_t(x)$, and interval bounds for extracted clustering at time $t=2^{24}$. 2 clusters, total VI = 23.77. }
  \label{fig:bottleneck4}
\end{minipage}
\caption{Diffusion on data with bottlenecks in $\R^2$ ($n=6550$). Red points indicate cluster modes. The number of estimated clusters monotonically decreases with $t$, and clusterings are observed to transition as components of the diffusion map are annihilated. The clusterings in Figures \ref{fig:bottleneck2}-\ref{fig:bottleneck3} are $\epsilon$-separable by diffusion distances for choices of $\epsilon>0$. The intervals $\mathcal{I}_\epsilon^{(\ell)}$ do not intersect for any choice of $\epsilon>0$.}\label{fig: Bottleneck Diffusion}
\end{figure}

\subsection{Synthetic Bottleneck Data}
\label{sec: Bottleneck 5}

In this section, we analyze a dataset with bottlenecks. Each bottleneck consists of two Gaussians of the same radius, connected by data sampled from a uniform distribution. Density is the same for all Gaussians but is higher than the density of data sampled from uniform distributions. We have added an additional Gaussian that is slightly closer to the right bottleneck. We implemented the M-LUND algorithm using a complete graph with a Gaussian kernel and diffusion scale $\sigma =0.86$. The parameters we used for the KDE were $N=200$ nearest neighbors and a KDE bandwidth of $\sigma_0=0.50$.  We visualize the performance of the M-LUND algorithm in Figure \ref{fig: Bottleneck Diffusion}. For $t$ small, many higher-frequency eigenfunctions contribute to diffusion distance computations and the LUND algorithm estimates ten latent clusters with significant overlap (Figure \ref{fig:bottleneck1}, $t\in [0,2^1]$). As $t$ becomes larger, fewer higher-frequency eigenfunctions contribute to diffusion distances, and $X$ is partitioned into five clusters (Figure \ref{fig:bottleneck2}, $t\in [2^2, 2^8]$). Both of these clusterings have $\mathcal{I}_\epsilon^{(\ell)}$ empty due to poor separation between clusters.

Once higher-frequency components of the diffusion map decay to zero, the LUND algorithm detects a 3-cluster clustering in which each bottleneck is assigned to a cluster and the separated Gaussian is assigned to a cluster (Figure \ref{fig:bottleneck3}, $t\in [2^9, 2^{23}]$). After the third diffusion map coordinate is annihilated, the LUND algorithm groups the separated Gaussian with the right bottleneck (Figure \ref{fig:bottleneck4}, $t\in [2^{24}, 2^{25}]$). The minimizer of total VI is the 3-cluster clustering, which is assigned a total VI of 14.63. The clusterings in Figures \ref{fig:bottleneck3}-\ref{fig:bottleneck4} are well-separated within the original graph, and we observe that there are choices $\epsilon\in\Big(0,\frac{1}{\sqrt{n}}\Big)$ for which the intervals $\mathcal{I}_\epsilon^{(\ell)}$ are nonempty. In this sense, the unique clusterings in $\MELD$, for $\epsilon$ sufficiently large, consist of the 2-cluster and 3-cluster clusterings of $X$. As was the case for the nonlinear dataset discussed in Section \ref{sec: Nonlinear 5}, the intervals $\mathcal{I}_\epsilon^{(\ell)}$ do not intersect for any choice of $\epsilon\in\Big(0,\frac{1}{\sqrt{n}}\Big)$.

\subsection{Benchmark Real Data} \label{sec: benchmark}

In this section, we present analysis of multiscale clustering algorithms on eleven publicly-available, real-world datasets that are frequently used as benchmarks for clustering. These datasets and their ground truth labels (denoted $\mathcal{C}_G$) were obtained from the University of California, Irvine's Machine Learning Repository~\cite{DuaUCI2019}. This choice of eleven real datasets was proposed by \cite{liu2020MarkovStability}, wherein MMS clustering was compared against conventional clustering schemes. Attributes of these datasets and the parameters used to generate $\mathbf{P}$ and $p(x)$ are summarized in Table \ref{tab: benchmark summary}. 

\begin{table}[t]
    \centering
    \begin{tabular}{|l|c|c|c|c|c|}
    \hline
        \textbf{Dataset} & \textbf{Number of } & \textbf{Ambient data} & \textbf{Number of ground} & \hspace{15pt}\textbf{ Parameters }\hspace{15pt} \\
        & \textbf{samples}, $n$ &  \textbf{dimensionality,} $D$ & \textbf{truth classes}, $K$ & $N$ \hfill $\sigma$ \hfill $\sigma_0$ \\\hline
        Breast Tissue & 106 & 9 & 6 & 5 \hspace{17pt} 16 140 \hfill 1230 \\
        Control Chart & 600 & 60 & 6 & 200 \hfill 58.97 \hfill 45.05 \\
        Glass & 214 & 9 & 6 & 5 \hspace{23pt} 1.07 \hfill 0.41  \\
        Image Seg.  & 2310 & 19 & 7 & 5 \hspace{25pt} 748 \hfill 15.50  \\
        Iris & 150 & 4 & 3 & 50 \hspace{18.5pt} 1.34 \hfill 0.457 \\
        Parkinsons & 195 & 22 & 2 & 5 \hspace{25.5pt} 111 \hfill 9.96 \\
        Seeds  & 210 & 7 & 3 & 100 \hspace{14.5pt} 0.91 \hfill 1.09 \\
        Vertebral  & 310 & 6 & 3 & 5 \hspace{21.5pt}  18.15 \hfill 12.39\\
        WBCD  & 569 & 30 & 2 & 20 \hfill 234 \hfill 283 \\
        Wine & 178 & 13 & 3 &   50 \hspace{16.5pt} 78.57 \hfill 117.56 \\
        Yeast  & 1484 & 8 & 10 & 10  \hspace{20pt} 0.69 \hfill 0.35  \\\hline 
    \end{tabular}
    \caption{Summary of the eleven benchmark datasets analyzed. All datasets were obtained from the University of California, Irvine's Machine Learning Repository~\cite{DuaUCI2019}. Graph and KDE parameters for this section's analysis are stored in the rightmost column. Here, $N$ denotes the number of nearest neighbors, while $\sigma$ and $\sigma_0$ are the diffusion scale and KDE bandwidth respectively.}
    \label{tab: benchmark summary}
\end{table}

The \emph{normalized mutual information} (\emph{NMI}) between an estimated clustering and the ground truth labels is used as the performance measure of the clusterings in the dataset in this section. NMI, which is defined by $NMI(\mathcal{C}, \mathcal{C}') =\sqrt{\frac{I(\mathcal{C}, \mathcal{C}')^2}{H(\mathcal{C})H( \mathcal{C}')}}$, is a measure of similarity between two clusterings, ranging $[0,1]$. NMI is closely related to VI, and it can be shown that $NMI(\mathcal{C}, \mathcal{C}') = 1$ if and only if $VI(\mathcal{C}, \mathcal{C}')=0$ (i.e., $\mathcal{C}=\mathcal{C}'$). Thus, if $NMI(\mathcal{C}, \mathcal{C}_G)$ is near 1, the clustering $\mathcal{C}$ is very close in VI to the ground truth labels. Conversely, $NMI(\mathcal{C}, \mathcal{C}')=0$ if and only if the random variables associated with the clusterings $\mathcal{C}$ and $\mathcal{C}'$ are independent; i.e., observing $\mathcal{C}$ yields no new information about the clustering $\mathcal{C}'$. Thus, if  $NMI(\mathcal{C}, \mathcal{C}_G)$ is near 0, there is only a weak relationship between $\mathcal{C}$ and the ground truth labels, and $\text{VI}(\mathcal{C}, \mathcal{C}_G)$ will be large. 

We first compare M-LUND against related algorithms (MMS, HSC, SLC, and $K$-means~\cite{liu2020MarkovStability, ng2002spectralclustering, friedman2001elements}) at a fixed scale by setting the number of clusters $K$ to be the number of ground truth classes in $\mathcal{C}_G$.  Diffusion-based algorithms (M-LUND, MMS, and HSC) were implemented using the same KNN graph with edges weighted with a Gaussian kernel. Graph parameters are summarized in Table \ref{tab: benchmark summary}, and the results of this analysis are provided in Table \ref{tab: benchmark fixed scale}. We compared performances using the two-sided paired-sample $t$-test~\cite{rice2006mathematical}, which tests the null hypothesis that the difference in performance between the M-LUND algorithm and its competitors is distributed normally with mean zero. Under this null hypothesis, the test statistic (denoted $t_S$) follows a Student's $t$-distribution with $n_D-1$ degrees of freedom, where $n_D$ reflects the number of datasets on which these algorithms were evaluated. We rejected the null hypothesis when comparing the M-LUND algorithm against each of MMS, HSC, SLC, and $K$-means at the $\alpha=0.05$ significance level. Thus, the M-LUND algorithm produces clusterings that are significantly closer to ground truth labels than those produced by the other four algorithms. 

\begin{table}[t]
    \begin{tabular}{|l|c|c|c|c|c|}
    \hline
        \textbf{Dataset} & \textbf{M-LUND} & \textbf{MMS} & \textbf{HSC} & \textbf{SLC} & \textbf{$K$-Means }  \\ \hline
        Breast Tissue &             \textbf{0.415} & 0.402 & 0.375 & 0.122 & 0.290 \\ 
	    Control Chart &             \textbf{0.806} & ---   & 0.714 & 0.695 & 0.690 \\ 
        Glass &                     \textbf{0.427} & 0.400 & 0.254 & 0.072 & 0.378 \\
        Image Seg. &                \textbf{0.706} & 0.638 & 0.543 & 0.366 & 0.523 \\
        Iris &                      \textbf{0.901} & 0.743 & 0.766 & 0.718 & 0.758 \\
        Parkinsons &                \textbf{0.235} & ---   & \textbf{0.235} & 0.005 & 0.126 \\ 
        Seeds &                     \textbf{0.739} & 0.732 & 0.667 & 0.067 & 0.695 \\
    Vertebral  &                    \textbf{0.574}  & 0.550 & 0.515 & 0.009 & 0.407 \\ 
	    WBCD &                      \textbf{0.498} & 0.403 & 0.473 & 0.005 & 0.465 \\
        Wine &                      \textbf{0.450} & 0.435 & 0.405 & 0.062 & 0.242 \\ 
        Yeast &                     \textbf{0.360} & 0.283 & 0.270 & 0.066 & 0.242 \\  
\hline    \textbf{Average} &        \textbf{0.563} & 0.510 & 0.474 & 0.165 & 0.465 \\ 
\hline
\end{tabular}
\caption{Comparison of clustering algorithms' performance on eleven benchmark datasets when $K$ is fixed to be the number of clusters in $\mathcal{C}_G$. MMS clustering did not learn a $K$-cluster clustering from the Control Chart and Parkinsons datasets, so we did not include performances on these datasets in averages or statistical tests.  We used the NMI between the outputted $K$-cluster clustering and ground truth labels to measure performance on a dataset. Thus, a high value reflects more similarity to the ground truth labels in the dataset.  On average, the $K$-cluster M-LUND clustering is significantly closer to the ground truth labels than the $K$-cluster clusterings produced by the algorithms we compare against: MMS~($p=~1.25 \times~10^{-2}$,~$t_S=~3.21$), HSC~($p=~1.09\times 10^{-3}$, $t_S= 4.97$), SLC~($p=~5.01\times~10^{-5}$, $t_S=7.85$), and $K$-means~($p=~5.69\times 10^{-3}$, $t_S=3.74$).} 
 \label{tab: benchmark fixed scale}
\end{table}

We next compare the M-LUND algorithm against related algorithms (MMS, HSC, and SLC~\cite{liu2020MarkovStability, ng2002spectralclustering, friedman2001elements}) in a multiscale setting. As before, diffusion-based algorithms were implemented using the same graph as well as the same exponential time sampling of the diffusion process. To evaluate an algorithm's performance, we compare the optimal clustering it outputs to the ground truth labels. For M-LUND, MMS, and HSC, we select the clustering that minimizes total VI as the optimal outputted clustering. SLC does not rely on a diffusion process to generate its clusterings, so we select  $\mathcal{C}^{(\ell^*)}= \text{argmin}_{2\leq \ell \leq n/2} L^{\text{in}}(\mathcal{C}^{(\ell)})/L^{\text{btw}}(\mathcal{C}^{(\ell)})$, where $L^{\text{in}}(\mathcal{C}^{(\ell)})~=~\max_{1\leq k \leq \ell}\max_{x,y\in X_k^{(\ell)}} \|x-y\|_2$ is the maximum within-cluster Euclidean distance for the $\ell$-cluster clustering in the dendogram, denoted $\mathcal{C}^{(\ell)}$, and $L^{\text{btw}}(\mathcal{C}^{(\ell)}) = \min_{1\leq k <k' \leq \ell} \mathcal{L}_{SLC}(X_k^{(\ell)}, X_{k'}^{(\ell)})$ is the minimum between-cluster value taken by the SLC linkage function. 

Table \ref{tab: benchmark multiscale} indicates that the M-LUND algorithm generates clusterings that are, on average, closer to the ground truth labels than those assigned by the algorithms it is compared against. Indeed, the performance achieved by the M-LUND algorithm is greater than or equal to its competitors across all datasets. As before, we compared performances using the two-sided, paired-sample $t$-test~\cite{rice2006mathematical}, testing the null hypothesis that the difference in performance between the M-LUND algorithm and its competitors is distributed normally with mean zero. We again rejected this null hypothesis when comparing the M-LUND algorithm against each of MMS, HSC, and SLC at the $\alpha=0.05$ significance level. Thus, in both the fixed-scale and multiscale settings, the M-LUND algorithm produces clusterings that are significantly closer to the ground truth labels compared to those generated by related algorithms.

\begin{table}[t]
    \begin{tabular}{|l|c|c|c|c|}
    \hline
        \textbf{Dataset} & \textbf{M-LUND} & \textbf{MMS} & \textbf{HSC} & \textbf{SLC}   \\ \hline
        Breast Tissue &        \textbf{0.480} & 0.465 & 0.000 & 0.016 \\
	    Control Chart &        \textbf{0.760} & \textbf{0.760} & 0.712 & 0.571 \\ 
        Glass &                \textbf{0.468} & 0.365 & 0.034 & 0.344 \\
        Image Seg. &           \textbf{0.630} & 0.023 & 0.437 & 0.009 \\ 
        Iris &                 \textbf{0.734} & \textbf{0.734} & \textbf{0.734 } & \textbf{0.734} \\
        Parkinsons &           \textbf{0.235} & 0.114 & \textbf{0.235} & 0.013 \\ 
        Seeds &                \textbf{0.734} & 0.551 & 0.635 & 0.400 \\ 
        Vertebral  &          \textbf{ 0.623} & 0.465 & 0.515 & 0.004 \\
	    WBCD &                 \textbf{0.443} & 0.358 & 0.363 & 0.005 \\ 
    Wine &                      \textbf{0.448} & 0.375& 0.379 & 0.314 \\ 
        Yeast &                 \textbf{0.294} & 0.195& 0.035 & 0.035 \\
\hline    \textbf{Average} &    \textbf{0.532} & 0.393& 0.371 & 0.194 \\
\hline
\end{tabular}
\caption{Comparison of clustering algorithms' performance on eleven benchmark datasets. We used the NMI between the optimal outputted clustering and the ground truth labels to measure an algorithm's performance on a dataset. Thus, a high value reflects more similarity to the ground truth labels in the dataset. The highest NMI for each dataset is marked in bold. On average, the optimal M-LUND clustering is significantly closer to the ground truth labels than the optimal clusterings produced by the algorithms we compare against: MMS~($1.82\times~10^{-2}$,~$t_S=~2.82$), HSC~($p=~8.34\times 10^{-3}$,~$t_S=~3.28$), and SLC~($p=~2.07\times 10^{-4}$,~$t_S=~5.67$).}
\label{tab: benchmark multiscale}
\end{table}

\begin{figure}[b] 
\floatbox[{\capbeside\thisfloatsetup{capbesideposition={right,center},capbesidewidth=2.4in}}]{figure}[\FBwidth]
{\caption{ Visualization of Salinas A scene. In the left panel, the ground truth labels for the pixels are provided. Black indicates bare soil and other colors indicate crop type. Specifically, dark blue indicates broccoli greens, teal indicates corn senesced greens, green indicates 5-week romaine, orange indicates 6-week romaine, yellow indicates 7-week romaine, and light blue indicates 8-week romaine. In the right panel, the spectra of a random subset of the Salinas A HSI are provided. Each pixel is colored by its ground truth label. }\label{fig: Salinas A Data}}
{\begin{minipage}{0.62\textwidth}
    \begin{subfigure}[t]{0.5\textwidth}
        \centering
        \includegraphics[height = 2in]{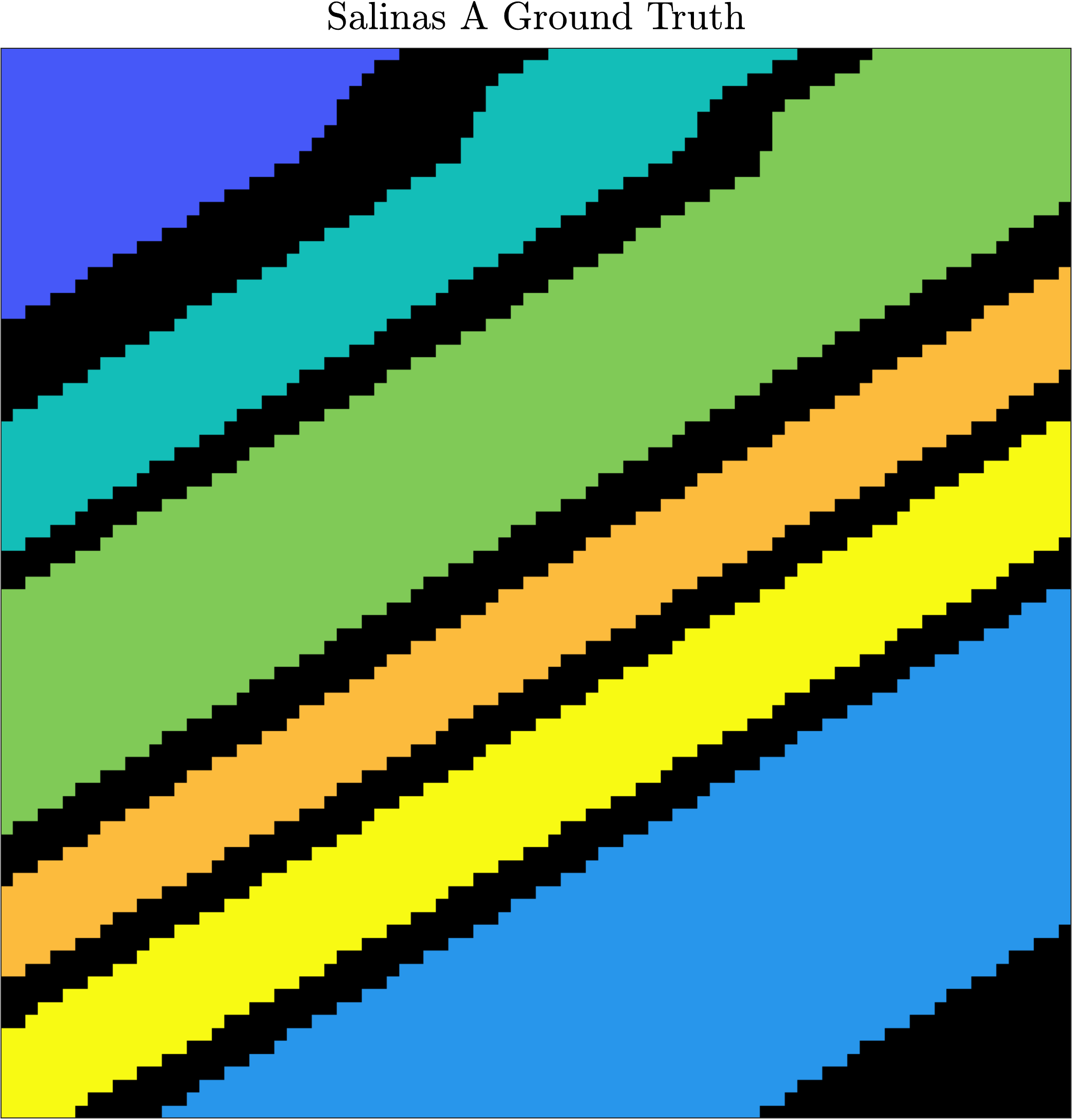}
    \end{subfigure}%
    \begin{subfigure}[t]{0.5\textwidth}
        \centering
        \includegraphics[height =2in]{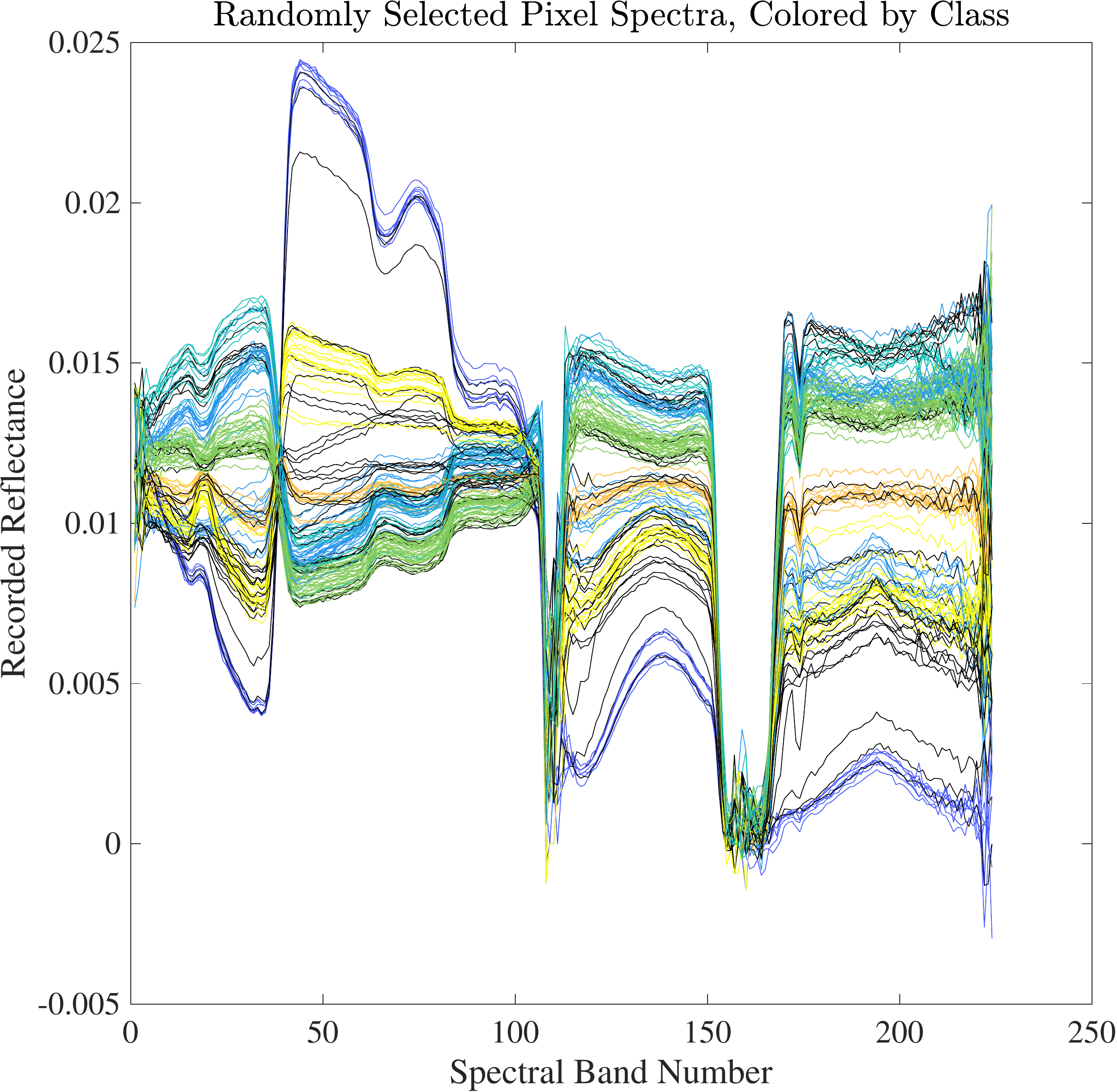}
    \end{subfigure}%
\end{minipage}}
\end{figure}

\subsection{Salinas A Hyperspectral Image}\label{sec: SA}

HSIs are images of a scene, typically generated by airborne sensors or satellites in orbit, that encode information about a hundred or more bands of electromagnetic activity. While HSIs are very high-dimensional and encode rich information about a scene, they often exhibit intrinsically low-dimensional structure~\cite{murphy2018unsupervised, murphy2019spectralspatial, zhang2020hyperspectral}. In this section, we show that the M-LUND algorithm detects latent multiscale structure in the Salinas A HSI~\cite{gualtieri1999salinasA}. The Salinas scene was generated using the Airborne Visible/Infrared Imaging Spectrometer Sensor over Salinas Valley, California, United States in 1998. We examined the Salinas A subset of the Salinas scene, which is $83\times 86$ pixels with $D=224$ spectral bands per pixel. To differentiate two pixels exhibiting the same exact value in each spectral band, we added Gaussian noise with variance $= 10^{-4}$ to the Salinas A HSI as a preprocessing step~\cite{murphy2019spectralspatial}.  In Figure \ref{fig: Salinas A Data}, we visualize the ground truth labels for the Salinas A image and the spectra of a random subset of the image. Black pixels reflect bare soil, while other colors reflect different crop types. Notably, the spectra of broccoli greens pixels, which are the dark blue crop type, is highly distinct from the spectra of all other ground truth classes. 

\begin{figure}[!t] 
\begin{minipage}{\textwidth}
  \centering
    \centering
    \begin{subfigure}[t]{0.2\textwidth}
        \centering
        \includegraphics[height = 1.1in]{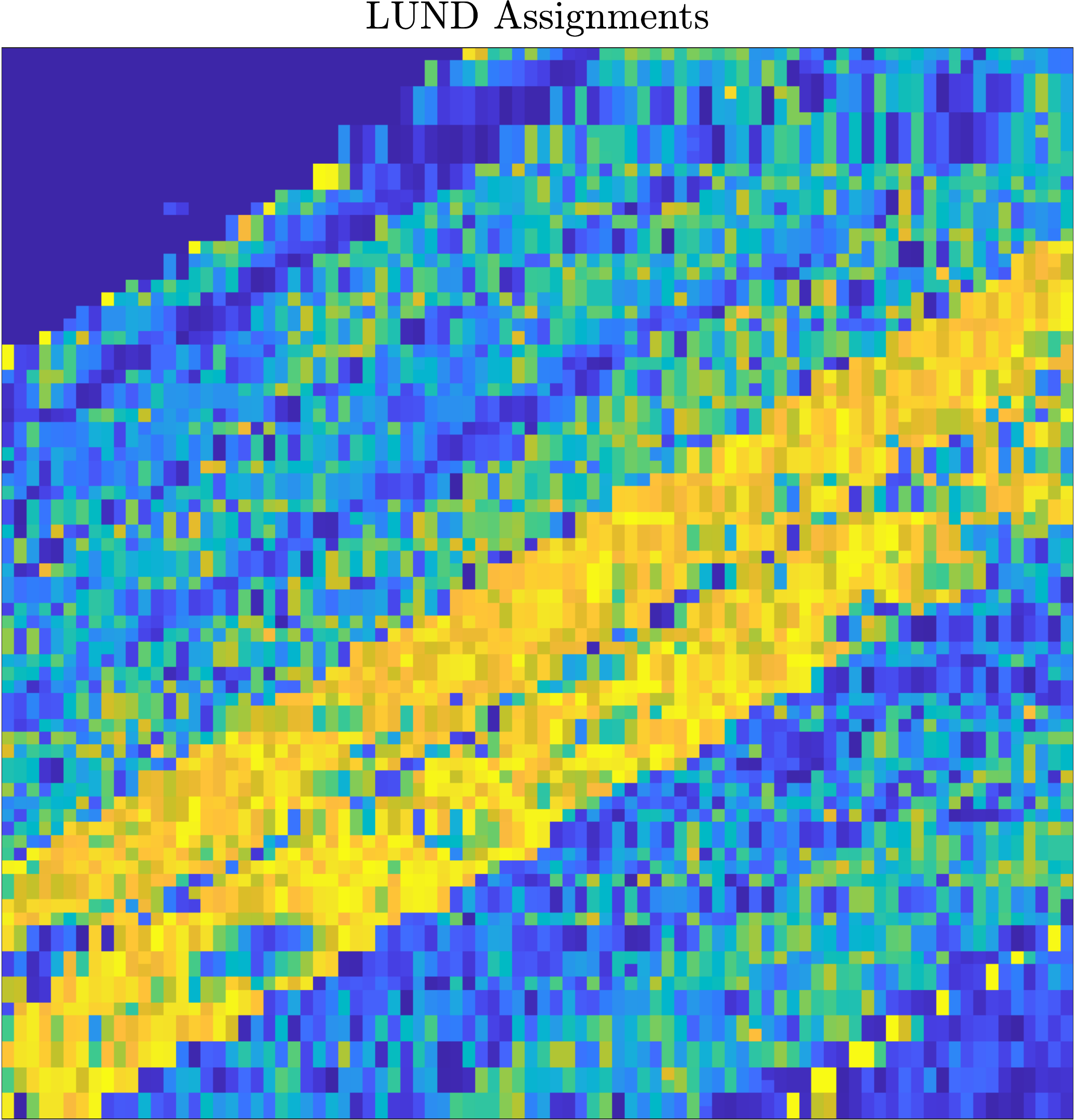}
    \end{subfigure}%
    \begin{subfigure}[t]{0.2\textwidth}
        \centering
        \includegraphics[height = 1.1in]{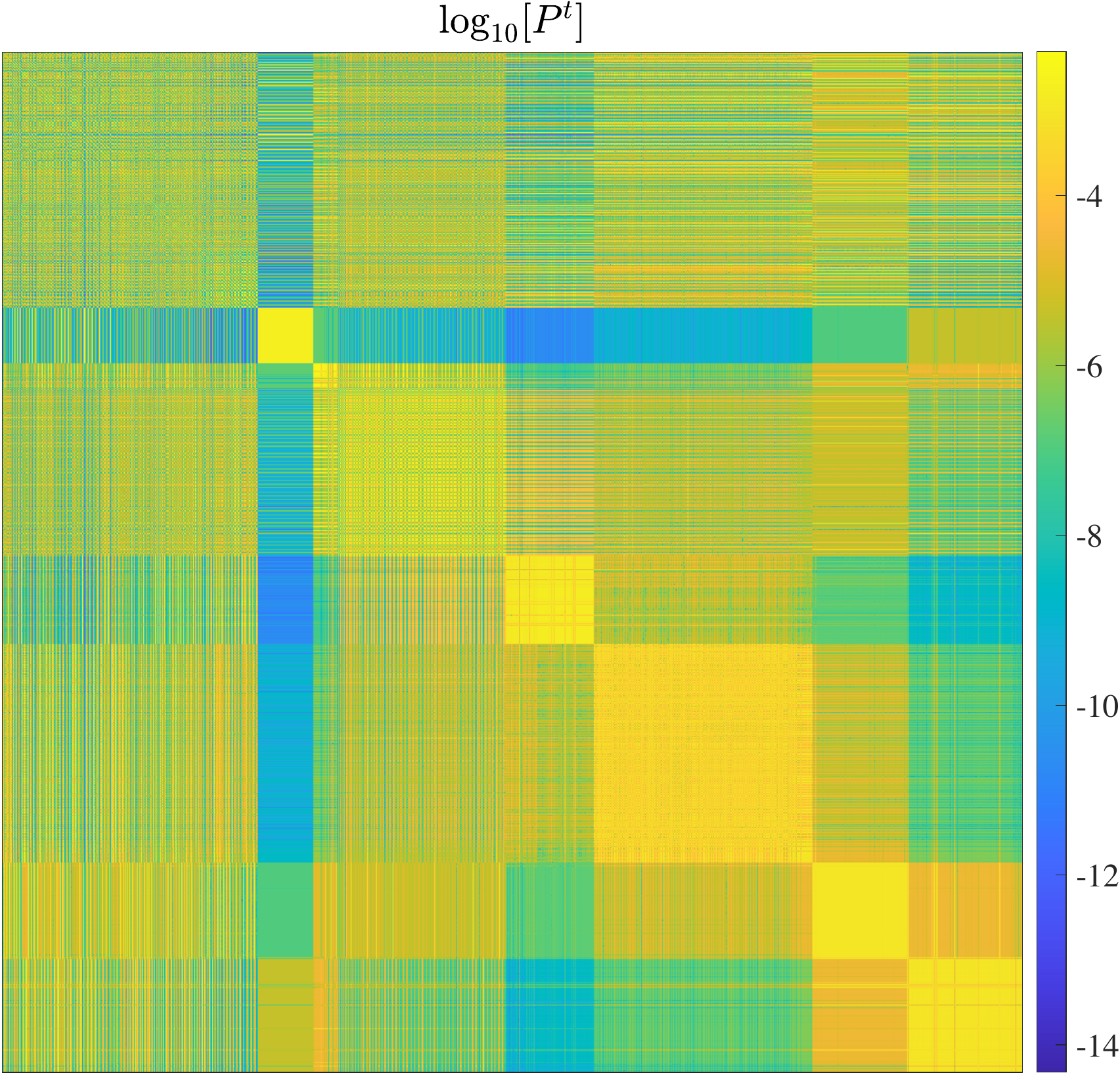}
    \end{subfigure}%
    \begin{subfigure}[t]{0.2\textwidth}
        \centering
        \includegraphics[height = 1.1in]{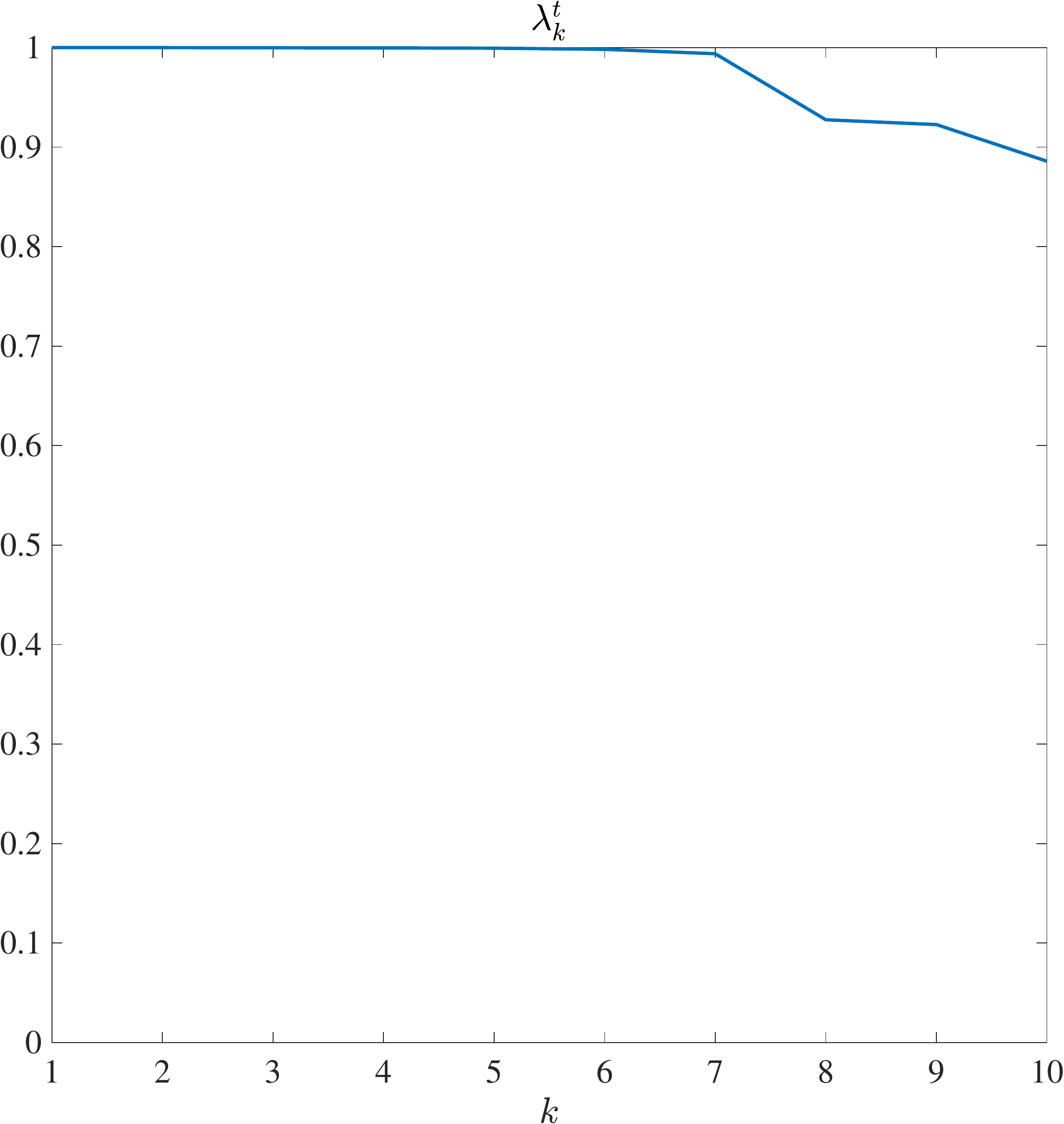}
    \end{subfigure}%
    \begin{subfigure}[t]{0.2\textwidth}
        \centering
        \includegraphics[height = 1.1in]{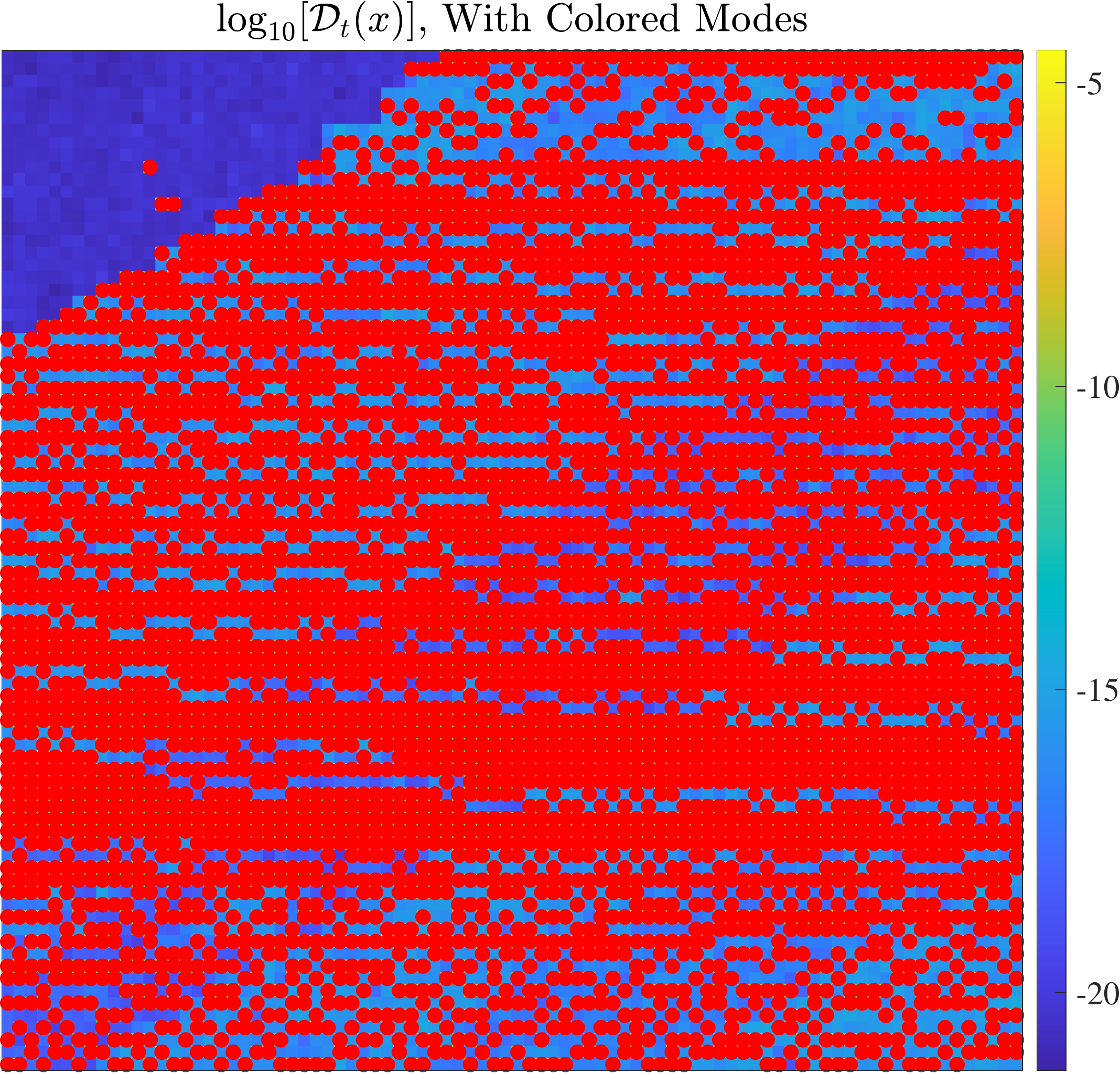}
    \end{subfigure}%
    \begin{subfigure}[t]{0.2\textwidth}
        \centering
        \includegraphics[height = 1.1in]{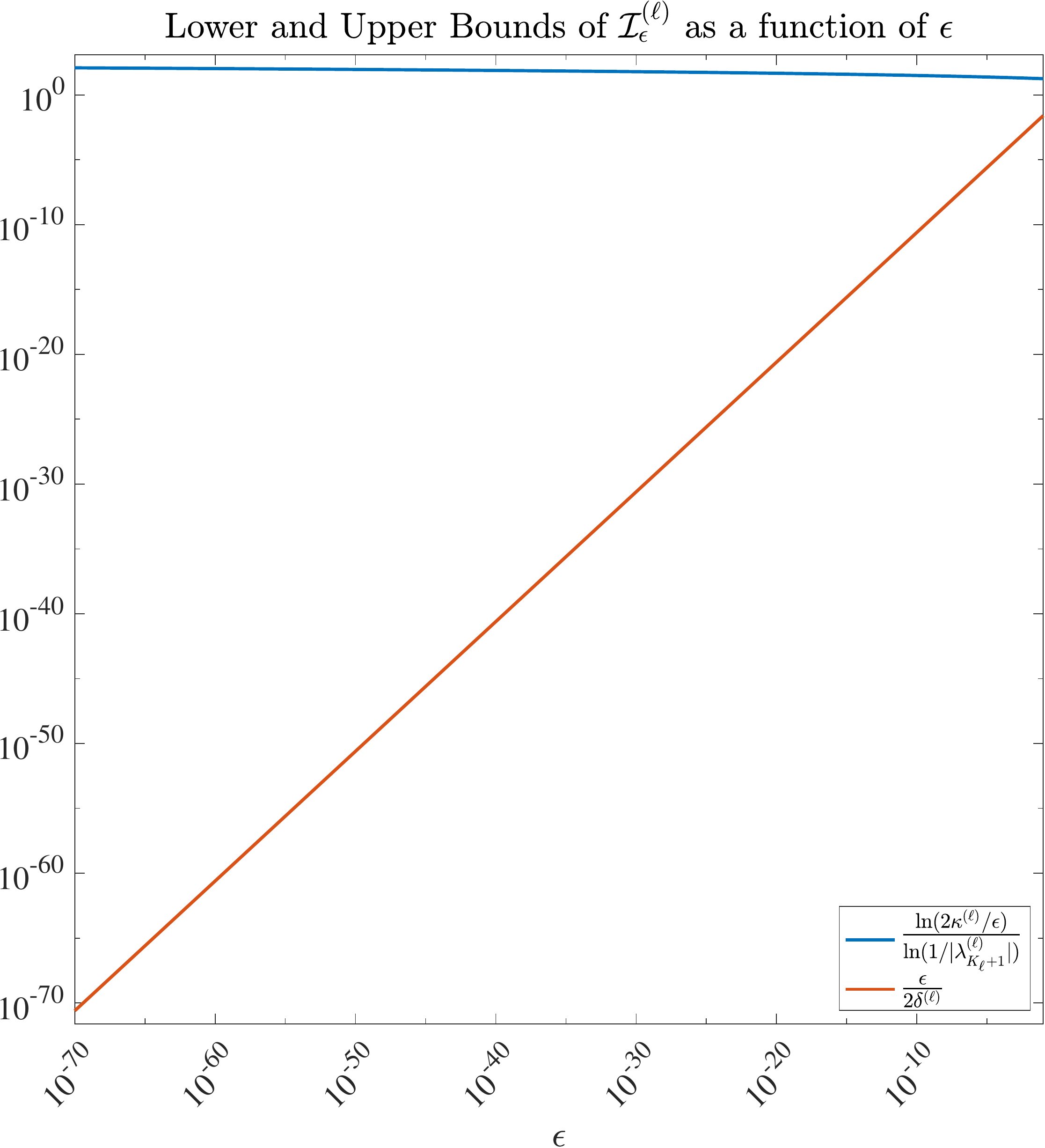}
    \end{subfigure}%
    \subcaption{LUND assignments, transition matrix, spectrum, $\mathcal{D}_t(x)$, and interval bounds for extracted clustering at time $t=2^1$. 4569 clusters.}
  \label{fig:SA1}\par\vspace{0.0625in}
    \centering
    \begin{subfigure}[t]{0.2\textwidth}
        \centering
        \includegraphics[height = 1.1in]{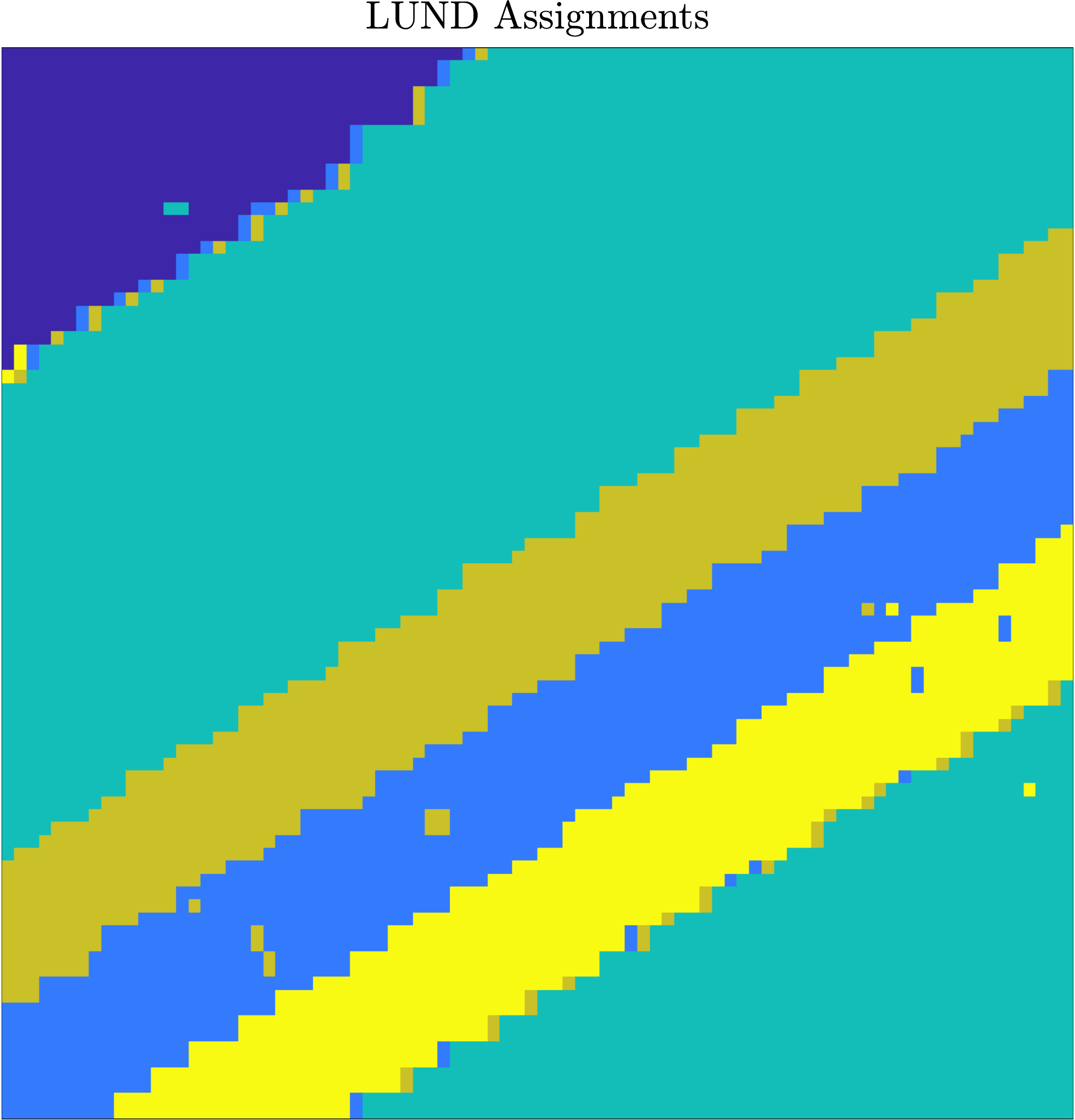}
    \end{subfigure}%
    \begin{subfigure}[t]{0.2\textwidth}
        \centering
        \includegraphics[height = 1.1in]{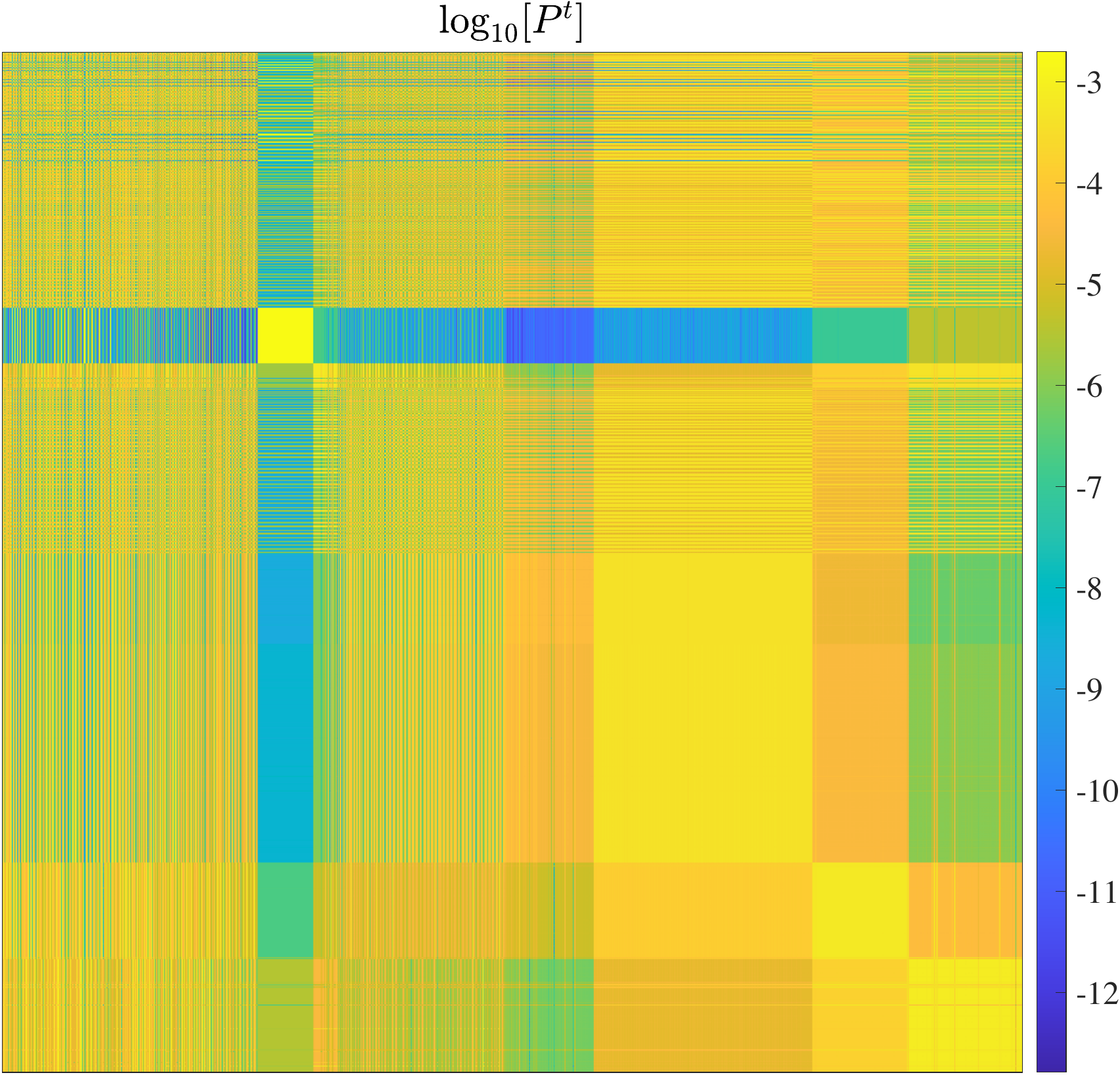}
    \end{subfigure}%
    \begin{subfigure}[t]{0.2\textwidth}
        \centering
        \includegraphics[height = 1.1in]{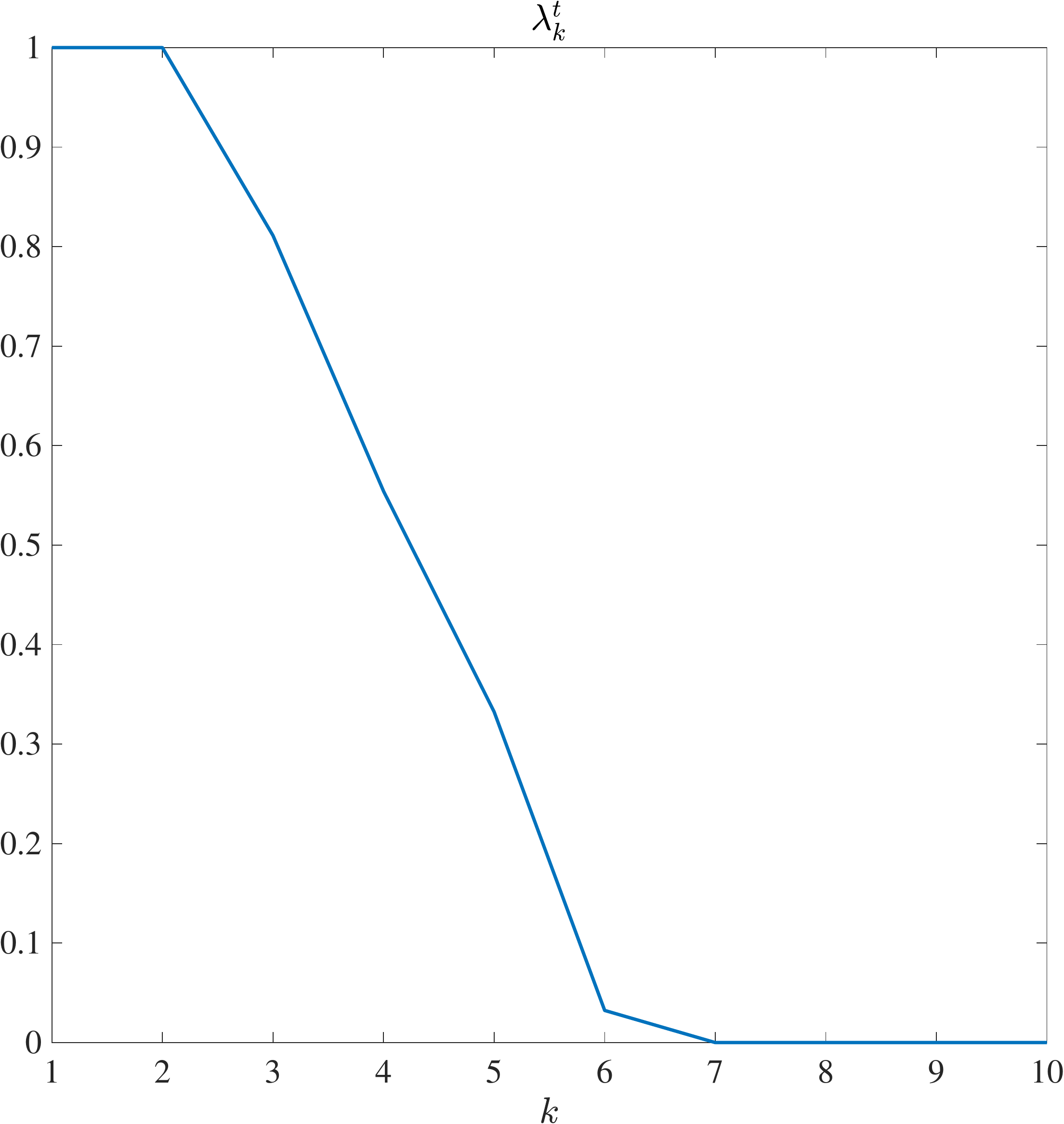}
    \end{subfigure}%
    \begin{subfigure}[t]{0.2\textwidth}
        \centering
        \includegraphics[height = 1.1in]{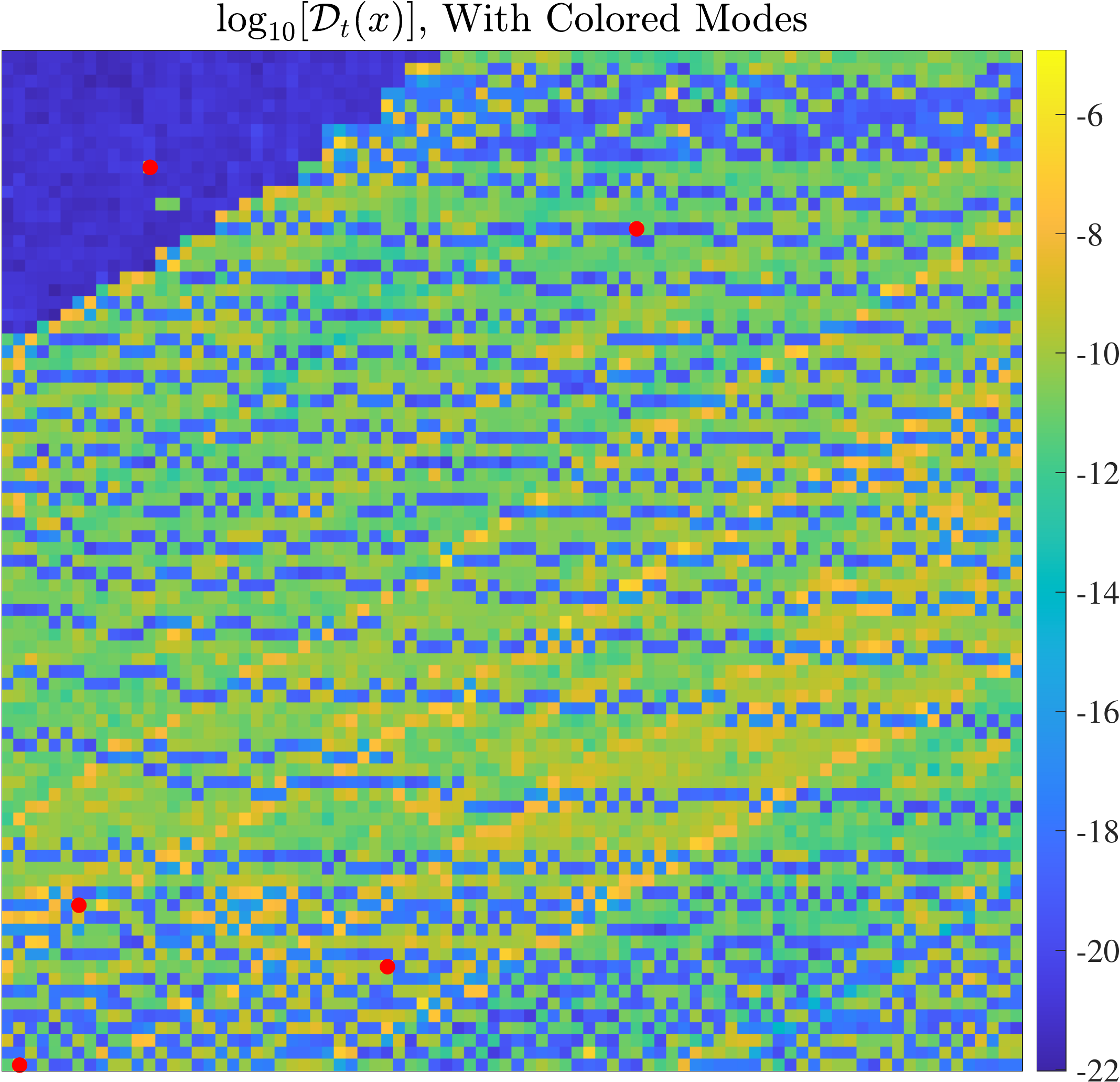}
    \end{subfigure}%
    \begin{subfigure}[t]{0.2\textwidth}
        \centering
        \includegraphics[height = 1.1in]{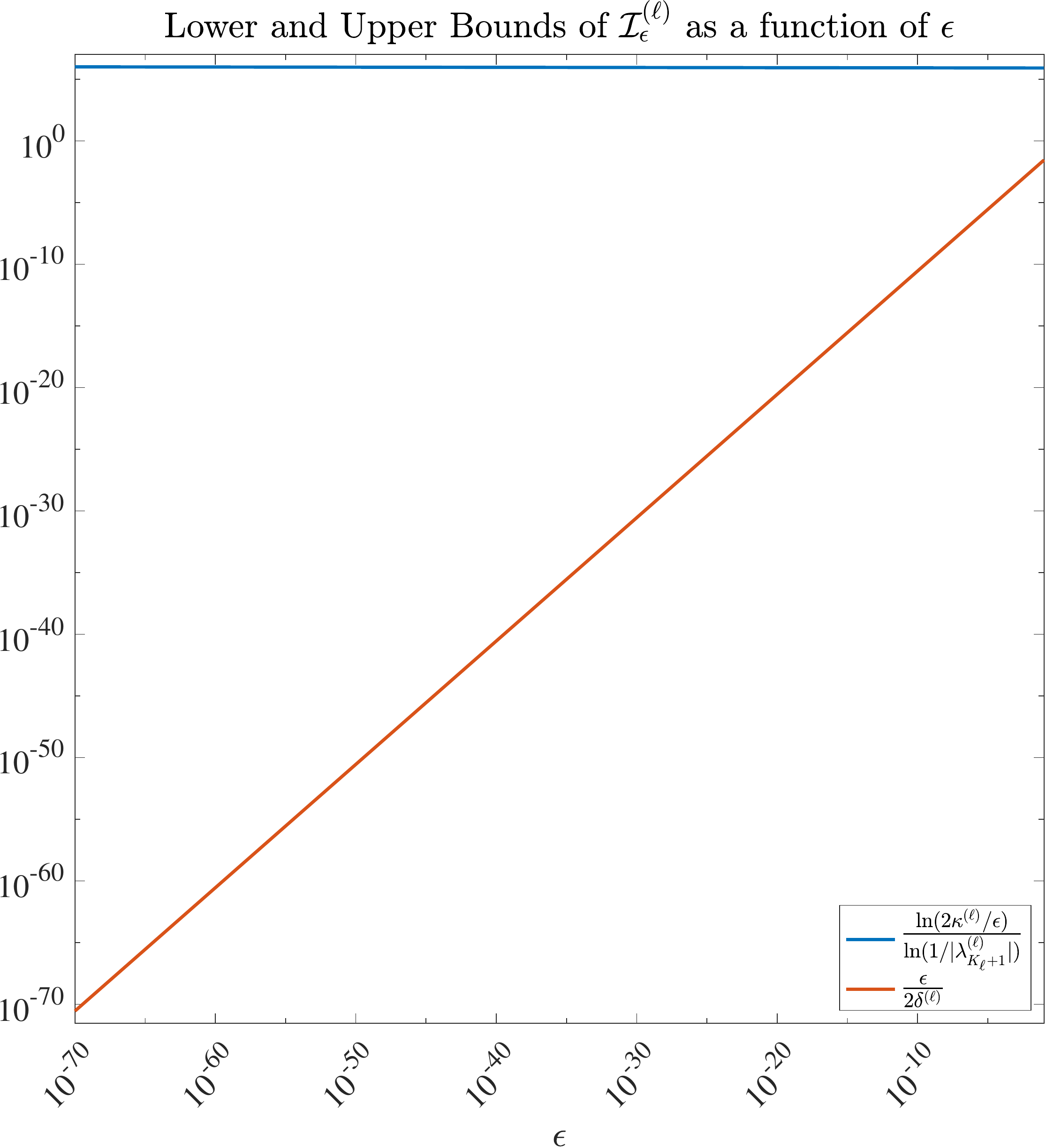}
    \end{subfigure}%
      \subcaption{LUND assignments, transition matrix, spectrum, $\mathcal{D}_t(x)$, and interval bounds for extracted clustering at time $t=2^{12}$. 5 clusters, total VI = 6.50. }
  \label{fig:SA2}\par\vspace{0.0625in}
    \centering
    \begin{subfigure}[t]{0.2\textwidth}
        \centering
        \includegraphics[height = 1.1in]{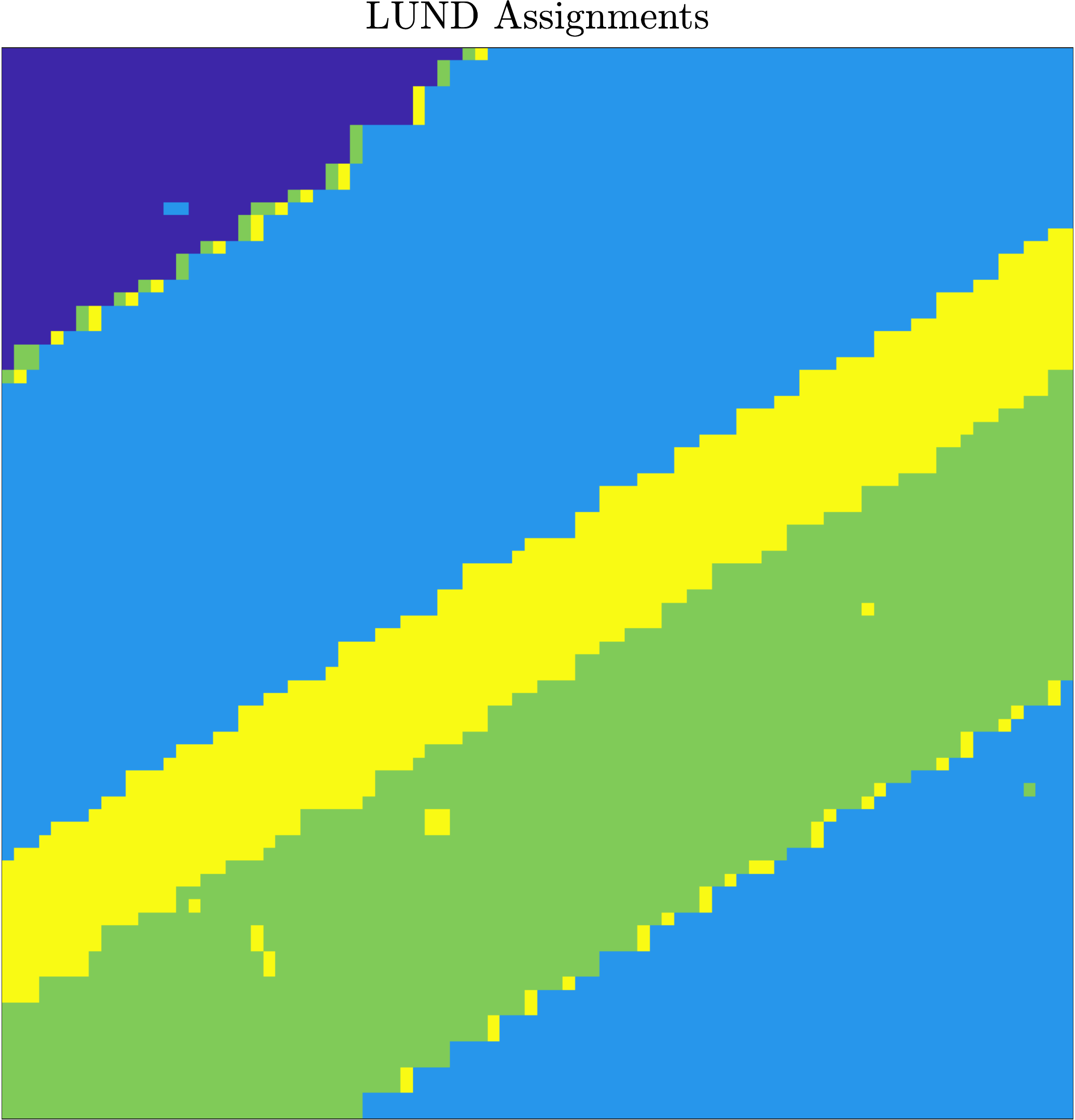}
    \end{subfigure}%
    \begin{subfigure}[t]{0.2\textwidth}
        \centering
        \includegraphics[height = 1.1in]{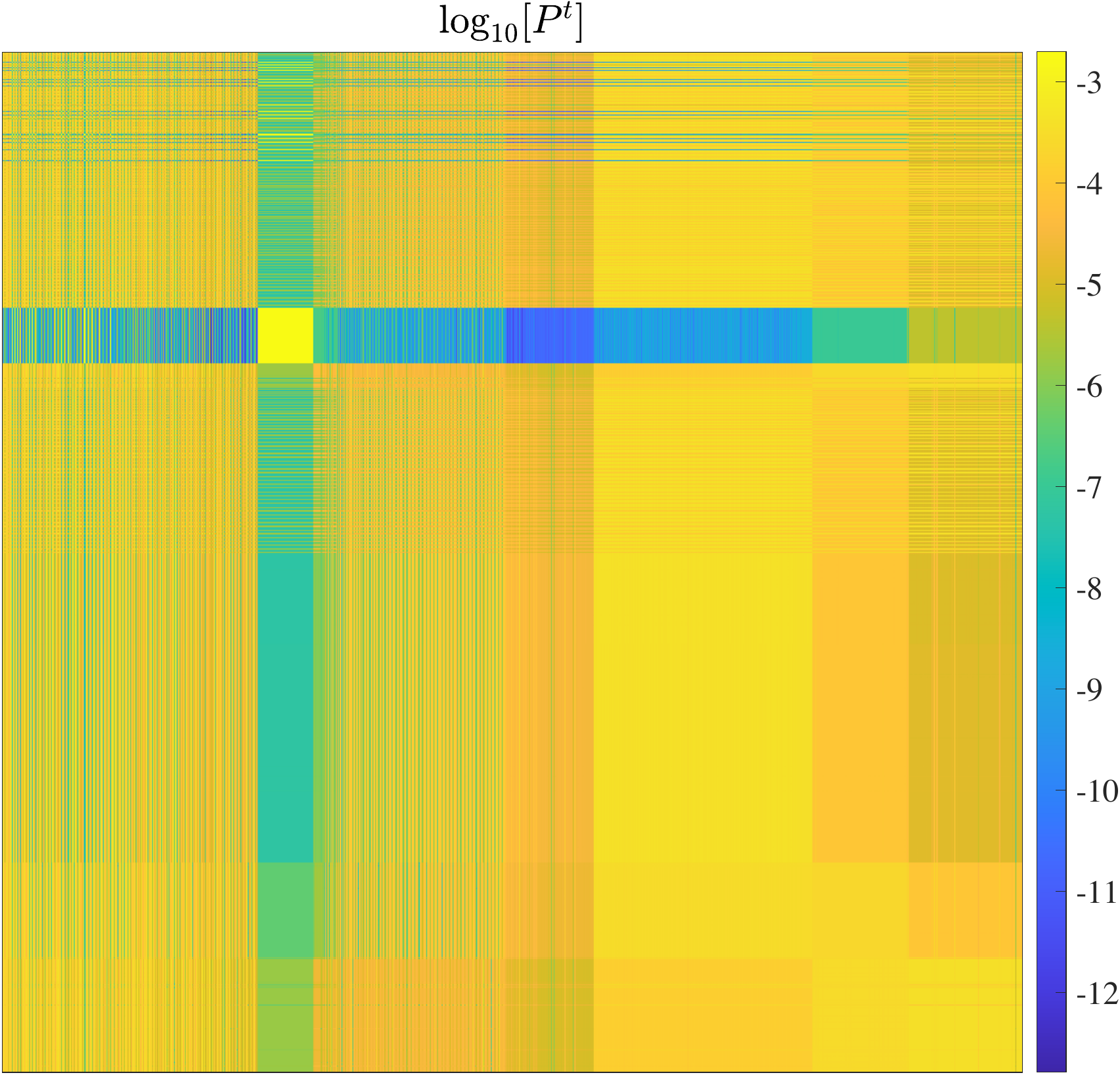}
    \end{subfigure}%
    \begin{subfigure}[t]{0.2\textwidth}
        \centering
        \includegraphics[height = 1.1in]{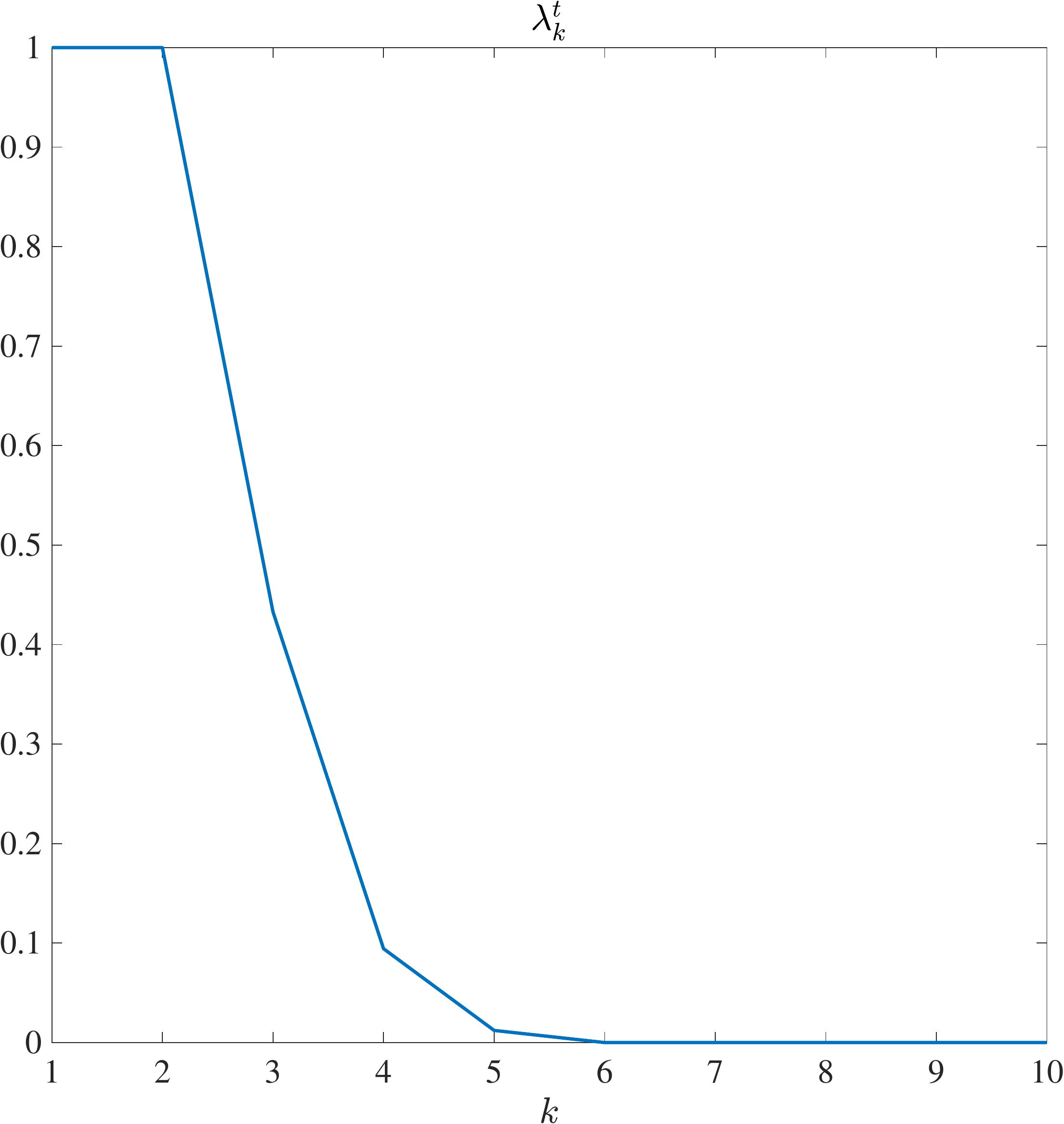}
    \end{subfigure}%
    \begin{subfigure}[t]{0.2\textwidth}
        \centering
        \includegraphics[height = 1.1in]{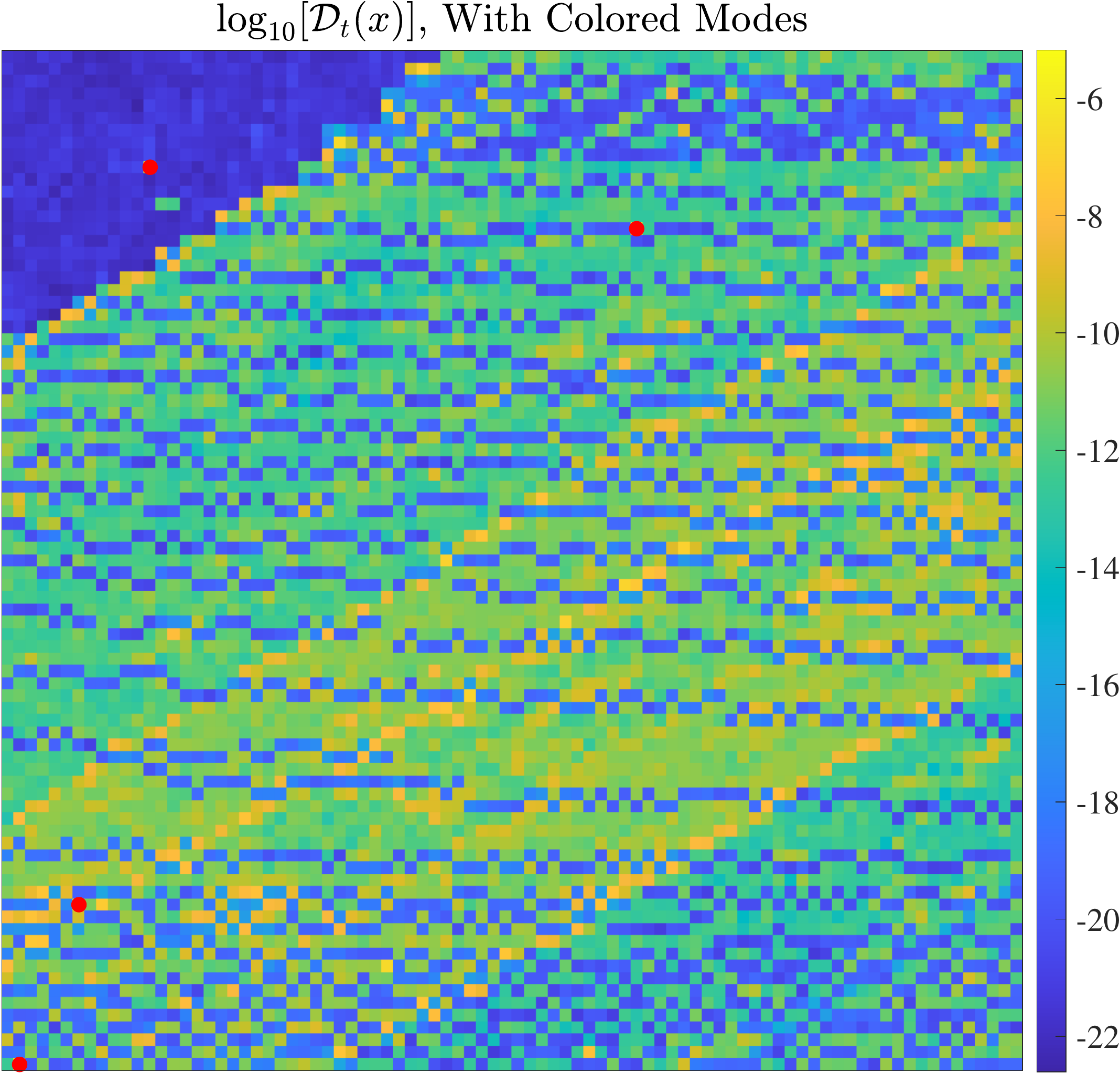}
    \end{subfigure}%
          \begin{subfigure}[t]{0.2\textwidth}
        \centering
        \includegraphics[height = 1.1in]{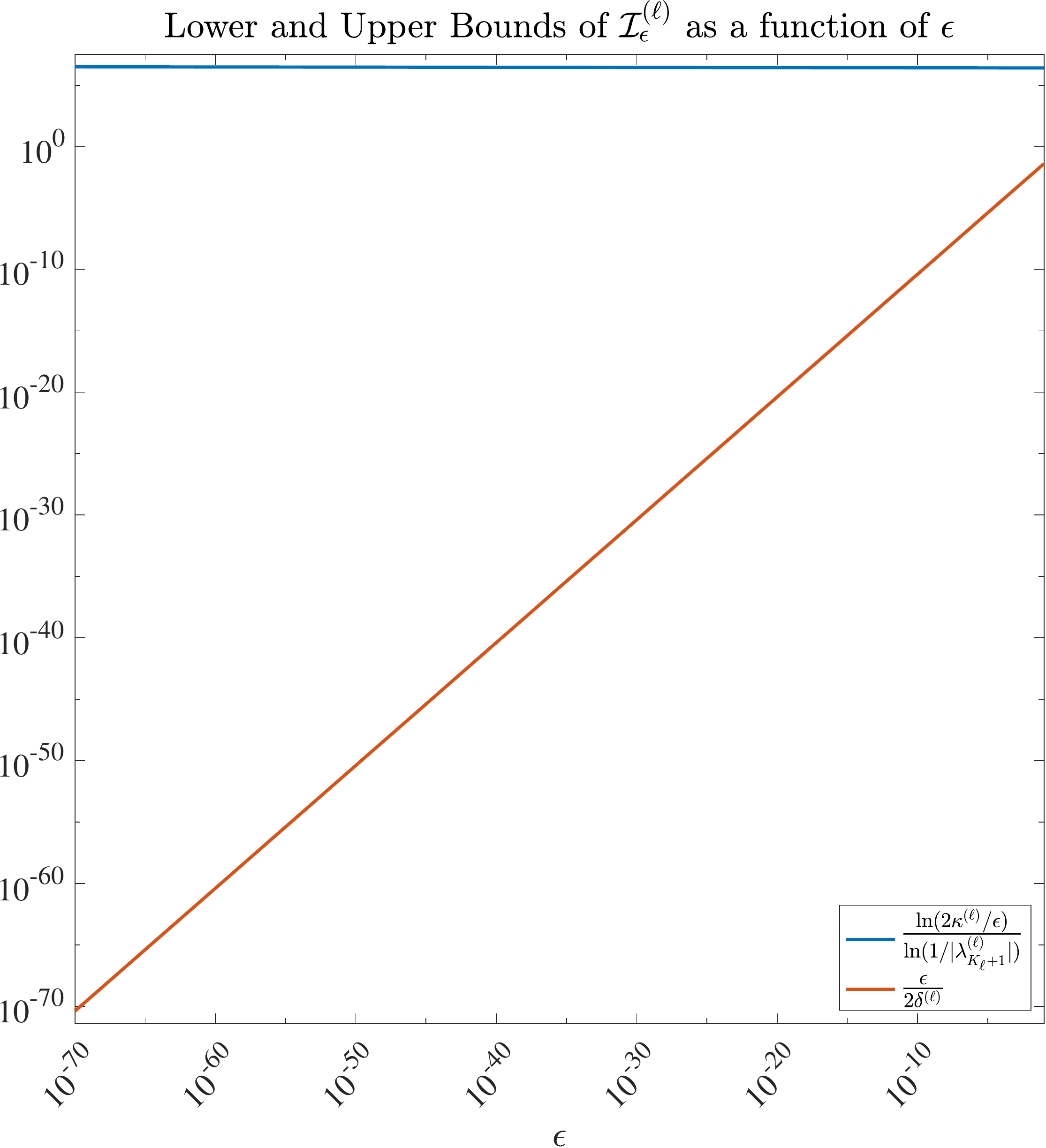}
    \end{subfigure}%
      \subcaption{LUND assignments, transition matrix, spectrum, $\mathcal{D}_t(x)$, and interval bounds for extracted clustering at time $t=2^{14}$. 4 clusters, total VI = 5.47. }
  \label{fig:SA3}\par\vspace{0.0625in}
    \centering
    \begin{subfigure}[t]{0.2\textwidth}
        \centering
        \includegraphics[height = 1.1in]{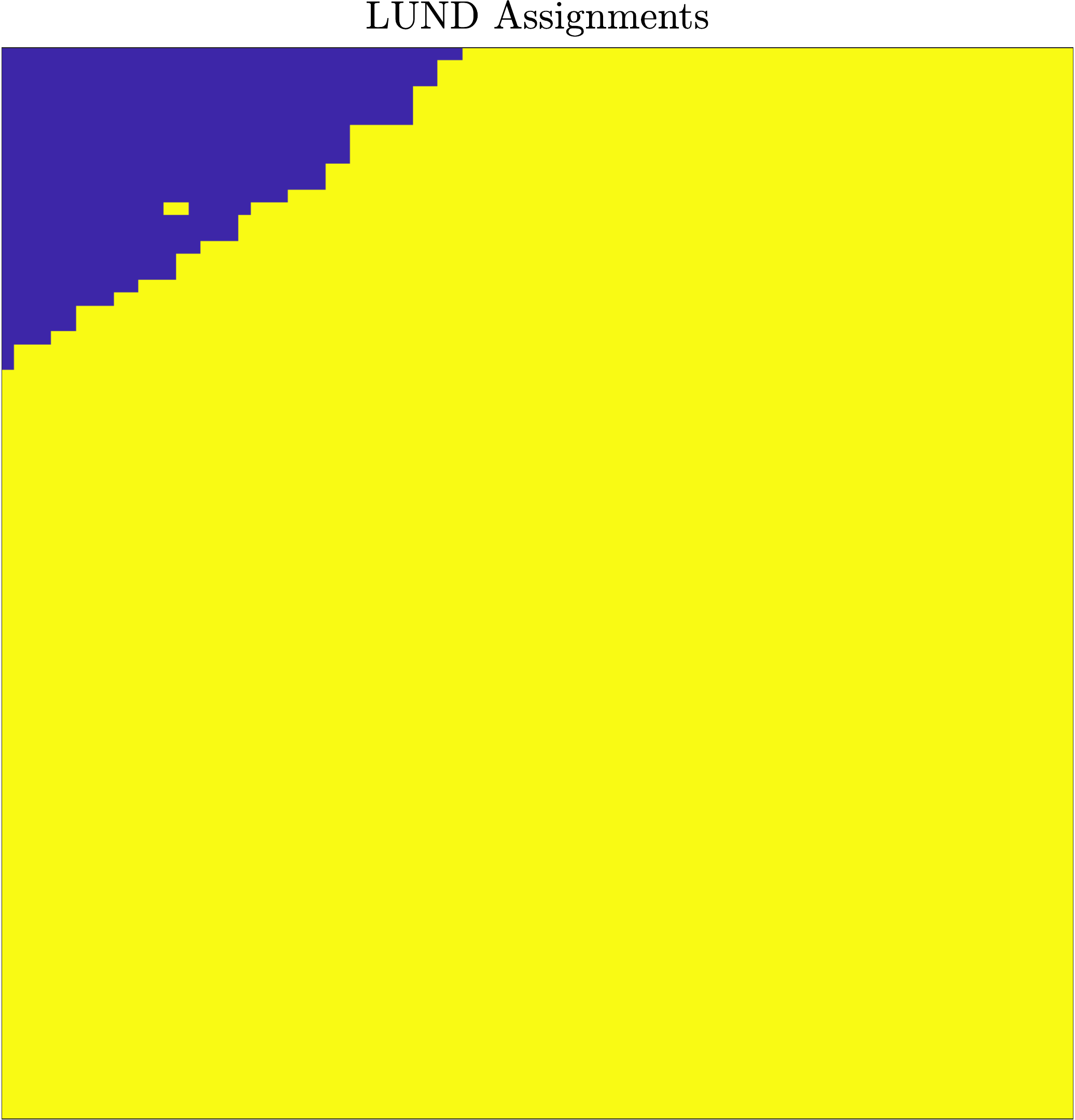}
    \end{subfigure}%
    \begin{subfigure}[t]{0.2\textwidth}
        \centering
        \includegraphics[height = 1.1in]{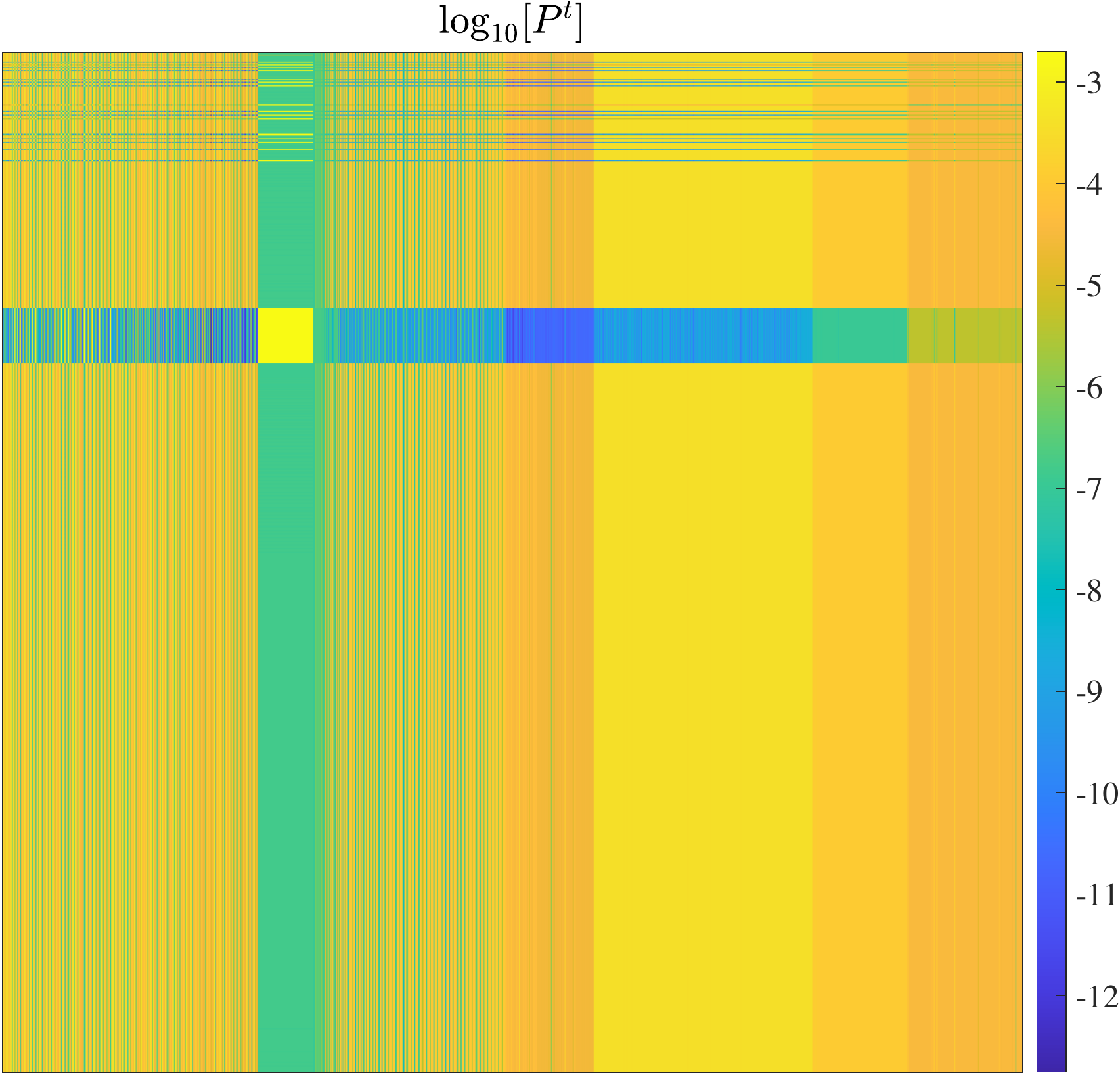}
    \end{subfigure}%
    \begin{subfigure}[t]{0.2\textwidth}
        \centering
        \includegraphics[height = 1.1in]{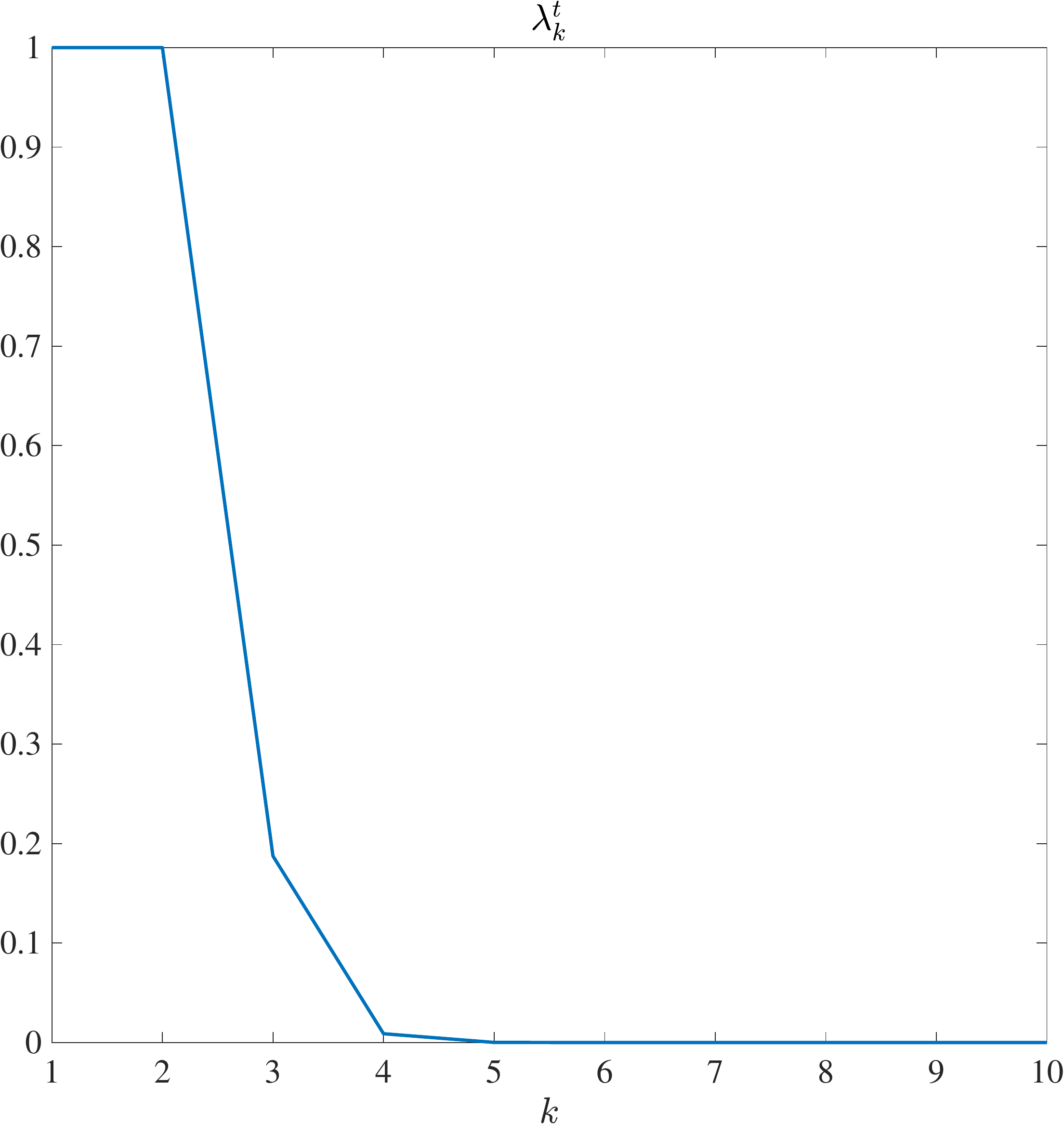}
    \end{subfigure}%
    \begin{subfigure}[t]{0.2\textwidth}
        \centering
        \includegraphics[height = 1.1in]{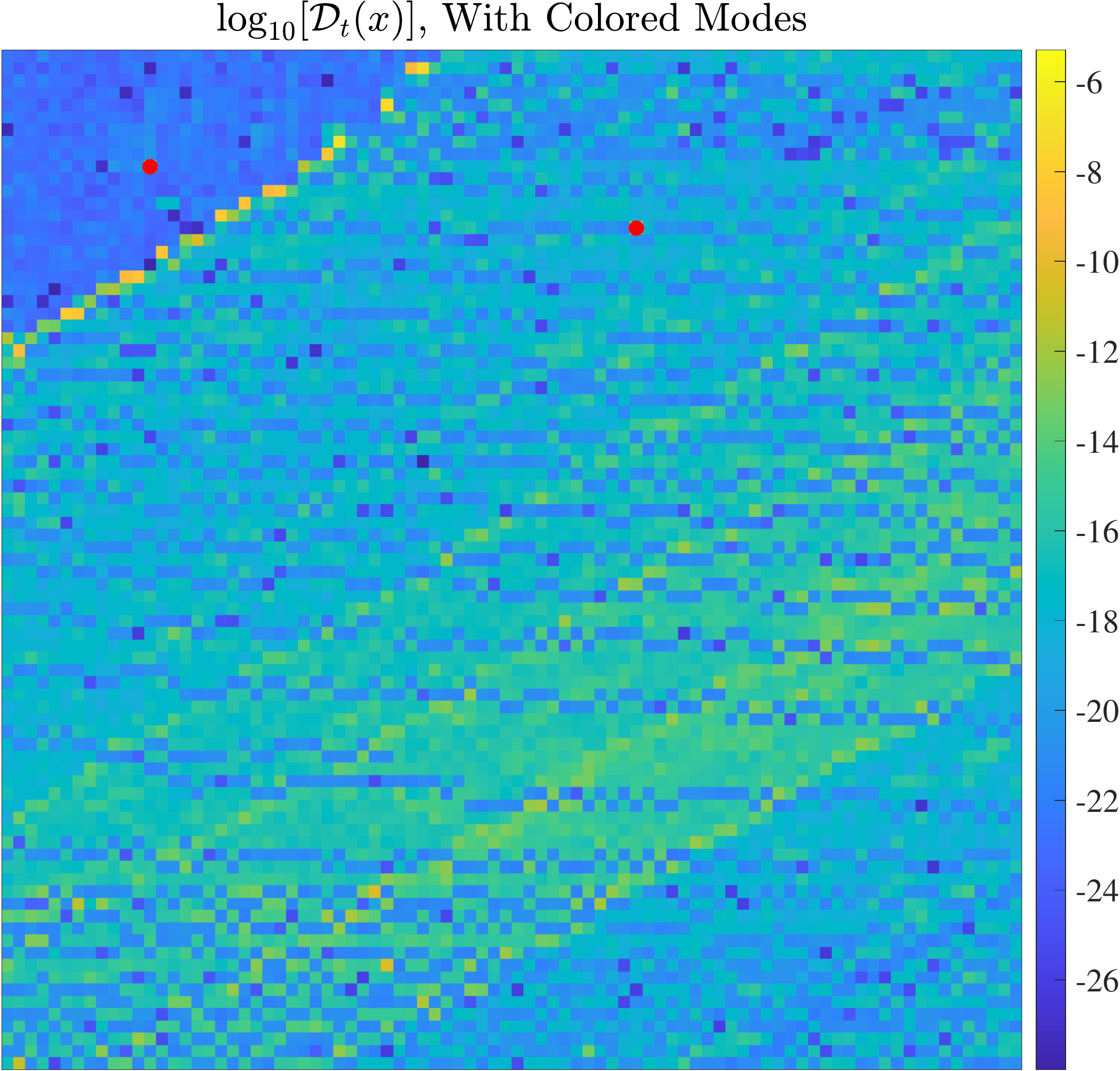}
    \end{subfigure}%
          \begin{subfigure}[t]{0.2\textwidth}
        \centering
        \includegraphics[height = 1.1in]{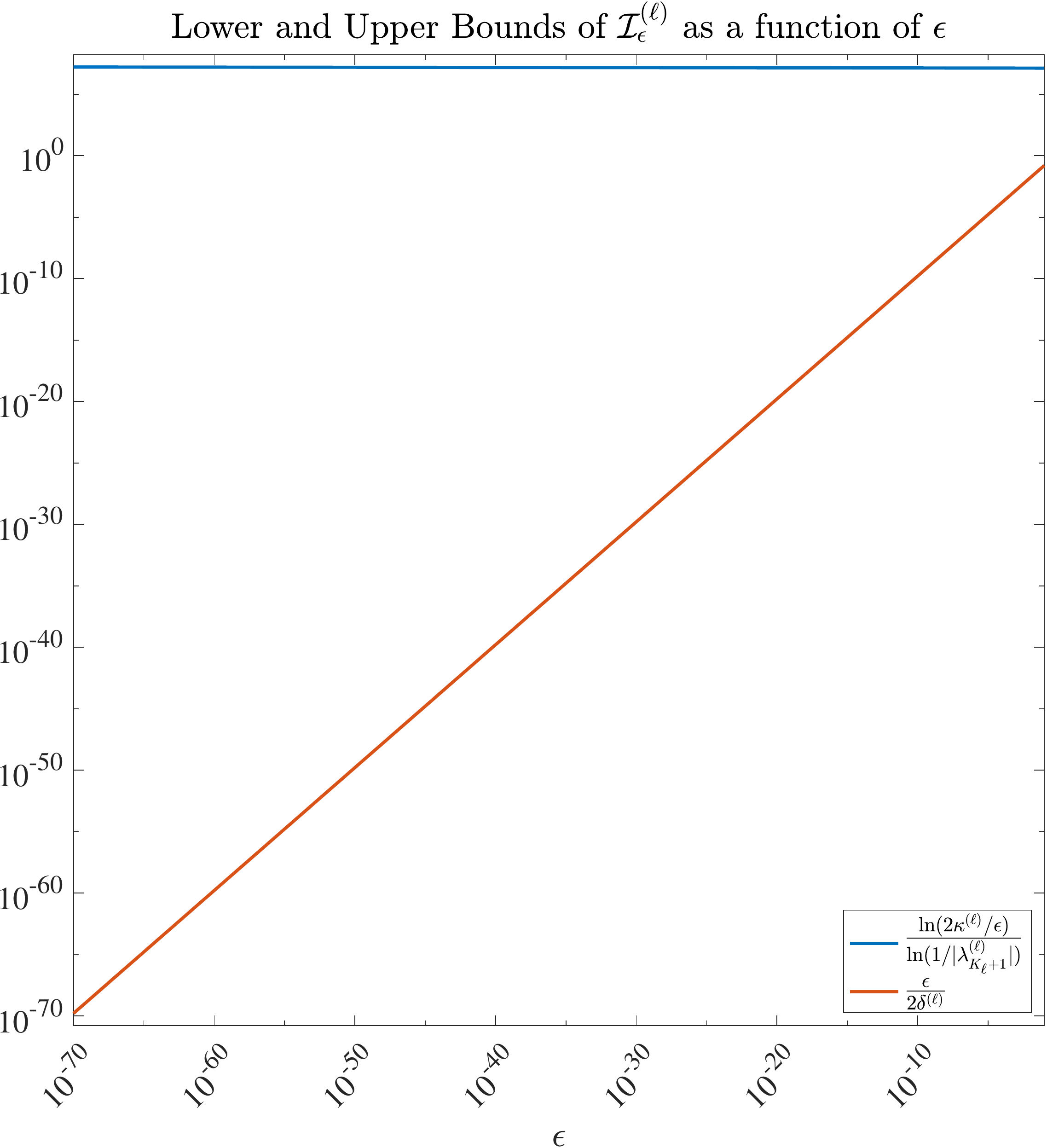}
    \end{subfigure}%
      \subcaption{LUND assignments, transition matrix, spectrum, $\mathcal{D}_t(x)$, and interval bounds for extracted clustering at time $t=2^{16}$. 2 clusters, total VI = 2.86. Optimal clustering. }
  \label{fig:SA4}
\end{minipage}
\caption{Diffusion process on the Salinas A HSI ($n=7138$)~\cite{gualtieri1999salinasA}. Red points indicate cluster modes. Multiscale structure is detected. The number of estimated clusters monotonically decreases with $t$. Data indices in $\textbf{P}$ are ordered by their ground truth class to illustrate multiscale structure. }\label{fig: Salinas A Diffusion}
\end{figure}

In Figure \ref{fig: Salinas A Diffusion}, we show how the pixel labels assigned by the LUND algorithm change as a function of the diffusion time parameter. The parameters we used were $N=75$ nearest neighbors, diffusion scale $\sigma =1.90$ and KDE bandwidth $\sigma_0=4.25\times 10^{-3}$. Early in the diffusion process, broccoli greens pixels are grouped together, while all other pixels are assigned singleton clusters (Figure \ref{fig:SA1}, $t\in [0, 2^{11}]$). Once higher-frequency coordinates of the diffusion map have been annihilated, the LUND algorithm detects 5 clusters (Figure \ref{fig:SA2}, $t\in [2^{12}, 2^{13}]$), reflecting fine-scale structure in the HSI. The clusters in this clustering correspond to broccoli greens, corn-senesced greens grouped with 5-week maturity romaine lettuce crops, 6-week maturity romaine lettuce crops, 7-week maturity romaine lettuce crops, and 8-week maturity romaine lettuce crops.  Later in the diffusion process, mature romaine lettuce (7-8 week maturity) clusters merge in a 4-cluster clustering with total VI = 5.47 (Figure \ref{fig:SA3}, $t=2^{14}$).  Finally, all romaine lettuce clusters merge, leaving only the broccoli greens crops separated (Figure \ref{fig:SA4}, $t\in [2^{15}, 2^{18}]$). This highly stable, 2-cluster clustering is the total VI minimizer for the HSI (total VI = 2.86). Due to poor separation between clusters, no clustering of the HSI is guaranteed to be $\epsilon$-separable by diffusion distances at any interval in the diffusion process. 

\begin{figure}[t] 
\begin{minipage}{\textwidth}
  \centering
    \centering
    \begin{subfigure}[t]{0.333\textwidth}
        \centering
        \includegraphics[height = 1.9in]{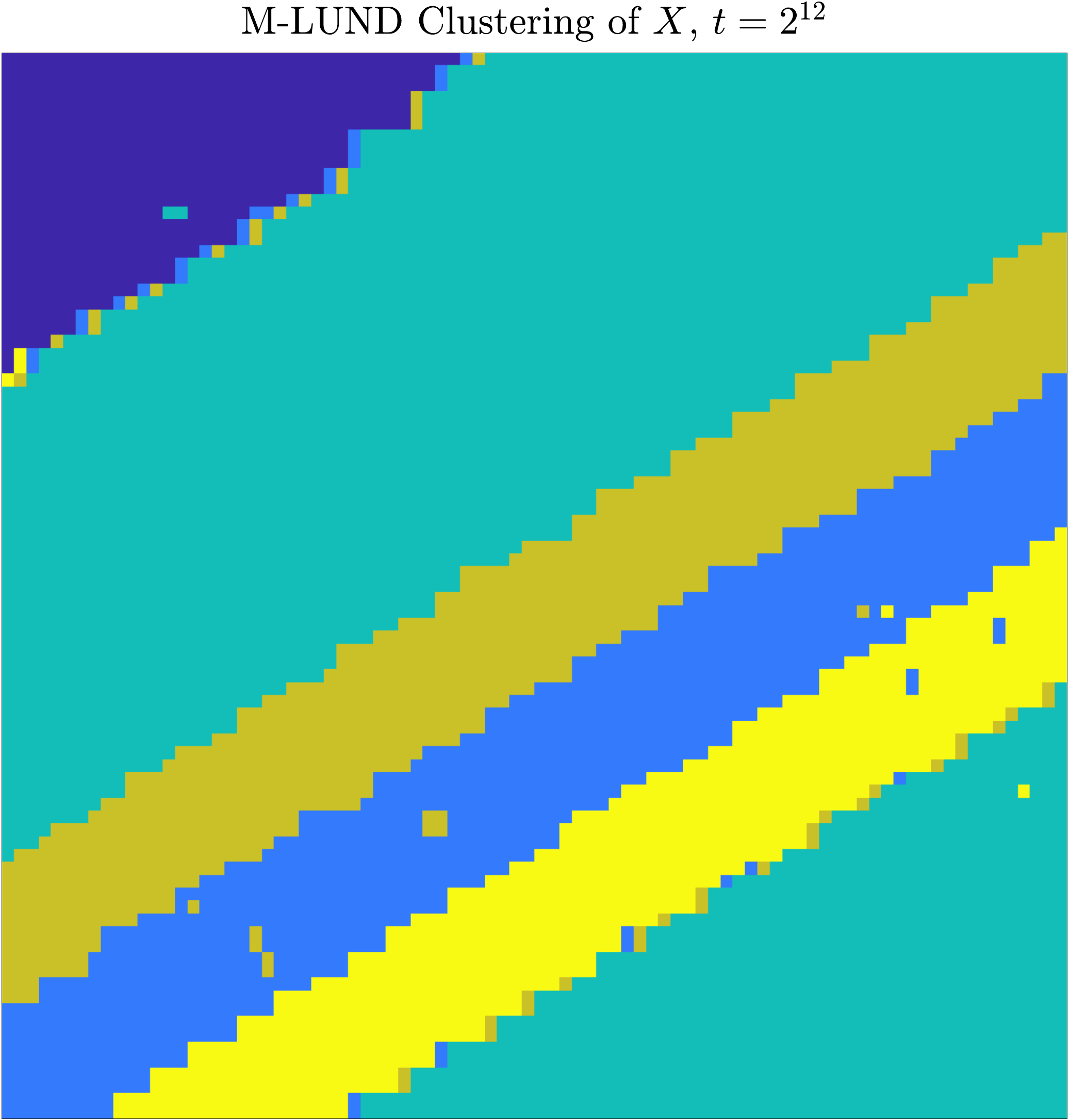}
    \end{subfigure}%
    \begin{subfigure}[t]{0.333\textwidth}
        \centering
        \includegraphics[height = 1.9in]{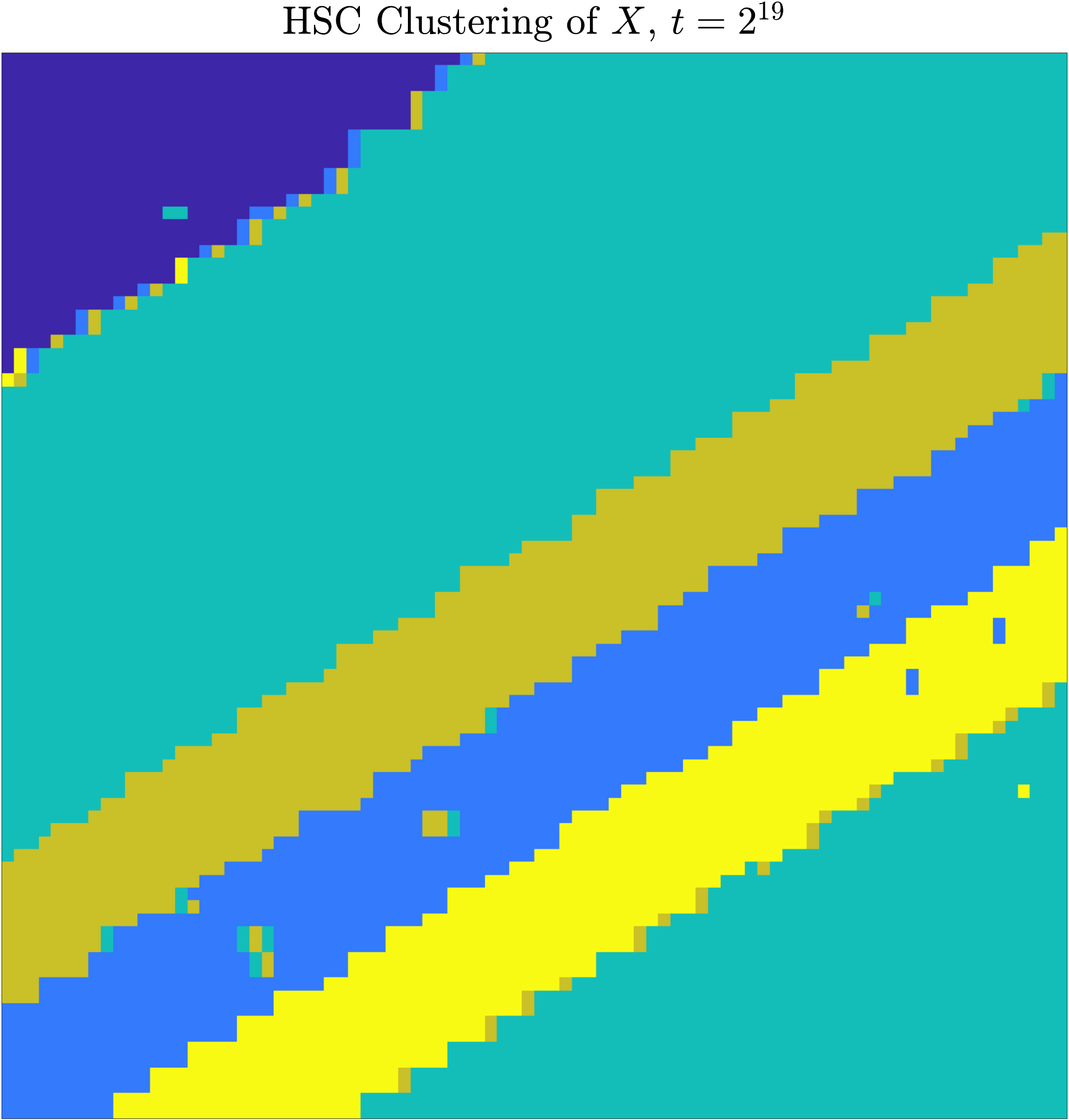}
    \end{subfigure}%
    \begin{subfigure}[t]{0.333\textwidth}
        \centering
        \includegraphics[height = 1.9in]{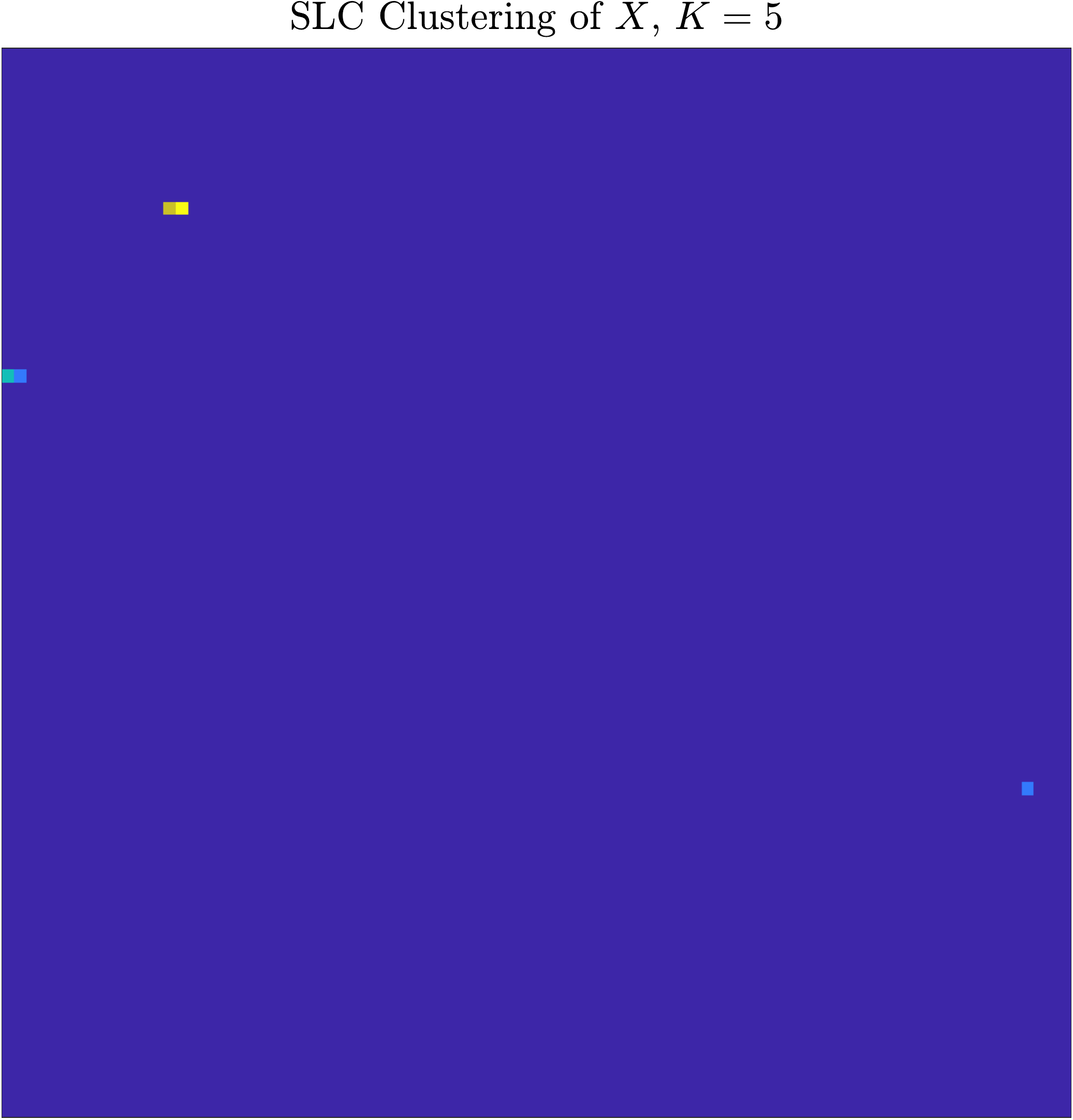}
    \end{subfigure}%
  \subcaption{$K=5$ Clustering assigned by M-LUND, HSC, and SLC.}
    \label{fig:comparison5}
  \centering
    \centering
    \begin{subfigure}[t]{0.333\textwidth}
        \centering
        \includegraphics[height = 1.9in]{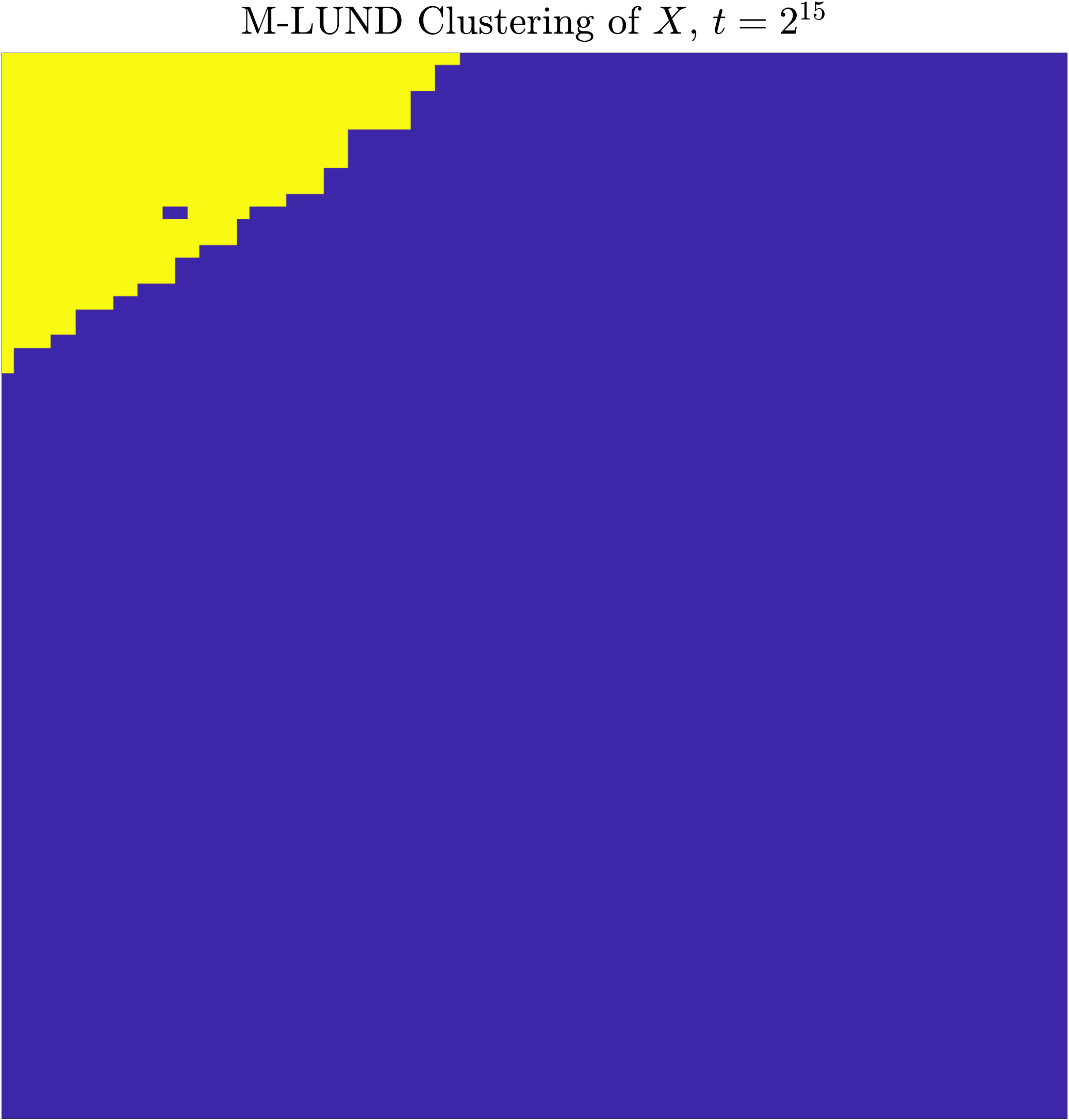}
    \end{subfigure}%
    \begin{subfigure}[t]{0.333\textwidth}
        \centering
        \includegraphics[height = 1.9in]{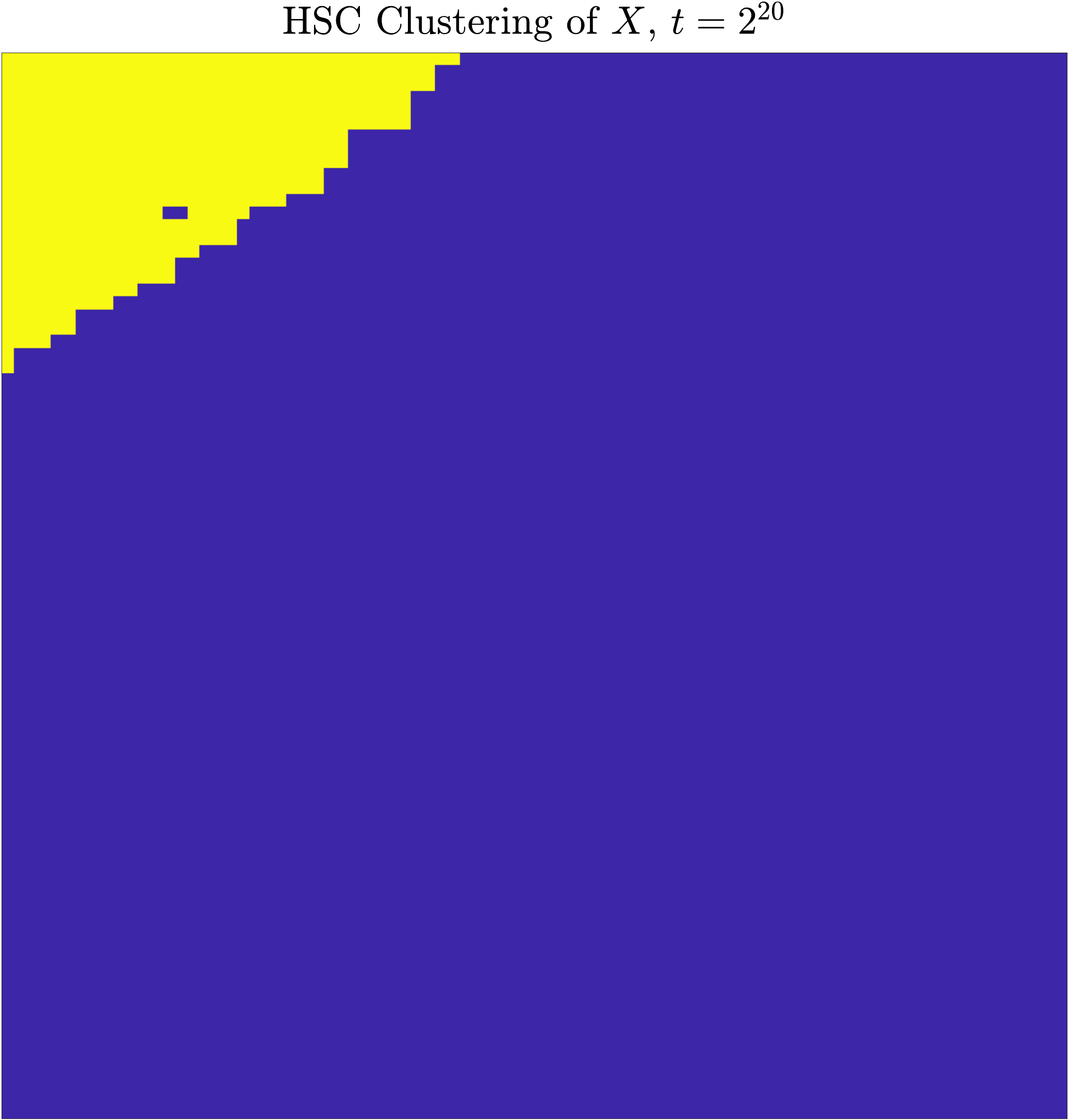}
    \end{subfigure}%
    \begin{subfigure}[t]{0.333\textwidth}
        \centering
\includegraphics[height = 1.9in]{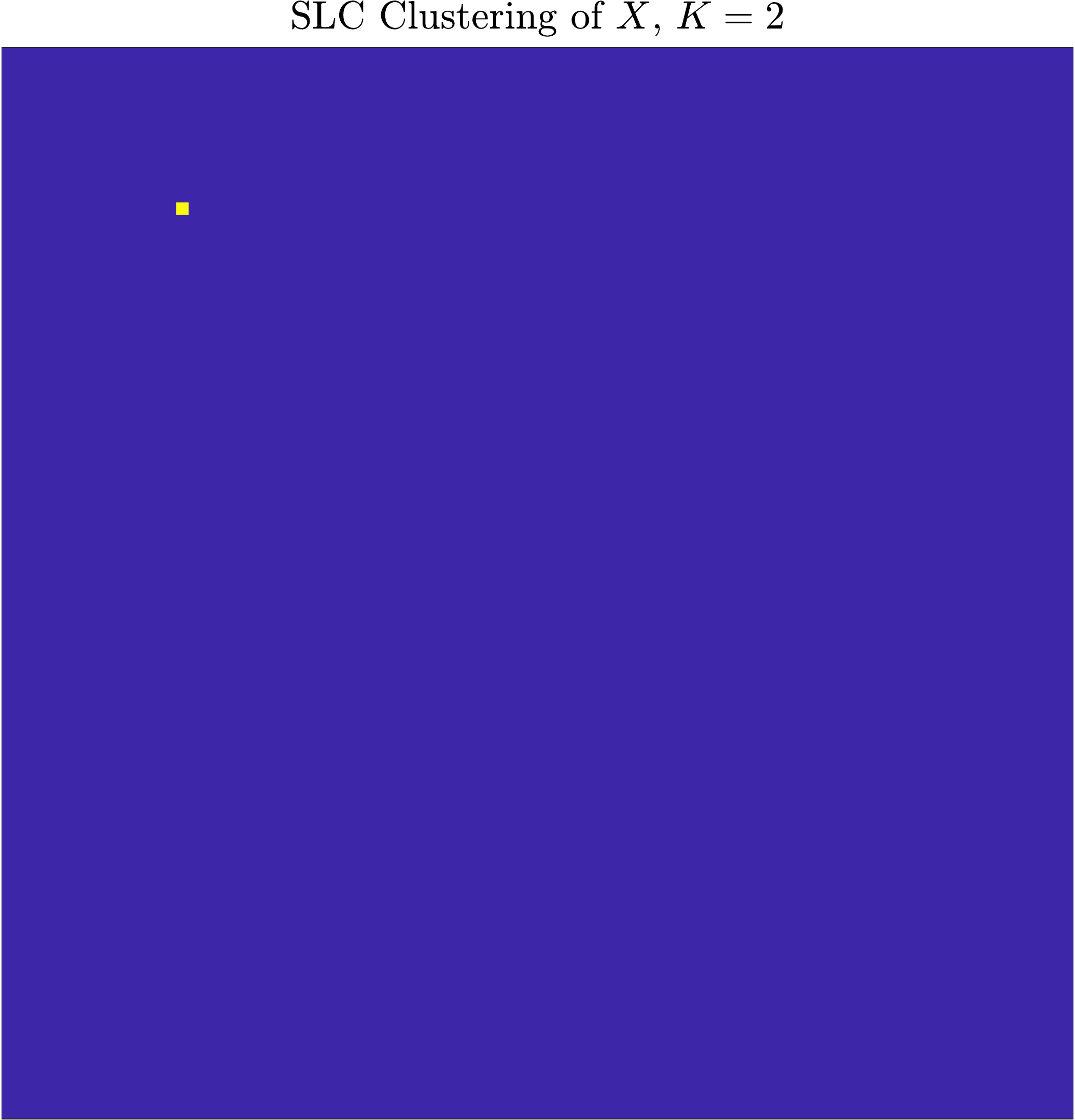}
    \end{subfigure}%
  \subcaption{$K=2$ Clustering assigned by the M-LUND algorithm, HSC, and SLC.}    \label{fig:comparison2}
  \end{minipage}
\caption{Comparison of the $K=5$ and $K=2$ clusterings  extracted by the M-LUND algorithm from the Salinas A HSI~\cite{gualtieri1999salinasA} against the $K=5$ and $K=2$ clusterings assigned by HSC and SLC~\cite{gower1969MST,  friedman2001elements, azran2006spectralclustering}. 
}\label{fig: Salinas A Comparison}
\end{figure}

\begin{figure}[t]
    \begin{subfigure}[t]{0.25\textwidth}
        \centering
        \includegraphics[height = 1.4in]{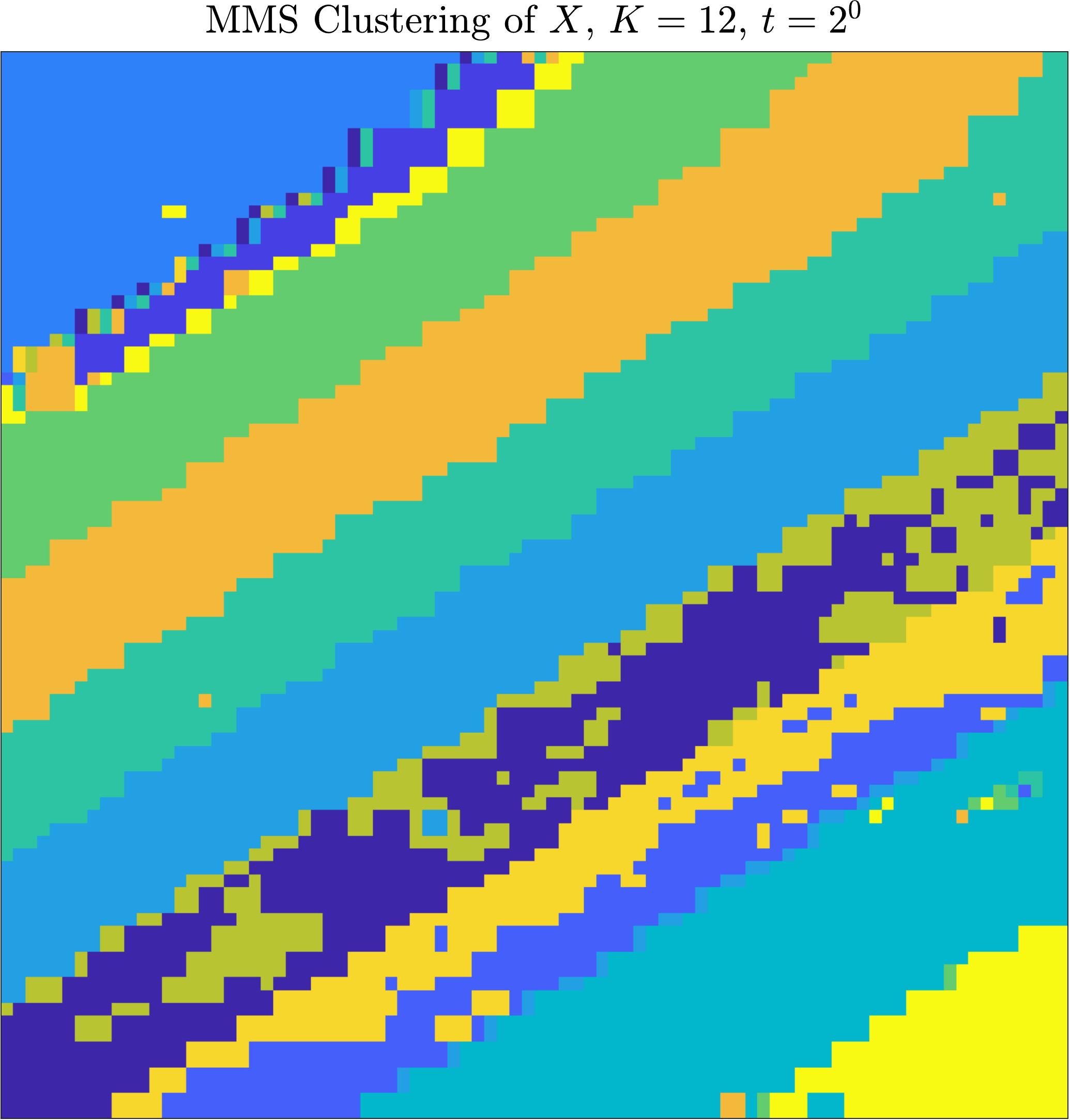}
    \end{subfigure}%
    \begin{subfigure}[t]{0.25\textwidth}
        \centering
        \includegraphics[height = 1.4in]{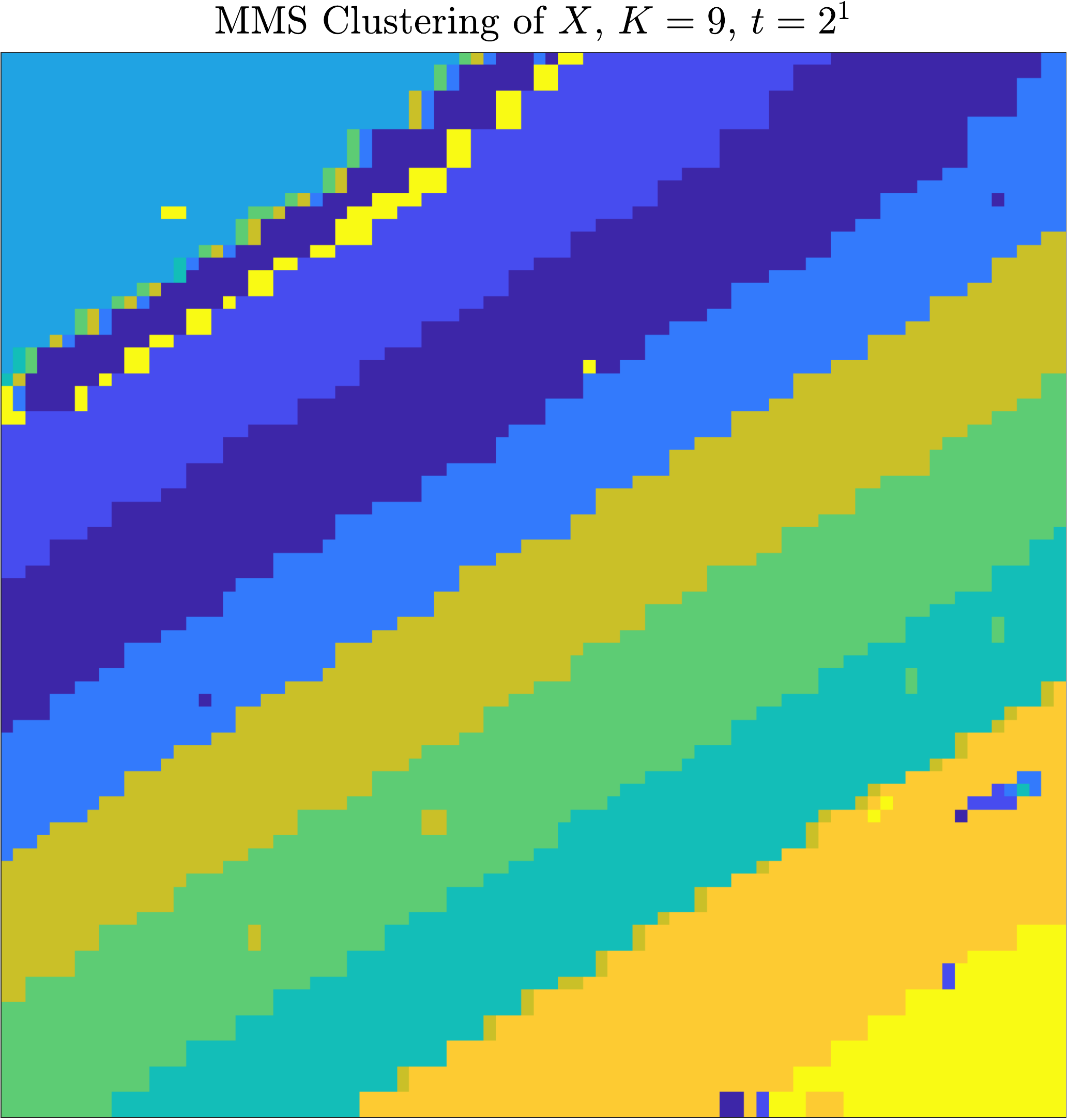}
    \end{subfigure}%
    \begin{subfigure}[t]{0.25\textwidth}
        \centering
        \includegraphics[height = 1.4in]{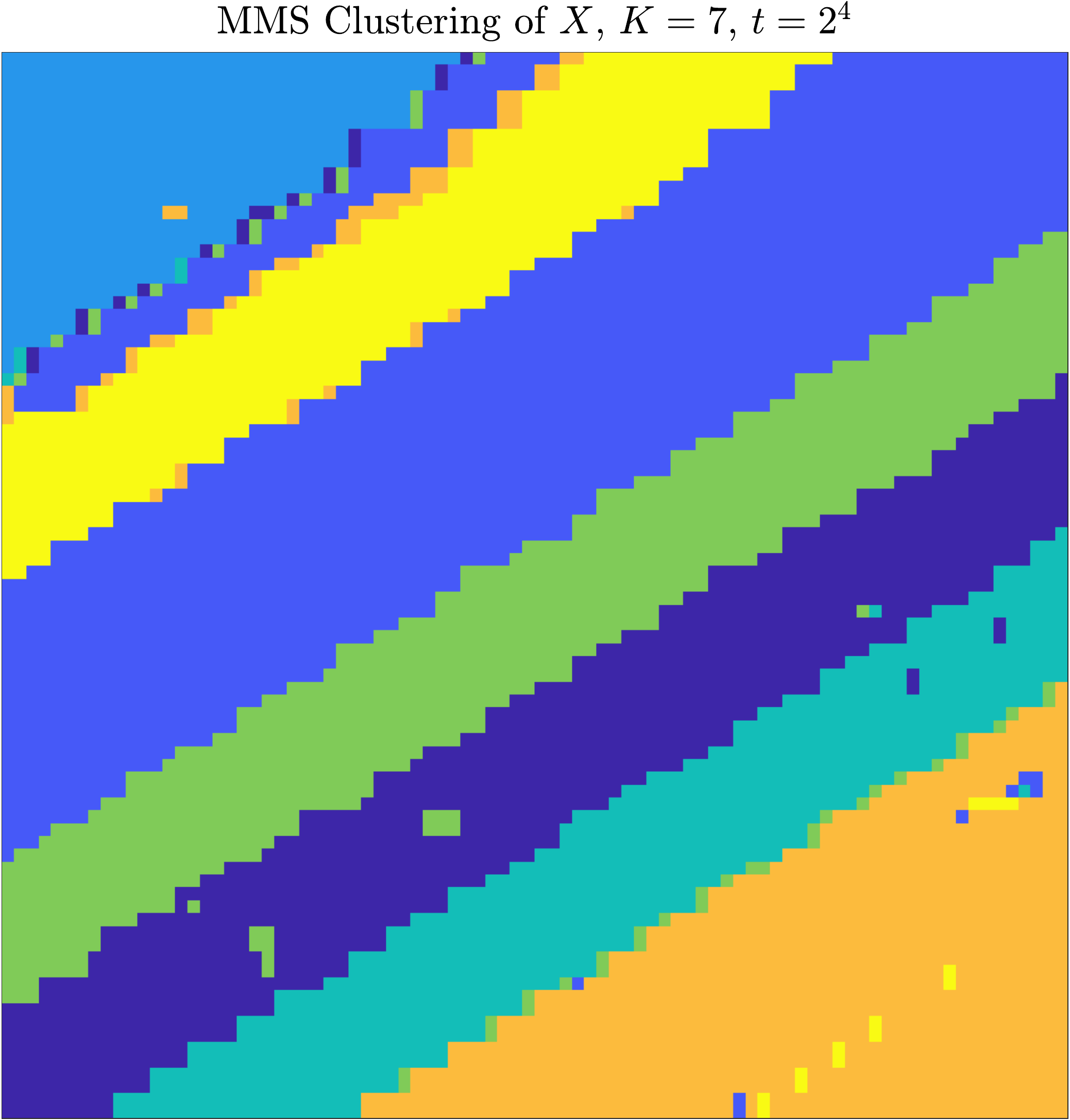}
    \end{subfigure}%
      \begin{subfigure}[t]{0.25\textwidth}
        \centering
        \includegraphics[height = 1.4in]{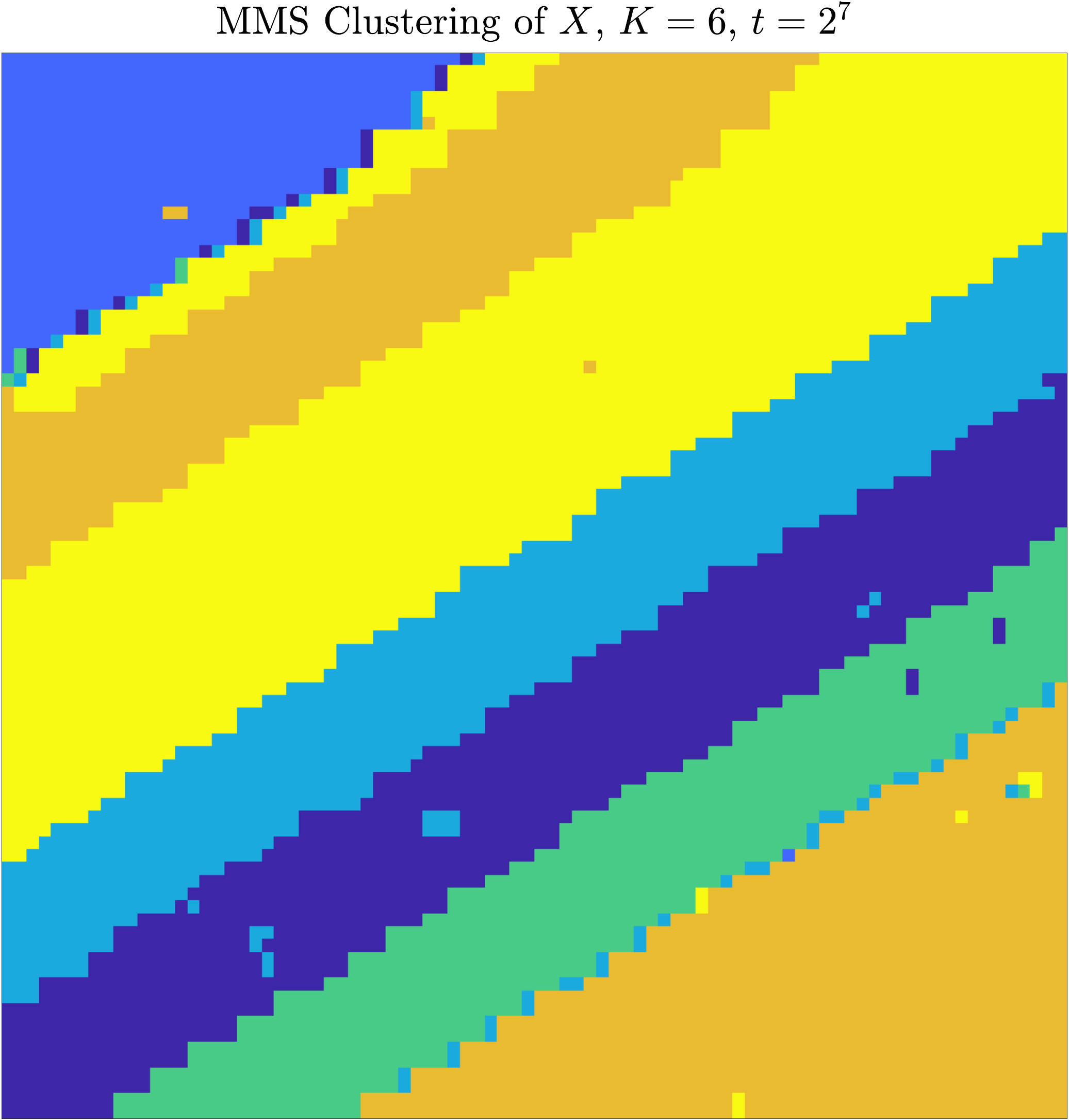}
    \end{subfigure}%
    \caption{Nontrivial clusterings of the Salinas A HSI~\cite{gualtieri1999salinasA} extracted by MMS clustering~\cite{liu2020MarkovStability}. Fine-scale multiscale structure is extracted from the HSI, but MMS clusterings did not return any clusterings coarser than the $K=6$ clustering in the rightmost panel.}\label{fig: Salinas A MMS}
\end{figure}

As in Section \ref{sec: benchmark}, we compare the performance of the M-LUND clustering algorithm against MMS clustering, HSC, and SLC. The same KNN graph with Gaussian kernel and same exponential sampling of the diffusion process were used for diffusion-based algorithms (M-LUND, MMS, and HSC). All algorithms except for MMS clustering produce 5-cluster and 2-cluster clusterings of the Salinas A image (Figure \ref{fig: Salinas A Comparison}). The clusterings produced by SLC do not meaningfully correspond to ground truth labels. In contrast, the M-LUND algorithm and HSC extract clusterings that can be related to the ground truth labels in Figure \ref{fig: Salinas A Data}. The $K=5$ clusterings generated by the two algorithms are similar, but romaine lettuce clusters estimated by the M-LUND algorithm are slightly more coherent than those estimated by HSC (Figure \ref{fig:comparison5}). M-LUND and HSC produce identical $K=2$ clusterings, wherin broccoli greens pixels are separated from all else in the scene (Figure \ref{fig:comparison2}). We observe a different trend in the outputs of MMS clustering (Figure \ref{fig: Salinas A MMS}). Indeed, clusterings rapidly transition early in the diffusion process from $K=12$ to $K=6$. This different trend may be due to MMS clustering not explicitly relying on the spectral decomposition of $\mathbf{P}$. In contrast, the M-LUND algorithm and HSC directly rely on the eigenfunctions of $\mathbf{P}$ to learn multiscale structure from $X$. Regardless, it is clear from the results of this section that a nonlinear diffusion-based clustering scheme is able to extract latent multiscale structure from the Salinas A HSI. 

\section{Conclusions and Future Work} \label{sec: Conclusion}

We have shown that Markov chains derived from a data-generated graph facilitate the detection of clusterings at many scales, and the specific scale at which one wishes to cluster is tightly linked to the diffusion time parameter. With this in mind, we introduced the Multiscale Environment for Learning by Diffusion (MELD) data model: a family of latent clusterings of the dataset, parameterized by the diffusion time parameter. We have shown that each clustering in the MELD data model can be separated by diffusion distances during an interval of time that depends on the geometry of the dataset and that clustering. We showed that clusterings that consist of well-separated, coherent clusters are more stable in the diffusion process and will occur more frequently in the MELD data model. 

We introduced the Multiscale Learning by Unsupervised Nonlinear Diffusion (M-LUND) clustering algorithm. The M-LUND algorithm is a multiscale extension of the LUND algorithm, which was introduced to leverage the attractive theoretical properties of diffusion distances~\cite{murphy2019LUND, nadler2006diffusion_maps_ACHA, coifman2006diffusionmaps, coifman2005PNAS}. M-LUND learns multiscale cluster structure from data by varying a time parameter in the LUND algorithm across an exponential sampling of the diffusion process.  It was proved that under reasonable assumptions on density and cluster structure, the M-LUND algorithm is guaranteed to extract an exponential sampling of the MELD data model from the dataset and choose a clustering from it as the minimizer of total VI.  Our theoretical results were corroborated on synthetic and real data experiments.

The reliance of the MELD data model on $\epsilon$ results in a tension between the choice of an exponential sampling rate $\beta$ that will sample the intervals during which MELD clusterings are $\epsilon$-separable by diffusion distances and the guarantee that diffusion distances produce strong enough separation for M-LUND to recover the MELD data model. For $\epsilon$ small, diffusion distances are guaranteed to provide strong separation on MELD clusterings during MELD intervals, but the range of $\beta$ that is guaranteed to sample these intervals is small. On the other hand, for $\epsilon$ large, there is a wide range of $\beta$ that are suitable for sampling these intervals, but in this case, $\epsilon$-separation by diffusion distances may not be strong enough to guarantee that the M-LUND algorithm will recover the MELD data model.

A limitation of the MELD data model is its reliance on the geometric constant $\delta^{(\ell)}$: the maximum probability, across all points in $X$, of transitioning between clusters in a single time step. In a dataset in which outliers in one cluster overlap with outliers in another, $\delta^{(\ell)}$ can be quite pessimistic. For such a dataset, $\delta^{(\ell)}$ will be large, but diffusion is not likely to spread between cluster cores~\cite{murphy2019LUND}. Indeed, in Section \ref{sec: 3D Gaussians 5}, we showed that the LUND algorithm performed well on overlapping Gaussians even though the separation parameter $\delta^{(\ell)}$ was large across the extracted clusterings. This suggests that the reliance of the MELD data model on the geometric constant $\delta^{(\ell)}$ forces it to exclude latent partitions of $X$ that lack sufficiently strong separation within the original graph. Thus, the MELD data model may be improved by future work to allow for weaker separation between clusters. 

To capture all scales of latent structure in a dataset, the M-LUND algorithm implements the LUND algorithm across an exponential sampling of the diffusion process. However, our numerical experiments suggest that latent clusterings are only extracted during a subset of the diffusion process. If this subset could be more precisely estimated before cluster analysis, the LUND algorithm could be implemented at those time steps alone, resulting in a decrease in complexity for the M-LUND algorithm. We hope to study this problem in future work as well. 

All results in the present manuscript are for finite samples, and there is a dependence on $n$ in many results.  It is natural to consider continuum formulations of the data model and associated clustering algorithms, which may necessitate new models for multiscale mixtures of manifold data. 
 
\section*{Acknowledgements}

This research is partially supported by the US National Science Foundation grants NSF-DMS 1912737, NSF-DMS 1924513, and NSF-CCF 1934553. We thank Z. Liu and M. Barahona (Imperial College London) for their code.

\section*{Declaration of Interests}
The authors declare no competing interests.

\section*{Availability of Data and Code}

Code to replicate results is available at \noindent \url{https://github.com/sampolk/MultiscaleDiffusionClustering}.

\bibliographystyle{alpha} 
\bibliography{biblio}
\end{document}